\documentclass[11pt]{book}              

\usepackage{xcolor}
\usepackage{amsmath}
\usepackage{amssymb}
\usepackage[round]{natbib}
\usepackage{graphicx}
\usepackage{amsthm}
\usepackage{tgpagella}
\usepackage{emptypage}
\usepackage{multirow}
\usepackage{booktabs}
\usepackage[LGR,T1]{fontenc}
\newcommand{\textgreek}[1]{\begingroup\fontencoding{LGR}\selectfont#1\endgroup}

\usepackage{tcolorbox}

\usepackage{microtype}

\definecolor{echodrk}{HTML}{0099cc}
\definecolor{mymauve}{rgb}{0.58,0,0.82}
\usepackage[colorlinks,linkcolor=mymauve,citecolor=echodrk]{hyperref}

\definecolor{boxgray}{rgb}{0.9,0.9,0.9}
\definecolor{boxpink}{rgb}{0.9,0.7,0.7}

\renewcommand\vec{\mathbf}

\makeatletter
\renewcommand{\thesection}{%
  \ifnum\c@chapter<1 \@arabic\c@section
  \else \thechapter.\@arabic\c@section
  \fi
}
\makeatother

\usepackage{marginnote}

\usepackage{amsmath,amsfonts,bm}

\def\eqref#1{equation~\ref{#1}}

\def\1{\bm{1}}

\DeclareMathAlphabet{\mathsfit}{\encodingdefault}{\sfdefault}{m}{sl}
\SetMathAlphabet{\mathsfit}{bold}{\encodingdefault}{\sfdefault}{bx}{n}

\def\fG{{\mathfrak{G}}}
\def\fH{{\mathfrak{H}}}
\def\fg{{\mathfrak{g}}}
\def\fh{{\mathfrak{h}}}

\def\fe{{\mathfrak{e}}}
\def\fk{{\mathfrak{k}}}

\newcommand{\GL}[1]{\ensuremath{\operatorname{GL}(#1)}}

\newcommand{\E}[1]{\ensuremath{\operatorname{E}(#1)}}
\newcommand{\SE}[1]{\ensuremath{\operatorname{SE}(#1)}}
\newcommand{\Orth}[1]{\ensuremath{\operatorname{O}(#1)}}
\newcommand{\SO}[1]{\ensuremath{\operatorname{SO}(#1)}}

\newcommand{\diff}[1]{\ensuremath{\operatorname{Diff}(#1)}}

\def\gC{{\mathcal{C}}}
\def\gD{{\mathcal{D}}}
\def\gE{{\mathcal{E}}}
\def\gF{{\mathcal{F}}}
\def\gG{{\mathcal{G}}}
\def\gH{{\mathcal{H}}}

\def\gR{{\mathcal{R}}}

\def\gV{{\mathcal{V}}}

\def\gX{{\mathcal{X}}}
\def\gY{{\mathcal{Y}}}

\newcommand{\R}{\mathbb{R}}
\newcommand{\Z}{\mathbb{Z}}

\def\thetab{\boldsymbol{\theta}}

\newcommand*{\ldblbrace}{\{\mskip-5mu\{}
\newcommand*{\rdblbrace}{\}\mskip-5mu\}}

\newcommand\mi{\mathrm{i}}

\definecolor{olivegreen}{HTML}{006400}
\definecolor{echoblue}{HTML}{0099CC}
\definecolor{gold}{HTML}{F09A00}
\definecolor{vividred}{HTML}{E60B42}
\definecolor{echonavy}{HTML}{0054B2}
\definecolor{darkgry}{HTML}{333333}
\definecolor{echopurple}{HTML}{9400D1}

\newcommand*{\horzbar}{\rule[.5ex]{2.5ex}{0.5pt}}

\newtheoremstyle{break}
  {\topsep}{\topsep}%
  {\itshape}{}%
  {\bfseries}{}%
  {\newline}{}%

\usepackage{fancyhdr}

\fancypagestyle{plain}{ %
  \fancyhf{} 
  
}

\fancypagestyle{fancybook}{%
    \fancyhf{}%
    \renewcommand*{\sectionmark}[1]{ \markright{\thesection\ ##1} }%
    \fancyhfoffset[LE]{6mm}
    \fancyhfoffset[RO]{6mm}%
    \fancyhead[LE]{\thepage\hskip3mm\vrule\hskip3mm BRONSTEIN, BRUNA, COHEN \& VELI\v{C}KOVI\'{C}}%
    \fancyhead[RO]{\rightmark\hskip3mm\vrule\hskip3mm\thepage}%
}
\usepackage[titletoc]{appendix}

\title{\bf Geometric Deep Learning \\
Grids, Groups, Graphs,\\ Geodesics, and Gauges}    
\author{Michael M. Bronstein\footnote{Imperial College London / USI IDSIA / Twitter}, Joan Bruna\footnote{New York University}, Taco Cohen\footnote{Qualcomm AI Research. Qualcomm AI Research is an initiative of Qualcomm Technologies, Inc.}, Petar Veli\v{c}kovi\'{c}\footnote{DeepMind}}              
\date{\today}                           

\begin{document}                        
\frontmatter                            
\maketitle                              
{\hypersetup{linkcolor=black}
\tableofcontents
}        
\mainmatter                             

\section*{Preface}  
\addcontentsline{toc}{section}{Preface}
\sectionmark{Preface}

For nearly two millenia since Euclid's \emph{Elements}, the word\marginnote{
\includegraphics[width=0.9\linewidth]{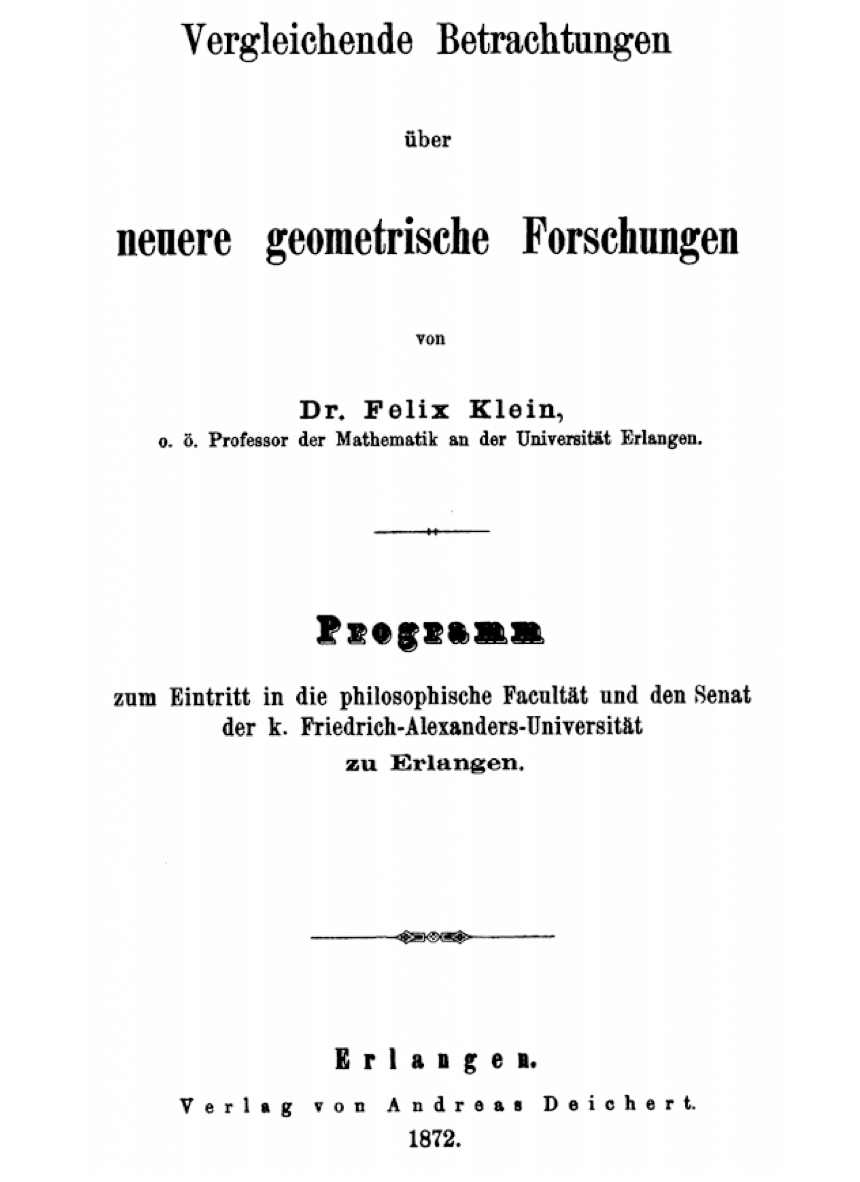}
According to a popular belief, the Erlangen Programme was delivered in Klein's inaugural address in October 1872. Klein indeed gave such a talk (though on December 7 of the same year), but it was for a non-mathematical audience and concerned primarily his ideas of mathematical education. What is now called the `Erlangen Programme' was actually a research prospectus brochure {\em Vergleichende Betrachtungen über neuere geometrische Forschungen} (``A comparative review of recent researches in geometry'') he prepared as part of his professor appointment. See \cite{tobies2019felix}. 
} `geometry' has been synonymous with \emph{Euclidean geometry}, as no other types of geometry existed. 
Euclid's monopoly came to an end in the nineteenth century, with examples of non-Euclidean geometries constructed by Lobachevesky, Bolyai, Gauss, and Riemann. 
%
%
Towards the end of that century, these studies had diverged into disparate fields, with 
mathematicians and philosophers debating the validity of and relations between these geometries as well as the nature of the ``one true geometry''.

A way out of this pickle was shown by a young mathematician Felix Klein, appointed in 1872 as professor in the small Bavarian University of Erlangen. In a research prospectus, which entered the annals of mathematics as the \emph{Erlangen Programme}, Klein proposed approaching geometry as the study of {\em invariants}, i.e. properties unchanged under some class of transformations, called the {\em symmetries} of the geometry.
%
This approach created clarity by showing that various geometries known at the time could be defined by an appropriate choice of symmetry transformations, formalized using the language of group theory. 
%
For instance, Euclidean geometry is concerned with lengths and angles, because these properties are preserved by the group of Euclidean transformations (rotations and translations), while affine geometry studies parallelism, which is preserved by the group of affine transformations. 
The relation between these geometries is immediately apparent when considering the respective groups, because the Euclidean group is a subgroup of the affine group, which in turn is a subgroup of the group of projective transformations. 

The impact of the Erlangen Programme on geometry was very profound.
Furthermore, it spilled to other fields, especially physics, where symmetry principles allowed to derive conservation laws from first principles of symmetry (an astonishing result known as Noether's Theorem), 
and even enabled the classification of elementary particles as irreducible representations of the symmetry group.
{\em Category theory}, 
now pervasive in pure mathematics,
can be ``regarded as a continuation of the Klein Erlangen Programme, in the sense that a geometrical space with its group of transformations is generalized to a category with its algebra of mappings'', in the words of its creators Samuel Eilenber and Saunders Mac Lane. \marginnote{See \cite{marquis2009category}. }


At the time of writing, the state of the field of deep learning is somewhat reminiscent of the field of geometry in the nineteenth century. 
There is a veritable zoo of neural network architectures for various kinds of data, but few unifying principles. 
As in times past, this makes it difficult to understand the relations between various methods, inevitably resulting in the reinvention and re-branding of the same concepts in different application domains.  
For a novice trying to learn the field, absorbing the sheer volume of redundant ideas is a true nightmare.

In this text, we make a modest attempt to apply the Erlangen Programme mindset to the domain of deep learning, with the ultimate goal of obtaining a systematisation of this field and `connecting the dots'. 
We call this geometrisation attempt `Geometric Deep Learning', and true to the spirit of Felix Klein, propose to derive different inductive biases and network architectures implementing them from first principles of symmetry and invariance. 
%
%
In particular, we focus on a large class of neural networks designed for analysing unstructured sets, grids, graphs, and manifolds, and show that they can be understood in a unified manner as methods that respect the structure and symmetries of these domains. 
%

We believe this text would appeal to a broad audience of deep learning researchers, practitioners, and enthusiasts. 
A novice may use it as an overview and introduction to Geometric Deep Learning. 
A seasoned deep learning expert may discover new ways of deriving familiar architectures from basic principles and perhaps some surprising connections.  
Practitioners may get new insights on how to solve problems in their respective fields.

With such a fast-paced field as modern machine learning, the risk of writing a text like this is that it becomes obsolete and irrelevant before it sees the light of day. Having focused on foundations, our hope is that the key concepts we discuss will transcend their specific realisations\marginnote{``The knowledge of certain principles easily compensates the lack of knowledge of certain facts.'' \citep{helvetius1759esprit}} --- or, as Claude Adrien Helvétius put it, {\em ``la connaissance de certains principes supplée facilement à la connoissance de certains faits.'' }

\section*{Notation}
\addcontentsline{toc}{section}{Notation}
\sectionmark{Notation}

\begin{minipage}{\textwidth}
\def\arraystretch{1.5}
\begin{tabular}{lp{3.25in}}
$\Omega,u$ & Domain, point on domain\\
$x(u) \in \mathcal{X}(\Omega,\mathcal{C})$ & Signal on the domain of the form $x:\Omega\rightarrow \mathcal{C}$\\
$f(x) \in \mathcal{F}(\mathcal{X}(\Omega))$ & Functions on signals on the domain of the form $f:\mathcal{X}(\Omega) \rightarrow \mathcal{Y}$\\
$\fG,\fg$ & Group, element of the group\\
$\fg.u, \rho(\fg)$ & Group action, group representation\\
$\vec{X}\in\mathcal{C}^{|\Omega|\times s}$ & Matrix representing a signal on a discrete domain\\
$\vec{x}_u\in\mathcal{C}^{s}$ & Vector representing a discrete domain signal $\vec{X}$ on element $u\in\Omega$\\
$x_{uj}\in\mathcal{C}$ & Scalar representing the $j$th component of a discrete domain signal $\vec{X}$ on element $u\in\Omega$\\
$\vec{F}(\vec{X})$ & Function on discrete domain signals that returns another discrete domain signal, as a matrix\\
$\tau:\Omega\rightarrow\Omega$ & Automorphism of the domain \\
$\eta:\Omega\rightarrow\Omega'$ & Isomorphism between two different domains \\
$\sigma: \mathcal{C}\rightarrow\mathcal{C}'$ & Activation function (point-wise non-linearity)\\
$G=(\mathcal{V},\mathcal{E})$ & Graph with nodes $\mathcal{V}$ and edges $\mathcal{E}$ \\
$\mathcal{T}=(\mathcal{V},\mathcal{E},\mathcal{F})$ & Mesh with nodes $\mathcal{V}$,  edges $\mathcal{E}$, and faces $\mathcal{F}$ \\
$x\star \theta$ & Convolution with filter $\theta$ \\
$S_v$ & Shift operator \\
$\varphi_i$ & Basis function \\
$T_u\Omega, T\Omega$ & Tangent space at $u$, tangent bundle \\
$X \in T_u\Omega$ & Tangent vector\\
$g_u(X,Y) = \langle X, Y\rangle_u$ & Riemannian metric\\
$\ell(\gamma), \ell_{uv}$ & Length of a curve $\gamma$, discrete metric on edge $(u,v)$ \\

\end{tabular}
\end{minipage}
\pagebreak

\section{Introduction}                

The last decade has witnessed an experimental revolution in data science and machine learning, epitomised by deep learning methods. Indeed, many high-dimensional learning tasks previously thought to be beyond reach -- such as computer vision, playing Go, or protein folding -- are in fact feasible with appropriate computational scale.  
 Remarkably, the essence of deep learning is built from two simple algorithmic principles: 
 first, the notion of representation or \emph{feature learning}, whereby adapted, often hierarchical, features  capture the appropriate notion of regularity for each task, and second, learning by local gradient-descent, typically implemented as {\em backpropagation}. 

While learning generic functions in high dimensions is a cursed estimation problem, most tasks of interest are not generic, and come with essential pre-defined regularities arising from the underlying low-dimensionality and structure of the physical world. This text is concerned with exposing these regularities through unified geometric principles that can be applied throughout a wide spectrum of applications. 

Exploiting the known symmetries of a large system is a powerful and classical remedy against the curse of dimensionality, and forms the basis of most physical theories. Deep learning systems are no exception, and since the early days researchers have adapted neural networks to exploit the low-dimensional geometry arising from physical measurements, e.g. grids in images, sequences in time-series, or position and momentum in molecules, and their associated symmetries, such as translation or rotation. 
Throughout our exposition, we will describe these models, as well as many others, as natural instances of the same underlying principle of geometric regularity.

Such a `geometric unification' endeavour 
in the spirit of the Erlangen Program 
serves a dual purpose: on one hand, it provides a common mathematical framework to study the most successful neural network architectures, such as CNNs, RNNs, GNNs, 
and Transformers. On the other, it gives a constructive procedure to incorporate prior physical knowledge into neural architectures 
and provide principled way to build future architectures yet to be invented.  

Before proceeding, it is worth noting that our work concerns \emph{representation learning architectures}  and exploiting the symmetries of data therein. The many exciting \emph{pipelines} where such representations may be used (such as self-supervised learning, generative modelling, or reinforcement learning) are \emph{not} our central focus\marginnote{The same applies for techniques used for \emph{optimising} or \emph{regularising} our architectures, such as Adam \citep{kingma2014adam}, dropout \citep{srivastava2014dropout} or batch normalisation \citep{ioffe2015batch}.}. Hence, we will not review in depth influential neural pipelines such as variational autoencoders \citep{kingma2013auto}, generative adversarial networks \citep{goodfellow2014generative}, normalising flows  \citep{rezende2015variational}, deep Q-networks \citep{mnih2015human}, proximal policy optimisation \citep{schulman2017proximal}, or deep mutual information maximisation \citep{hjelm2018learning}. That being said, we believe that the principles we will focus on are of significant importance in all of these areas.

Further, while we have attempted to cast a reasonably wide net in order to illustrate the power of our geometric blueprint, our work does not attempt to accurately summarise the \emph{entire} existing wealth of research on Geometric Deep Learning. Rather, we study several well-known architectures in-depth in order to demonstrate the principles and ground them in existing research, with the hope that we have left sufficient references for the reader to meaningfully apply these principles to any future geometric deep architecture they encounter or devise.

\section{Learning in High Dimensions}


    

Supervised machine learning, in its simplest formalisation, considers a set of $N$ observations $\gD=\{(x_i, y_i)\}_{i=1}^{ N}$ drawn \emph{i.i.d.} 
from an underlying data distribution $P$ defined over $\gX \times \gY$, where $\gX$ and $\gY$ are respectively the data and the label domains. The defining feature in this setup is that $\gX$ is a {\em high-dimensional space}: one typically assumes $\gX= \R^d$ to be a  Euclidean space  of large dimension $d$.


Let us further assume that the labels $y$ are generated by an unknown function $f$, such that $y_i = f(x_i)$, 
and the learning problem reduces to estimating the function $f$ using a parametrised function class $\gF=\{ f_{\thetab \in \Theta}\}$. Neural networks are a common realisation of such parametric function classes, in which case $\thetab \in \Theta$ corresponds to the network weights. 
In this idealised setup, there is no noise in the labels, and modern deep learning systems typically operate in the so-called \emph{interpolating regime}, where the estimated $\tilde{f} \in \gF$ satisfies $\tilde{f}(x_i) = f(x_i)$ for all $i=1,\hdots, N$.  
The performance of a learning algorithm is measured in terms of the \emph{expected performance} \marginnote{Statistical learning theory is concerned with more refined notions of generalisation based on {\em concentration inequalities}; we will review some of these in future work.} on new samples drawn from ${P}$, using some {\em loss} $L(\cdot,\cdot)$ 
$$\gR(\tilde{f}):= \mathbb{E}_{P}\,\, L(\tilde{f}(x), f(x)),$$
with the squared-loss $L(y,y')=\frac{1}{2}|y-y'|^2$ being among the most commonly used ones.

A successful learning scheme thus needs to encode the appropriate notion of regularity or  \emph{inductive bias} for $f$, imposed through the construction of the function class $\mathcal{F}$ and the use of {\em regularisation}. We briefly introduce this concept in the following section. 

\subsection{Inductive Bias via Function Regularity}
\label{sec:inductive}

Modern machine learning operates with large, high-quality datasets, which, together with appropriate computational resources, motivate the design of rich function classes $\gF$ with the capacity to interpolate such large data. 
This 
mindset plays well with neural networks, since even the simplest choices of architecture yields a {\em dense} class of functions.\marginnote{A set $\mathcal{A}\subset \mathcal{X}$ is said to be {\em dense} in $\mathcal{X}$ if its closure
$$
\mathcal{A}\cup \{ \displaystyle \lim_{i\rightarrow \infty} a_i : a_i \in \mathcal{A}\} = \mathcal{X}.
$$ 
This implies that any point in $\mathcal{X}$ is arbitrarily close to a point in $\mathcal{A}$. A typical Universal Approximation result shows that the class of functions represented e.g. by a two-layer perceptron, $f(\mathbf{x}) = \mathbf{c}^\top\mathrm{sign}(\mathbf{A}\mathbf{x}+\mathbf{b})$ is dense in the space of continuous functions on $\mathbb{R}^d$. 
} 
The capacity to approximate almost arbitrary functions
is the subject of various \emph{Universal Approximation Theorems}; several such results were proved and popularised in the 1990s by applied mathematicians and computer scientists (see e.g. \cite{cybenko1989approximation,hornik1991approximation,barron1993universal,leshno1993multilayer,maiorov1999best,pinkus1999approximation}). 

\begin{figure}[h!]
    \centering
    \includegraphics[width=1\linewidth]{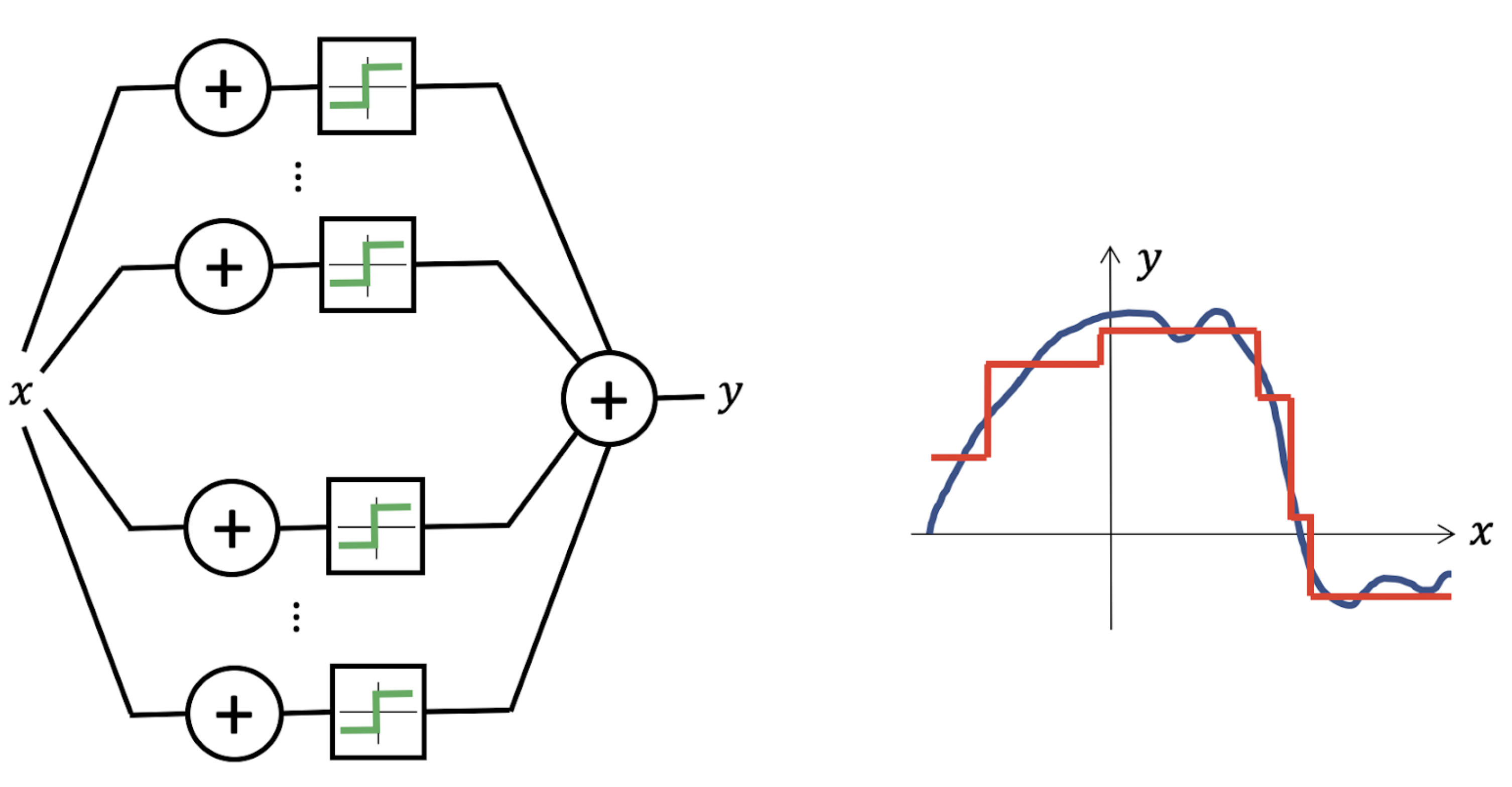}
    \caption{Multilayer Perceptrons \citep{rosenblatt1958perceptron}, the simplest feed-forward neural networks, are universal approximators: with just one hidden layer, they can represent combinations of step functions, allowing to approximate any continuous function with arbitrary precision.}
    \label{fig:ua_mlp}
\end{figure}%


Universal Approximation, however, does not imply an {\em absence} of inductive bias. Given a hypothesis space $\gF$ with universal approximation, we can define a complexity measure $c: \gF \to \R_{+}$ and redefine our interpolation problem as 
$$
\tilde{f} \in \arg\min_{g \in \gF} c(g) \quad \mathrm{s.t.} \quad g(x_i) = f(x_i)  \quad \mathrm{for} \,\,\, i=1, \hdots, N,
$$
i.e., we are looking for the most regular functions within our hypothesis class. 
For standard function spaces, this complexity measure can be defined as a {\em norm},\marginnote{
Informally, a norm $\|x\|$ can be regarded as a ``length'' of vector $x$. 
A {\em Banach space} is a complete vector space equipped with a norm. 
} making $\gF$ a {\em Banach space} and allowing to leverage a plethora of theoretical results in functional analysis. 
In low dimensions, splines are a workhorse for function approximation. They can be formulated as above, with a norm capturing 
the classical notion of smoothness, such as the squared-norm of second-derivatives $\int_{-\infty}^{+\infty} |f''(x)|^2 \mathrm{d}x$ for cubic splines. 


In the case of neural networks,  the complexity measure $c$ can be expressed in terms of the network weights, i.e. $c(f_{\boldsymbol{\theta}}) = {c}(\boldsymbol{\theta})$.
The $L_2$-norm of the network weights, known as \emph{weight decay}, or the so-called \emph{path-norm} \citep{neyshabur2015norm} are popular choices in deep learning literature.  
%
%
From a Bayesian perspective, such complexity measures can also be interpreted as the negative log of the prior for the function of interest. More generally, this complexity can be enforced \emph{explicitly} by incorporating it into the empirical loss (resulting in the so-called Structural Risk Minimisation), or \emph{implicitly}, as a result of a certain optimisation scheme. For example, it is well-known that gradient-descent on an under-determined least-squares objective will choose interpolating solutions with minimal $L_2$ norm. The extension of such implicit regularisation results to modern neural networks is the subject of current studies (see e.g. \cite{blanc2020implicit, shamir2020implicit, razin2020implicit, gunasekar2017implicit}).   
All in all, a natural question arises: how to define effective priors that capture the expected regularities and complexities of real-world prediction tasks? 

\subsection{The Curse of Dimensionality}

While interpolation in low-dimensions 
(with $d=1,2$ or $3$) is a classic signal processing task with very precise mathematical control of estimation errors using increasingly sophisticated regularity classes (such as spline interpolants, wavelets, curvelets, or ridgelets), the situation for high-dimensional problems is entirely different. 
%



\begin{figure}[h!]
    \centering
       \includegraphics[width=\linewidth]{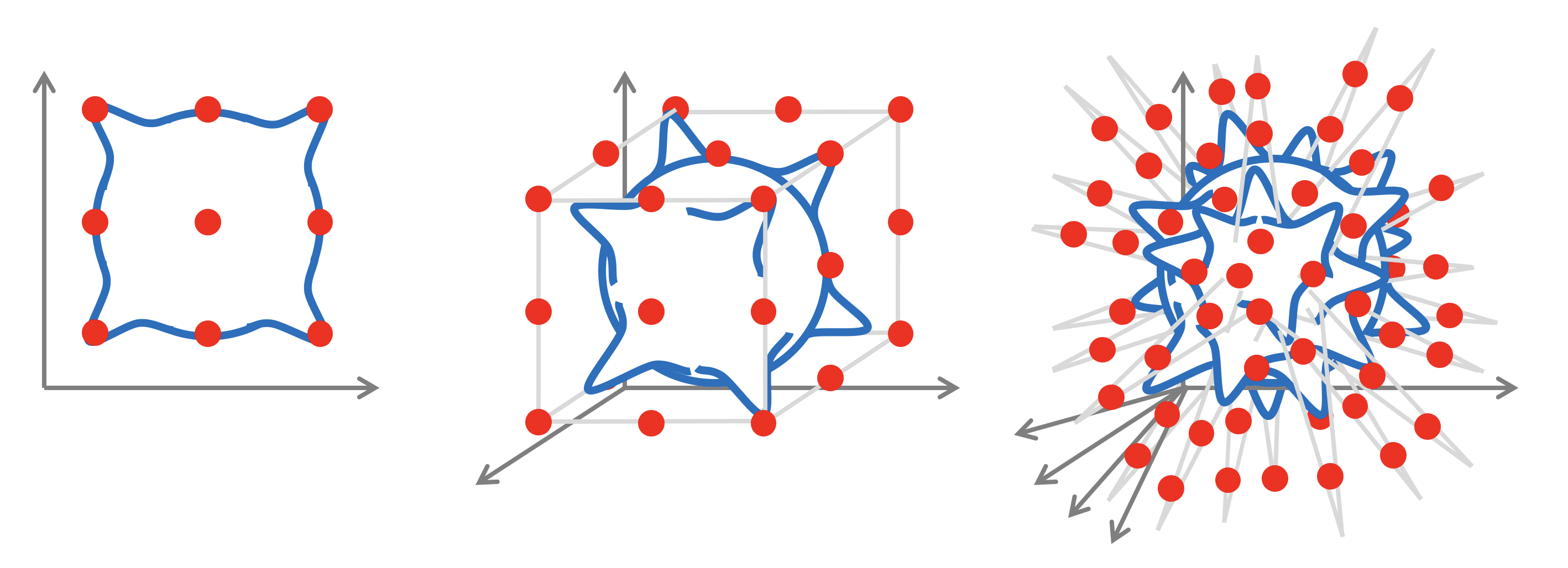} 
    \caption{We consider a Lipschitz function $f(x) = \sum_{j=1}^{2^d} z_j \phi(x-x_j)$ where $z_j=\pm 1$, $x_j \in \R^d$ is placed in each quadrant, and $\phi$ a locally supported Lipschitz `bump'. Unless we observe the function in most of the $2^d$ quadrants, we will incur in a constant error in predicting it. This simple geometric argument can be formalised through the notion of \emph{Maximum Discrepancy} \citep{von2004distance}, defined for the Lipschitz class as $\kappa(d)=\mathbb{E}_{x,x'} \sup_{f \in \mathrm{Lip}(1)} \left| \frac{1}{N}\sum_{l} f(x_l) - \frac{1}{N}\sum_{l} f(x'_l) \right| \simeq N^{-1/d}$, which measures the largest expected discrepancy between two independent $N$-sample expectations. Ensuring that $\kappa(d) \simeq \epsilon$ requires $N = \Theta(\epsilon^{-d})$; the corresponding sample $\{x_l\}_l$ defines an $\epsilon$-net of the domain. For a $d$-dimensional Euclidean domain of diameter $1$, its size grows exponentially as $\epsilon^{-d}$.}
    \label{fig:curseofdim}
\end{figure}

In order to convey the essence of the idea, let us consider a classical notion of regularity that can be easily extended to high dimensions: 1-Lipschitz- functions $f:\mathcal{X} \to \R$, i.e. functions satisfying  $|f(x) - f(x')| \leq \|x - x'\|$ for all $x, x' \in \mathcal{X}$. This hypothesis only asks the target function to be \emph{locally} smooth, i.e., if we perturb the input $x$ slightly (as measured by the norm $\|x - x'\|$), the output $f(x)$ is not allowed to change much. 
If our only knowledge of the target function $f$ is that it is $1$-Lipschitz, how many observations do we expect to require to ensure that our estimate $\tilde{f}$ will be close to $f$? 
Figure \ref{fig:curseofdim} reveals that the general answer is necessarily exponential in the dimension $d$, signaling that the Lipschitz class grows `too quickly' as the input dimension increases: in many applications with even modest dimension $d$, the number of samples would be bigger than the number of atoms in the universe.
 The situation is not better if one replaces the Lipschitz class by a global smoothness hypothesis, such as the Sobolev Class $\gH^{s}(\Omega_d)$\marginnote{A function $f$ is in the {\em Sobolev class} $\gH^{s}(\Omega_d)$ if $f \in L^2(\Omega_d)$ and the generalised $s$-th order derivative is square-integrable: $\int |\omega|^{2s+1} |\hat{f}(\omega)|^2 d\omega < \infty$, where 
$\hat{f}$ is the Fourier transform of $f$; see Section~\ref{sec:grids_euclidean}. 
}. Indeed, classic results \citep{tsybakov2008introduction} establish a minimax rate of approximation and learning for the Sobolev class of the order $\epsilon^{-d/s}$, showing that the extra smoothness assumptions on $f$ only improve the statistical picture when $s \propto d$, an unrealistic assumption in practice. 



\begin{figure}[htbp]
    \centering
\includegraphics[width=0.6\linewidth]{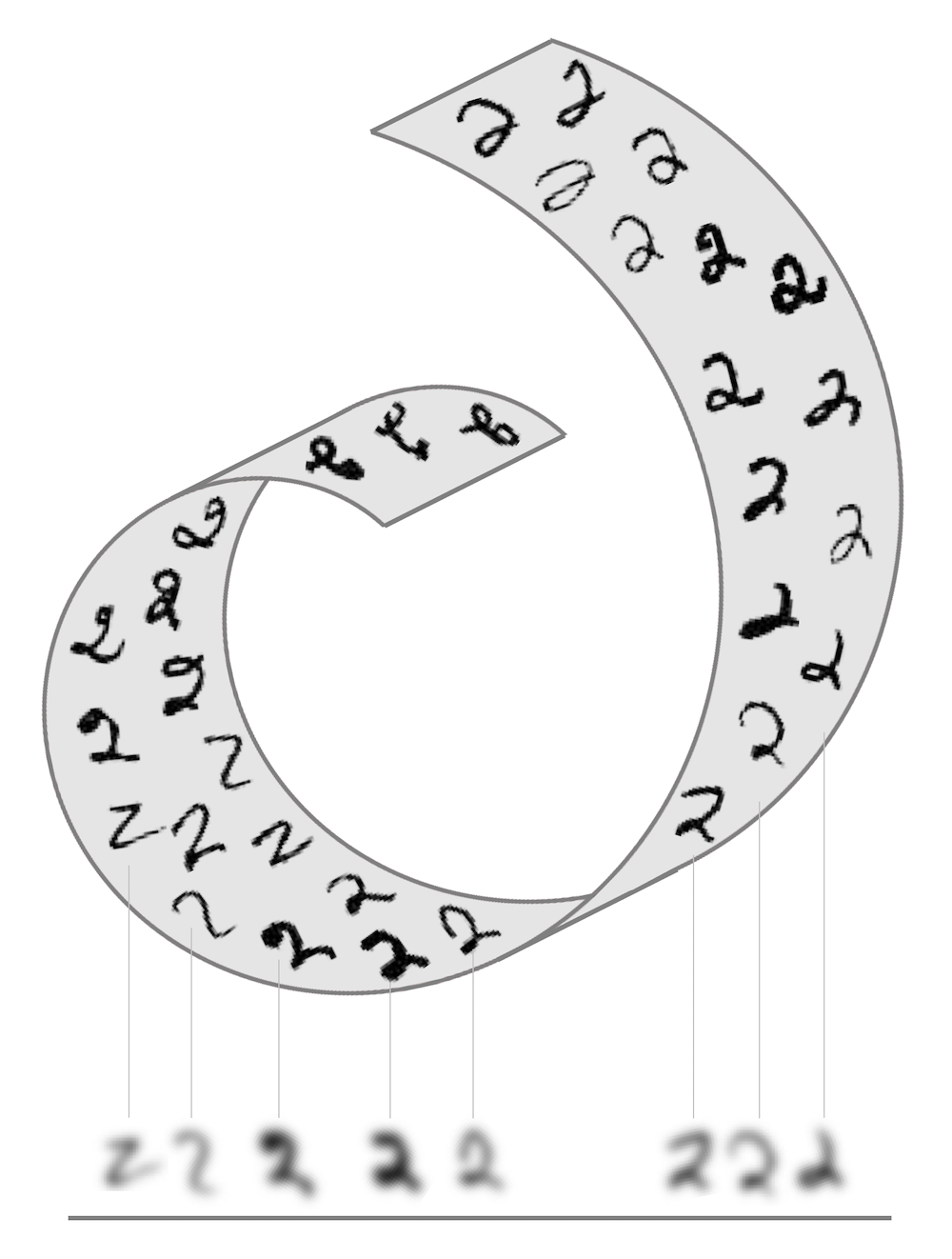}
    \caption{
    If the unknown function $f$ 
    is presumed to be well approximated as $f(\mathbf{x}) \approx g(\mathbf{A}\mathbf{x})$ for some unknown $\mathbf{A} \in \mathbb{R}^{k \times d}$ with $k \ll d$, then shallow neural networks can capture this inductive bias, see e.g.  \cite{bach2017breaking}. In typical applications, such dependency on low-dimensional projections is unrealistic, as illustrated in this example: a low-pass filter projects the input images to a low-dimensional subspace; while it conveys most of the semantics, substantial information is lost.
    }
    \label{fig:curseofdim2}
\end{figure}

Fully-connected neural networks define function spaces that enable more flexible notions of regularity, obtained by considering complexity functions $c$ on their weights. In particular, by choosing a sparsity-promoting regularisation, they have the ability to break this curse of dimensionality \citep{bach2017breaking}. However, this comes at the expense of making strong assumptions on the nature of the target function $f$, such as that $f$ depends on a collection of low-dimensional projections of the input (see Figure \ref{fig:curseofdim2}).
In most real-world applications (such as computer vision, speech analysis, physics, or  chemistry), functions of interest 
tend to exhibits complex long-range correlations that cannot be expressed with low-dimensional projections (Figure \ref{fig:curseofdim2}), making this hypothesis unrealistic.  
It is thus necessary to define an alternative source of regularity, by exploiting the spatial structure of the physical domain and the geometric priors of $f$, as we describe in the next Section~\ref{sec:geom_priors}. 

\newpage
\section{Geometric Priors}
\label{sec:geom_priors}
Modern data analysis is synonymous with high-dimensional learning.  
While the simple arguments of Section~\ref{sec:inductive} reveal the impossibility of learning 
from generic high-dimensional data as a result of the curse of dimensionality, 
there is hope for physically-structured data, where we can employ two fundamental principles: {\em symmetry} and {\em scale separation}. 
In the settings considered in this text, this additional structure will usually come from the structure of the domain underlying the input signals: 
we will assume that our machine learning system operates on \emph{signals} (functions) on some domain $\Omega$. 
While in many cases linear combinations of points on $\Omega$ is not well-defined\marginnote{$\Omega$ must be a vector space in order for an expression $\alpha u + \beta v$ to make sense.}, we can linearly combine signals on it, i.e., the space of signals forms a vector space. 
Moreover, since we can define an inner product between signals, this space is a {\em Hilbert space}.

\begin{tcolorbox}[width=\linewidth,
                  boxsep=0pt,
                  left=7.5pt,
                  right=7.5pt,
                  top=7.5pt,
                  bottom=7.5pt,
                  arc=0pt,
                  boxrule=0pt,toprule=0pt,
                  colback=boxgray,
                  ]

The space of $\mathcal{C}$-valued signals on $\Omega$ 
\marginnote{When $\Omega$ has some additional structure, we may further restrict the kinds of signals in $\mathcal{X}(\Omega, \mathcal{C})$. For example, when $\Omega$ is a smooth manifold, we may require the signals to be smooth. 
Whenever possible, we will omit the range $\mathcal{C}$ for brevity. 
} 
(for $\Omega$ a set, possibly with additional structure, and $\mathcal{C}$ a vector space, whose dimensions are called \emph{channels}) 
\begin{equation}
    \mathcal{X}(\Omega, \mathcal{C}) = \{ x : \Omega \rightarrow \mathcal{C} \}
\end{equation}
is a function space that has a vector space structure. 
Addition and scalar multiplication of signals is defined as:
\begin{equation*}
    (\alpha x + \beta y)(u) = \alpha x(u) + \beta y(u) \quad \text{for all} \quad u\in \Omega,
\end{equation*}
with real scalars $\alpha, \beta$. 
Given an inner product $\langle v, w \rangle_\mathcal{C}$ on $\mathcal{C}$ and a measure\marginnote{When the domain $\Omega$ is discrete, $\mu$ can be chosen as the {\em counting measure}, in which case the integral becomes a sum. In the following, we will omit the measure and use $\mathrm{d}u$ for brevity.  } $\mu$ on $\Omega$ (with respect to which we can define an integral), we can define an inner product on $\mathcal{X}(\Omega, \mathcal{C})$ as 
\begin{equation}
    \langle x, y \rangle = \int_{\Omega} \langle x(u), \, y(u) \rangle_{\mathcal{C}} \; \mathrm{d}\mu(u).
    \label{eqn:innerprod}
\end{equation}
\end{tcolorbox}


As a typical illustration, take $\Omega = \mathbb{Z}_n\times \mathbb{Z}_n$ to be a two-dimensional $n\times n$ grid, $x$ an RGB image (i.e. a signal $x : \Omega \rightarrow \R^3$), and $f$ a function (such as a single-layer Perceptron) operating on $3n^2$-dimensional inputs. 
As we will see in the following with greater detail, the domain $\Omega$ is usually endowed with certain geometric structure and symmetries. 
Scale separation results from our ability to preserve important characteristics of the signal when transferring it onto a coarser version of the domain (in our example, subsampling the image by coarsening the underlying grid).

We will show that both principles, to which we will generically refer as {\em geometric priors}, are prominent in most modern deep learning architectures. In the case of images considered above, geometric priors are built into Convolutional Neural Networks (CNNs) in the form of convolutional filters with shared weights (exploiting translational symmetry) and pooling (exploiting scale separation). 
Extending these ideas to other domains such as graphs and manifolds and showing how geometric priors emerge from fundamental principles is the main goal of Geometric Deep Learning and the {\em leitmotif} of our text.

\subsection{Symmetries, Representations, and Invariance}
\label{sec:symmetries}


Informally, a {\em symmetry} of an object or system is a transformation that leaves a certain property of said object or system unchanged or {\em invariant}.
Such transformations may be either smooth, continuous, or discrete.
%
Symmetries are ubiquitous in many machine learning tasks.
For example, in computer vision the object category is unchanged by shifts, so shifts are symmetries in the problem of visual object classification.
%
In computational chemistry, the task of predicting properties of molecules independently of their orientation in space requires {\em rotational invariance}. 
Discrete symmetries emerge naturally when describing particle systems where particles do not have canonical ordering and thus can be arbitrarily permuted, as well as
in many dynamical systems, via the time-reversal symmetry (such as systems in detailed balance or the Newton's second law of motion). 
As we will see in Section~\ref{sec:proto-graphs}, permutation symmetries are also central to the analysis of graph-structured data. 

\paragraph{Symmetry groups}
The set of symmetries of an object satisfies a number of properties. 
First, symmetries may be combined to obtain new symmetries: if $\fg$ and $\fh$ are two symmetries,
then their  compositions $\fg \circ \fh$ and $\fh \circ \fg$
\marginnote{We will follow the juxtaposition notation convention used in group theory, $\fg \circ \fh = \fg \fh$, which should be read right-to-left: we first apply $\fh$ and then $\fg$. The order is important, as in many cases symmetries are non-commutative. 
Readers familiar with Lie groups might be disturbed by our choice to use the Fraktur font to denote group elements, as it is a  common notation of Lie algebras.
} are also symmetries.
The reason is that if both transformations leave the object invariant, then so does the composition of transformations, and hence the composition is also a symmetry.
Furthermore, symmetries are always invertible, and the inverse is also a symmetry.
This shows that the collection of all symmetries form an algebraic object known as a {\em group}. Since these objects will be a centerpiece of the mathematical model of Geometric Deep Learning, they deserve a formal definition and detailed discussion:


\begin{tcolorbox}[width=\linewidth,
                  boxsep=0pt,
                  left=7.5pt,
                  right=7.5pt,
                  top=7.5pt,
                  bottom=7.5pt,
                  arc=0pt,
                  boxrule=0pt,toprule=0pt,
                  colback=boxgray,
                  ]
A {\em group} is a set $\fG$ along with a binary operation $\circ : \fG \times \fG \rightarrow \fG$ called {\em composition} (for brevity, denoted by juxtaposition $\fg \circ \fh = \fg \fh$) 
satisfying the following axioms: \vspace{3mm}\\
    \noindent {\em Associativity:} $(\fg \fh) \fk = \fg (\fh \fk)$ for all $\fg, \fh, \fk \in \fG$.\vspace{2mm}\\
    \noindent  {\em Identity:} there exists a unique $\fe \in \fG$ satisfying $\fe \fg = \fg \fe = \fg$ for all $\fg \in \fG$.\vspace{2mm}\\
    \noindent  {\em Inverse:} For each $\fg \in \fG$ there is a unique inverse $\fg^{-1} \in \fG$ such that $\fg \fg^{-1} = \fg^{-1} \fg = \fe$.\vspace{2mm}\\
    \noindent  {\em Closure:} The group is closed under composition, i.e., for every $\fg, \fh \in \fG$, we have $\fg \fh \ \in \fG$.

\end{tcolorbox}
Note that \emph{commutativity} is not part of this definition, i.e. we may have $\fg \fh \neq \fh \fg$.
Groups for which $\fg \fh = \fh \fg$ for all $\fg, \fh \in \fG$ are called commutative or {\em Abelian}\marginnote{After the Norwegian mathematician Niels Henrik Abel (1802--1829).}.

\begin{figure}
    \centering
\raisebox{-0.5\height}{    \includegraphics[width=0.45\linewidth]{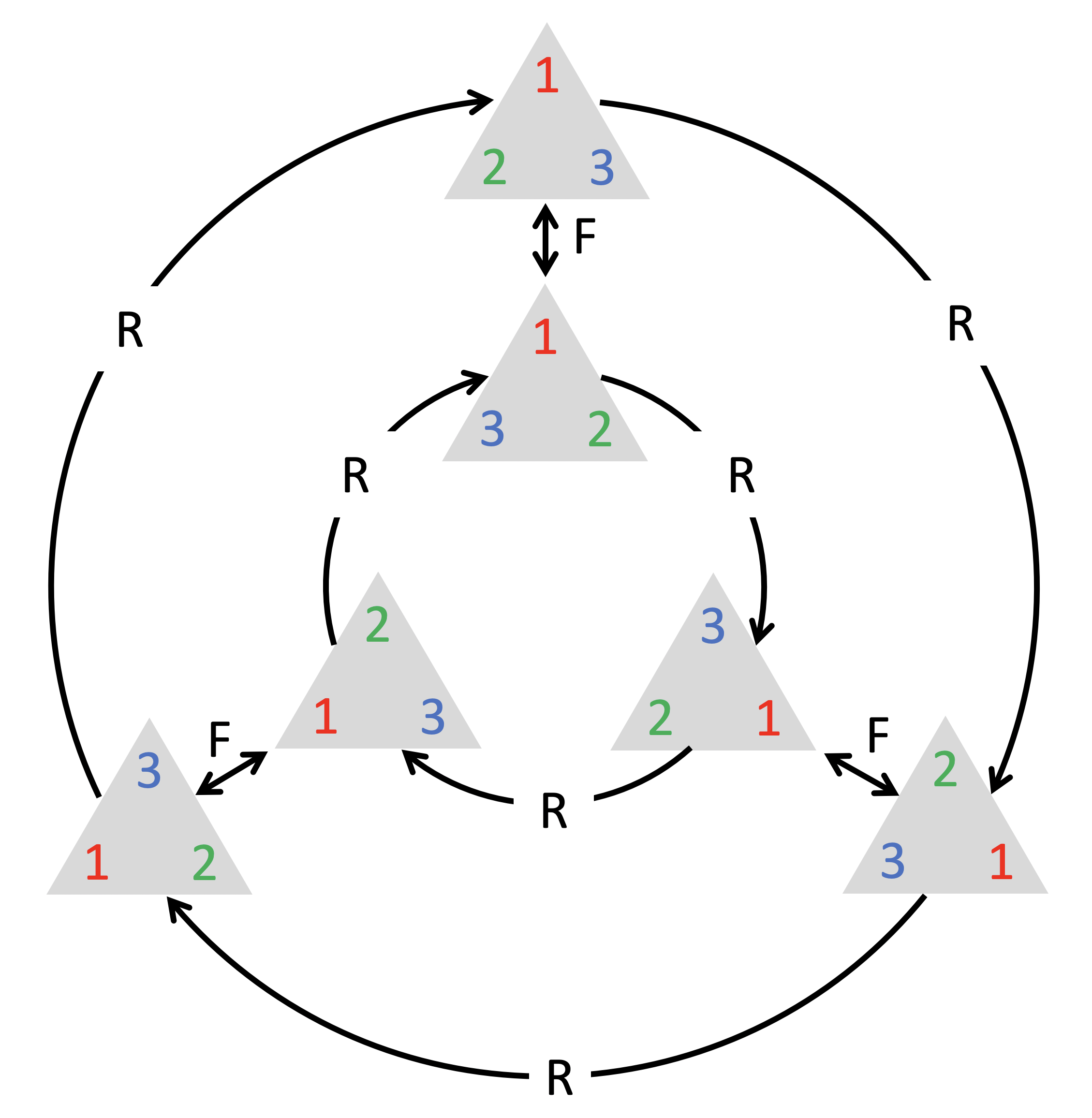}
}
    \hspace{2mm}
    \raisebox{-0.5\height}{
        \includegraphics[width=0.48\linewidth]{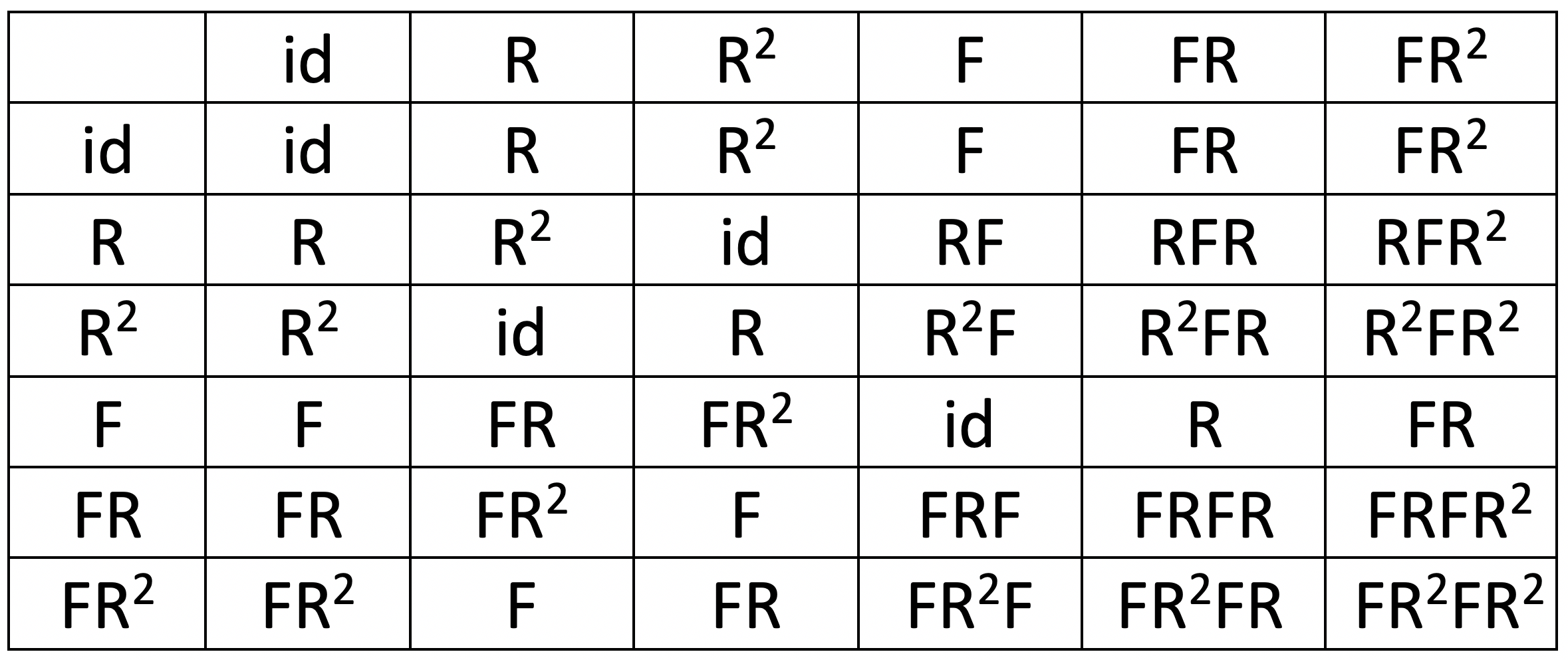}
        }
    \caption{Left: an equilateral triangle with corners labelled by $1, 2, 3$, and all possible rotations and reflections of the triangle. The group $\mathrm{D}_3$ of rotation/reflection symmetries of the triangle is generated by only two elements (rotation by $60^\circ$ R and reflection F) and is the same as the group $\Sigma_3$ of permutations of three elements. 
    Right: the multiplication table of the group $\mathrm{D}_3$. The element in the row $\fg$ and column $\fh$ corresponds to the element $\fg \fh$. 
    \label{fig:group-example-d3-s3}
    }
\end{figure}

Though some groups can be very large and even infinite, they often arise from compositions of just a few elements, called {\em group generators}. Formally, $\mathfrak{G}$ is said to be {\em generated} by a subset $S \subseteq \mathfrak{G}$ (called the group {\em generator}) if every element $\fg \in \fG$ can be written as a finite composition of the elements of $S$ and their inverses. 
For instance, the symmetry group of an equilateral triangle (dihedral group $\mathrm{D}_3$) is generated by a $60^\circ$ rotation and a reflection (Figure \ref{fig:group-example-d3-s3}). 
The 1D {\em translation group}, which we will discuss in detail in the following, is generated by infinitesimal displacements; this is an example of a {\em Lie group} of differentiable symmetries.\marginnote{Lie groups have a differentiable manifold structure. One such example that  we will study in Section~\ref{sec:groups} is the special orthogonal group $\mathrm{SO}(3)$, which is a 3-dimensional manifold.}

Note that here we have defined a group as an abstract object, without saying what the group elements \emph{are} (e.g. transformations of some domain), only how they \emph{compose}.
Hence, very different kinds of objects may have the same symmetry group.
For instance, the aforementioned group of rotational and reflection symmetries of a triangle is the same as the group of permutations of a sequence of three elements (we can permute the corners in the triangle in any way using a rotation and reflection -- see Figure \ref{fig:group-example-d3-s3})\marginnote{The diagram shown in Figure \ref{fig:group-example-d3-s3} (where each node is associated with a group element, and each arrow with a generator), is known as the {\em Cayley diagram}.}.

\paragraph{Group Actions and Group Representations}
Rather than considering groups as abstract entities, 
we are mostly interested in how groups \emph{act} on data.
%
Since we assumed that there is some domain $\Omega$ underlying our data, we will study how the group acts on $\Omega$ 
(e.g. translation of points of the plane), and from there obtain actions of the same group on the space of signals $\mathcal{X}(\Omega)$
(e.g. translations of planar images and feature maps).

A \emph{group action} \marginnote{Technically, what we define here is a {\em left} group action.} of $\fG$ on a set $\Omega$ 
is defined as a mapping $(\fg, u) \mapsto \fg.u$ associating a group element $\fg \in \fG$ and a point $u\in \Omega$ with some other point on $\Omega$ in a way that 
is compatible with the group operations, i.e., $\fg.(\fh.u) = (\fg \fh).u$ for all $\fg, \fh \in \fG$ and $u \in \Omega$. 
We shall see numerous instances of group actions in the following sections. 
%
For example, in the plane the {\em Euclidean group} $\mathrm{E}(2)$ is the group of transformations of $\R^2$ that preserves Euclidean distances\marginnote{Distance-preserving transformations are called {\em isometries}. According to Klein's Erlangen Programme, the classical Euclidean geometry arises from this group.}, and consists of translations, rotations, and reflections.
The same group, however, can also act on the space of \emph{images} on the plane (by translating, rotating and flipping the grid of pixels), as well as on the representation spaces learned by a neural network.
More precisely, if we have a group $\fG$ acting on $\Omega$, we automatically obtain an action of $\fG$ on the space $\mathcal{X}(\Omega)$: 
\begin{equation}
    (\fg . x)(u) = x(\fg^{-1} u).
    \label{eq:group_action}
\end{equation}
Due to the inverse on $\fg$, this is indeed a valid group action, in that we have $(\fg . (\fh . x))(u) = ((\fg \fh) . x)(u)$.  

%

The most important kind of group actions, which we will encounter repeatedly throughout this text, are \emph{linear} group actions, also known as {\em group representations}.
The action on signals in equation~(\ref{eq:group_action}) is indeed linear, in the sense that 
$$
\fg . (\alpha x + \beta x') = \alpha (\fg . x) + \beta (\fg . x') $$
for any scalars $\alpha, \beta$ and signals $x, x' \in \mathcal{X}(\Omega)$. 
We can describe linear actions either as maps $(\fg, x) \mapsto \fg.x$ that are linear in $x$, or equivalently, by currying, 
as a map $\rho : \fG \rightarrow \R^{n \times n}$\marginnote{When $\Omega$ is infinte, the space of signals $\mathcal{X}(\Omega)$ is infinite dimensional, in which case $\rho(\fg)$ is a linear operator on this space, rather than a finite dimensional matrix. In practice, one must always discretise to a finite grid, though.} that assigns to each group element $\fg$ an (invertible) matrix $\rho(\fg)$.
The dimension $n$ of the matrix is in general arbitrary and not necessarily related to the dimensionality of the group or the dimensionality of $\Omega$, but in applications to deep learning $n$ will usually be the dimensionality of the feature space on which the group acts.
For instance, we may have the group of 2D translations acting on a space of images with $n$ pixels.

As with a general group action, the assignment of matrices to group elements should be compatible with the group action.
More specifically, the matrix representing a composite group element $\fg \fh$ should equal the matrix product of the representation of $\fg$ and $\fh$:


\begin{tcolorbox}[width=\linewidth,
                  boxsep=0pt,
                  left=7.5pt,
                  right=7.5pt,
                  top=7.5pt,
                  bottom=7.5pt,
                  arc=0pt,
                  boxrule=0pt,toprule=0pt,
                  colback=boxgray,
                  ]
    A $n$-dimensional real \emph{representation} of a group $\fG$ is a map $\rho : \fG \rightarrow \R^{n \times n}$, assigning to each $\fg \in \fG$ an \emph{invertible} matrix $\rho(\fg)$, and satisfying the condition $\rho(\fg \fh) = \rho(\fg) \rho(\fh)$ for all $\fg, \fh \in \fG$.
    \marginnote{Similarly, a complex representation is a map $\rho : \fG \rightarrow \mathbb{C}^{n \times n}$ satisfying the same equation.}
    A representation is called \emph{unitary} or \emph{orthogonal} if the matrix $\rho(\fg)$ is unitary or orthogonal for all $\fg \in \fG$.
\end{tcolorbox}

Written in the language of group representations, the action of $\fG$ on signals $x \in \mathcal{X}(\Omega)$ is defined as $\rho(\fg) x(u) = x(\fg^{-1} u)$.
We again verify that 
$$
(\rho(\fg) (\rho(\fh) x))(u) = (\rho(\fg \fh) x)(u).
$$
%



\begin{figure}
    \centering
    \includegraphics[width=0.75\linewidth]{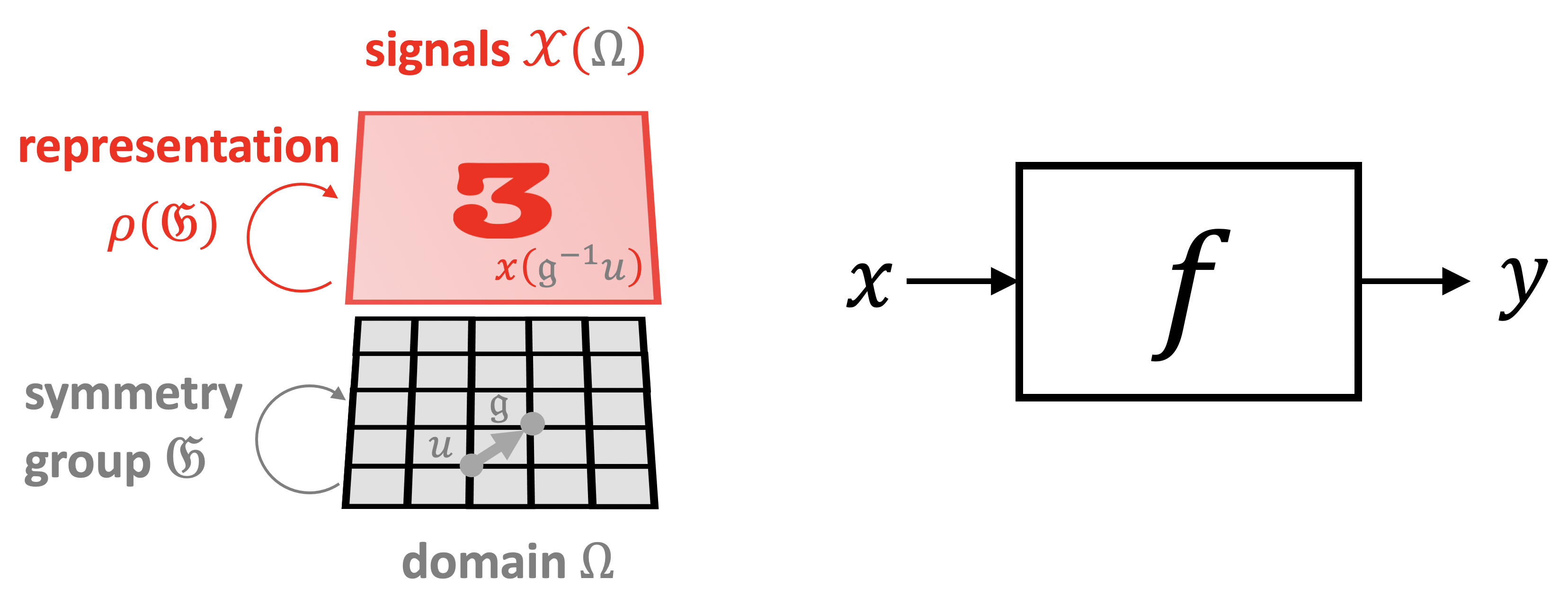}
    \caption{
    Three spaces of interest in Geometric Deep Learning: the (physical) {\em domain} $\Omega$, the space of {\em signals} $\mathcal{X}(\Omega)$, and the {\em hypothesis} class $\mathcal{F}(\mathcal{X}(\Omega))$. 
    Symmetries of the domain $\Omega$ (captured by the group $\fG$) act on signals $x\in \mathcal{X}(\Omega)$ through group representations $\rho(\fg)$, imposing structure on the functions $f\in \mathcal{F}(\mathcal{X}(\Omega))$ acting on such signals.
    }
    \label{fig:symmetryactors}
\end{figure}

\paragraph{Invariant and Equivariant functions}

The symmetry of the domain $\Omega$ underlying the signals $\mathcal{X}(\Omega)$ imposes structure on the function $f$ defined on such signals. 
It turns out to be a powerful inductive bias, improving learning\marginnote{In general, $f$ depends both on the signal an the domain, i.e., $\mathcal{F}(\mathcal{X}(\Omega), \Omega)$. We will often omit the latter dependency for brevity. } efficiency by reducing the space of possible interpolants, $\mathcal{F}(\mathcal{X}(\Omega))$, to those which satisfy the symmetry priors. 
%
%
Two important cases we will be exploring in this text are {\em invariant} and {\em equivariant} functions.


\begin{tcolorbox}[width=\linewidth,
                  boxsep=0pt,
                  left=7.5pt,
                  right=7.5pt,
                  top=7.5pt,
                  bottom=7.5pt,
                  arc=0pt,
                  boxrule=0pt,toprule=0pt,
                  colback=boxgray,
                  ]
    A function $f: \mathcal{X}(\Omega) \rightarrow \mathcal{Y}$ is $\fG$-{\em invariant} if 
    $
    f(\rho(\fg)x) = f(x)$  for all $\fg \in \fG$ and $x \in \mathcal{X}(\Omega)$, i.e., its output is unaffected by the group action on the input. 
\end{tcolorbox}

A classical example of invariance is {\em shift-invariance},\marginnote{Note that signal processing books routinely use the term `shift-invariance' referring to shift-equivariance, e.g. Linear Shift-invariant Systems. } 
arising in computer vision and pattern recognition applications such as image classification. The function $f$ in this case (typically implemented as a Convolutional Neural Network) inputs an image and outputs the probability of the image to contain an object from a certain class (e.g. cat or dog). %
It is often reasonably assumed that the classification result should not be affected by the position of the object in the image, i.e., the function $f$ must be shift-invariant. 
Multi-layer Perceptrons, which can approximate any smooth function, do not have this property -- one of the reasons why early attempts to apply these architectures to problems of pattern recognition in the 1970s failed. 
The development of neural network architectures with local weight sharing, as epitomised by Convolutional Neural Networks, was, among other reasons, motivated by the need for shift-invariant object classification.

If we however take a closer look at the convolutional layers of CNNs, we will find that they are not shift-invariant but {\em shift-equivariant}: in other words, a shift of the input to a convolutional layer produces a shift in the output feature maps by the same amount. 

\begin{tcolorbox}[width=\linewidth,
                  boxsep=0pt,
                  left=7.5pt,
                  right=7.5pt,
                  top=7.5pt,
                  bottom=7.5pt,
                  arc=0pt,
                  boxrule=0pt,toprule=0pt,
                  colback=boxgray,
                  ]
    A function $f: \mathcal{X}(\Omega) \rightarrow \mathcal{X}(\Omega)$ is $\fG$-{\em equivariant} if\marginnote{
    More generally, we might have $f: \mathcal{X}(\Omega) \rightarrow \mathcal{X}(\Omega')$ with input and output spaces having different domains $\Omega, \Omega'$ and representations $\rho$, $\rho'$ of the same group $\fG$.
    In this case, equivariance is defined as $f(\rho(\fg)x) = \rho'(\fg) f(x)$.
    } 
    $f(\rho(\fg)x) = \rho(\fg) f(x)$  for all $\fg \in \fG$, i.e., group action on the input affects the output in the same way. 
\end{tcolorbox}

Resorting again to computer vision, a prototypical application requiring shift-equivariance is image segmentation, where the output of $f$ is a pixel-wise image mask. 
Obviously, the segmentation mask must follow shifts in the input image.  
In this example, the domains of the input and output are the same, but since the input has three color channels while the output has one channel per class, the representations $(\rho, \mathcal{X}(\Omega, \mathcal{C}))$ and $(\rho', \mathcal{X}(\Omega, \mathcal{C}'))$ are somewhat different.

However, even the previous use case of image classification is usually implemented as a sequence of convolutional (shift-equivariant) layers, followed by global pooling (which is shift-invariant). 
As we will see in Section~\ref{sec:gdl_blueprint}, this is a general blueprint of a majority of deep learning architectures, including CNNs and Graph Neural Networks (GNNs). 

\subsection{Isomorphisms and Automorphisms}
\label{sec:isomorphism}

\paragraph{Subgroups and Levels of structure}

As mentioned before, a symmetry\marginnote{Invertible and structure-preserving maps between different objects often go under the generic name of {\em isomorphisms} (Greek for `equal shape'). An isomorphism from an object to itself is called an {\em automorphism}, or symmetry.}
is a transformation that preserves some property or structure, and the set of all such transformations for a given structure forms a symmetry group.
It happens often that there is not one but multiple structures of interest, and so we can consider several {\em levels of structure} on our domain $\Omega$. 
%
Hence, what counts as a symmetry depends on the structure under consideration, but in all cases a symmetry is an invertible map that respects this structure.

On the most basic level, the domain $\Omega$ is a \emph{set}, which has a minimal amount of structure: all we can say is that the set has some \emph{cardinality}\marginnote{For a finite set, the cardinality is the number of elements (`size') of the set, and for infinite sets the cardinality indicates different kinds of infinities, such as the countable infinity of the natural numbers, or the uncountable infinity of the continuum $\R$.}. 
Self-maps that preserve this structure are  \emph{bijections} (invertible maps), which we may consider as set-level symmetries.
%
One can easily verify that this is a group by checking the axioms: a compositions of two bijections is also a bijection (closure), the associativity stems from the associativity of the function composition, the map $\tau(u)=u$ is the identity element, and for every $\tau$ the inverse exists by definition, satisfying $(\tau \circ \tau^{-1})(u) = (\tau^{-1} \circ \tau)(u) =u$.

Depending on the application, there may be further levels of structure.  
For instance, if $\Omega$ is a topological space, we can consider maps that preserve {\em continuity}: such maps are called {\em homeomorphisms} and in addition to simple bijections between sets, are also continuous and have continuous inverse.  
Intuitively, continuous functions are well-behaved and map points in a neighbourhood (open set) around a point $u$ to a neighbourhood around $\tau(u)$.  

%
%
One can further demand that the map and its inverse are (continuously) {\em  differentiable},\marginnote{Every differentiable function is continuous. If the map is continuously differentiable `sufficiently many times', it is said to be {\em smooth}.} i.e., the map and its inverse have a derivative at every point (and the derivative is also continuous). 
This requires further differentiable structure that comes with differentiable manifolds, where such maps are called {\em diffeomorphisms} and denoted by $\mathrm{Diff}(\Omega)$. 
%
%
Additional examples of structures we will encounter include {\em distances} or {\em metrics} (maps preserving them are called {\em isometries}) or {\em orientation} (to the best of our knowledge, orientation-preserving maps do not have a common Greek name).

\begin{tcolorbox}[width=\linewidth,
                  boxsep=0pt,
                  left=7.5pt,
                  right=7.5pt,
                  top=7.5pt,
                  bottom=7.5pt,
                  arc=0pt,
                  boxrule=0pt,toprule=0pt,
                  colback=boxgray,
                  ]
A {\em metric} or {\em distance} is a function $d:\Omega\times\Omega \rightarrow [0,\infty)$ satisfying for all $u,v,w \in \Omega$:  \vspace{3mm}\\
    \noindent {\em 	Identity of indiscernibles:} $d(u,v) =0$ iff $u=v$.\vspace{2mm}\\
    \noindent  {\em Symmetry:} $d(u,v) = d(v,u)$.\vspace{2mm}\\
    \noindent  {\em Triangle inequality:} $d(u,v) \leq  d(u,w) + d(w,v)$.\vspace{2mm}\\

A space equipped with a metric $(\Omega,d)$ is called a {\em metric space}. 

\end{tcolorbox}


The right level of structure to consider depends on the problem.
For example, when segmenting histopathology slide images, we may wish to consider flipped versions of an image as equivalent (as the sample can be flipped when put under the microscope), but if we are trying to classify road signs, we would only want to consider orientation-preserving transformations as symmetries (since reflections could change the meaning of the sign).

As we add levels of structure to be preserved, the symmetry group will get smaller.
Indeed, adding structure is equivalent to selecting a \emph{subgroup}, which is a subset of the larger group that satisfies the axioms of a group by itself:  
 
 \begin{tcolorbox}[width=\linewidth,
                   boxsep=0pt,
                   left=7.5pt,
                   right=7.5pt,
                   top=7.5pt,
                   bottom=7.5pt,
                   arc=0pt,
                   boxrule=0pt,toprule=0pt,
                   colback=boxgray,
                   ]
 Let $(\fG,\circ)$ be a group and $\mathfrak{H} \subseteq \fG$ a subset. $\mathfrak{H}$ is said to be a {\em subgroup} of $\fG$ if $(\mathfrak{H},\circ)$ constitutes a group with the same operation. 
 \end{tcolorbox}
 
For instance, the group of Euclidean isometries $\E{2}$ is a subgroup of the group of planar diffeomorphisms $\diff{2}$, and in turn the group of orientation-preserving isometries $\SE{2}$ is a subgroup of $\E{2}$.
This hierarchy of structure follows the Erlangen Programme philosophy outlined in the Preface: in Klein's construction, the Projective, Affine, and Euclidean geometries have increasingly more invariants and correspond to progressively smaller groups.

\paragraph{Isomorphisms and Automorphisms}

We have described symmetries as structure preserving and invertible maps \emph{from an object to itself}.
Such maps are also known as \emph{automorphisms}, and describe a way in which an object is equivalent it itself.
However, an equally important class of maps are the so-called \emph{isomorphisms}, which exhibit an equivalence between two non-identical objects.
These concepts are often conflated, but distinguishing them is necessary to create  clarity for our following discussion.

To understand the difference, consider a set $\Omega = \{0,1,2\}$.
An automorphism of the set $\Omega$ is a bijection $\tau : \Omega \rightarrow \Omega$ such as a cyclic shift $\tau(u) = u + 1 \mod 3$.
Such a map preserves the cardinality property, and maps $\Omega$ onto itself.
If we have another set $\Omega' = \{a, b, c\}$ with the same number of elements, then a bijection $\eta : \Omega \rightarrow \Omega'$ such as $\eta(0) = a$, $\eta(1) = b$, $\eta(2) = c$ is a {\em set isomorphism}.

As we will see in Section~\ref{sec:proto-graphs}
for graphs, 
the notion of structure includes not just the number of nodes, but also the connectivity.
An isomorphism $\eta: \mathcal{V} \rightarrow \mathcal{V}'$ between two graphs $\mathcal{G}=(\mathcal{V},\mathcal{E})$ and $\mathcal{G}'=(\mathcal{V}',\mathcal{E}')$ is thus a bijection between the nodes that maps pairs of connected nodes to pairs of connected nodes, and likewise for pairs of non-connected nodes.\marginnote{I.e., $(\eta(u),\eta(v)) \in \mathcal{V}'$ iff $(u,v) \in \mathcal{V}$. } Two isomorphic graphs are thus structurally identical, and differ only in the way their nodes are ordered. \marginnote{\includegraphics[width=\linewidth]{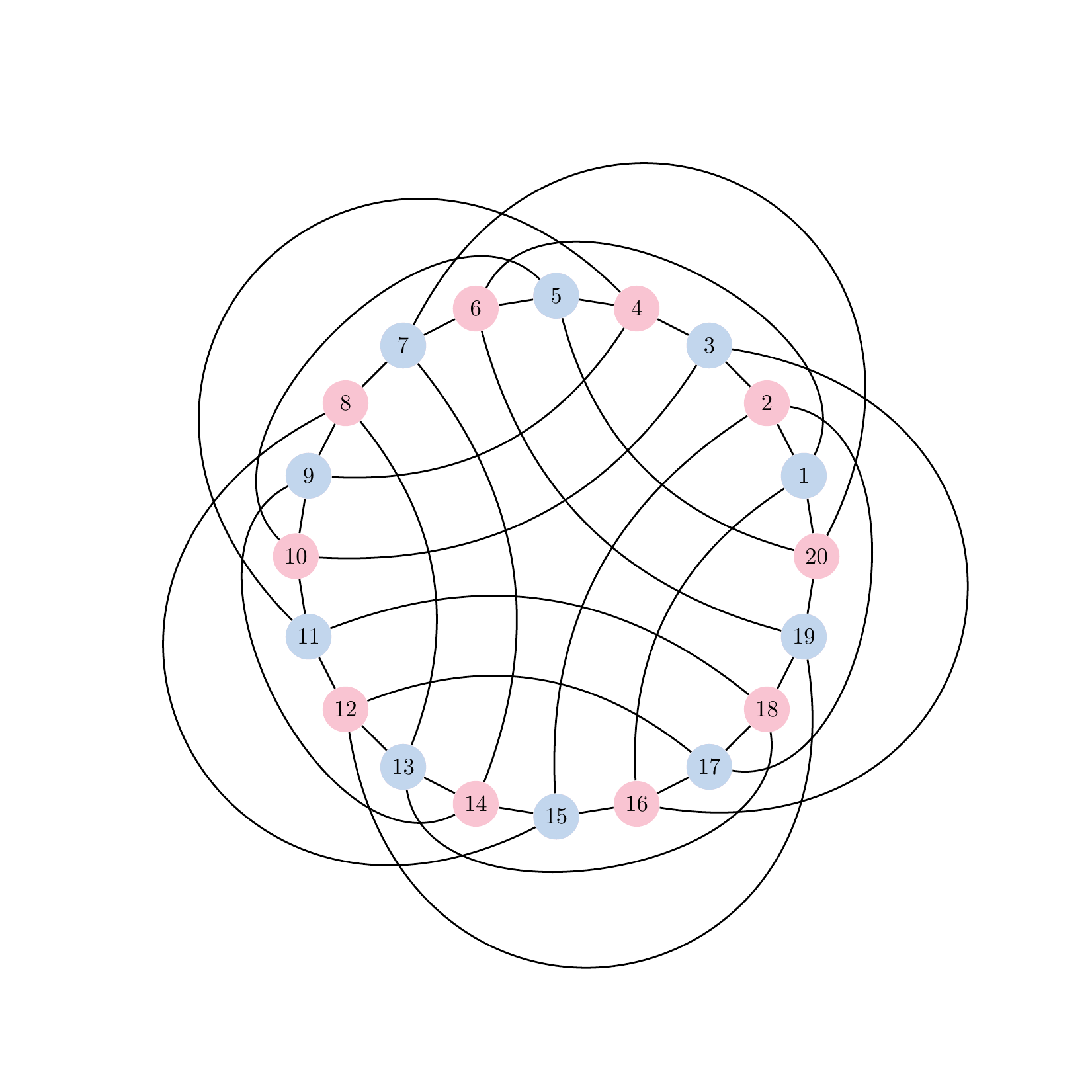}\vspace{-2mm}\\ The \emph{Folkman graph} \citep{folkman1967regular} is a beautiful example of a graph with 3840  automorphisms, exemplified by the many symmetric ways to draw it.}
On the other hand, a graph automorphism or symmetry is a map $\tau : \mathcal{V} \rightarrow \mathcal{V}$ maps the nodes of the graph back to itself, while preserving the connectivity. A graph with a non-trivial automorphism (i.e., $\tau \neq \mathrm{id}$) presents symmetries.



\subsection{Deformation Stability}
\label{sec:geom_stab}
%
%
The symmetry formalism introduced in Sections~\ref{sec:symmetries}--\ref{sec:isomorphism} captures an idealised world where we know exactly which transformations are to be considered as symmetries, and we want to respect these symmetries {\em exactly}. 
For instance in computer vision, we might assume that planar translations are exact symmetries.
However, the real world is noisy and this model falls short in two ways. 

\marginnote{
\includegraphics[width=0.95\linewidth]{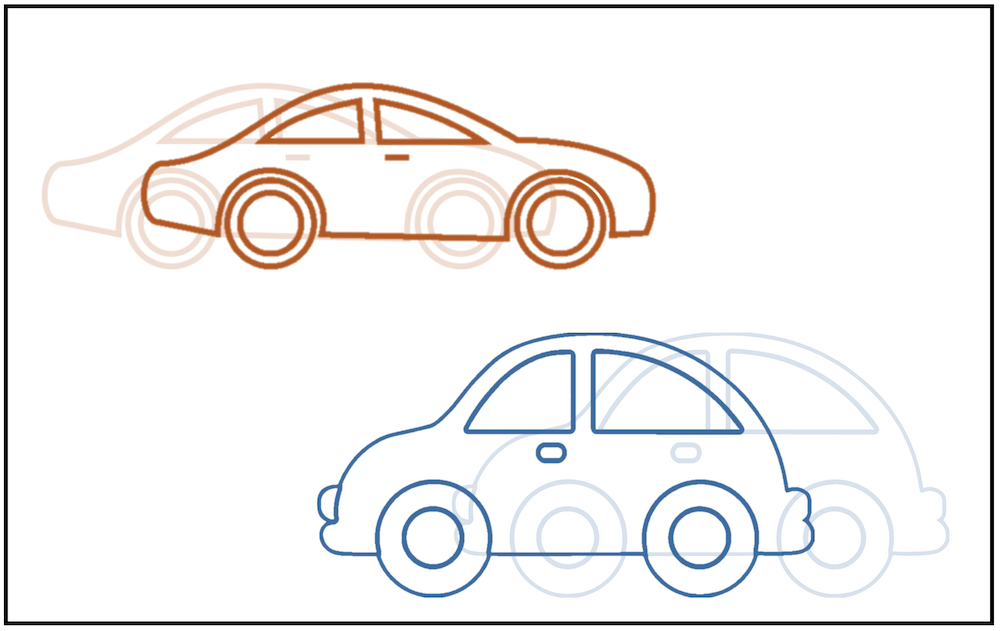}
Two objects moving at different velocities in a video define 
a transformation outside the translation group.
}
Firstly, while these simple groups provide a way to understand \emph{global} symmetries of the domain $\Omega$ (and by extension, of signals on it, $\gX(\Omega)$), they do not capture \emph{local} symmetries well. 
For instance, consider a video scene 
with several objects, each moving along its own different direction.
At subsequent frames, the resulting scene 
will contain approximately the same semantic information, yet no global translation explains the transformation from one frame to another.
In other cases, such as a deformable 3D object viewed by a camera, it is simply very hard to describe the group of transformations that preserve the object identity.
These examples illustrate that in reality we are more interested in a 
far larger set of transformations where global, exact invariance is replaced by a local, inexact one.
In our discussion, we will distinguish between two scenarios: the setting where the domain $\Omega$ is fixed, and signals $x \in \gX(\Omega)$ are undergoing deformations, and the setting where the domain $\Omega$ itself may be deformed.

\paragraph{Stability to signal deformations}
In many applications, we know a priori that a small deformation of the signal $x$ should not change the output of $f(x)$, so it is tempting to consider such deformations as symmetries.
For instance, we could view small diffeomorphisms $\tau \in \diff{\Omega}$, or even small bijections, as symmetries.
However, small deformations can be composed to form large deformations, so ``small deformations'' do not form a group,\marginnote{E.g., the composition of two $\epsilon$-isometries is a $2\epsilon$-isometry, violating the closure property. } and we cannot ask for invariance or equivariance to small deformations only.
Since large deformations can can actually materially change the semantic content of the input, it is not a good idea to use the full group $\diff{\Omega}$ as symmetry group either.

A better approach is to quantify how ``far'' a given $\tau \in \diff{\Omega}$ is from a given symmetry subgroup $\fG \subset \diff{\Omega}$ (e.g. translations)
with a complexity measure $c(\tau)$, so that $c(\tau) = 0$ whenever $\tau \in \fG$. 
We can now replace our previous  definition of exact invariance and equivarance under group actions with a `softer' notion of  
\emph{deformation stability} (or {\em approximate invariance}):
\begin{equation}
\label{eq:defstability1}
\| f(\rho(\tau) x) - f(x)\| \leq C c(\tau) \|x\|,~,~    \forall x\in \gX(\Omega)
\end{equation}
where $\rho(\tau)x(u) = x(\tau^{-1} u)$ as before, and where $C$ is some constant independent of the signal $x$. 
A function $f\in \mathcal{F}(\mathcal{X}(\Omega))$ satisfying the above equation is said to be {\em geometrically stable}. 
We will see examples of such functions in the next Section~\ref{sec:scale_separation}.

Since $c(\tau)=0$ for $\tau \in \fG$, this definition generalises the $\fG$-invariance property defined above. Its utility in applications depends on introducing an appropriate deformation cost. In the case of images defined over a continuous Euclidean plane, 
a popular choice is $c^2(\tau) := \int_\Omega \| \nabla \tau(u)\|^2 \mathrm{d}u$, which measures the `elasticity' of $\tau$, i.e., how different it is from the displacement by a constant vector field. 
This deformation cost is in fact a norm often called the {\em Dirichlet energy}, and can be used to quantify how far $\tau$ is from the translation group.

\begin{figure}[!htbp]
    \centering
    \includegraphics[width=1\textwidth]{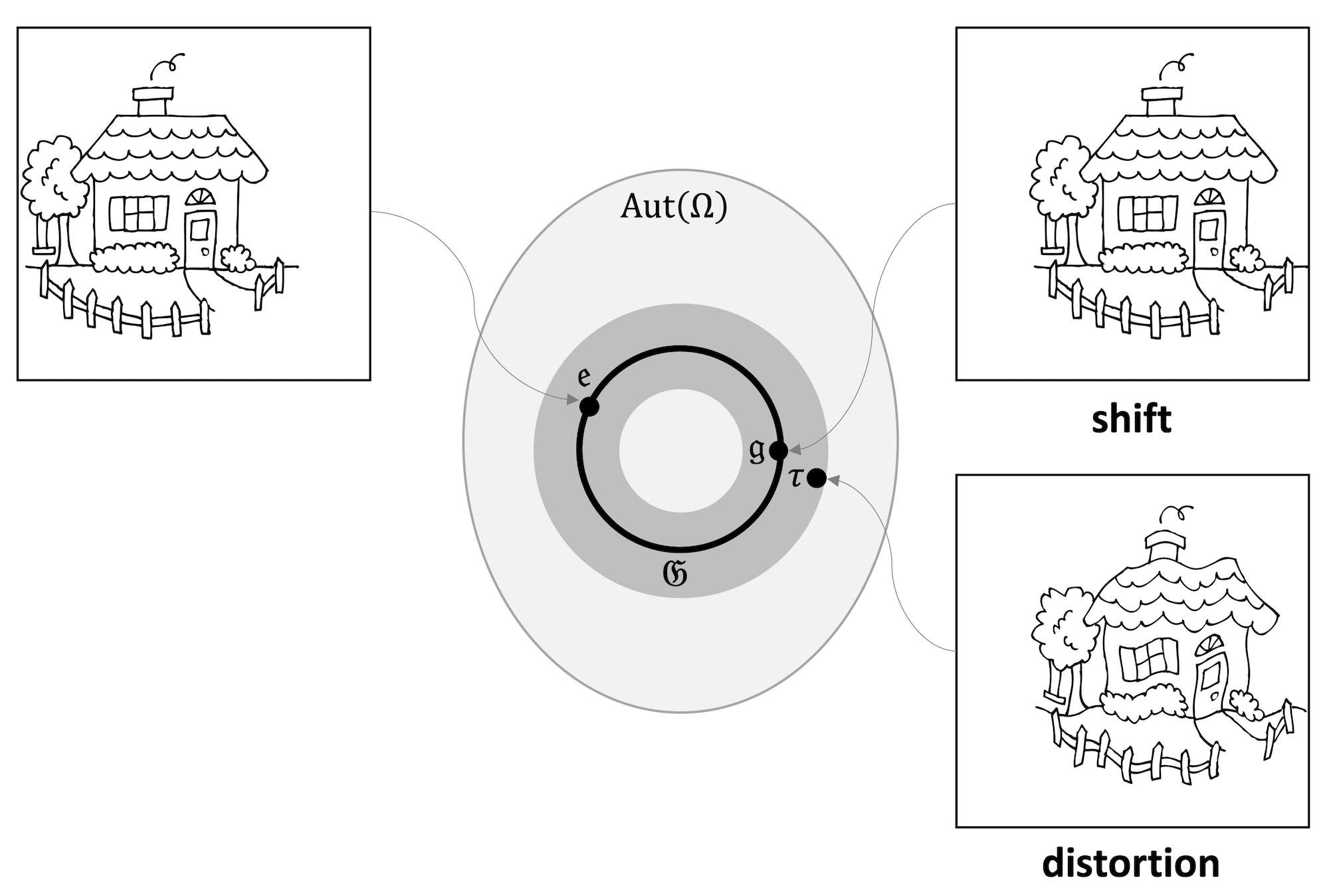}
    \caption{The set of all bijective mappings from $\Omega$ into itself forms the \emph{set automorphism group}  $\mathrm{Aut}(\Omega)$, of which a symmetry group $\fG$ (shown as a circle) is a subgroup. Geometric Stability extends the notion of $\fG$-invariance and equivariance to `transformations around $\fG$' (shown as gray ring), quantified in the sense of some metric between transformations.
    In this example, a smooth distortion of the image is close to a shift. 
    }
    \label{fig:groupdeformation}
\end{figure}


\paragraph{Stability to domain deformations}
In many applications, the object being deformed is not the signal, but the geometric domain $\Omega$ itself. Canonical instances of this are applications dealing with graphs and manifolds: a graph can model a social network at different instance of time containing slightly different social relations (follow graph), or a manifold can model a 3D object undergoing non-rigid deformations. 
This deformation can be quantified as follows. If $\mathcal{D}$ denotes the space of all possible variable domains (such as the space of all graphs, or the space of Riemannian manifolds), one can define for $\Omega, \tilde{\Omega} \in \mathcal{D}$   an appropriate metric (`distance') $d(\Omega, \tilde{\Omega})$ satisfying $d(\Omega,\tilde{\Omega})=0$ if $\Omega$ and $\tilde{\Omega}$ are equivalent in some sense: for example, the graph edit distance vanishes when the graphs are isomorphic, and the Gromov-Hausdorff distance between Riemannian manifolds equipped with geodesic distances vanishes when two manifolds are isometric.\marginnote{The graph edit distance measures the minimal cost of making two graphs isomorphic by a sequences of graph edit operations. The Gromov-Hausdorff distance measures the smallest possible metric distortion of a correspondence between two metric spaces, see \cite{gromov1981structures}. }

A common construction of such distances between domains relies on some family of 
invertible mapping $\eta: \Omega \to \tilde{\Omega}$ that try to `align' the domains in a way that the corresponding structures are best preserved. 
For example, in the case of graphs or Riemannian manifolds (regarded as metric spaces with the geodesic distance), this alignment can compare pair-wise adjacency or distance structures ($d$ and $\tilde{d}$, respectively),  
%
$$d_{\gD}(\Omega, \tilde{\Omega}) =  \inf_{\eta \in \fG}\|d - \tilde{d}\circ (\eta \times \eta)\|$$
%
where $\fG$ is the group of isomorphisms such as bijections or isometries, and the norm is defined over the product space $\Omega \times \Omega$. In other words, a distance between elements of $\Omega,\tilde{\Omega}$ is `lifted' to a distance between the domains themselves, by accounting for all the possible alignments that preserve the internal structure.   
\marginnote{Two graphs can be aligned by the Quadratic Assignment Problem (QAP), which considers in its simplest form two graphs $G,\tilde{G}$ of the same size $n$, and solves $\min_{\mathbf{P} \in \Sigma_n} \mathrm{trace}(\mathbf{A P \tilde{A} P}^\top)$, where $\mathbf{A}, \tilde{\mathbf{A}}$ are the respective adjacency matrices and $\Sigma_n$ is the group of $n \times n$ permutation matrices. The graph edit distance can be associated with such QAP \citep{bougleux2015quadratic}.
}
Given a signal $x \in \gX(\Omega)$ and a deformed domain $\tilde{\Omega}$, one can then consider the deformed signal $\tilde{x} = x \circ \eta^{-1} \in \gX(\tilde{\Omega})$. 


By slightly abusing the notation, we define $\gX(\mathcal{D}) = \{ (\gX(\Omega), \Omega) \, : \, \Omega \in \mathcal{D} \}$ as the ensemble of possible input signals defined over a varying domain. 
A function $f : \gX(\mathcal{D}) \to \gY$ is stable to domain deformations if 
\begin{equation}
\| f( x, \Omega) - f(\tilde{x}, \tilde{\Omega}) \| \leq C \|x \| d_{\gD}(\Omega, \tilde{\Omega})~
\label{eqn:domain_def_stability}
\end{equation}
for all $\Omega, \tilde{\Omega} \in \mathcal{D}$, and  $x \in \mathcal{X}(\Omega)$. 
%
We will discuss this notion of stability in the context of manifolds in Sections~\ref{sec:manifolds}--\ref{sec:meshes}, where isometric deformations play a crucial role.
Furthermore, it can be shown that the stability to domain deformations is a natural generalisation of the stability to signal deformations, by viewing the latter in terms of deformations of the volume form \cite{gama2019diffusion}. 



\subsection{Scale Separation}
\label{sec:scale_separation}

While deformation stability substantially strengthens the global symmetry priors, it is not sufficient in itself to overcome the curse of dimensionality, in the sense that, informally speaking, there are still ``too many" functions that respect (\ref{eq:defstability1}) as the size of the domain grows. A key insight to overcome this curse is to exploit the multiscale structure of physical tasks. Before describing multiscale representations, we need to introduce the main elements of Fourier transforms, which rely on frequency rather than scale. 

\paragraph{Fourier Transform and Global invariants}
Arguably 
\marginnote{\includegraphics[width=\linewidth]{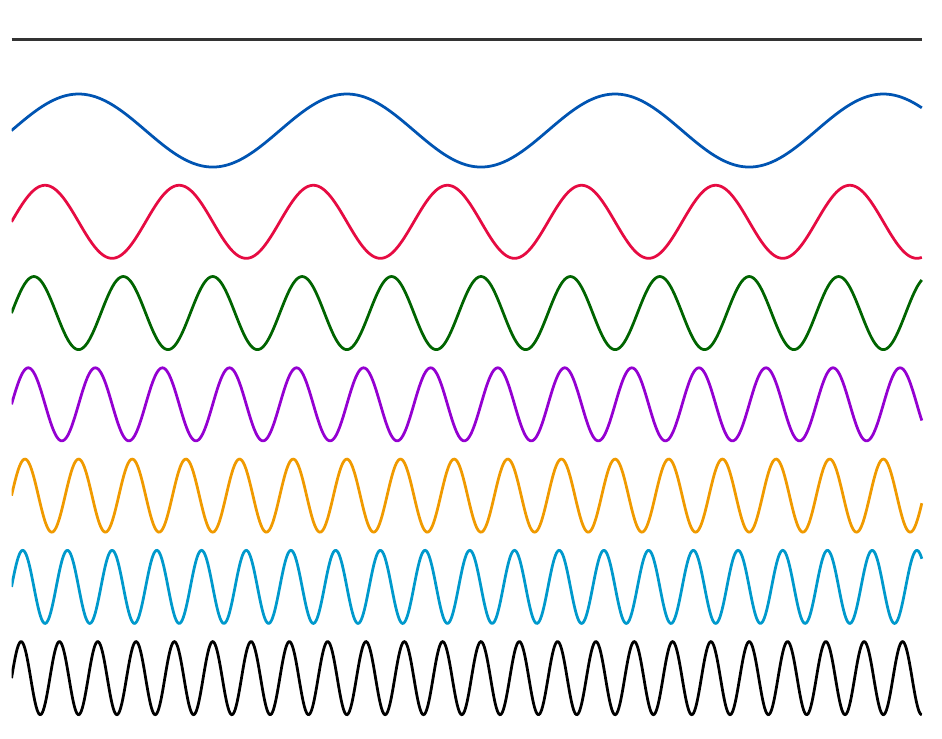}
Fourier basis functions have global support. As a result, local signals produce energy across all frequencies. 
}
the most famous signal decomposition is the {\em Fourier transform}, the cornerstone of harmonic analysis. 
The classical one-dimensional Fourier transform  
$$
\hat{x}(\xi) = \int_{-\infty}^{+\infty} x(u) e^{-\mi \xi u} \mathrm{d}u
$$
expresses the function $x(u) \in L^2(\Omega)$ on the domain $\Omega = \mathbb{R}$ as a linear combination 
of orthogonal oscillating {\em basis functions} $\varphi_\xi(u) = e^{\mi\xi u}$, 
indexed 
by their rate of oscillation (or \emph{frequency}) $\xi$. Such an organisation into frequencies reveals important information about the signal, e.g. its smoothness and localisation. 
%
The Fourier basis itself has a deep geometric foundation and can be interpreted as the natural vibrations of the domain, related to its geometric structure (see e.g. \cite{berger2012panoramic}). 

%


%

The Fourier transform\marginnote{In the following, we will use convolution and {\em (cross-)correlation} 
$$
(x \, \star \,\theta)(u) = \int_{-\infty}^{+\infty} \hspace{-2mm} x(v)\theta(u+v) \mathrm{d}v
$$
interchangeably, as it is common in machine learning: the difference between the two is whether the filter is reflected, and since the filter is typically learnable, the distinction is purely notational.
} plays a crucial role in signal processing as it offers a dual formulation of {\em convolution}, 
$$
(x\star \theta)(u) = \int_{-\infty}^{+\infty} x(v)\theta(u-v) \mathrm{d}v
$$
a standard model of linear signal filtering (here and in the following, $x$ denotes the signal and $\theta$ the filter). 
As we will show in the following, the convolution operator is diagonalised in the Fourier basis, making it possible to express convolution as the product of the respective Fourier transforms, 
$$
\widehat{(x\star \theta)}(\xi) = \hat{x}(\xi) \cdot \hat{\theta}(\xi),
$$
a fact known in signal processing as the Convolution Theorem. 
%

As it turns out, many fundamental differential operators such as the Laplacian are described as convolutions on Euclidean domains. 
Since such differential operators can be defined intrinsically  over very general geometries, this provides a formal procedure to extend Fourier transforms beyond Euclidean domains, including graphs, groups and manifolds. We will discuss this in detail in Section~\ref{sec:manifolds}. 

An essential aspect of Fourier transforms is that they reveal \emph{global} properties of the signal and the domain, such as smoothness or conductance. Such global behavior is convenient in presence of global symmetries of the domain such as translation, but not to study more general diffeomorphisms. This requires a representation that trades off spatial and frequential localisation, as we see next.

\paragraph{Multiscale representations}
The 
notion of local invariance can be articulated by switching from a Fourier frequency-based representation to a {\em scale-based} representation, the cornerstone of multi-scale decomposition methods such as {\em wavelets}.\marginnote{See \cite{mallat1999wavelet} for a comperehensive introduction. } 
%
The essential insight of multi-scale methods is to decompose functions defined over the domain $\Omega$ into elementary functions that are localised \emph{both in space and frequency}.\marginnote{
\includegraphics[width=0.9\linewidth]{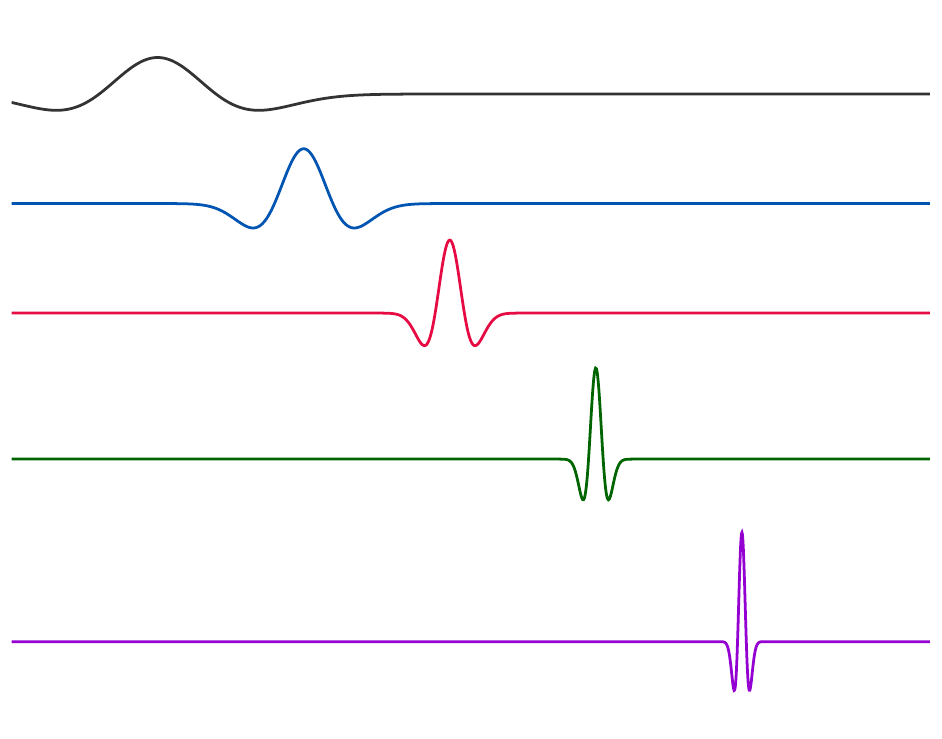}
Contrary to Fourier, wavelet atoms are localised and multi-scale, allowing to capture 
fine details of the signal with atoms having small spatial
support and coarse details with atoms having large
spatial support.  
The term \emph{ atom} here is synonymous with `basis element' in Fourier analysis, with the caveat that wavelets are redundant (over-complete).
}
In the case of wavelets, this is achieved by correlating a translated and dilated filter ({\em mother wavelet}) $\psi$, producing a combined spatio-frequency representation called a  {\em continuous wavelet transform} 
$$
(W_\psi x)(u,\xi) = \xi^{-1/2} \int_{-\infty}^{+\infty} 
\psi\left(\frac{v-u}{\xi}\right) x(v) \mathrm{d}v.
$$
The translated and dilated filters are called {\em wavelet atoms}; 
their spatial position and dilation correspond to the coordinates $u$ and $\xi$ of the wavelet transform. 
These coordinates are usually sampled dyadically ($\xi=2^{-j}$ and $u = 2^{-j}k$), with $j$ referred to as {\em scale}.  
%
%
%
Multi-scale signal representations bring important benefits in terms of capturing regularity properties beyond global smoothness, such as piece-wise smoothness, which made them a popular tool in signal and image processing and numerical analysis in the 90s. 

\paragraph{Deformation stability of Multiscale representations:}
The benefit of multiscale localised wavelet decompositions over Fourier decompositions is revealed when considering the effect of small deformations `nearby' the underlying symmetry group. Let us illustrate this important concept in the Euclidean domain and the translation group.   
Since the Fourier representation diagonalises the shift operator (which can be thought of as convolution, as we will see in more detail in  Section~\ref{sec:grids_euclidean}), it is an efficient representation for translation transformations. However, Fourier decompositions are unstable under high-frequency deformations. 
In contrast, wavelet decompositions offer a stable representation in such cases. 

Indeed, let us consider $\tau \in \mathrm{Aut}(\Omega)$ and its associated linear representation $\rho(\tau)$. When $\tau(u) = u - v$ is a shift, 
as we will verify in Section \ref{sec:grids_euclidean}, the operator $\rho(\tau) = S_v$ is a {\em shift operator} that commutes with convolution. 
Since convolution operators are diagonalised by the Fourier transform, the action of shift in the frequency domain amounts to shifting the complex phase of the Fourier transform,  
%
$$
(\widehat{S_v x})(\xi) = e^{-\mi \xi v} \hat{x}(\xi). 
$$ 
Thus, the {\em Fourier modulus} $f(x) = |\hat{x}|$ removing the complex phase is a simple shift-invariant function, $f(S_v x) = f(x)$. 
%
However, if we have only approximate translation, 
$\tau(u) = u - \tilde{\tau}(u)$ with $\|\nabla \tau \|_\infty = \sup_{u\in \Omega} \| \nabla \tilde{\tau}(u)\| \leq \epsilon$,  
the situation is entirely different: it is possible to show that 
$$
\frac{\|f(\rho(\tau) x) - f(x) \| }{ \|x\| }= \mathcal{O}(1)
$$ 
irrespective of how small $\epsilon$ is (i.e., how close is $\tau$ to being a shift). Consequently, such Fourier representation is {\em unstable under deformations}, however small. This unstability is manifested in general domains and non-rigid transformations; we will see another instance of this unstability in the analysis of 3d shapes using the natural extension of Fourier transforms described in Section \ref{sec:geomanifoldsec}. 

Wavelets offer a remedy to this problem that also reveals the power of multi-scale representations. In the above example, we can show \citep{mallat2012group} that the wavelet decomposition $W_\psi x$ is {\em approximately equivariant} to deformations,
$$
\frac{\| \rho(\tau) (W_\psi x) - W_\psi (\rho(\tau) x) \|}{\|x\|} = \mathcal{O}(\epsilon).
\marginnote{This notation implies that $\rho(\tau)$ acts on the spatial coordinate of $(W_\psi x)(u,\xi)$.
}
$$
%
In other words, decomposing the signal information into scales using localised filters rather than frequencies turns a global unstable representation into a family of locally stable features. Importantly, such measurements at different scales are not yet invariant, and need to be progressively processed towards the low frequencies, hinting at the deep compositional nature of modern neural networks, and captured in our Blueprint for Geometric Deep Learning, presented next.



\paragraph{Scale Separation Prior: }
We can build from this insight by considering a multiscale coarsening of the data domain $\Omega$ into a hierarchy $\Omega_1, \hdots, \Omega_J$. 
As it turns out, such coarsening can be defined on very general domains, including grids, graphs, and manifolds. 
Informally, a coarsening assimilates nearby points $u, u' \in \Omega$ together, and thus only requires an appropriate notion of {\em metric} in the domain. 
If $\gX_{j}(\Omega_j,\mathcal{C}_j) := \{x_j: \Omega_j \to \mathcal{C}_j \}$ 
denotes signals defined over the coarsened domain $\Omega_j$, we informally say that a function $f : \gX(\Omega) \to \gY$ is \emph{locally stable} at scale $j$ if it admits a factorisation of the form $f \approx f_j \circ P_j $, where $P_j : \gX(\Omega) \to \gX_{j}(\Omega_j)$ is a non-linear \emph{coarse graining} and $f_j : \gX_{j}(\Omega_j) \to \gY$. In other words, while the target function $f$ might depend on complex long-range interactions between features over the whole domain, in locally-stable functions it is possible to \emph{separate} the interactions across scales, by first focusing on localised interactions that are then propagated towards the coarse scales. 

\begin{figure}
    \centering
    \includegraphics[width=0.75\textwidth]{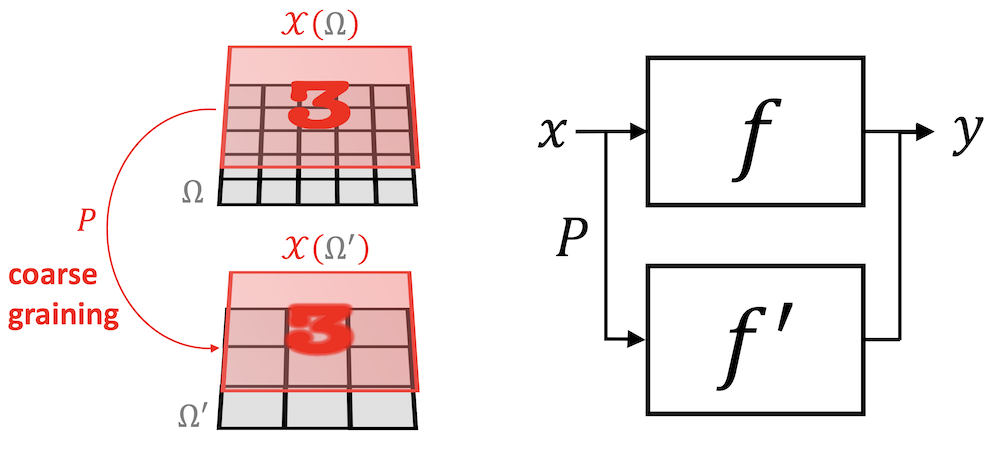}
    \caption{Illustration of Scale Separation for image classification tasks. The classifier $f'$ defined on signals on the coarse grid $\mathcal{X}(\Omega')$ should satisfy $f \approx f'\circ P$, where $P: \mathcal{X}(\Omega) \rightarrow \mathcal{X}(\Omega')$. }
    \label{fig:scale_separation}
\end{figure}

Such principles\marginnote{Fast Multipole Method (FMM) is a numerical technique originally developed to speed up the calculation of long-ranged forces in $n$-body problems. FMM groups sources that lie close together and treats them as a single source.
} are of fundamental importance in many areas of physics and mathematics, as manifested for instance in statistical physics in the so-called renormalisation group, or leveraged in important numerical algorithms such as the Fast Multipole Method. In machine learning, multiscale representations and local invariance are the fundamental mathematical principles underpinning the efficiency of Convolutional Neural Networks and Graph Neural Networks and are typically implemented in the form of {\em local pooling}. In future work, we will further develop tools from computational harmonic analysis that unify these principles across our geometric domains and will shed light onto the statistical learning benefits of scale separation.

\subsection{The Blueprint of Geometric Deep Learning}
\label{sec:gdl_blueprint}

The geometric principles of Symmetry, Geometric Stability, and Scale Separation discussed in Sections~\ref{sec:symmetries}--\ref{sec:scale_separation} can be combined to provide a universal blueprint for learning stable representations of high-dimensional data. 
These representations will be produced by functions $f$ operating on signals $\mathcal{X}(\Omega,\mathcal{C})$ defined on the domain $\Omega$, which is endowed with a symmetry group $\fG$.

The geometric priors we have described so far do not prescribe a specific {\em architecture} for building such representation, but rather a series of necessary conditions. 
However, they hint at an axiomatic construction that provably satisfies these geometric priors, while ensuring a highly expressive representation that can approximate any target function satisfying such priors. 

A simple initial observation is that, in order to obtain a highly expressive representation, we are required to introduce a non-linear element, since
if $f$ is linear and $\fG$-invariant, then for all $x \in \gX(\Omega)$, \marginnote{Here, $\mu(\fg)$ is known as the \emph{Haar measure} of the group $\fG$, and the integral is performed over the entire group.}
$$
f(x) =  \frac{1}{\mu(\fG)} \int_{\fG} f( \fg. x) \mathrm{d}\mu(\fg) = f\left(\frac{1}{\mu(\fG)} \int_{\fG} (\fg.x) \mathrm{d}\mu(\fg) \right),
$$
which indicates that $F$ only depends on $x$ through the  \emph{$\fG$-average} $A{x} = \frac{1}{\mu(\fG)} \int_{\fG} (\fg.x) \mathrm{d}\mu(\fg)$. In the case of images and translation, this would entail using only the average RGB color of the input! 

While this reasoning shows that the family of {\em linear invariants} is not a very rich object, the family of {\em linear equivariants} provides a much more powerful tool, since it enables the construction of rich and stable features by composition with appropriate non-linear maps, as we will now explain. 
Indeed, if $B: \gX(\Omega, \gC) \to \gX( \Omega, \gC')$ is $\fG$-equivariant satisfying $B(\fg.x) = \fg.B(x)$ for all $x \in \gX$ and $\fg \in \fG$, and $\sigma: \gC' \to \gC''$ is an arbitrary (non-linear) map, then we easily verify that the composition $U := (\bm{\sigma} \circ B): \gX(\Omega, \gC) \to \gX( \Omega, \gC'')$ is also $\fG$-equivariant, where $\bm{\sigma}: \gX(\Omega,\gC') \to \gX(\Omega, \gC'')$ is the element-wise instantiation of $\sigma$ given as $(\bm{\sigma}(x))(u) := \sigma( x(u))$.

This simple property allows us to define a very general family of $\fG$-invariants, by composing $U$ with the group averages $A \circ U : \gX(\Omega, \gC) \to \gC''$. A natural question is thus whether any $\fG$-invariant function can be approximated at arbitrary precision by such a model, for appropriate choices of $B$ and $\sigma$. 
It is not hard to adapt the standard Universal Approximation Theorems from unstructured vector inputs to show that shallow `geometric' networks are also universal approximators, by properly generalising the group average to a general non-linear invariant. 
\marginnote{Such proofs have been demonstrated, for example, for the Deep Sets model by \citet{zaheer2017deep}.}
However, as already described in the case of Fourier versus Wavelet invariants, there is a 
fundamental tension between shallow global invariance and deformation stability.
This motivates an alternative representation, which considers instead \emph{localised} equivariant maps.\marginnote{Meaningful metrics can be defined on grids, graphs, manifolds, and groups. A notable exception are sets, where there is no predefined notion of metric. 
} 
Assuming that $\Omega$ is further equipped with a distance metric $d$, we call an equivariant map $U$ localised if $(Ux)(u)$ depends only on the values of $x(v)$ for $\mathcal{N}_u = \{v : d(u,v) \leq r\}$, for some small radius $r$; the latter set $\mathcal{N}_u$ is called the {\em receptive field}. 

A single layer of local equivariant map $U$ cannot approximate functions with long-range interactions, but a composition of several local equivariant maps $U_J \circ U_{J-1} \dots \circ U_1$ increases the receptive field\marginnote{The term `receptive field' originated in the  neuroscience literature, referring to the spatial domain that affects the output of a given neuron.}  while preserving the stability properties of local equivariants. The receptive field is further increased by interleaving downsampling operators that coarsen the domain (again assuming a metric structure), completing the parallel with Multiresolution Analysis (MRA, see e.g. \cite{mallat1999wavelet}). 

\begin{figure}
    \centering
    \includegraphics[width=1\textwidth]{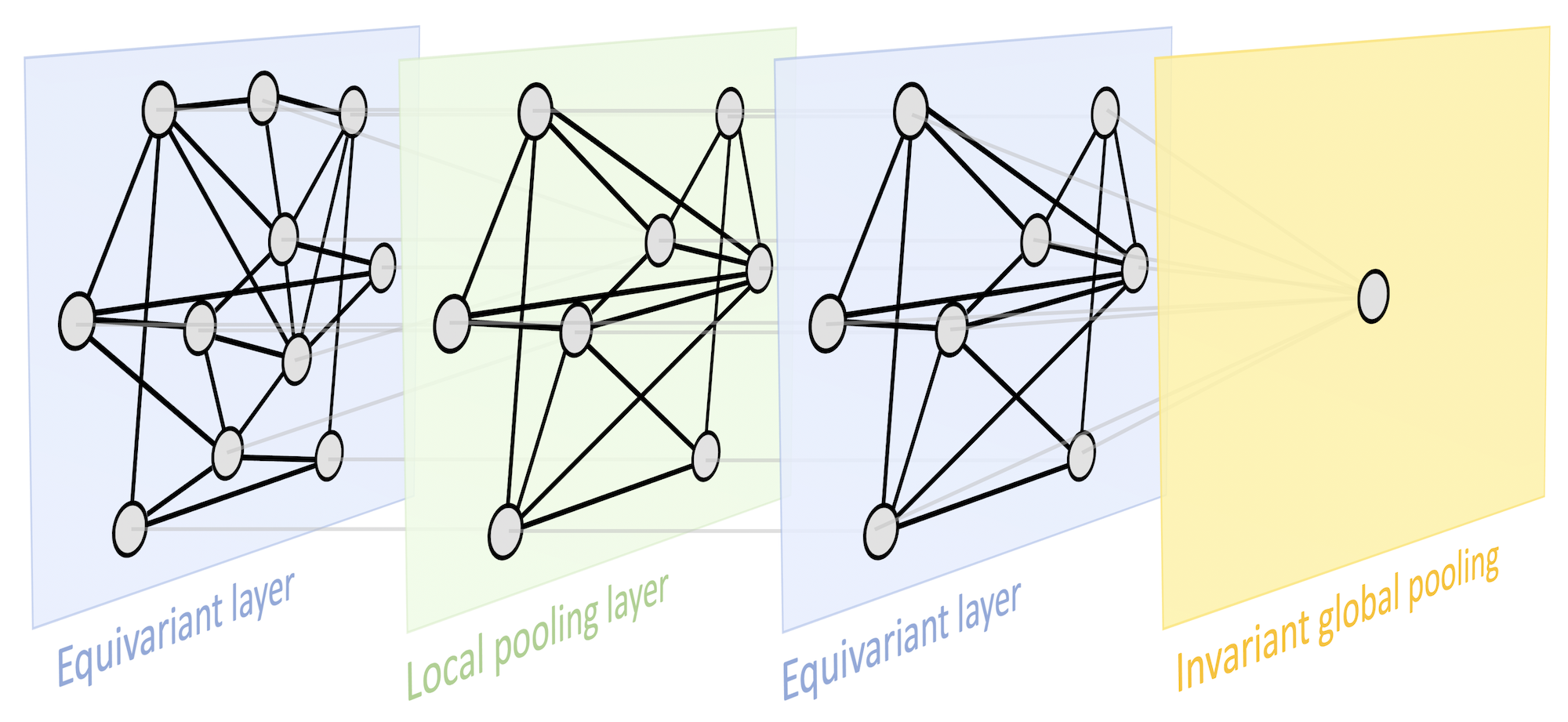}
    \caption{Geometric Deep Learning blueprint, exemplified on a graph. A typical Graph Neural Network architecture may contain permutation equivariant layers (computing node-wise features), local pooling (graph coarsening), and a permutation-invariant global pooling layer (readout layer). }
    \label{fig:blueprint}
\end{figure}

In summary, the geometry of the input domain, with knowledge of an underyling symmetry group, provides three key building blocks: (i) a local equivariant map, (ii) a global invariant map, and (iii) a coarsening operator. These building blocks provide a rich function approximation space with prescribed invariance and stability properties by combining them together in a scheme we refer to as the  
\emph{Geometric Deep Learning Blueprint} (Figure~\ref{fig:blueprint}). 

\begin{tcolorbox}[width=\linewidth,
                  boxsep=0pt,
                  left=7.5pt,
                  right=7.5pt,
                  top=7.5pt,
                  bottom=7.5pt,
                  arc=0pt,
                  boxrule=0pt,toprule=0pt,
                  colback=boxgray,
                  ]
    
\begin{center}
\textbf{Geometric Deep Learning Blueprint}
\end{center}
Let $\Omega$ and $\Omega'$ be domains, $\fG$ a symmetry group over $\Omega$, and write $\Omega' \subseteq \Omega$ if $\Omega'$ can be considered a compact version of $\Omega$. \\

We define the following building blocks:\vspace{2mm}\\

    \noindent {\em Linear $\fG$-equivariant layer} $B: \gX(\Omega, \gC) \to \gX( \Omega', \gC')$ satisfying $B(\fg.x) = \fg.B(x)$ for all $\fg \in \fG$ and $x\in \gX(\Omega,\mathcal{C})$.\vspace{2mm}\\

    \noindent {\em Nonlinearity} $\sigma: \gC \to \gC'$ applied element-wise as $(\bm{\sigma}(x))(u) = \sigma( x(u))$.\vspace{2mm}\\

    \noindent {\em Local pooling (coarsening)}  $P : \gX(\Omega, \gC) \rightarrow \gX(\Omega', \gC) $, such that $\Omega'\subseteq\Omega$.\vspace{2mm}\\
    
    \noindent {\em $\fG$-invariant layer (global pooling) } $A: \gX(\Omega, \gC) \rightarrow \mathcal{Y}$ satisfying $A(\fg .x) = A(x)$ for all $\fg \in \fG$ and $x\in \gX(\Omega,\mathcal{C})$.\vspace{2mm}\\

Using these blocks allows constructing $\fG$-invariant functions $f:\mathcal{X}(\Omega,\mathcal{C}) \rightarrow \mathcal{Y}$ of the form 
$$
f = A \circ \boldsymbol{\sigma}_J \circ B_J \circ P_{J-1} \circ \hdots \circ P_1 \circ \boldsymbol{\sigma}_1 \circ B_1
$$
where the blocks are selected such that the output space of each block matches the input space of the next one. Different blocks may exploit different choices of symmetry groups $\fG$.
\end{tcolorbox}

\paragraph{Different settings of Geometric Deep Learning}
One can make an important distinction between the setting when the domain $\Omega$ is assumed to be {\em fixed} and one is only interested in varying input signals defined on that domain, or the domain is part of the input as {\em varies} together with signals defined on it. 
A classical instance of the former case is encountered in computer vision applications, where images are assumed to be defined on a fixed domain (grid). 
Graph classification is an example of the latter setting, where both the structure of the graph as well as the signal defined on it (e.g. node features) are important. 
In the case of varying domain, geometric stability (in the sense of insensitivity to the deformation of $\Omega$) plays a crucial role in Geometric Deep Learning architectures.

This blueprint  has the right level of generality to be used across a wide range of geometric domains. 
Different Geometric Deep Learning methods thus differ in their choice of the domain, symmetry group, and the specific implementation details of the aforementioned building blocks. 
As we will see in the following, a large class of deep learning architectures currently in use fall into this scheme and can thus be derived from common geometric principles.  
%

In the following sections~(\ref{sec:proto-graphs}--\ref{sec:meshes}) we will describe the various geometric domains focusing on the `5G', and in Sections~\ref{sec:cnnsec}--\ref{sec:lstm} the  specific implementations of Geometric Deep Learning on these domains.

\begin{tcolorbox}[width=\linewidth,
                  boxsep=0pt,
                  left=7.5pt,
                  right=7.5pt,
                  top=7.5pt,
                  bottom=7.5pt,
                  arc=0pt,
                  boxrule=0pt,toprule=0pt,
                  colback=boxgray,
                  ]
    
\begin{center}
\begin{tabular}{lll}
     {\bf Architecture} & {\bf Domain} $\Omega$ & {\bf Symmetry group} $\mathfrak{G}$\vspace{2.5mm}\\
     {\em CNN} & Grid & Translation\vspace{2mm}\\
     {\em Spherical CNN} & Sphere / $\mathrm{SO}({3})$ & Rotation $\mathrm{SO}({3})$\vspace{2mm}\\

     {\em Intrinsic / Mesh CNN} & Manifold & Isometry $\mathrm{Iso}(\Omega)$ / \\
     & & Gauge symmetry $\mathrm{SO}(2)$\vspace{2mm}\\

     {\em GNN} & Graph & Permutation  $\Sigma_n$\vspace{2mm}\\     

     {\em Deep Sets} & Set & Permutation $\Sigma_n$\vspace{2mm}\\     
     {\em Transformer} & Complete Graph & Permutation $\Sigma_n$\vspace{2mm}\\
     {\em LSTM} & 1D Grid & Time warping\\

\end{tabular}
\end{center}
\end{tcolorbox}

\section{Geometric Domains: the 5 Gs}

%

The main focus of our text will be on graphs, grids, groups, geodesics, and  gauges. In this context, by `groups' we mean global symmetry transformations in homogeneous space, by `geodesics' metric structures on manifolds, and by `gauges' local reference frames defined on tangent bundles (and vector bundles in general).  
These notions will be explained in more detail later. 
%
%
In the next sections, we will discuss in detail the main elements in common and the key distinguishing features between these structures and describe the symmetry groups associated with them.
Our exposition is not in the order of generality -- in fact, grids are particular cases of graphs -- but a way to highlight important concepts underlying our Geometric Deep Learning blueprint.

\begin{figure}
    \centering
    \includegraphics[width=1\textwidth]{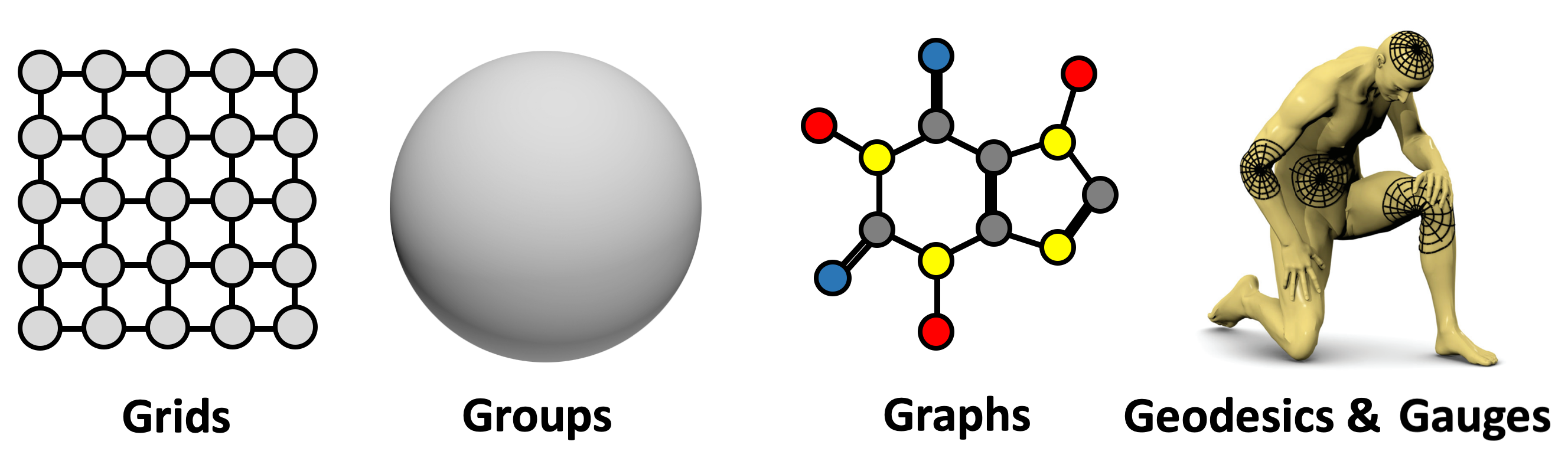}
    \caption{The 5G of Geometric Deep Learning: grids, groups \& homogeneous spaces with global symmetry, graphs, geodesics \& metrics on manifolds, and gauges (frames for tangent or feature spaces). }
    \label{fig:4g}
\end{figure}

\subsection{Graphs and Sets}\label{sec:proto-graphs}

In multiple branches of science, from sociology to particle physics, graphs are used as models of systems of relations and interactions. From our perspective, graphs give rise to a very basic type of invariance modelled by the group of permutations. 
Furthermore, other objects of interest to us, such as grids and sets, can be obtained as a particular case of graphs.  
%


A {\em graph} $\gG = (\gV, \gE)$ is a collection of {\em nodes}\marginnote{Depending on the application field, nodes may also be called \emph{vertices}, and edges are often referred to as \emph{links} or \emph{relations}. We will use these terms interchangeably.} $\gV$  and {\em edges} $\gE \subseteq \gV\times \gV$ between pairs of nodes. For the purpose of the following discussion, we will further assume the nodes to be endowed with $s$-dimensional {\em node features}, 
denoted by $\mathbf{x}_u$ for all $u \in \gV$. 
Social networks are perhaps among the most commonly studied examples of graphs, where nodes represent users, edges correspond to friendship relations between them, and node features model user properties such as age, profile picture, etc. It is also often possible to endow the edges, or entire graphs, with features;\marginnote{
\includegraphics[width=0.8\linewidth]{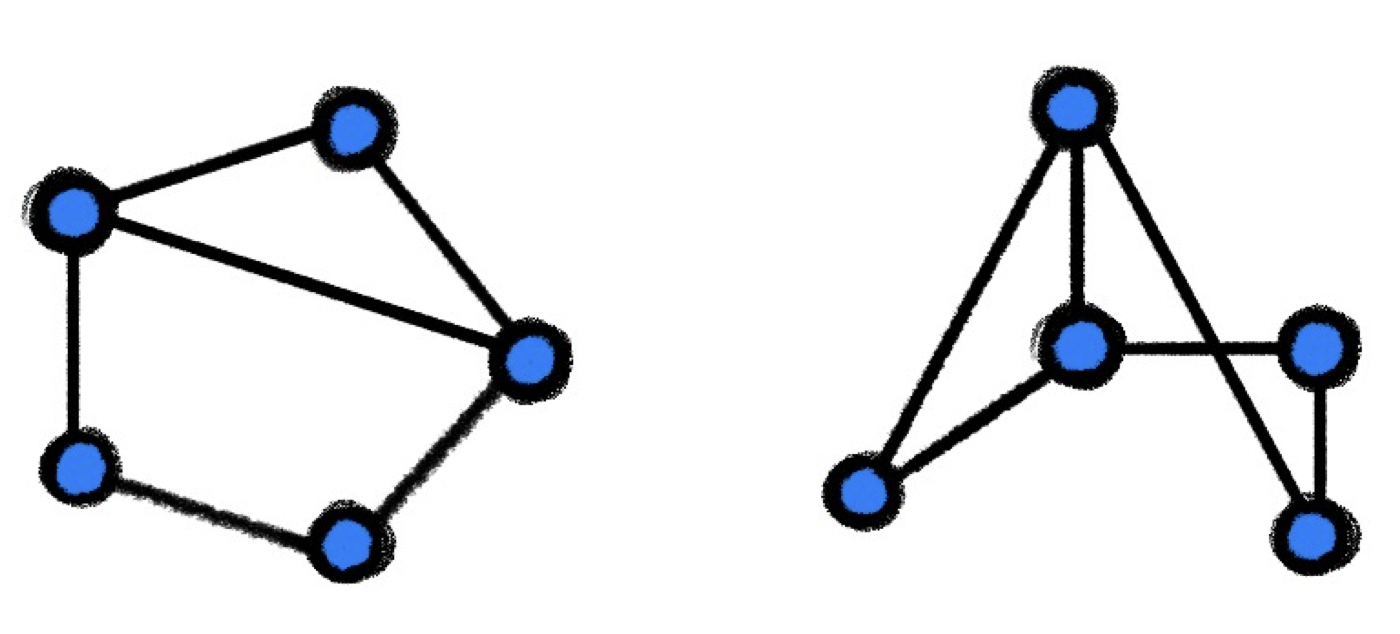}
{\em Isomorphism} is an edge-preserving bijection between two graphs. Two isomorphic graphs shown here are identical up to reordering of their nodes. 
} but as this does not alter the main findings of this section, we will defer discussing it to future work.

The key structural property of graphs is that the nodes in $\gV$ are usually not assumed to be provided in any particular order, and thus any operations performed on graphs should not depend on the ordering of nodes. The desirable property  that functions acting on graphs should satisfy is thus {\em permutation invariance}, and it implies that for any two \emph{isomorphic} graphs,  the outcomes of these functions are identical. 
%
%
We can see this as a 
particular setting of our blueprint, where the domain $\Omega = \mathcal{G}$ and the space $\mathcal{X}(\mathcal{G},\mathbb{R}^d)$ is that of $d$-dimensional node-wise signals. 
The symmetry we consider is given by the \emph{permutation group} $\mathfrak{G} = \Sigma_n$, whose elements are all the possible orderings of the set of node indices $\{1,\hdots, n\}$.

Let us first illustrate the concept of permutation invariance on \emph{sets}, a special case of graphs without edges (i.e., $\gE=\emptyset$). 
By stacking the node features as rows of the $n\times d$ matrix
 $\mathbf{X} = (\mathbf{x}_1, \hdots, \mathbf{x}_n)^\top$, we do effectively specify an ordering of the nodes. 
The action of the permutation $\mathfrak{g}\in\Sigma_n$ on  the set of nodes amounts to the reordering of the rows of $\mathbf{X}$, which can be represented as an $n\times n$ {\em permutation matrix} $\rho(\mathfrak{g}) = \mathbf{P}$,\marginnote{There are exactly $n!$ such permutations, so $\Sigma_n$ is, even for modest $n$, a very large group. } where each row and column contains exactly one $1$ and all the other entries are zeros.

 
 


A function $f$ operating on this set is then said to be \emph{permutation invariant} if, for any such permutation matrix ${\bf P}$, it holds that $f({\bf P}{\bf X}) = f({\bf X})$. One simple such function is 
\begin{equation}
\label{eq:basicinv}
    f({\bf X}) = \phi\left(\sum_{u\in \gV} \psi\left(\mathbf{ x}_u\right)\right)~,
\end{equation}
where the function $\psi$ is independently applied to every node's features, and $\phi$ is applied on its \emph{sum-aggregated} outputs: as sum is independent of the order in which its inputs are provided, such a function is  invariant with respect to the permutation of the node set, and is hence guaranteed to always return the same output, no matter how the nodes are permuted.



Functions like the above provide a `global' graph-wise output, but very often, we will be interested in functions that act `locally', in a node-wise manner. 
For example, we may want to apply some function to \emph{update} the features in every node, obtaining the set of \emph{latent} node features. 
%
If we stack these latent features into a matrix $\mathbf{H} = \mathbf{F}({\bf X})$\marginnote{We use the bold notation  for our function $\mathbf{F}(\mathbf{X})$ to emphasise it outputs node-wise vector features and is hence a matrix-valued function.} is no longer permutation invariant: the order of the rows of ${\bf H}$ should be \emph{tied} to the order of the rows of ${\bf X}$, so that we know which output node feature corresponds to which input node. We need instead  a more fine-grained notion of {\em permutation equivariance}, stating that, once we ``commit'' to a permutation of inputs, it consistently permutes the resulting objects. 
Formally, $\mathbf{F}(\mathbf{X})$ is a \emph{permutation equivariant} function if, for any permutation matrix ${\bf P}$, it holds that $\mathbf{F}({\bf P}{\bf X}) = {\bf P}\mathbf{F}({\bf X})$. A shared node-wise linear transform 
\begin{equation}
    \mathbf{F}_{\mathbf\Theta}({\bf X}) = {\bf X}{\mathbf \Theta}
\end{equation}
specified by a weight matrix $\mathbf\Theta\in\mathbb{R}^{d\times d'}$, is one possible construction of such a permutation equivariant function, producing in our example 
latent features of the form 
$\mathbf{h}_u = \boldsymbol{\Theta}^\top\mathbf{x}_u$.


This construction arises naturally from our Geometric Deep Learning blueprint. 
We can first attempt to characterise {\em linear equivariants} (functions of the form $\mathbf{F} {\bf P X} = \bf{P} \mathbf{FX}$), for which it is easy to verify 
that any such map can be written as a linear combination of two \emph{generators}, the identity $\mathbf{F}_1 \mathbf{X} = {\bf X}$ and the average ${\mathbf{F}_2 \mathbf{X}}= \frac{1}{n}\boldsymbol{1}\boldsymbol{1}^\top \mathbf{X} = \frac{1}{n} \sum_{u=1}^n {\bf x}_u$. As will be described in Section \ref{sec:deepset}, the popular Deep Sets \citep{zaheer2017deep} architecture follows precisely this blueprint.


We can now generalise the notions of permutation invariance and equivariance from sets to graphs. 
In the generic setting $\gE\neq\emptyset$, the graph connectivity can be represented by the $n\times n$ {\em adjacency matrix} $\mathbf{A}$,\marginnote{When the graph is {\em undirected}, i.e. $(u,v) \in \gE$ iff $(v,u) \in \gE$, the adjacency matrix is {\em symmetric}, $\mathbf{A}= \mathbf{A}^\top$. } defined as  
\begin{equation}
    a_{uv} = \begin{cases}
    1 & (u, v)\in \gE\\
    0 & \text{otherwise}.
    \end{cases}
\end{equation}
Note that now the adjacency and feature matrices $\mathbf{A}$ and $\mathbf{X}$ are ``synchronised'', in the sense that $a_{uv}$ specifies the adjacency information between the nodes described by the $u$th and $v$th rows of $\mathbf{X}$. Therefore, applying a permutation matrix $\mathbf{P}$ to the node features $\mathbf{X}$ automatically implies applying it to $\mathbf{A}$'s rows and columns, $\mathbf{P}\mathbf{A}\mathbf{P}^\top$. \marginnote{$\mathbf{P}\mathbf{A}\mathbf{P}^\top$ is the representation of $\Sigma_n$ acting on matrices. }
We say that (a graph-wise function) $f$ is \emph{permutation invariant} if  
\begin{equation}
f({\bf PX}, {\bf PAP}^\top) = f({\bf X}, {\bf A})\marginnote{As a way to emphasise the fact that our functions operating over graphs now need to take into account the adjacency information, we use the notation $f({\bf X}, {\bf A})$.}
\end{equation}
and (a node-wise function) $\mathbf{F}$ is \emph{permutation equivariant} if 
\begin{equation}
\label{eq:permequivgraph}
\mathbf{F}({\bf PX}, {\bf PAP}^\top) = {\bf P}\mathbf{F}({\bf X}, {\bf A})
\end{equation}
for any permutation matrix  ${\bf P}$.
%
%
%
%
%

Here again, we can first characterise linear equivariant functions.\marginnote{This corresponds to the {\em Bell number} $B_4$, which counts the number of ways to partition a set of $4$ elements, in this case given by the 4-indices $(u,v), (u',v')$ indexing a linear map acting on the adjacency matrix.  
}  
As observed by \cite{maron2018invariant}, any linear $\mathbf{F}$ satisfying equation (\ref{eq:permequivgraph}) can be expressed as a linear combination 
of fifteen linear generators; remarkably, this family of generators is {\em independent of} $n$. 
Amongst these generators, our blueprint specifically advocates for those that are also {\em local}, i.e., whereby  
the output on node $u$ directly depends on its neighbouring nodes in the graph. We can formalise this constraint explicitly in our model construction, by defining what it means for a node to be neighbouring another.

A (undirected) {\em neighbourhood} of node $u$, sometimes also called {\em 1-hop},  is defined as \marginnote{
Often, the node $u$ itself is included in its own neighbourhood.}
\begin{equation}
    \mathcal{N}_u = \{ v : (u,v) \in \gE \,\mathrm{or}\, (v,u) \in \gE \}
\end{equation}
and the {\em neighbourhood features} as the multiset 
\begin{equation}
    \mathbf{X}_{\mathcal{N}_u} = \ldblbrace \mathbf{x}_v : v\in\mathcal{N}_u \rdblbrace.
\marginnote{A {\em multiset}, denoted $\ldblbrace \, \dots \, \rdblbrace$, is a set where the same element can appear more than once. This is the case here because the features of different nodes can be equal.}
\end{equation}
Operating on 1-hop neighbourhoods aligns well with the \emph{locality} aspect of our blueprint: namely, defining our metric over graphs as the \emph{shortest path distance} between nodes using edges in $\mathcal{E}$.

\begin{figure}
    \centering
    \includegraphics[width=\linewidth]{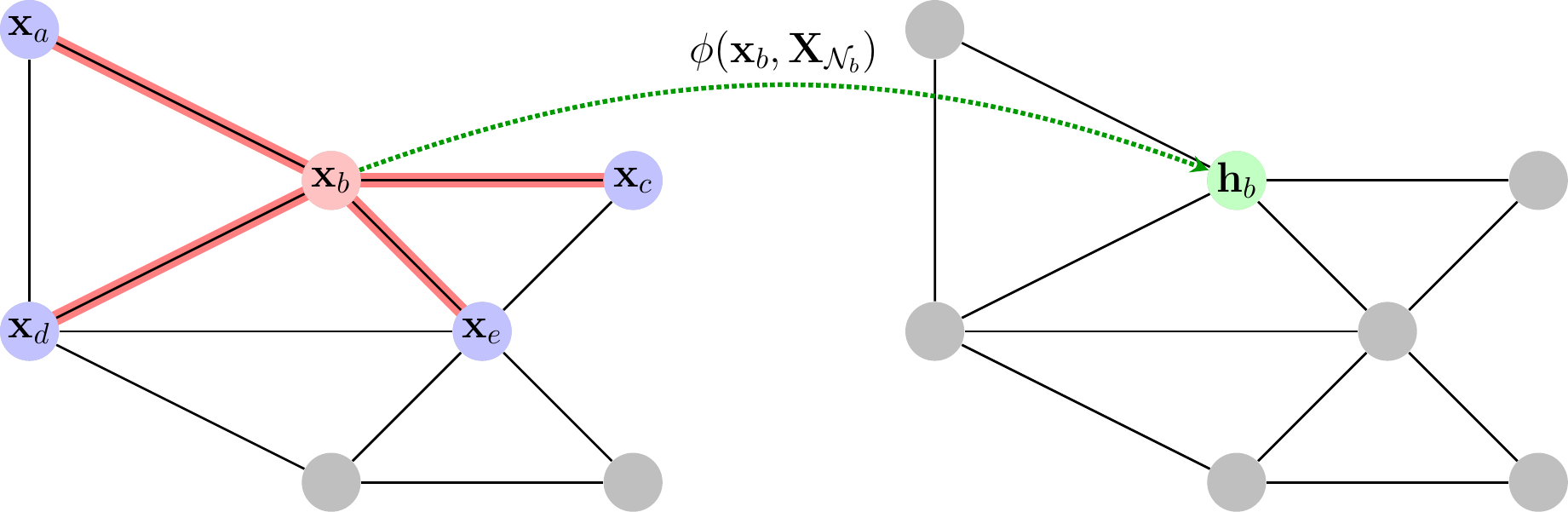}
    \caption{An illustration of constructing permutation-equivariant functions over graphs, by applying a permutation-invariant function $\phi$ to every neighbourhood. In this case, $\phi$ is applied to the features $\mathbf{x}_b$ of node $b$ as well as the multiset  of its neighbourhood features, $\mathbf{X}_{\mathcal{N}_b} = \ldblbrace\mathbf{x}_a, \mathbf{x}_b, \mathbf{x}_c, \mathbf{x}_d, \mathbf{x}_e\rdblbrace$. Applying $\phi$ in this manner to every node's neighbourhood recovers the rows of the resulting matrix of latents features $\mathbf{H}=\mathbf{F}(\mathbf{X}, \mathbf{A})$.}
    \label{fig:gc_gdl}
\end{figure}

%
The GDL blueprint thus yields a general recipe for constructing permutation equivariant functions on graphs, by specifying a \emph{local} function $\phi$ that operates over the features of a node and its neighbourhood, $\phi(\mathbf{x}_u, \mathbf{X}_{\mathcal{N}_u})$. Then, a permutation equivariant function $\mathbf{F}$ can be constructed by applying $\phi$ to every node's neighbourhood in isolation (see Figure \ref{fig:gc_gdl}):
\begin{equation}
    \mathbf{F}({\bf X}, {\bf A}) =
\left[
  \begin{array}{ccc}
    \horzbar & \phi(\mathbf{x}_1, \mathbf{X}_{\mathcal{N}_1}) & \horzbar \\
    \horzbar & \phi(\mathbf{x}_2, \mathbf{X}_{\mathcal{N}_2}) & \horzbar \\
             & \vdots    &          \\
    \horzbar & \phi(\mathbf{x}_n, \mathbf{X}_{\mathcal{N}_n}) & \horzbar
  \end{array}
\right]
\label{eq:graph_equivariant}
\end{equation}
As $\mathbf{F}$ is constructed by applying a shared function $\phi$ to each node locally, its permutation equivariance rests on $\phi$'s output being independent on the ordering of the nodes in $\mathcal{N}_u$. Thus, if $\phi$ is built to be permutation invariant, then this property is satisfied.
As we will see in future work, the choice of $\phi$ plays a crucial role in the expressive power of such a  scheme. When $\phi$ is injective, it is equivalent to one step of the {\em Weisfeiler-Lehman graph isomorphism test}, a classical algorithm in graph theory providing a necessary condition for two graphs to be isomorphic by an iterative color refinement procedure.



%


It is also worth noticing that the difference between functions defined on sets and more general graphs in this example is that in the latter case we need to explicitly account for the structure of the domain. 
As a consequence, graphs stand apart in the sense that the domain becomes {\em part of the input} in machine learning problems, whereas when dealing with sets and grids (both particular cases of graphs) we can specify only the features and assume the domain to be {\em fixed}. 
This distinction will be a recurring motif in our discussion. 
As a result, the notion of geometric stability (invariance to domain deformation) is crucial in most problems of learning on graphs. It straightforwardly follows from our construction that permutation invariant and equivariant functions produce identical outputs on isomorphic (topologically-equivalent) graphs. These results can be generalised to approximately isomorphic graphs, and several results on stability under graph perturbations exist  \citep{levie2018cayleynets}.  We will return to this important point in our discussion on manifolds, which we will use as an vehicle to study such invariance in further detail.

Second, due to their additional structure, graphs and grids, unlike sets, can be coarsened in a non-trivial way\marginnote{More precisely, we cannot define a non-trivial coarsening assuming set structure alone. There exist established approaches that infer topological structure from unordered sets, and those can admit non-trivial coarsening.}, giving rise to a variety of pooling operations. 
%

\subsection{Grids and Euclidean spaces} 
\label{sec:grids_euclidean}



The second type of objects we consider are grids. 
It is fair to say that the impact of deep learning was particularly dramatic in computer vision,  natural language processing, and speech recognition. These applications all share a 
geometric common denominator: an underlying grid structure. 
%
As already mentioned, grids are a particular case of graphs with special adjacency. However, since the order of nodes in a grid is fixed, machine learning models for signals defined on grids are no longer required to account for permutation invariance, and have a stronger geometric prior: translation invariance. 


\paragraph{Circulant matrices and Convolutions}
Let us dwell on this point in more detail. 
Assuming for simplicity periodic boundary conditions, we can think of a one-dimensional grid as a {\em ring graph}\marginnote{
    \includegraphics[width=0.9\linewidth]{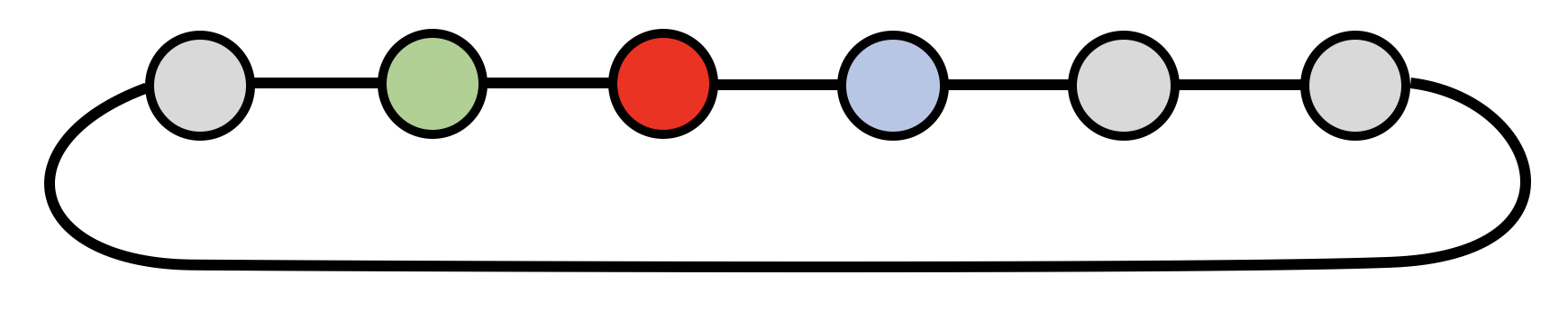}
} with nodes indexed by $0, 1,\hdots, n-1$ modulo $n$ (which we will omit for notation brevity) and
the adjacency matrix with elements $a_{u,u+1 \, \mathrm{mod} \, n} = 1$ and zero otherwise. There are two main differences from the general graph case we have discussed before. 
First, each node $u$ has identical connectivity, to its neighbours $u-1$ and $u+1$, and thus structure-wise indistinguishable from the others. 
\marginnote{As we will see later, this makes the grid a homogeneous space.
}
Second and more importantly, since the nodes of the grid have a fixed ordering, we also have a fixed ordering of the {\em neighbours:} we can call $u-1$ the `left neighbour' and $u+1 $ the `right neighbour'.
If we use our previous recipe for designing a equivariant function $\mathbf{F}$ using a local aggregation function $\phi$, we now have $\mathbf{f}(\mathbf{x}_u) = \phi(\mathbf{x}_{u-1}, \mathbf{x}_{u}, \mathbf{x}_{u+1})$ at every node of the grid: $\phi$ does not need to be permutation invariant anymore.  
For a particular choice of a linear transformation $\phi(\mathbf{x}_{u-1}, \mathbf{x}_{u}, \mathbf{x}_{u+1}) = \theta_{-1}\mathbf{x}_{u-1} + \theta_0 \mathbf{x}_{u} + \theta_1 \mathbf{x}_{u+1}$, we can write $\vec{F}(\mathbf{X})$ as a matrix product, 
$$
\vec{F}(\mathbf{X}) = 
\left[
\begin{array}{ccccc}
\theta_0 & \theta_1  & & & \theta_{-1}\\
 \theta_{-1} & \theta_0 & \theta_1 & & \\
&  \ddots & \ddots & \ddots & \\
 &  & \theta_{-1} & \theta_0 & \theta_1\\
\theta_1 &  &  & \theta_{-1} & \theta_0

 \end{array}
  \right]
\left[  
    \begin{array}{ccc}
    \horzbar & \mathbf{x}_0 & \horzbar \\
    \horzbar & \mathbf{x}_1 & \horzbar \\
             & \vdots    &          \\
    \horzbar & \mathbf{x}_{n-2} & \horzbar  \\           
    \horzbar & \mathbf{x}_{n-1} & \horzbar
  \end{array}
   \right]
$$
Note this very special multi-diagonal structure with one element repeated along each diagonal, sometimes referred to as ``weight sharing'' in the machine learning literature. 
%


More generally, given a vector $\boldsymbol{\theta} = (\theta_0, \hdots, \theta_{n-1})$, a {\em circulant matrix} 
$\mathbf{C}(\boldsymbol{\theta}) = (\theta_{u-v \, \mathrm{mod} \, n})$ is obtained by appending circularly shifted versions of the vector $\boldsymbol{\theta}$.  
Circulant matrices are synonymous with discrete convolutions, \marginnote{Because of the periodic boundary conditions, it is a {\em circular} or {\em cyclic convolution}. In signal processing, $\boldsymbol{\theta}$ is often referred to as the ``filter,'' and in CNNs, its coefficients are learnable. } 
$$
(\mathbf{x} \star \boldsymbol{\theta})_u = \sum_{v=0}^{n-1}  x_{v \, \mathrm{mod}\, n} \,\, \theta_{u-v \, \mathrm{mod}\, n}
$$
as one has $\mathbf{C}(\boldsymbol{\theta})\mathbf{x} = \mathbf{x} \star \boldsymbol{\theta}$. 
%
A particular choice of $\boldsymbol{\theta}=(0,1,0,\hdots, 0)^\top$ yields a special circulant matrix that shifts vectors to the right by one position. This matrix is called the (right) {\em shift} or {\em translation operator}  and denoted by $\mathbf{S}$.\marginnote{The left shift operator is given by $\mathbf{S}^\top$. Obviously, shifting left and then right (or vice versa) does not do anything, which means $\mathbf{S}$ is {\em orthogonal}: $\mathbf{S}^\top \mathbf{S} = \mathbf{S} \mathbf{S}^\top = \mathbf{I}$.}



Circulant matrices can be characterised by their {\em commutativity} property: the product of circulant matrices is commutative, i.e. 
$\mathbf{C}(\boldsymbol{\theta}) \mathbf{C}(\boldsymbol{\eta}) = \mathbf{C}(\boldsymbol{\eta}) \mathbf{C}(\boldsymbol{\theta})$ for any $\boldsymbol{\theta}$ and $\boldsymbol{\eta}$.
Since the shift is a circulant matrix, we get the familiar {\em translation} or {\em shift equivariance} of the convolution operator, 
$$
\mathbf{S} \mathbf{C}(\boldsymbol{\theta}) \mathbf{x} = 
\mathbf{C}(\boldsymbol{\theta}) \mathbf{S} \mathbf{x}.
$$
Such commutativity property should not be surprising, since the underlying symmetry group (the translation group) is Abelian. 
Moreover, the opposite direction appears to be true as well, i.e. 
a matrix is circulant iff it commutes with shift.
This, in turn, allows us to {\em define} convolution as a translation equivariant linear operation, and is a nice illustration of the power of geometric priors and the overall philosophy of Geometric ML: convolution emerges from the first principle of translational symmetry.

Note that unlike the situation on sets and graphs, the number of linearly independent shift-equivariant functions (convolutions) \emph{grows} with the size of the domain (since we have one degree of freedom in each diagonal of a circulant matrix). However, the scale separation prior guarantees filters can be {\em local}, resulting in the same $\Theta(1)$-parameter complexity per layer, as we will verify in Section \ref{sec:cnnsec} when discussing the use of these principles in the implementation of Convolutional Neural Network architectures.


\paragraph{Derivation of the discrete Fourier transform}
We have already mentioned the Fourier transform and its connection to convolution: the fact that the Fourier transform diagonalises the convolution operation is an important property used in signal processing to perform convolution in the frequency domain as an element-wise product of the Fourier transforms. 
However, textbooks usually only state this fact, rarely explaining {\em where} the Fourier transform comes from and what is so {\em special} about the Fourier basis. 
Here we can show it, demonstrating once more how foundational are the basic principles of symmetry.

For this purpose, recall a fact from linear\marginnote{We must additionally assume distinct eigenvalues, otherwise there might be multiple possible diagonalisations. This assumption is satisfied with our choice of $\mathbf{S}$.}  algebra that (diagonalisable) matrices are {\em joinly diagonalisable} iff they mutually commute. 
In other words, there exists a common eigenbasis for all the circulant matrices, in which they differ only by their eigenvalues. 
We can therefore pick one circulant matrix and compute its eigenvectors---we are assured that these will be the eigenvectors of all other circulant matrices as well.
It is convenient to pick the shift operator, for which the eigenvectors happen to be the discrete Fourier basis \marginnote{
$\mathbf{S}$ is orthogonal but non-symmetric, hence, its eigenvectors are orthogonal but the eigenvalues are complex (roots of unity). 
}
$$
\boldsymbol{\varphi}_k = \frac{1}{\sqrt{n}}\left (1, e^{ \frac{2\pi \mi k}{n}}, e^{ \frac{4\pi \mi k}{n} }, \hdots , e^{ \frac{2\pi \mi  (n-1)k}{n} }
\right)^\top, \hspace{5mm} k = 0, 1, \hdots, n-1,  
$$
which we can arrange into an $n\times n$ Fourier matrix $\boldsymbol{\Phi} = (\boldsymbol{\varphi}_0, \hdots, \boldsymbol{\varphi}_{n-1})$.
Multiplication by $\boldsymbol{\Phi}^*$\marginnote{Note that the eigenvectors are complex, so we need to take complex conjugation when transposing $\boldsymbol{\Phi}$.} 
gives the Discrete Fourier Transform (DFT), and by $\boldsymbol{\Phi}$ the inverse DFT, 
$$
\hat{x}_k = \frac{1}{\sqrt{n}}\sum_{u = 0}^{n-1}x_u e^{-\frac{2\pi \mi k u}{n} } \hspace{15mm}
{x}_u = \frac{1}{\sqrt{n}} \sum_{k = 0}^{n-1}\hat{x}_k e^{+\frac{2\pi \mi k u}{n} }.
$$

Since all circulant matrices are jointly diagonalisable,\marginnote{
Since the Fourier transform is an orthogonal matrix ($\boldsymbol{\Phi}^* \boldsymbol{\Phi} = \mathbf{I}$), geometrically it acts as a change of the system of coordinates that amounts to an $n$-dimensional rotation. In this system of coordinates (``Fourier domain''), the action of a circulant $\mathbf{C}$ matrix becomes element-wise product.
}  
they are also diagonalised by the Fourier transform 
and differ only in their eigenvalues. Since the eigenvalues of the circulant matrix $\mathbf{C}(\boldsymbol{\theta})$ are the Fourier transform of the filter (see e.g. \cite{bamieh2018discovering}), $\hat{\boldsymbol{\theta}} = \boldsymbol{\Phi}^* \boldsymbol{\theta}$, we obtain the Convolution Theorem: 
$$
\mathbf{C}(\boldsymbol{\theta}) \mathbf{x} = \boldsymbol{\Phi}
\left[  
    \begin{array}{ccc}
    \hat{\theta}_0 &  &  \\
    & \ddots & \\
    & & \hat{\theta}_{n-1}
  \end{array}
   \right]\boldsymbol{\Phi}^*\mathbf{x}
   = \boldsymbol{\Phi} (\hat{\boldsymbol{\theta}} \odot \hat{\mathbf{x}} )
$$

Because the Fourier matrix $\boldsymbol{\Phi}$ has a special algebraic structure, the products $\boldsymbol{\Phi}^\star \mathbf{x}$ and $\boldsymbol{\Phi} \mathbf{x}$ can be computed with $\mathcal{O}(n \log n)$ complexity using a Fast Fourier Transform (FFT) algorithm. This is one of the reasons why frequency-domain filtering is so popular in signal processing; furthermore, the filter is typically designed directly in the frequency domain, so the Fourier transform $\hat{\boldsymbol{\theta}}$ is never explicitly computed.

Besides the didactic value of the derivation of the Fourier transform
and convolution we have done here, it provides a scheme to generalise these concepts to graphs. Realising that the adjacency matrix of the ring graph is exactly the shift operator, one can can develop the graph Fourier transform and an analogy of the convolution operator by computing the eigenvectors of the adjacency matrix (see e.g. \cite{sandryhaila2013discrete}).
Early attempts to develop graph neural networks by analogy to CNNs, sometimes termed `spectral GNNs', exploited this exact blueprint.\marginnote{In graph signal processing, the eigenvectors of the {\em graph Laplacian} are often used as an alternative of the adjacency matrix to construct the graph Fourier transform, see \cite{shuman2013emerging}. On grids, both matrices have joint eigenvectors, but on graphs they results in somewhat different though related constructions.} 
We will see in 
Sections~\ref{sec:manifolds}--\ref{sec:meshes}
that this analogy has some important limitations. 
The first limitation comes from the fact that a grid is fixed, and hence all signals on it can be represented in the same Fourier basis. In contrast, on general graphs, the Fourier basis depends on the structure of the graph. Hence, we cannot directly compare Fourier transforms on two different graphs --- a problem that translated into a lack of generalisation in machine learning problems. 
Secondly, multi-dimensional grids, which are constructed as tensor products of one-dimensional grids, retain the underlying structure: the Fourier basis elements and the corresponding frequencies (eigenvalues) can be organised in multiple dimensions. In images, for example, we can naturally talk about horizontal and vertical frequency and filters have a notion of {\em direction}. On graphs, the structure of the Fourier domain is one-dimensional, as we can only organise the Fourier basis functions by the magnitude of the corresponding frequencies. As a result, graph filters are oblivious of direction or {\em isotropic}.

\paragraph{Derivation of the continuous Fourier transform}
For the sake of completeness, and as a segway for the next discussion, we repeat our analysis in the continuous setting. 
Like in Section~\ref{sec:scale_separation}, consider functions defined on $\Omega = \mathbb{R}$ and the translation operator $(S_v f)(u) = f(u-v)$ shifting $f$ by some position $v$. 
 Applying $S_v$ to the Fourier basis functions $\varphi_\xi(u) = e^{\mi \xi u}$  yields, by associativity of the exponent,  
$$
S_v e^{\mi\xi u} = e^{\mi\xi (u-v)} = e^{-\mi\xi v} e^{\mi\xi u},
$$
i.e., $\varphi{u}_\xi(u)$ is the complex eigenvector of $S_v$ with the complex eigenvalue  $e^{-\mi\xi v} $ -- exactly mirroring the situation we had in the discrete setting. 
Since $S_v$ is a unitary operator (i.e., $\| S_v x \|_p = \| x \|_p$ for any $p$ and $x \in L_p(\R)$), any eigenvalue $\lambda$ must satisfy $|\lambda|=1$, which corresponds precisely to the eigenvalues $e^{-i\xi v}$ found above. 
Moreover, the spectrum of the translation operator is \emph{simple}, meaning that two functions sharing the same eigenvalue must necessarily be collinear. Indeed, suppose that $S_v f = e^{-\mi \xi_0 v} f$ for some $\xi_0$. Taking the Fourier transform in both sides, we obtain 
$$\forall ~\xi~,~e^{-\mi \xi v} \hat{f}(\xi) = e^{-\mi \xi_0 v} \hat{f}(\xi)~,$$
which implies that $\hat{f}(\xi)=0$ for $\xi \neq \xi_0$, thus $f = \alpha \varphi_{\xi_0}$.

For a general linear operator $C$ that is translation equivariant ($S_v C = C S_v$), we have   
$$
S_v C e^{\mi\xi u} = C S_v e^{\mi\xi u} = e^{-\mi\xi v} C e^{\mi\xi u}, 
$$
implying that $C e^{\mi\xi u}$ is also an eigenfunction\marginnote{{\em Eigenfunction} is synonymous with `eigenvector' and is used when referring to eigenvectors of continuous operators.} of $S_v$ with eigenvalue $e^{-\mi\xi v}$, 
from where it follows from the simplicity of spectrum that $C e^{\mi\xi u} = \beta \varphi_{\xi}(u)$; in other words, the Fourier basis is the eigenbasis of all translation equivariant operators.  
As a result, $C$ is \emph{diagonal} in the Fourier domain and can be  expressed as $C e^{\mi\xi u} = \hat{p}_{C}(\xi) e^{\mi\xi u}$, where $\hat{p}_{C}(\xi)$ is a {\em transfer function} acting on different frequencies $\xi$.  
Finally, for an arbitrary function $x(u)$, by linearity, 
%
\begin{eqnarray*}
(C x) (u) &=& C \int_{-\infty}^{+\infty} \hat{x}(\xi) e^{\mi \xi u} \mathrm{d}\xi = \int_{-\infty}^{+\infty} \hat{x}(\xi) \hat{p}_C(\xi) e^{\mi \xi u} \mathrm{d}\xi \\
&=& \int_{-\infty}^{+\infty} p_C(v) x(u-v) \mathrm{d}v~ = (x \star p_C)(u),\marginnote{The spectral characterisation of the translation group is a particular case of a more general result in Functional Analysis, the \emph{Stone's Theorem}, which derives an equivalent characterisation for any one-parameter unitary group.}
\end{eqnarray*}
where $p_C(u)$ is the inverse Fourier transform of $\hat{p}_C(\xi)$. It thus follows that every linear translation equivariant operator is a convolution. 

\subsection{Groups and Homogeneous spaces}
\label{sec:groups}






Our discussion of grids highlighted how shifts and convolutions are intimately connected: convolutions are linear shift-equivariant\marginnote{Technically, we need the group to be {\em locally compact}, so that there exists a left-invariant Haar measure. Integrating with respect to this measure, we can ``shift'' the integrand by any group element and obtain the same result, just as how we have 
$$
\int_{-\infty}^{+\infty} \hspace{-2mm} x(u) \mathrm{d}u = \int_{-\infty}^{+\infty} \hspace{-2mm} x(u - v) \mathrm{d}u
$$ 
for functions $x : \R \rightarrow \R$.} operations, and vice versa, any shift-equivariant linear operator is a convolution.
Furthermore, shift operators can be jointly diagonalised by the Fourier transform. 
As it turns out, this is part of a far larger story: both convolution and the Fourier transform  can be defined {\em for any group of symmetries} that we can sum or integrate
over.

Consider the Euclidean domain $\Omega = \mathbb{R}$.
We can understand the convolution as a pattern matching operation: we match shifted copies of a filter $\theta(u)$ with an input signal $x(u)$.
The value of the convolution $(x \star \theta)(u)$ at a point $u$ is the inner product of the signal $x$ with the filter \emph{shifted by $u$},
$$
(x \star \theta)(u) =  \langle x, S_u \theta \rangle  = \int_\R x(v)\theta(u + v)  \mathrm{d}v.\marginnote{Note that what we define here is not convolution but {\em cross-correlation}, which is tacitly used in deep learning under the name `convolution'. We do it for consistency with the following discussion, since in our notation $(\rho(\fg)x)(u) = x(u-v)$ and $(\rho(\fg^{-1})x)(u) = x(u+v)$.}
$$
Note that in this case $u$ is both {\em a point on the domain $\Omega = \R$} and also {\em an element of the translation group}, which we can identify with the domain itself, $\fG = \R$.
We will now show how to generalise this construction, by simply replacing the translation group by another group $\fG$ acting on $\Omega$.

\paragraph{Group convolution}
As discussed in Section~\ref{sec:geom_priors}, the action of the group $\fG$ on the domain $\Omega$
induces a representation $\rho$ of $\fG$ on the space of signals $\mathcal{X}(\Omega)$ 
via $\rho(\fg) x(u) = x(\fg^{-1} u)$.
In the above example, $\fG$ is the translation group whose elements act by shifting the coordinates, $u+v$, whereas $\rho(\fg)$ is the shift operator acting on signals as $(S_v x)(u) = x(u-v)$. 
Finally, in order to apply a filter to the signal, we invoke our assumption of $\mathcal{X}(\Omega)$ being a Hilbert space, with an inner product 
\begin{equation*}
    \langle x, \theta \rangle = \int_{\Omega} x(u) \theta(u) \mathrm{d}u, \marginnote{The integration is done w.r.t. an invariant measure $\mu$ on $\Omega$. In case $\mu$ is discrete, this means summing over $\Omega$. } 
\end{equation*}
where we assumed, for the sake of simplicity, scalar-valued signals, $\mathcal{X}(\Omega,\mathbb{R})$; in general the inner product has the form of equation~(\ref{eqn:innerprod}). 

Having thus defined how to transform signals and match them with filters, we can define the {\em group convolution} for signals on $\Omega$, 
\begin{equation}
    \label{eq:group-conv}
    (x \star \theta)(\fg) = \langle x, \rho(\fg) \theta \rangle = \int_\Omega x(u) \theta(\fg^{-1} u) \mathrm{d}u. 
\end{equation}
Note that $x \star \theta$ takes values on the {\em elements} $\fg$ {\em of our group} $\fG$ rather than points on the domain $\Omega$.
Hence, the next layer, which takes $x \star \theta$ as input, should 
act on signals defined 
{\em on to the group} $\fG$, 
a point we will return to shortly.

Just like how the traditional Euclidean convolution is shift-equivariant, 
the more general group convolution is $\fG$-{\em equivariant}. 
The key observation is that matching the signal $x$ with a $\fg$-transformed filter $\rho(\fg) \theta$ is the same as matching the inverse transformed signal $\rho(\fg^{-1}) x$ with the untransformed filter $\theta$.
Mathematically, this can be expressed as $\langle x, \rho(\fg) \theta \rangle = \langle \rho(\fg^{-1}) x, \theta \rangle$.
%
%
With this insight, $\fG$-equivariance of the group convolution~(\ref{eq:group-conv}) follows immediately from its definition and the defining property $\rho(\fh^{-1}) \rho(\fg) = \rho(\fh^{-1} \fg)$ of group representations, 
\begin{equation*}
    (\rho(\fh) x \star \theta)(\fg)
    = \langle \rho(\fh) x, \rho(\fg) \theta \rangle
    = \langle x, \rho(\fh^{-1} \fg) \theta \rangle
    = \rho(\fh) (x \star \theta)(\fg).
\end{equation*}


Let us look at some examples. 
The case of one-dimensional grid we have studied above is obtained with the choice 
$\Omega = \Z_n = \{0, \ldots, n-1\}$ and the cyclic shift group $\fG = \Z_n$. 
The group elements in this case are cyclic shifts of indices, i.e., an element $\fg \in \fG$ can be identified with some $u = 0,\hdots, n-1$ such that $\fg. v = v - u \, \mathrm{mod} \, n$, whereas the inverse element is $\fg^{-1}. v = v + u \, \mathrm{mod} \, n$. Importantly, in this example the elements of the {\em group} (shifts) are also elements of the {\em domain} (indices).   
%
%
We thus can, with some abuse of notation, identify the two structures (i.e., $\Omega = \fG$); our expression for the group convolution in this case 
$$
(x \star \theta)(\fg) =  \sum_{v=0}^{n-1} x_v \, \theta_{\fg^{-1} v},
$$
leads to the familiar convolution \marginnote{Actually here again, this is cross-correlation.} $\displaystyle (x\star \theta)_u = \sum_{v=0}^{n-1} x_v \, \theta_{v + u\,\, \mathrm{mod} \, n}$.

\paragraph{Spherical convolution}
Now consider
\marginnote{
\includegraphics[width=0.9\linewidth]{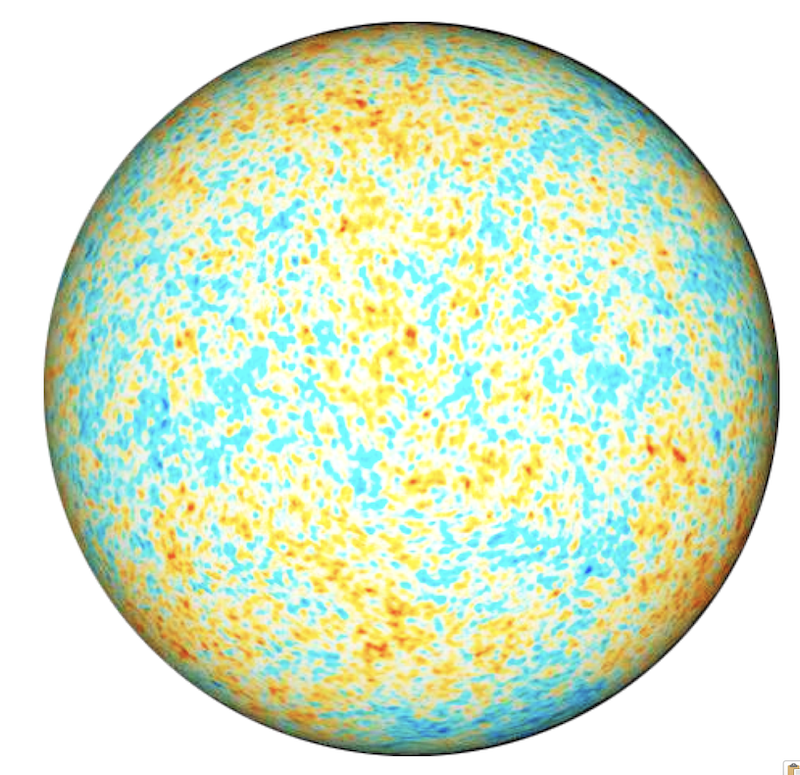}
Cosmic microwave background radiation, captured by the Planck space observatory, is a signal on $\mathbb{S}^2$. 
}
the two-dimensional sphere $\Omega = \mathbb{S}^2$ with the group of rotations, the {\em special orthogonal group} $\fG = \mathrm{SO}(3)$.  
While chosen for pedagogical reason, this example is actually very practical and arises in numerous applications. In astrophysics, for example, observational data often naturally has spherical geometry. Furthermore, spherical symmetries are very important in applications in chemistry when modeling molecules and trying to predict their properties, e.g. for the purpose of virtual drug screening.

Representing a point on the sphere as a three-dimensional unit vector $\mathbf{u} : \|\mathbf{u}\|=1$, the action of the group can be represented as a $3\times 3$ orthogonal matrix $\mathbf{R}$ with $\mathrm{det}(\mathbf{R}) = 1$. The spherical convolution can thus be written as the inner product between the signal and the rotated filter,
$$
(x\star \theta)(\mathbf{R}) = \int_{\mathbb{S}^2} x(\textbf{u}) \theta(\mathbf{R}^{-1}\mathbf{u}) \mathrm{d}\mathbf{u}.
\marginnote{
\includegraphics[width=0.9\linewidth]{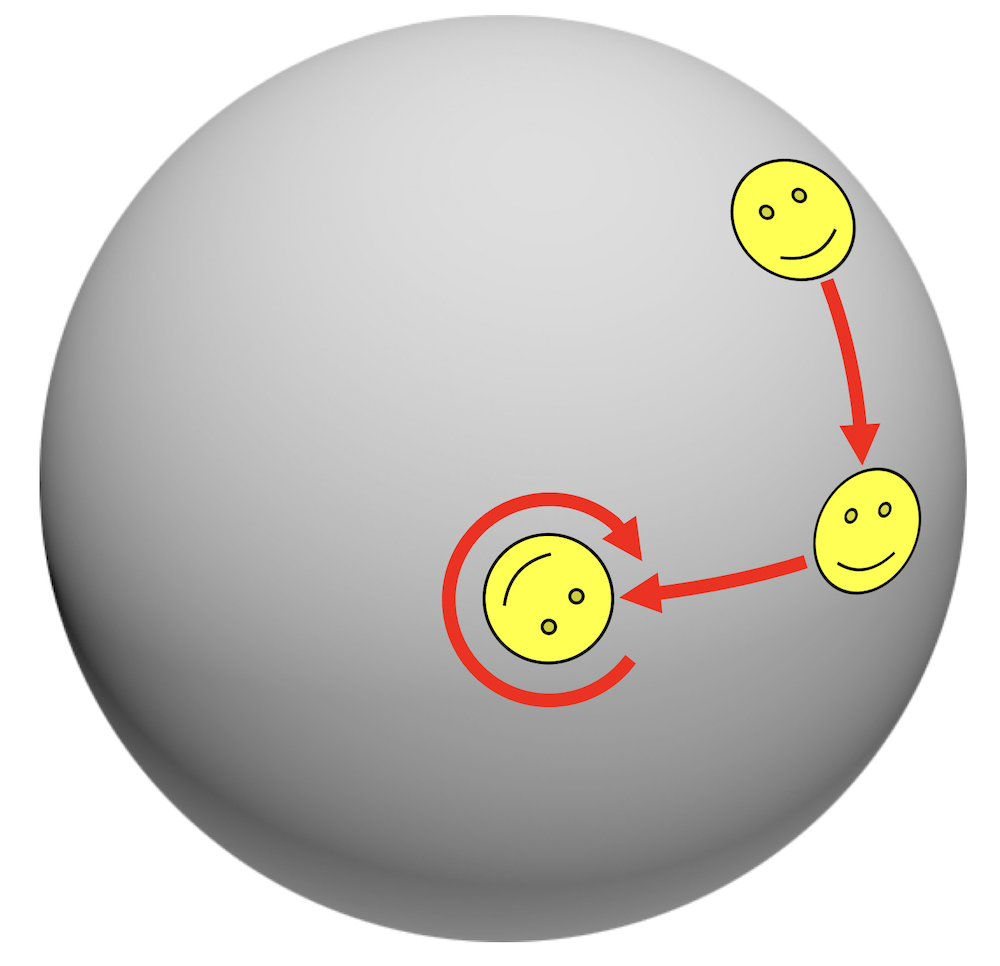}
The action of $\mathrm{SO}(3)$ group on $\mathbb{S}^2$. Note that three types of rotation are possible; the $\mathrm{SO}(3)$ is a three-dimensional manifold.  
}
$$

The first thing to note is than now the group is not identical to the domain: the group $\mathrm{SO}(3)$ is a Lie group that is in fact a three-dimensional manifold, whereas $\mathbb{S}^2$ is a two-dimensional one.  
Consequently, in this case, unlike the previous example, the  convolution is a function {\em on} $\SO{3}$ {\em rather than on} $\Omega$. 


This has important practical consequences: in our Geometric Deep Learning blueprint, we concatenate multiple equivariant maps (``layers'' in deep learning jargon) by applying a subsequent operator to the output of the previous one. 
In the case of translations, we can apply multiple convolutions in sequence, since their outputs are all defined on the same domain $\Omega$. 
In the general setting, since $x \star \theta$ is a function on $\fG$ rather than on $\Omega$, we cannot use exactly the same operation subsequently---it means that the next operation has to deal with {\em signals on $\fG$}, i.e. $x \in \mathcal{X}(\fG)$.  
Our definition of group convolution allows this case: 
we take as domain $\Omega = \fG$ acted on by $\fG$ itself via the group action $(\fg, \fh) \mapsto \fg \fh$ defined by the composition operation of $\fG$.
This yields the representation $\rho(\fg)$ acting on $x \in \mathcal{X}(\fG)$  by $(\rho(\fg) x)(\fh) = x(\fg^{-1}\fh)$\marginnote{The representation of $\fG$ acting on functions defined on $\fG$ itself is called the \emph{regular representation} of $\fG$.}.
Just like before, the inner product is defined by integrating the 
point-wise product of the signal and the filter 
over the domain, which now equals  $\Omega = \fG$.
In our example of spherical convolution, a second layer of convolution would thus have the form
$$
((x\star \theta)\star \phi)(\mathbf{R}) = \int_{\mathrm{SO}(3)} (x\star \theta)(\textbf{Q}) \phi(\mathbf{R}^{-1}\mathbf{Q}) \mathrm{d}\mathbf{Q}.
$$



%

Since convolution involves inner product that in turn requires integrating over the domain $\Omega$, we can only use it on domains $\Omega$ that are small (in the discrete case) or low-dimensional (in the continuous case).  
For instance, we can use convolutions on the plane $\R^2$ (two dimensional) or special orthogonal group $\SE{3}$ (three dimensional), or on the finite set of nodes of a graph ($n$-dimensional), but we cannot in practice perform convolution on the group of permutations $\Sigma_n$, which has $n!$ elements. 
Likewise, integrating over higher-dimensional groups like the affine group (containing  translations, rotations, shearing and scaling, for a total of $6$ dimensions) is not feasible in practice. 
Nevertheless, as we have seen in Section \ref{sec:gnn-intro}, we can still build equivariant convolutions for large groups $\fG$ by working with signals defined on low-dimensional spaces $\Omega$ on which $\fG$ acts.
Indeed, it is possible to show that any equivariant linear map $f : \mathcal{X}(\Omega) \rightarrow \mathcal{X}(\Omega')$ between two domains $\Omega, \Omega'$ can be written as a generalised convolution similar to the group convolution discussed here. 

Second, we note that the Fourier transform we derived in the previous section from the shift-equivariance property of the convolution can also be extended to a more general case by projecting the signal onto the matrix elements of irreducible representations of the symmetry group. We will discuss this in future work. 
In the case of $\SO{3}$ studied here, this gives rise to the {\em spherical harmonics} and {\em Wigner D-functions}, which find wide applications in quantum mechanics and chemistry.

Finally, we point to the assumption that has so far underpinned our discussion in this section: whether $\Omega$ was a grid, plane, or the sphere, we could transform every point into any other point, intuitively meaning that all the points on the domain ``look the same.'' 
A domain $\Omega$ with such property is called a {\em homogeneous space}, where for any $u, v\in \Omega$ there exists $\mathfrak{g}\in\fG$ such that 
$
\mathfrak{g}.u = v\marginnote{The additional properties, $\mathfrak{e}.u = u$ and $\mathfrak{g} (\mathfrak{h}. u) = (\mathfrak{gh}). u$ are tacitly assumed here. }
$.
In the next section we will try to relax this assumption.

\subsection{Geodesics and Manifolds}
\label{sec:geomanifoldsec}

In our last example, the 
sphere $\mathbb{S}^2$ \marginnote{As well as the group of rotations $\mathrm{SO}(3)$, by virtue of it being a {\em Lie group}. }
was a {\em manifold}, albeit a special one 
%
with a global symmetry group due to its homogeneous structure. 
Unfortunately, this is not the case for the majority of manifolds, which typically do not have global symmetries. 
In this case, we cannot straightforwardly define an action of $\fG$ on the space of signals on $\Omega$ and use it to `slide' filters around in order to define a convolution as a direct generalisation of the classical construction. 
%
Nevertheless, manifolds do have two types of invariance 
that we will explore in this section: transformations preserving metric structure and local reference frame change.  

While for many machine learning readers manifolds might appear as somewhat exotic objects, they are in fact very common in various scientific domains. In physics, manifolds play a central role as the model of our Universe --- according to Einstein's General Relativity Theory, gravity arises from the curvature of the space-time, modeled as a pseudo-Riemannian manifold.  
In more `prosaic' fields such as computer graphics and vision, manifolds are a common mathematical model of 3D shapes.\marginnote{The term `3D' is somewhat misleading and refers to the {\em embedding space}. The shapes themselves are 2D manifolds (surfaces).}  %
The broad spectrum of applications of such models ranges from virtual and augmented reality and 
special effects obtained by means of `motion capture' to structural biology dealing with protein interactions that stick together (`bind' in chemical jargon) like pieces of 3D puzzle. 
The common denominator of these applications is the use of a manifold to represent the boundary surface of some 3D object.

There are several reasons why such models are convenient.\marginnote{
\includegraphics[width=0.9\linewidth]{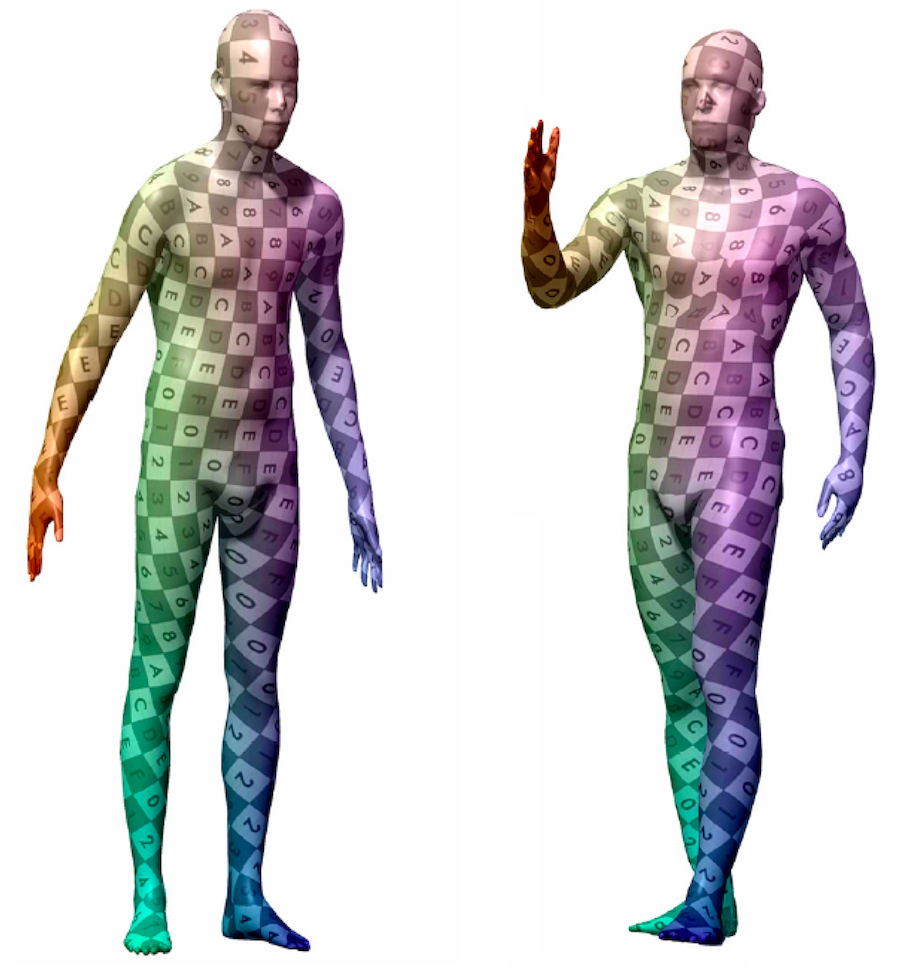}
The human body is an example of a non-rigid object deforming in a  nearly-isometric way. 
} First, they offer a compact description of the 3D object, eliminating the need to allocate memory to `empty space' as is required in grid-based representations. 
Second, they allow to ignore the internal structure of the object. This is a handy property for example in structural biology where the internal folding of a protein molecule is often irrelevant for interactions that happen on the molecular surface. 
Third and most importantly, one often needs to deal with {\em deformable objects} that undergo non-rigid deformations. Our own body is one such example, and many applications in computer graphics and vision, such as the aforementioned motion capture and virtual avatars, require {\em deformation invariance}.
Such deformations can be modelled very well as transformations that preserve the intrinsic structure of a (Riemannian) manifold, namely the distances between points measured \emph{along} the manifold, without regard to the way the manifold is embedded in the ambient space.

We should emphasise that 
manifolds fall under the setting of {\em varying domains} in our Geometric Deep Learning blueprint, and in this sense are similar to graphs. We will highlight the importance of the notion of invariance to domain deformations -- what we called `geometric stability' in Section~\ref{sec:geom_stab}. 
Since differential geometry is perhaps less familiar to the machine learning audience, we will introduce the basic concepts required for our discussion and refer the reader to \cite{penrose2005road} for their detailed exposition.

\paragraph{Riemannian manifolds}
\label{sec:manifolds}

\begin{figure}[h!]
    \centering
    \includegraphics[width=\linewidth]{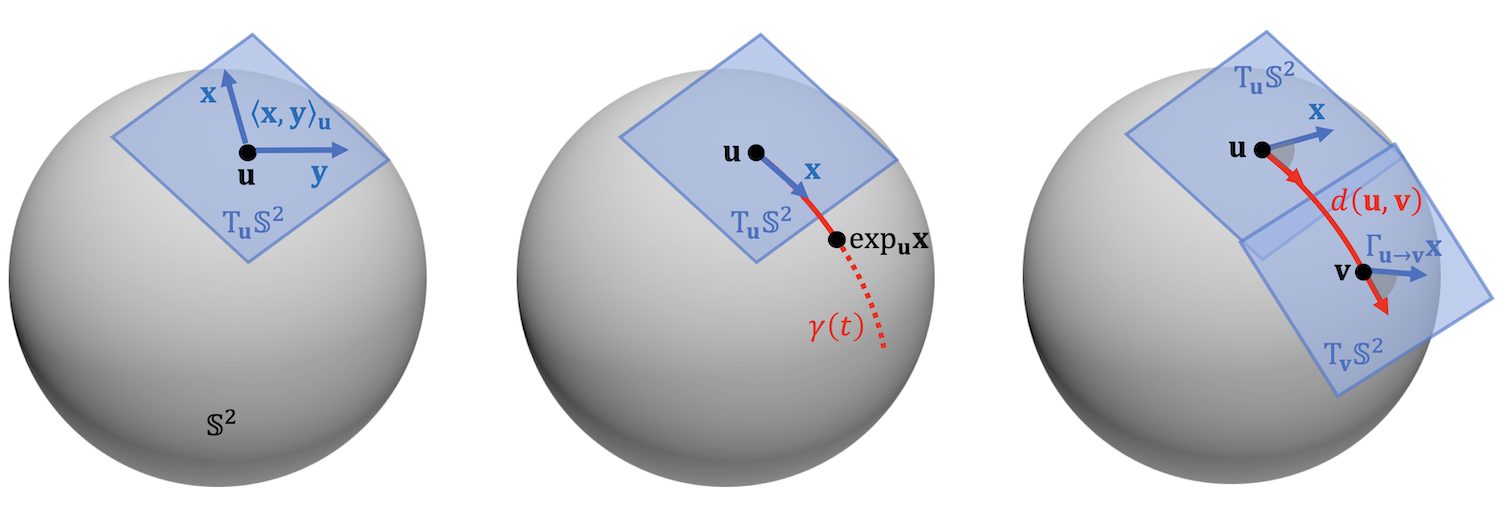}
    \caption{
    Basic notions of Riemannian geometry illustrated on the example of the two-dimensional sphere $\mathbb{S}^2 = \{\mathbf{u}\in \mathbb{R}^3 : \|\mathbf{u}\|=1 \}$, realised a subset (sub-manifold) of $\mathbb{R}^3$. 
The tangent space to the sphere is given as $T_\mathbf{u}\mathbb{S}^2 = \{ \mathbf{x}\in \mathbb{R}^3 : \mathbf{x}^\top \mathbf{u} = 0\}$ and is a 2D plane -- hence this is a 2-dimensional manifold. The Riemannian metric is simply the Euclidean inner product restricted to the tangent plane, $\langle \mathbf{x}, \mathbf{y}\rangle_\mathbf{u} = \mathbf{x}^\top \mathbf{y}$ for any $\mathbf{x},\mathbf{x}\in T_\mathbf{u}\mathbb{S}^2$. 
The exponential map is given by 
$\exp_\mathbf{u}(\mathbf{x}) = \cos(\| \mathbf{x}\|)\mathbf{u} + \frac{\sin(\| \mathbf{x}\|)}{\|\mathbf{x}\|} \mathbf{x}$, for $\mathbf{x} \in T_\mathbf{u}\mathbb{S}^2$. 
Geodesics are great arcs of length $d(\mathbf{u},\mathbf{v}) = \cos^{-1}(\mathbf{u}^\top \mathbf{v})$.
    }
    \label{fig:sphere_metric}
\end{figure}

Since the formal definition of a manifold\marginnote{By `smooth' we mean differentiable suffient number of times, which is tacitly assumed for convenience. `Deformed' here means \emph{diffeomorphic}, i.e., we can map between the two neighbourhoods using a smooth and invertible map with smooth inverse. 
} is somewhat involved, we 
prefer to provide an intuitive picture at the expense of some precision. 
In this context, we can think of a (differentiable or smooth) manifold as a smooth multidimensional curved surface 
that is \emph{locally Euclidean}, in the sense that any small neighbourhood around any point it can be deformed
to a neighbourhood of $\R^s$; 
%
in this case the manifold is said to be $s$-{\em dimensional}. 
%
%
%
This allows us to 
locally approximate the manifold around point $u$ through the \emph{tangent space} $T_u\Omega$.  
The latter can be visualised by thinking of a prototypical two-dimensional manifold, the sphere, and attaching a plane to it at a point: with sufficient zoom, the spherical surface will seem planar (Figure~\ref{fig:sphere_metric}). 
%
\marginnote{Formally, the tangent bundle is the {\em disjoint union} $\displaystyle T\Omega = \bigsqcup_{u\in \Omega} T_u\Omega$.} The collection of all tangent spaces is called the {\em tangent bundle}, denoted $T\Omega$; we will dwell on the concept of bundles in more detail in Section~\ref{sec:gauges}. 

A {\em tangent vector}, which we denote by $X\in T_u\Omega$, can be thought of as a local displacement from point $u$. 
In order to measure the \emph{lengths} of tangent vectors and  \emph{angles} between them, 
\marginnote{A bilinear function $g$ is said to be {\em positive-definite} if $g(X,X)>0$ for any non-zero vector $X\neq 0$. If $g$ is expressed as a matrix $\mathbf{G}$, it means $\mathbf{G} \succ 0$. The determinant $|\mathbf{G}|^{1/2}$ provides a local volume element, which does not depend on the choice of the basis. } 
we need to equip the tangent space with additional structure, expressed as a 
positive-definite bilinear function
$g_u: T_u\Omega \times T_u\Omega \rightarrow \mathbb{R}$ depending smoothly on $u$. Such a function is 
called a {\em Riemannian metric}, in honour of Bernhardt Riemann who introduced the concept in 1856, and can be thought of as an inner product on the tangent space, $\langle X, Y \rangle_u = g_u(X,Y)$, which is an expression of the angle between any two tangent vectors $X, Y \in T_u\Omega$. 
The metric also induces a norm $\| X \|_u = g_u^{1/2}(X,X)$ allowing to locally measure lengths of vectors.

We must stress that tangent vectors 
are abstract geometric entities that exists in their own right and are {\em coordinate-free}. If we are to express a tangent vector $X$ numerically as an array of numbers, we can only represent it as a list of coordinates $\mathbf{x}=(x_1, \hdots, x_s)$ {\em relative to some local basis}\marginnote{Unfortunately, too often vectors are identified with their coordinates. 
To emphasise this important difference, we use $X$ to denote a tangent vector and $\mathbf{x}$ to denote its coordinates. } $\{ X_1, \hdots X_s \} \subseteq T_u\Omega$.   
Similarly, the metric can be expressed as an $s\times s$ matrix $\mathbf{G}$ with elements $g_{ij} = g_u(X_i,X_j)$ in that basis. 
We will return to this point in Section~\ref{sec:gauges}. 

%


%

%
%

A manifold equipped with a metric is called a
{\em Riemannian manifold} and properties that can be expressed entirely in terms of the metric are said to be {\em intrinsic}. %
This is a crucial notion for our discussion, as according to our template, we will be seeking to construct functions acting on signals defined on $\Omega$ that are invariant to metric-preserving transformations called {\em isometries} that deform the manifold without affecting its local structure.\marginnote{
\includegraphics[width=1\linewidth]{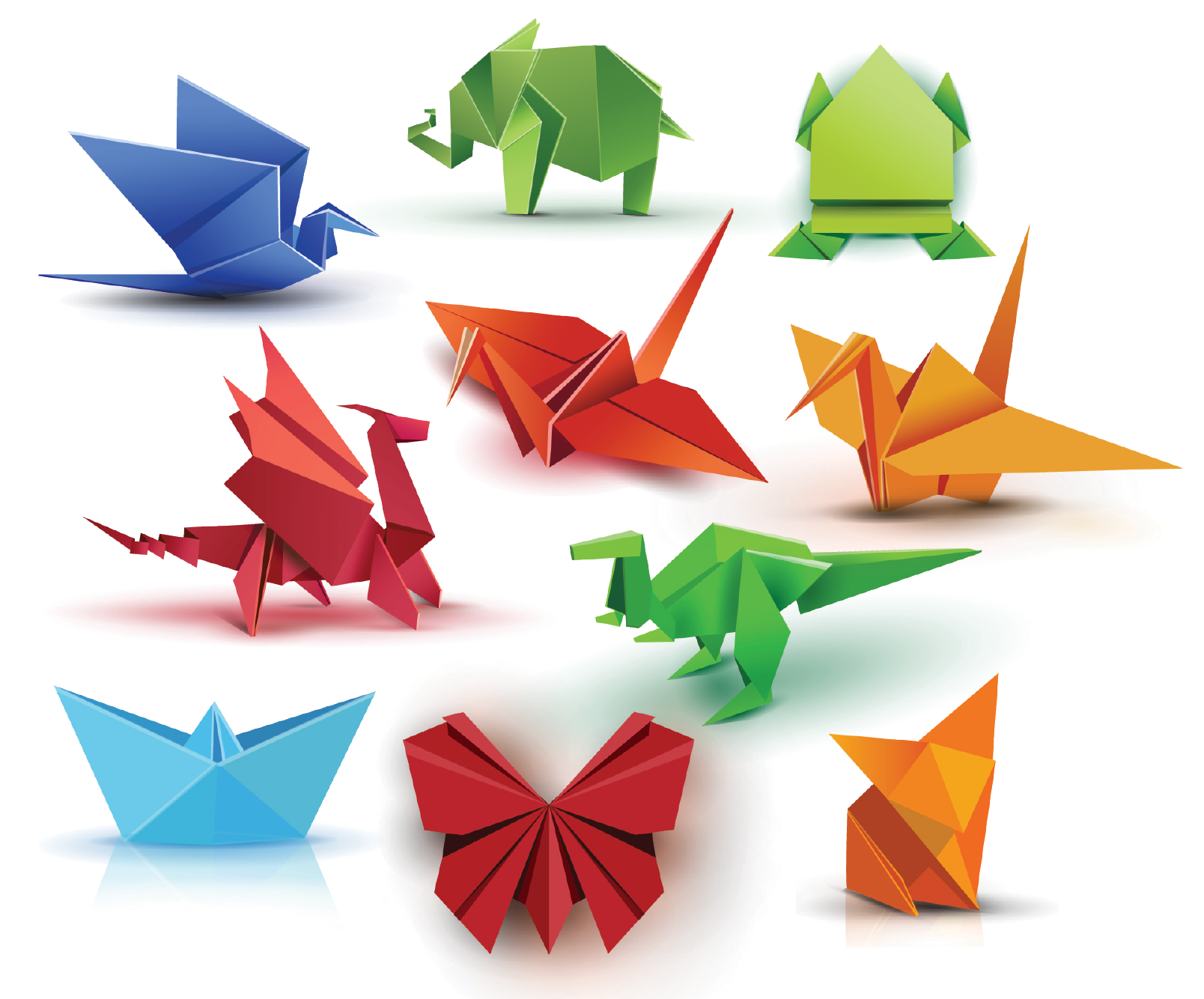}
This result is known as the {\em Embedding Theorem}, due to 
\cite{nash1971imbedding}.
The 
art of origami is a manifestation of different isometric embeddings of the planar surface in $\mathbb{R}^3$ (Figure: Shutterstock/300 librarians). 
}  If such functions can be expressed in terms of intrinsic quantities, they are automatically guaranteed to be isometry-invariant and thus unaffected by isometric deformations. 
These results can be further extended to dealing with approximate isometries; this is thus an instance of the geometric stability (domain deformation) discussed in our blueprint.



While, as we noted, the definition of a Riemannian manifold does not require a
geometric realisation in any space, it turns out that any smooth Riemannian manifold can be realised as a subset of a Euclidean space of sufficiently high dimension (in which case it is said to be `embedded' in that space) by using the structure of the Euclidean space to induce a Riemannian
metric. Such an embedding is however not necessarily unique -- as we will see, two different isometric 
realisations of a Riemannian metric are possible. 


\paragraph{Scalar and Vector fields}
Since we are interested in signals defined on $\Omega$, we need to provide the proper notion of scalar-
and vector-valued functions on manifolds. 
A (smooth) {\em scalar field} is a function of the form $x: \Omega\rightarrow \mathbb{R}$. \marginnote{\includegraphics[width=0.9\linewidth]{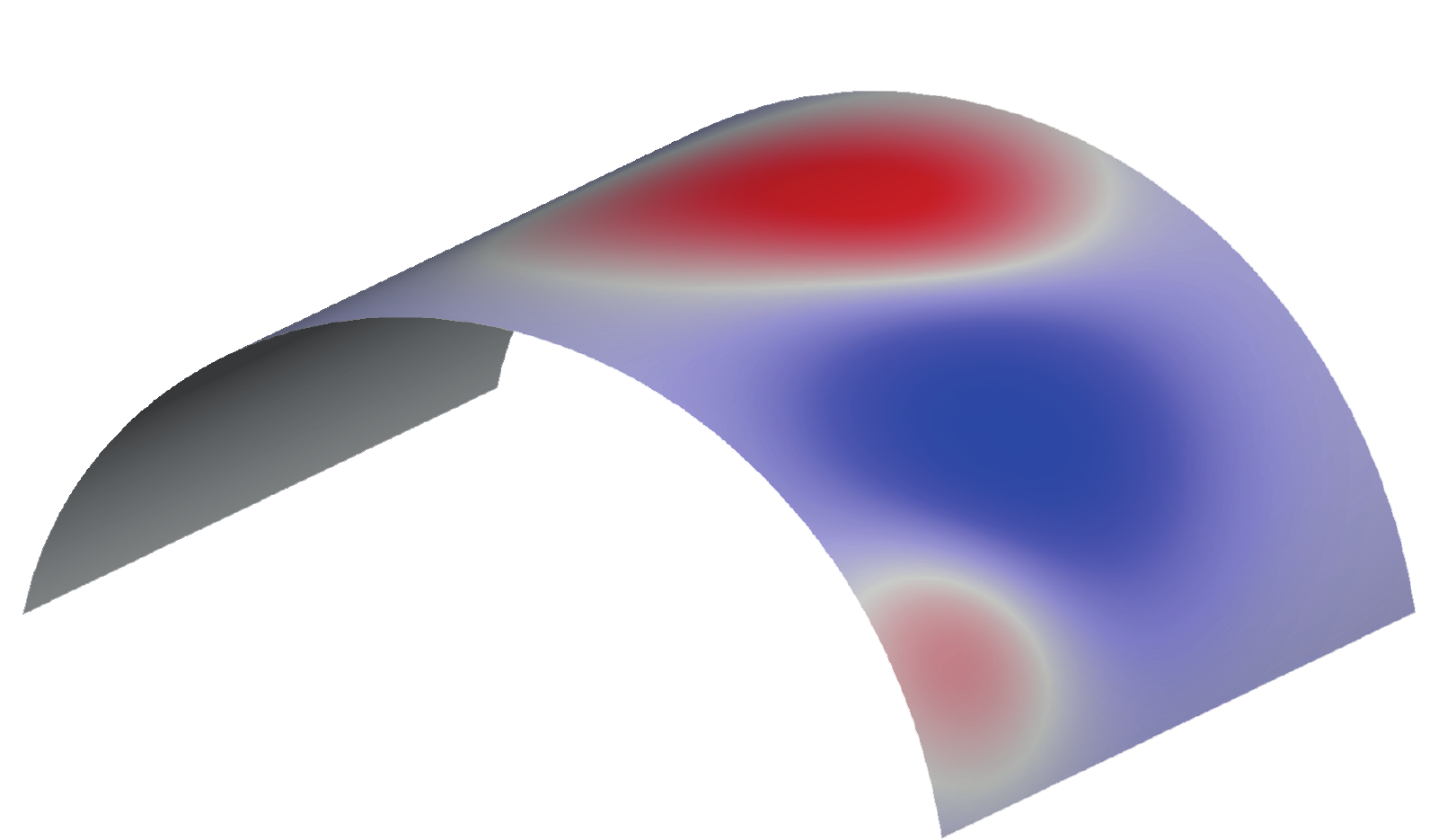}\\
Example of a scalar field. 
}
%
Scalar fields form a vector space $\mathcal{X}(\Omega, \mathbb{R})$ that can be equipped with the inner product 
\begin{equation}
\langle x, y \rangle = \int_\Omega x(u) y(u) \mathrm{d}u,
\label{eq:inner_scalar}   
\end{equation}
where $\mathrm{d}u$ 
is the volume element induced by the Riemannian metric. 
A (smooth) {\em tangent vector field} is a function of the form $X: \Omega \rightarrow T\Omega$  assigning to each point a tangent vector in the respective tangent space, $u\mapsto X(u)\in T_u\Omega$.  
Vector fields\marginnote{\includegraphics[width=0.9\linewidth]{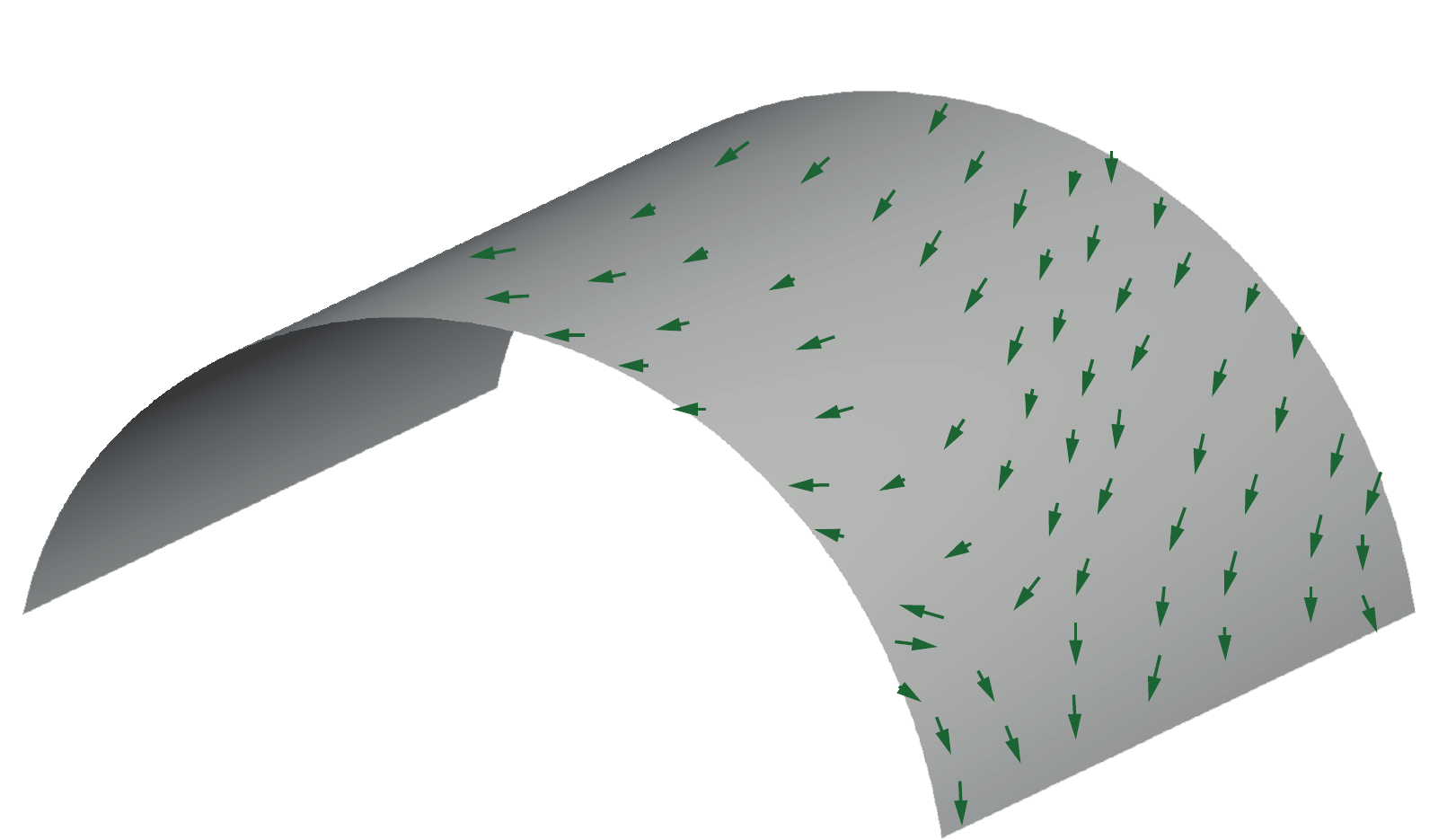}\\
Example of a vector field. The fields are typically assumed to be of the same regularity class (smoothness) as the manifold itself.
} also form a vector space $\mathcal{X}(\Omega, T\Omega)$ with the inner product defined through the Riemannian metric, 
\begin{equation}
\langle X, Y \rangle = \int_\Omega g_u(X(u), Y(u)) \mathrm{d}u. \label{eq:inner_vector}   
\end{equation}

\paragraph{Intrinsic gradient}
Another way to think of (and actually {\em define}) vector fields is as a generalised notion of derivative. 
In classical calculus, one can locally linearise a (smooth) function through the {\em differential} $\mathrm{d}x(u) = x(u+\mathrm{d}u) - x(u)$, which provides the change of the value of the function $x$ at point $u$ as a result of an inifinitesimal displacement $\mathrm{d}u$. 
However, in our case the na{\"i}ve use of this definition is impossible, since 
expressions of the form ``$u+\mathrm{d}u$'' are meaningless on manifolds due to the lack of a global vector space structure. 
%
%

The solution is 
to use tangent vectors as a model of local infinitesimal displacement. 
%
Given a smooth scalar field $x\in \mathcal{X}(\Omega,\mathbb{R})$, we can think of a (smooth) vector field as a linear map $Y:\mathcal{X}(\Omega,\mathbb{R}) \rightarrow \mathcal{X}(\Omega,\mathbb{R})$ satisfying the properties of a {\em derivation}: 
$Y(c) = 0$ for any constant $c$ (corresponding to the intuition that constant functions have vanishing derivatives), $Y(x+z) = Y(x) + Y(z)$ (linearity), and $Y(xz) = Y(x)z + xY(z)$ ({\em product} or {\em Leibniz rule}), for any smooth scalar fields $x,z \in \mathcal{X}(\Omega,\mathbb{R})$. 
It can be shown that one can use these properties to define vector fields axiomatically. 
The differential $\mathrm{d}x(Y) = Y(x)$ 
can be viewed as an operator $(u,Y) \mapsto Y(x)$   
%
and interpreted as follows: the change of $x$ as the result of displacement $Y \in T_u\Omega$ at point $u$ is given by $\mathrm{d}_ux(Y)$. \marginnote{Importantly, this construction {\em does not use the Riemannian metric whatsoever} and can thus can be extended to a more general construction of bundles discussed in the Section~\ref{sec:gauges}. }
It is thus an extension of the classical notion of {\em directional derivative}. 

Alternatively, at each point $u$ the differential  can be regarded as a {\em linear functional} $\mathrm{d}x_u : T_u\Omega \rightarrow \mathbb{R}$ acting on tangent vectors $X \in T_u\Omega$. Linear functionals on a vector space are called {\em dual vectors} or {\em covectors}; if in addition we are given an inner product (Riemannian metric), a dual vector can always be represented as 
$$
\mathrm{d}x_u (X)  = g_u(\nabla x(u), X).\marginnote{This is a consequence of the Riesz-Fr{\'e}chet Representation Theorem, by which every dual vector can be expressed as an inner product with a vector. } 
$$ 
The representation of the differential at point $u$ is a tangent vector $\nabla x(u) \in T_u \Omega$ called the (intrinsic) {\em gradient} of $x$; similarly to the gradient in classical calculus, it can be thought of as the direction of the steepest increase of $x$. 
The gradient considered as an {\em operator} $\nabla : \mathcal{X}(\Omega,\mathbb{R}) \rightarrow \mathcal{X}(\Omega,T\Omega)$ assigns at each point $x(u) \mapsto \nabla x(u) \in T_u \Omega$; thus, the gradient of a scalar field $x$ is a vector field $\nabla x$. 





\paragraph{Geodesics}

Now consider a smooth curve $\gamma : [0,T] \rightarrow \Omega$ on the manifold with endpoints $u = \gamma(0)$ and $v = \gamma(T)$. The derivative of the curve at point $t$ is a tangent vector $\gamma'(t) \in T_{\gamma(t)}\Omega$ called the {\em velocity vector}. \marginnote{It is tacitly assumed that curves are given in {\em arclength parametrisation}, such that $\| \gamma' \| = 1$ (constant velocity).} 
Among all the curves connecting points $u$ and $v$, we are interested in those of {\em minimum length}, i.e., we are seeking $\gamma$ minimising the length functional 
$$
\ell(\gamma) = \int_{0}^T \| \gamma'(t) \|_{\gamma(t)} \mathrm{d}t =  \int_{0}^T g_{\gamma(t)}^{1/2} (\gamma'(t), \gamma'(t)) \mathrm{d}t. 
$$
Such curves are called {\em geodesics} (from the Greek \textgreek{geodai\textsigma{}{\'i}a}, literally `division of Earth') and they play important role in differential geometry. 
Crucially to our discussion, the way we defined geodesics is intrinsic, as they depend solely on the Riemannian metric (through the length functional).

Readers familiar with differential geometry might recall that geodesics are a more general concept and their definition in fact does not necessarily require a Riemannian metric but a {\em connection} (also called a {\em covariant derivative}, as it generalises the notion of derivative to vector and tensor fields), which is defined axiomatically, similarly to our construction of the differential. 
Given a Riemannian metric, there exists a unique special connection called the\marginnote{The Levi-Civita connection is torsion-free and compatible with the metric. The Fundamental Theorem of Riemannian geometry guarantees its existence and uniqueness. } {\em Levi-Civita connection}  which is often tacitly assumed in Riemannian geometry. Geodesics arising from this connection are the length-minimising curves we have defined above.

We will show next how to use geodesics to define a way to transport tangent vectors on the manifold (parallel transport), create local intrinsic maps from the manifold to the tangent space (exponential map), and define distances (geodesic metric). This will allow us to construct convolution-like operations by applying a filter locally in the tangent space. %

\paragraph{Parallel transport}\marginnote{\includegraphics[width=0.9\linewidth]{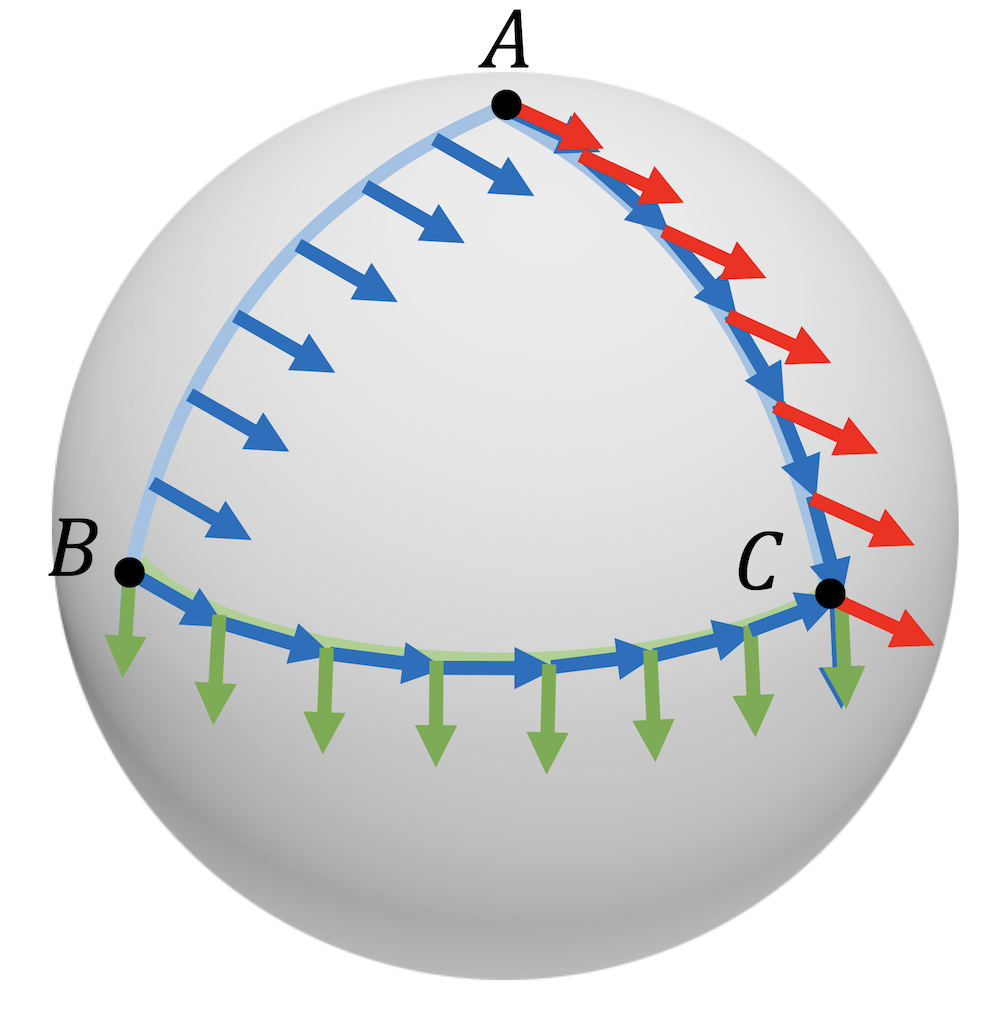}
Euclidean transport of a vector from A to C makes no sense on the sphere, as the resulting vectors (red) are not in the tangent plane. Parallel transport from A to C (blue) rotates the vector along the path. It is path dependent: going along the path BC and ABC produces different results.  
}
One issue we have already encountered when dealing with manifolds is that we cannot directly add or subtract two points $u,v \in \Omega$. 
The same problem arises when trying to compare tangent vectors at different points: though they have the same dimension, they belong to {\em different spaces}, e.g. $X\in T_u\Omega$ and $Y\in T_v\Omega$, and thus not directly comparable. 
Geodesics provide a mechanism to move vectors from one point to another, in the following way: let $\gamma$ be a geodesic connecting points $u=\gamma(0)$ and $v=\gamma(T)$ and let $X \in T_u\Omega$. We can define a new set of tangent vectors along the geodesic, $X(t) \in T_{\gamma(t)}\Omega$ such that the length of $X(t)$ and the angle (expressed through the Riemannian metric) between it and the velocity vector of the curve is constant,
$$
g_{\gamma(t)}(X(t),\gamma'(t) ) = g_{u}(X,\gamma'(0)) = \mathrm{const}, \quad\quad \|X(t)\|_{\gamma(t)} = \|X\|_u = \mathrm{const}.
$$
As a result, we get a unique vector $X(T) \in T_{v}\Omega$ at the end point $v$. 

The map $\Gamma_{u\rightarrow v}(X) : T_u\Omega \rightarrow T_u\Omega$ and $T_v\Omega$ defined as $\Gamma_{u\rightarrow v}(X) = X(T)$ using the above notation is called {\em parallel transport} or {\em connection}; the latter term implying it is a mechanism to `connect' between the tangent spaces $T_u\Omega$ and $T_v\Omega$. 
Due to the angle and length preservation conditions, parallel transport amounts to only rotation of the vector, so it can be associated with an element of the special orthogonal group $\mathrm{SO}(s)$ (called the {\em structure group} of the tangent bundle),\marginnote{Assuming that the manifold is orientable, otherwise $\mathrm{O}(s)$.} which we will denote by $\fg_{u\rightarrow v}$ and 
discuss in further detail in Section~\ref{sec:gauges}. 

As we mentioned before,  a connection can be defined axiomatically independently of the Riemannian metric, providing thus an abstract notion of parallel transport along any smooth curve. The result of such transport, however, depends on the path taken.  


\paragraph{Exponential map}
Locally around a point $u$, it is always possible to define a unique geodesic in a given direction $X \in T_u\Omega$, i.e. such that 
$\gamma(0) = u$ and $\gamma'(0) = X$. 
When $\gamma_X(t)$ is defined for all $t\geq 0$ (that is, we can shoot the geodesic from a point $u$ for as long as we like), the manifold is said to be {\em geodesically complete} and the exponential map is defined on the whole tangent space.
Since compact manifolds are geodesically complete, we can tacitly assume this convenient property.

This definition of geodesic provided a point and a direction gives a natural mapping from (a subset of) the tangent space $T_u \Omega$ to $\Omega$ called the {\em exponential map}\marginnote{
Note that geodesic completeness does not necessarily guarantee that $\exp$ is a global diffeomorphism -- the largest radius $r$ about $u$ for which $\exp_u(B_r(0) \subseteq T_u\Omega)$ is mapped diffeomorphically is called the {\em injectivity radius}. }  $\exp : B_r(0) \subset T_u\Omega \rightarrow \Omega$, which is defined by taking a unit step along the geodesic in the direction $X$, i.e., $\exp_u(X) = \gamma_X(1)$. %
The exponential map $\exp_u$ is a local diffeomorphism, as it deforms the neighbourhood $B_r(0)$ (a ball or radius $r$) of the origin on $T_u\Omega$ into a neighbourhood of $u$. Conversely, one can also regard the exponential map as an intrinsic local deformation (`flattening') of the manifold into the tangent space.

\paragraph{Geodesic distances}
A result known as the Hopf-Rinow Theorem 
\marginnote{Hopf-Rinow Theorem thus estabilishes the equivalence between geodesic and metric completeness, the latter meaning every Cauchy sequence converges in the geodesic distance metric.} guarantees that geodesically complete manifolds are also {\em complete metric spaces}, in which one can realise a distance (called the {\em geodesic distance} or {\em metric})  between any pair of points $u,v$ as the length of the shortest path between them
$$
d_g(u,v) = \min_{\gamma} \ell(\gamma) \quad\quad \text{s.t.} \quad\quad \gamma(0) = u, \,\, \gamma(T) = v,
$$
which exists (i.e., the minimum is attained).\marginnote{Note that the term `metric' is used in two senses: Riemannian metric $g$ and distance $d$. To avoid confusion, we will use the term `distance' referring to the latter. Our notation $d_g$ makes the distance depend on the Riemannian metric $g$, though the definition of geodesic length $L$. }



\paragraph{Isometries}
Consider now a deformation of our manifold $\Omega$ into another manifold $\tilde{\Omega}$ with a Riemannian metric $h$, which we assume to be 
%
a diffeomorphism $\eta: (\Omega, g) \rightarrow (\tilde{\Omega}, h)$ between the manifolds.   
Its differential $\mathrm{d}\eta : T\Omega \rightarrow T\tilde{\Omega}$ defines a map between the respective tangent bundles (referred to as {\em pushforward}), such that   
at a point $u$, we have $\mathrm{d}\eta_u : T_u \Omega \rightarrow T_{\eta(u)}\tilde{\Omega}$, 
interpreted as before: 
if we make a small displacement from point $u$ by tangent vector $X \in T_u \Omega$, the map $\eta$ will be displaced from point $\eta(u)$ by tangent vector $\mathrm{d}\eta_u(X) \in T_{\eta(u)}\tilde{\Omega}$.

Since the pushforward\marginnote{Pushforward and pullback are adjoint operators $\langle \eta^* \alpha, X \rangle = \langle \alpha, \eta_*X \rangle$ where $\alpha \in T^*\Omega$ is a {\em dual vector field}, defined at each point as a linear functional acting on $T_u\Omega$ and the inner products are defined respectively on vector and dual vector fields. } provides a mechanism to associate tangent vectors on the two manifolds, it allows to {\em pullback } the metric $h$ from $\tilde{\Omega}$ to $\Omega$, 
$$
(\eta^*h)_u(X,Y) = h_{\eta(u)}(\mathrm{d}\eta_u(X), \mathrm{d}\eta_u(Y))
$$ 
If the pullback metric coincides at every point with that of $\Omega$, i.e., $g = \eta^*h$, the map $\eta$ is called  (a Riemannian) {\em isometry}. 
For two-dimensional manifolds (surfaces), isometries can be intuitively understood as inelastic deformations that deform the manifold without `stretching' or `tearing' it.

By virtue of their definition, isometries preserve intrinsic structures such as geodesic distances, which are expressed entirely in terms of the Riemannian metric. 
Therefore, we can also understand isometries from the position of metric geometry, 
 as distance-preserving maps (`metric isometries') {\em between metric spaces} $\eta : (\Omega, d_g) \rightarrow (\tilde{\Omega}, d_h)$, in the sense that 
$$
d_g (u,v) = d_h(\eta(u),\eta(v)) 
$$
for all $u,v \in \Omega$, 
or more compactly, $d_g = d_h \circ (\eta \times \eta)$. In other words, Riemannian isometries are also metric isometries. 
On {\em connected} manifolds, the converse is also true: every metric isometry is also a Riemannian isometry. 
\marginnote{This result is known as the Myers–Steenrod Theorem. We tacitly assume our manifolds to be connected. }

In our Geometric Deep Learning blueprint, $\eta$ is a model of domain deformations. When $\eta$ is an isometry, 
any intrinsic quantities are unaffected by such deformations. 
%
%
One can generalise exact (metric) isometries through the notions of 
{\em metric dilation} 
$$
\mathrm{dil}(\eta) = \sup_{u\neq v \in \Omega}\frac{d_h(\eta(u),\eta(v))}{d_g (u,v)}
$$
or {\em metric distortion} 
$$
\mathrm{dis}(\eta) = \sup_{u, v \in \Omega}|d_h(\eta(u),\eta(v)) - d_g (u,v)|,
\marginnote{The Gromov-Hausdorff distance
between metric spaces, which we mentioned in Section~\ref{sec:isomorphism}, can be expressed as the smallest possible metric distortion.}
$$
which capture the relative and absolute change of the geodesic distances under $\eta$, respectively. 
The condition~(\ref{eqn:domain_def_stability}) for the stability of a function $f \in \mathcal{F}(\mathcal{X}(\Omega))$ 
under domain deformation
can be rewritten in this case as 
$$
\| f(x,\Omega) - f(x\circ \eta^{-1},\tilde{\Omega})\| \leq C \|x\| \mathrm{dis}(\eta). 
$$


\paragraph{Intrinsic symmetries}
A particular case of the above is a diffeomorphism  of the domain itself (what we termed {\em automorphism} in Section~\ref{sec:isomorphism}), which we will denote by $\tau \in \mathrm{Diff}(\Omega)$. We will call it a Riemannian  \mbox{(self-)isometry} if the pullback metric satisfies $\tau^* g = g$, or a metric (self-)isometry if $d_g = d_g\circ (\tau \times \tau)$.
Not surprisingly,\marginnote{Continuous symmetries on manifolds are infinitesimally generated by special tangent vector fields called {\em Killing fields}, named after Wilhelm Killing.} isometries form a group with the composition operator denoted by $\mathrm{Iso}(\Omega)$ and called the {\em isometry group}; the identity element is the map $\tau(u) = u$ and the inverse always exists (by definition of $\tau$ as a diffeomorphism). 
Self-isometries are thus {\em intrinsic symmetries} of manifolds.


\paragraph{Fourier analysis on Manifolds}
%
We will now show how to construct intrinsic convolution-like operations on manifolds, which, by construction, will be invariant to isometric deformations.    
For this purpose, 
we have two options: One is to use an analogy of the Fourier transform, and define the convolution as a product in the Fourier domain. 
The other is to define the convolution spatially, by correlating a filter locally with the signal. 
Let us discuss the spectral approach first.

We remind that in the Euclidean domain the Fourier transform is obtained as the eigenvectors of circulant matrices, which are jointly diagonalisable due to their commutativity. Thus, any circulant matrix and in particular, differential operator, can be used to define an analogy of the Fourier transform on general domains. 
In Riemannian geometry, it is common to use the orthogonal eigenbasis of the Laplacian operator, which we will define here.

For this purpose, recall our definition of the intrinsic gradient operator $\nabla : \mathcal{X}(\Omega,\mathbb{R}) \rightarrow \mathcal{X}(\Omega,T\Omega)$, 
producing a tangent vector field that indicates the local direction of steepest increase of a scalar field on the manifold. 
In a similar manner, we can define the {\em divergence operator} $\nabla^* : \mathcal{X}(\Omega,T\Omega) \rightarrow \mathcal{X}(\Omega,\mathbb{R})$. 
If we think of a tangent vector field as a
flow on the manifold, the divergence measures
the net flow of a field at a point, allowing to distinguish
between field `sources' and `sinks'. 
We use the notation $\nabla^*$ (as opposed to the common $\mathrm{div}$) to emphasise that the two operators are adjoint,
$$
\langle X, \nabla x\rangle = \langle \nabla^* X,  x\rangle,
$$
where we use the inner products (\ref{eq:inner_scalar}) and~(\ref{eq:inner_vector}) between scalar and vector fields. 

The {\em Laplacian} (also known as the {\em Laplace-Beltrami operator} in differential geometry) is an operator on $\mathcal{X}(\Omega)$ defined as $\Delta = \nabla^* \nabla$,  
%
which can be interpreted \marginnote{From this interpretation it is also clear that the Laplacian is isotropic. We will see in Section~\ref{sec:meshes} that it is possible to define {\em anisotropic Laplacians} (see \citep{andreux2014anisotropic,boscaini2016anisotropic}) of the form $\nabla^*(A(u)\nabla)$, where $A(u)$ is a position-dependent tensor determining local direction. 
}
as the difference between the average of a function on an
infinitesimal sphere around a point and the value of the
function at the point itself. 
It is one of the most important operators in mathematical physics, used to describe phenomena as diverse as heat diffusion, quantum oscillations,
and wave propagation.  
Importantly in our context, the Laplacian is intrinsic, and thus invariant under isometries of $\Omega$.

It is easy to see that the Laplacian is self-adjoint (`symmetric'),
$$
\langle \nabla x, \nabla x \rangle = \langle x , \Delta x\rangle = \langle \Delta x ,  x\rangle.
$$
The quadratic form on the left in the above expression is actually the already familiar Dirichlet energy,  
$$c^2(x) = \| \nabla x \|^2 = \langle \nabla x, \nabla x \rangle = \int_\Omega \| \nabla x(u) \|_u^2 \mathrm{d}u = \int_\Omega g_u( \nabla x(u),  \nabla x(u)) \mathrm{d}u
$$
measuring the smoothness of $x$. 
%
%
%
%
%
%

The Laplacian operator admits an eigedecomposition 
$$
\Delta \varphi_k = \lambda_k \varphi_k, \quad\quad k=0, 1, \hdots 
$$
%
with countable spectrum if the manifold is compact (which we tacitly assume), and orthogonal eigenfunctions, $\langle \varphi_k, \varphi_l \rangle = \delta_{kl}$, due to the self-adjointness of $\Delta$. 
%
The Laplacian eigenbasis can also be constructed as a set of orthogonal minimisers of the Dirichlet energy, 
$$
\varphi_{k+1} = \arg\min_{\varphi} \| \nabla \varphi \|^2      \quad\quad \text{s.t.}\quad\quad   \|\varphi \|=1 \,\,\, \text{and} \,\,\, \langle \varphi, \varphi_{j} \rangle = 0 
$$
for $j=0,\hdots, k$, allowing to interpret it as the smoothest orthogonal basis on $\Omega$. 
The eigenfunctions $\varphi_0, \varphi_1, \hdots$ and the corresponding eigenvalues $0 = \lambda_0 \leq \lambda_1 \leq \hdots $ can be interpreted as the analogy of the atoms and frequencies in the classical Fourier transform. \marginnote{In fact $e^{\mi \xi u}$ are the eigenfunctions of the Euclidean Laplacian $\tfrac{\mathrm{d}^2}{\mathrm{d} u^2}$.}

This orthogonal basis allows to expand square-integrable functions on $\Omega$ into {\em Fourier series}  
$$
x (u) = \sum_{k \geq 0} \langle x, \varphi_k \rangle \varphi_k (u)
$$
where $\hat{x}_k = \langle x, \varphi_k \rangle$ are referred to as the {\em Fourier coefficient} or the (generalised) Fourier transform of $x$. \marginnote{Note that this Fourier transform has a discrete index, since the spectrum is discrete due to the compactness of $\Omega$.}
Truncating the Fourier series results in an approximation error that can be bounded \citep{aflalo2013spectral} by 
$$
\left\| x - \sum_{k = 0}^N \langle x, \varphi_k \rangle \varphi_k \right\|^2 \leq \frac{\| \nabla x\|^2}{\lambda_{N+1}}.
$$
\cite{aflalo2015optimality} further showed that no other basis attains a better error, making the Laplacian eigenbasis  {\em optimal} for representing smooth signals on manifolds.

\paragraph{Spectral Convolution on Manifolds}

{\em Spectral convolution} can be defined as the product of Fourier transforms of the signal $x$ and the filter $\theta$, 
\begin{equation}
(x \star \theta)(u) = 
\sum_{k \geq 0} (\hat{x}_k \cdot \hat{\theta}_k ) \varphi_k (u).
\label{eqn:conv_spectral}
\end{equation}
Note that here we use what is a {\em property} of the classical Fourier transform (the Convolution Theorem) as a way to {\em define} a non-Euclidean convolution.
By virtue of its construction, the spectral convolution is intrinsic and thus isometry-invariant. 
Furthermore, since the Laplacian operator is isotropic, it has no sense of direction; in this sense, the situation is similar to that we had on graphs in Section~\ref{sec:proto-graphs} due to permutation invariance of neighbour aggregation.

\begin{figure}[h!]
    \centering
    \includegraphics[width=\linewidth]{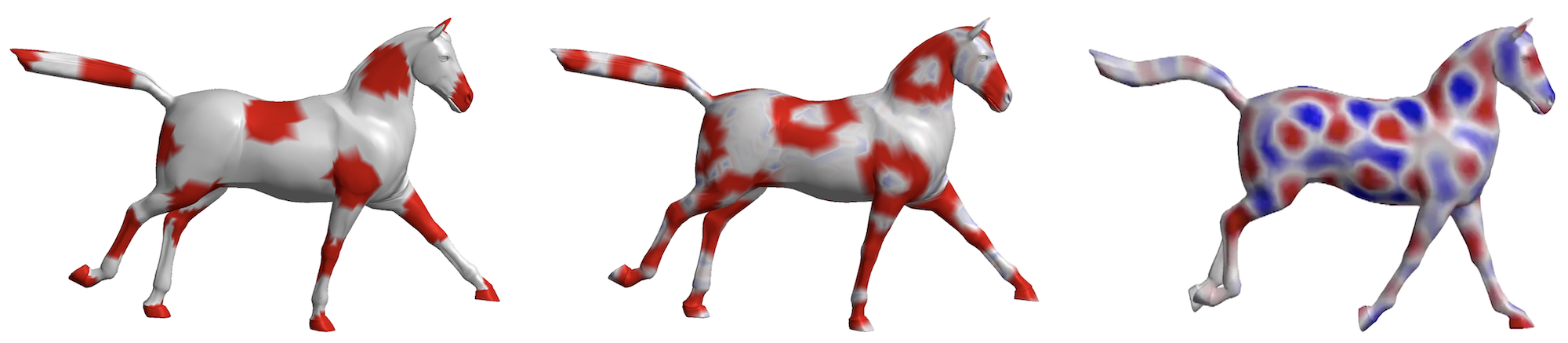}
    \caption{Instability of spectral filters under domain perturbation. Left: a signal $\mathbf{x}$ on the mesh $\Omega$. Middle: result of spectral filtering 
    in the eigenbasis 
    of the Laplacian $\Delta$ on $\Omega$. 
    Right: 
    the same spectral filter 
    applied to the eigenvectors 
    of the Laplacian $\tilde{\Delta}$ of a nearly-isometrically perturbed domain $\tilde{\Omega}$ 
    produces a very different result. 
    }
    \label{fig:mesh_horses}
\end{figure}%

In practice, a direct computation of~(\ref{eqn:conv_spectral}) appears to be prohibitively expensive due to the need to diagonalise the Laplacian. Even worse, 
it turns out geometrically unstable: the higher-frequency eigenfunctions of the Laplacian can change dramatically as a result of even small near-isometric perturbations of the domain $\Omega$ (see Figure~\ref{fig:mesh_horses}). 
A more stable solution is provided by realising the filter as a {\em spectral transfer function} of the form $\hat{p}(\Delta)$, 
\begin{eqnarray}
(\hat{p}(\Delta) x)(u) &=& \sum_{k\geq 0} \hat{p}(\lambda_k) \langle x, \varphi_k\rangle \varphi_k (u) \label{eqn:conv_spec} \\
&=& \int_\Omega x(v) \, \sum_{k\geq 0} \hat{p}(\lambda_k) \varphi_k(v) \varphi_k (u) \, \mathrm{d}v \label{eqn:conv_spat}
\end{eqnarray}
which can be interpreted in two manners: either as a spectral filter~(\ref{eqn:conv_spec}), where we identify $\hat{\theta}_k = \hat{p}(\lambda_k)$, or as a spatial filter~(\ref{eqn:conv_spat}) with a position-dependent kernel $\theta(u,v) = \sum_{k\geq 0} \hat{p}(\lambda_k) \varphi_k(v) \varphi_k (u)$.  
The advantage of this formulation is that $\hat{p}(\lambda)$ can be parametrised by a small number of coefficients, and choosing parametric functions such as polynomials\marginnote{Geometric Deep Learning methods based on spectral convolution expressed through the Fourier transform are often referred to as `spectral' and opposed to `spatial' methods we have seen before in the context of graphs. We see here that these two views may be equivalent, so this dichotomy is somewhat artificial and not completely appropriate.} $\hat{p}(\lambda) = \sum_{l=0}^r \alpha_l \lambda^l$ allows for efficiently computing the filter as 
$$
(\hat{p}(\Delta) x)(u) = \sum_{k\geq 0} \sum_{l= 0}^r \alpha_l \lambda_k^l \, \langle x, \varphi_k\rangle \varphi_k (u) = \sum_{l= 0}^r \alpha_l (\Delta^l x)(u),
$$
avoiding the spectral decomposition altogether. 
We will discuss this construction in further detail in Section~\ref{sec:meshes}. 



\paragraph{Spatial Convolution on Manifolds}

A second alternative is to attempt defining convolution on manifolds is by matching a filter at different points, like we did in formula~(\ref{eq:group-conv}), 
\begin{equation}
(x \star \theta)(u) = \int_{T_u \Omega} x(\exp_u Y) \theta_u(Y) \mathrm{d}Y, 
\label{eq:conv_tangent_space}
\end{equation}
where we now have to use the exponential map to access the values of the scalar field $x$ from the tangent space, and 
the filter $\theta_u$ is defined in the tangent space at each point and hence position-dependent. %
If one defines the filter intrinsically, such a convolution would be isometry-invariant, a property we mentioned as crucial in many computer vision and graphics applications. 

We need, however, to note several substantial differences from our previous construction in Sections~\ref{sec:grids_euclidean}--\ref{sec:groups}. 
First, because a manifold is generally not a homogeneous space, we do not have anymore a global group structure allowing us have a shared filter (i.e., the same $\theta$ at every $u$ rather than $\theta_u$ in expression~(\ref{eq:conv_tangent_space})) defined at one point and then move it around. 
%
An analogy of this operation on the manifold would require parallel transport, allowing to apply a shared $\theta$, defined as a function on $T_u\Omega$, at some other $T_v\Omega$.   
However, as we have seen, this in general will depend on the path between $u$ and $v$, so {\em the way we move the filter around matters.}
Third, since we can use the exponential map only locally, the filter must be {\em local}, with support bounded by the injectivity radius.  
Fourth 
and most crucially, we cannot work with $\theta(X)$, as $X$ is an abstract geometric object: in order for it to be used for computations, we must represent it {\em relative to some local basis}  
$\omega_u : \mathbb{R}^s \rightarrow T_u\Omega$, as an $s$-dimensional array of coordinates $\mathbf{x} = \omega^{-1}_u(X)$. 
This allows us to rewrite the convolution~(\ref{eq:conv_tangent_space}) as 
\begin{equation}
(x \star \theta)(u) = \int_{[0,1]^s} x(\exp_u (\omega_u \mathbf{y})) \theta(\mathbf{y}) \mathrm{d}\mathbf{y}, 
\label{eqn:conv_exp}
\end{equation}
%


with the filter defined on the unit cube. Since the exponential map is intrinsic (through the definition of geodesic), the resulting convolution is isometry-invariant.

Yet, this tacitly assumed  we can carry the frame $\omega_u$ along to another manifold, i.e. $\omega_u' = \mathrm{d}\eta_u \circ \omega_u$.
%
Obtaining
such a frame (or \emph{gauge}, in physics terminology) given only a the manifold $\Omega$ in a consistent manner is however fraught with difficulty.
%
First, a smooth global gauge may not exist: this is the situation on manifolds that are not \emph{parallelisable},\marginnote{The sphere $\mathbb{S}^2$ is an example of a non-parallelisable manifold, a result of    
the {\em Poincar{\'e}-Hopf Theorem}, which is colloquially stated as `one cannot comb a hairy ball without creating a cowlick.' \vspace{2mm}
}   in which case one cannot define a smooth non-vanishing tangent vector field.
%
%
Second, we do not have a canonical gauge on manifolds, so this choice is arbitrary; since our convolution depends on $\omega$, if one chose a different one, we would obtain different results. 

We should note that this is a case where practice diverges from theory: in practice, 
it is possible to build frames that are mostly smooth, with a limited number of singularities, e.g. by taking the intrinsic gradient of some intrinsic scalar field on the manifold.  \marginnote{
\includegraphics[width=1\linewidth]{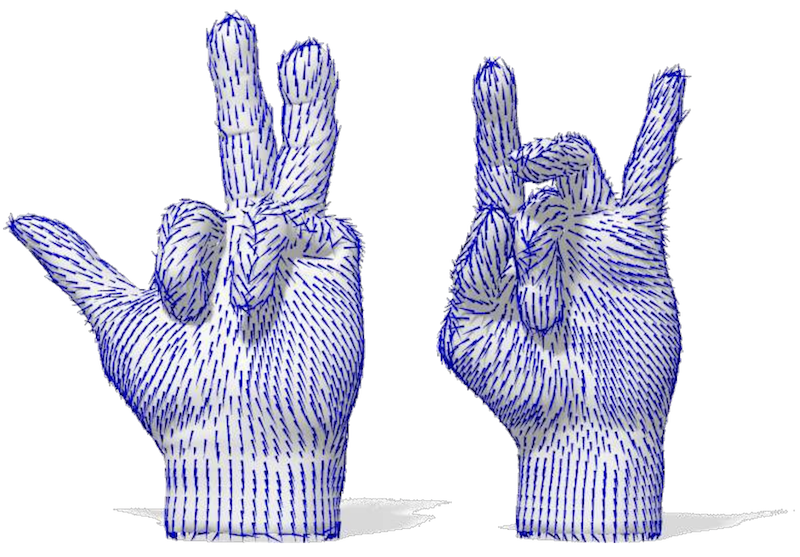} 
Example of stable gauges constructed on nearly-isometric manifolds (only one axis is shown) using the GFrames algorithm of \cite{melzi2019gframes}.
} 
%
Moreover, such constructions are stable, i.e., the frames constructed this way will be identical on isometric manifolds and similar on approximately isometric ones. 
Such approaches were in fact employed in the early works on deep learning on manifolds  \citep{masci2015geodesic,monti2017geometric}.

Nevertheless, this solution is not entirely satisfactory because near singularities, the filter orientation (being defined in a fixed manner relative to the gauge) will vary wildly, leading to a non-smooth feature map even if the input signal and filter are smooth. 
Moreover, there is no clear reason why a given direction at some point $u$ should be considered equivalent to another direction at an altogether different point $v$. 
Thus, despite {\em practical} alternatives, we will look next for 
a more {\em theoretically} well-founded approach that would be altogether independent on the choice of gauge. 

\subsection{Gauges and Bundles}
\label{sec:gauges}

The notion of gauge, which we have defined as a frame for the tangent space, is 
%
quite a bit more general in physics:  it can refer to a frame for any\marginnote{Historically, fibre bundles arose first in modern differential geometry of {\'E}lie Cartan (who however did not define them explicitly), and were then further developed as a standalone object in the field of topology in the 1930s.} \emph{vector bundle}, not just the tangent bundle. 
Informally, a vector bundle describes a family of vector spaces parametrised by another space and 
consists of a \emph{base space} $\Omega$ with an identical vector space $\mathbb{V}$ (called the \emph{fibre}) attached to each position $u \in \Omega$ (for the tangent bundle these are the tangent spaces $T_u\Omega$).  
%
Roughly speaking, a bundle looks as a product $\Omega \times \mathbb{V}$ locally around $u$, but globally might be `twisted' and have an overall different structure. 
%
%
%
%
%
In Geometric Deep Learning, fibres 
can be used to model the feature spaces at each point in the manifold $\Omega$, with the dimension of the fibre being equal to the number of feature channels. 
In this context, a new and fascinating kind of symmetry, called \emph{gauge symmetry} may present itself.




Let us consider again an $s$-dimensional manifold $\Omega$ with its tangent bundle $T\Omega$, and a vector field 
$X : \Omega \rightarrow T \Omega$ (which in this terminology is referred to as a {\em section} on the tangent bundle). 
Relative to a gauge $\omega$ for the tangent bundle, $X$ is represented as a function $\mathbf{x} : \Omega \rightarrow \R^s$.
However it is important to realise that what we are really interested in is the underlying geometrical object (vector field), whose representation as a function $\mathbf{x} \in \mathcal{X}(\Omega,\R^s)$ {\em depends on the choice of gauge} $\omega$. 
If we change the gauge, we also need to change $\mathbf{x}$ so as to preserve the underlying vector field being represented.




%


%

\paragraph{Tangent bundles and the Structure group}
When we change the gauge, we need to apply at each point an   invertible matrix that maps the old gauge to the new one.  This matrix is unique for every pair of gauges at each point, but possibly different at different points.  
%
In other words, a {\em gauge transformation} is a mapping $\fg : \Omega \rightarrow \GL{s}$, where $\GL{s}$ is the {\em general linear group} of invertible $s \times s$ matrices.
It acts on the gauge $\omega_u : \R^s \rightarrow T_u \Omega$ to produce a new gauge $\omega'_u = \omega_u \circ \fg_u : \R^s \rightarrow T_u \Omega$. 
The gauge transformation acts on a coordinate vector field 
at each point 
via $\mathbf{x}'(u) = \fg^{-1}_u \mathbf{x}(u)$ to produce the coordinate representation $\mathbf{x}'$ of $X$ relative to the new gauge.
The underlying vector field remains unchanged: 
$$
X(u) = \omega'_u(\mathbf{x'}(u)) = \omega_u( \fg_u \fg^{-1}_u \mathbf{x}(u)) = \omega_u(\mathbf{x}(u)) = X(u), 
$$
which is exactly the property we desired. 
More generally, we may have a field of geometric quantities that transform according to a representation $\rho$ of $\GL{s}$, e.g. a field of 2-tensors (matrices) $\mathbf{A}(u) \in \R^{s \times s}$ that transform like $\mathbf{A'}(u) = \rho_2(\fg^{-1}_u) \mathbf{A}(u) = \rho_1(\fg_u) \mathbf{A}(u) \rho_1(\fg^{-1}_u)$. 
In this case, the gauge transformation $\fg_u$ acts via $\rho(\fg_u)$.

Sometimes we may wish to restrict attention to frames with a certain property, such as orthogonal frames, right-handed frames, etc.
Unsurprisingly, we are interested in a set of  some property-preserving transformations that  form a group.
For instance, the group that preserves orthogonality is the orthogonal group $\Orth{s}$ (rotations and reflections), and the group that additionally preserves  orientation or `handedness' is $\SO{s}$ (pure rotations).
Thus, in general we have a group $\fG$ called the \emph{structure group} of the bundle, and a gauge transformation is a map $\fg : \Omega \rightarrow \fG$.
A key observation is that in all cases with the given property, for any two frames at a given point there exists exactly one gauge transformation relating them.


As mentioned before, gauge theory extends beyond tangent bundles, and in general, we can consider a bundle of vector spaces whose structure and dimensions are not necessarily related to those of the base space $\Omega$. \marginnote{We use $s$ to denote the dimension of the base space $\Omega$ and $d$ referring to the dimension of the fibre. For tangent bundles, $d=s$ is the dimension of the underlying manifold. For RGB image, $s=2$ and $d=3$.} 
For instance, a color image pixel has a position $u \in \Omega = \Z^2$ on a 2D grid and a value $\mathbf{x}(u) \in \R^3$ in the RGB space, so the space of pixels can be viewed as a vector bundle with base space $\Z^2$ and a fibre $\R^3$ attached at each point. 
It is customary to express an RGB image relative to a gauge that has basis vectors for R, G, and B (in that order), so that the coordinate representation of the image looks like $\mathbf{x}(u) = (r(u), g(u), b(u))^\top$.
But we may equally well permute the basis vectors (color channels) independently at each position, as long as we remember the frame (order of channels) in use at each point\marginnote{In this example we have chosen $\fG = \Sigma_3$ the permutations of the 3 color channels as the structure group of the bundle. Other choices, such as a Hue rotation $\fG = \SO{2}$ are also possible.}.
As a computational operation this is rather pointless, but as we will see shortly it is conceptually useful to think about gauge transformations for the space of RGB colors, because it allows us to express a gauge symmetry -- in this case, an equivalence between colors -- and make functions defined on images respect this symmetry (treating each color equivalently).

As in the case of a vector field on a manifold, an RGB gauge transformation changes the numerical representation of an image (permuting the RGB values independently at each pixel) but not the underlying image. 
In machine learning applications, we are interested in constructing functions $f \in \mathcal{F}(\mathcal{X}(\Omega))$ on such images (e.g. to perform image classification or segmentation), implemented as layers of a neural network. 
It follows that if, for whatever reason, we were to apply a gauge transformation to our image, we would need to also change the  function $f$ (network layers) so as to preserve their meaning. 
Consider for simplicity a $1 \times 1$ convolution, i.e. a map that takes an RGB pixel $\mathbf{x}(u) \in \R^3$ to a feature vector $\mathbf{y}(u) \in \R^C$.
According to our Geometric Deep Learning blueprint, the output is associated with a group representation $\rho_{\textup{out}}$, in this case a $C$-dimensional representation of the structure group $\fG = \Sigma_3$ (RGB channel permutations), and similarly the input is associated with a representation $\rho_{\textup{in}}(\fg) = \fg$. 
Then, if we apply a gauge transformation to the input, we would need to change the linear map ($1 \times 1$ convolution) $f : \R^3 \rightarrow \R^C$ to $f' = \rho^{-1}_{\textup{out}}(\fg) \circ f \circ \rho_{\textup{in}}(\fg)$ so that the output feature vector $\mathbf{y}(u) = f(\mathbf{x}(u))$ transforms like $\mathbf{y}'(u) = \rho_{\textup{out}}(\fg_u) \mathbf{y}(u)$ at every point.
Indeed we verify: 
\begin{equation*}
    \mathbf{y}' = f'(\mathbf{x}') = \rho^{-1}_{\textup{out}}(\fg) f(\rho_{\textup{in}}(\fg) \rho^{-1}_{\textup{in}}(\fg) \mathbf{x}) = \rho^{-1}_{\textup{out}}(\fg) f(\mathbf{x}).\marginnote{Here the notation $\rho^{-1}(\fg)$ should be understood as the inverse of the group representation (matrix) $\rho(\fg)$.}
\end{equation*}



\paragraph{Gauge Symmetries}
To say that we consider gauge transformations to be symmetries is to say that any two gauges related by a gauge transformation are to be considered equivalent. 
For instance, if we take $\fG = \SO{d}$, any two right-handed orthogonal frames are considered equivalent, because we can map any such frame to any other such frame by a rotation.
In other words, there are no distinguished local directions such as ``up'' or ``right''.
Similarly, if $\fG = \Orth{d}$ (the orthogonal group), then any left and right handed orthogonal frame are considered equivalent.
In this case, there is no preferred orientation either.
In general, we can consider a group $\fG$ 
and a collection of frames at every point $u$ such that for any two of them there is a unique $\fg(u) \in \fG$ that maps one frame onto the other. 
%


Regarding gauge transformations as symmetries in our Geometric Deep Learning blueprint, we are interested in making  
the functions $f$ acting on signals defined on $\Omega$ and expressed with respect to the gauge should equivariant to such transformation. 
Concretely, this means that if we apply a gauge transformation to the input, the output should undergo the same transformation (perhaps acting via a different representation of $\fG$).
We noted before that when we change the gauge, the function $f$ should be changed as well, but for a gauge equivariant map this is not the case: changing the gauge leaves the mapping invariant.
To see this, consider again the RGB color space example.
The map $f : \R^3 \rightarrow \R^C$ is equivariant if $f \circ \rho_{\textup{in}}(\fg) = \rho_{\textup{out}}(\fg) \circ f$, but in this case the gauge transformation applied to $f$ has no effect: $\rho_{\textup{out}}^{-1}(\fg) \circ f \circ \rho_{\textup{in}}(\fg) = f$.
In other words, the coordinate expression of a gauge equivariant map is independent of the gauge, in the same way that 
in the case of graph, we applied the same function 
regardless of how the input nodes were permuted. 
However, unlike the case of graphs and other examples covered so far, gauge transformations act {\em not on} $\Omega$ but separately {\em on each of the feature vectors} $\mathbf{x}(u)$ by a transformation $\fg(u) \in \fG$ for each $u \in \Omega$.

%

Further considerations enter the picture when we look at filters on manifolds with a larger spatial support. 
Let us first consider an easy example 
of a mapping $f : \mathcal{X}(\Omega, \mathbb{R}) \rightarrow \mathcal{X}(\Omega, \mathbb{R})$ from 
scalar fields to scalar fields on an $s$-dimensional manifold $\Omega$. 
Unlike vectors and other geometric quantities, scalars do not have an orientation, so a scalar field $x \in \mathcal{X}(\Omega, \R)$ is {\em invariant} to gauge transformations (it transforms according to the trivial representation $\rho(\fg) = 1$). 
Hence, any linear map from scalar fields to scalar fields is {\em gauge equivariant} (or invariant, which is the same in this case). 
For example, we could write $f$ similarly to~(\ref{eqn:conv_spat}), as a convolution-like operation with a position-dependent filter $\theta : \Omega \times \Omega \rightarrow \R$, 
\begin{equation}
    \label{eq:intro:general-linear-map-scalar-fields}
    (x \star \theta)(u) = \int_{\Omega} \theta(u, v) x(v) \mathrm{d}v.  
\end{equation}
This implies that we have a potentially different filter $\theta_u = \theta(u, \cdot)$ at each point, i.e., no spatial weight sharing --- which gauge symmetry alone does not provide. 

Consider now a more interesting case of a mapping $f : \mathcal{X}(\Omega, T\Omega) \rightarrow \mathcal{X}(\Omega, T\Omega)$ from vector fields to vector fields. 
Relative to a gauge, the input and output vector fields $X, Y \in \mathcal{X}(\Omega, T\Omega)$ are vector-valued functions $\mathbf{x}, \mathbf{y} \in \mathcal{X}(\Omega, \R^s)$. 
A general linear map between such functions can be written using the same equation we used for scalars  (\ref{eq:intro:general-linear-map-scalar-fields}), only replacing the scalar kernel by a matrix-valued one $\boldsymbol{\Theta} : \Omega \times \Omega \rightarrow \R^{s \times s}$.
The matrix $\boldsymbol{\Theta} (u, v)$ should map tangent vectors in $T_v\Omega$ to tangent vectors in $T_u\Omega$, but these points have {\em different gauges} that we may change {\em arbitrarily and independently.} 
 That is, the filter would have to satisfy $\boldsymbol{\Theta} (u, v) = \rho^{-1}(\fg(u)) \boldsymbol{\Theta}(u, v) \rho(\fg(v))$ for all $u, v \in \Omega$, where $\rho$ denotes the action of $\fG$ on vectors, given by an $s\times s$ rotation matrix. 
Since $\fg(u)$ and $\fg(v)$ can be chosen freely, this is an overly strong constraint on the filter. \marginnote{Indeed $\boldsymbol{\Theta}$ would have to be zero in this case}

A better approach is to first transport the vectors to a common tangent space by means of the connection,  and then impose gauge equivariance w.r.t. a single gauge transformation at one point only. 
Instead of (\ref{eq:intro:general-linear-map-scalar-fields}), we can then define the following map between vector fields,
\begin{equation}
    (\mathbf{x} \star \boldsymbol{\Theta})(u) = \int_\Omega \boldsymbol{\Theta}(u, v) \rho(\fg_{v \rightarrow u}) \mathbf{x}(v) \mathrm{d}v,\label{eqn:gauge_eq_conv}
\end{equation}
where $\fg_{v \rightarrow u} \in \fG$ denotes the parallel transport from $v$ to $u$ along the geodesic connecting these two points;  
its representation $\rho(\fg_{v \rightarrow u})$ is an $s \times s$ rotation matrix rotating the vector as it moves between the points. 
Note that this geodesic is assumed to be unique, which is true only locally and thus the filter must have a local support. 
Under a gauge transformation $\fg_u$, this element transforms as $\fg_{u \rightarrow v} \mapsto \fg_u^{-1} \fg_{u \rightarrow v} \fg_v$, and the field itself transforms as $\mathbf{x}(v) \mapsto \rho(\fg_v) \mathbf{x}(v)$. 
If the filter commutes with the structure group representation  $\boldsymbol{\Theta}(u, v) \rho(\fg_u) = \rho(\fg_u) \boldsymbol{\Theta}(u, v)$, 
equation~(\ref{eqn:gauge_eq_conv}) defines a {\em gauge-equivariant convolution}, which transforms as  
$$ 
(\mathbf{x}' \star \boldsymbol{\Theta})(u) = \rho^{-1}(\fg_u) (\mathbf{x} \star \boldsymbol{\Theta})(u). 
$$ 
under the aforementioned transformation. 




\subsection{Geometric graphs and Meshes}
\label{sec:meshes}

We will conclude our discussion of different geometric domains with {\em geometric graphs} (i.e., graphs that can be realised in some geometric space) and {\em meshes}. 
In our `5G' of geometric domains, meshes fall somewhere between graphs and manifolds: in many regards, they are similar to graphs, but their additional structure allows to also treat them similarly to continuous objects. 
For this reason, we do not consider meshes as a standalone object in our scheme, and in fact, will emphasise that many of the constructions we derive in this section for meshes are directly applicable to general graphs as well.   

As we already mentioned in Section~\ref{sec:manifolds}, two-dimensional manifolds (surfaces) are a common way of modelling 3D objects (or, better said, the boundary surfaces of such objects).  In computer graphics and vision applications, such surfaces are often discretised as {\em triangular meshes}, \marginnote{Triangular meshes are examples of topological structures known as {\em simplicial complexes}.} 
which can be roughly thought of as a piece-wise planar approximation of a surface obtained by gluing triangles together along their edges. 
Meshes are thus (undirected) {\em graphs with additional structure}: in addition to nodes and edges, a mesh 
$\mathcal{T} = (\mathcal{V},\mathcal{E}, \mathcal{F})$ 
also have ordered triplets of nodes forming {\em triangular faces} $\mathcal{F} =  \{ (u,v,q) : u,v,q \in \mathcal{V} \,\,\, \text{and} \,\,\, (u,v), (u,q), (q,v) \in  \mathcal{E}\}$; the order of the nodes defines the face {\em orientation}. \marginnote{
\includegraphics[width=0.9\linewidth]{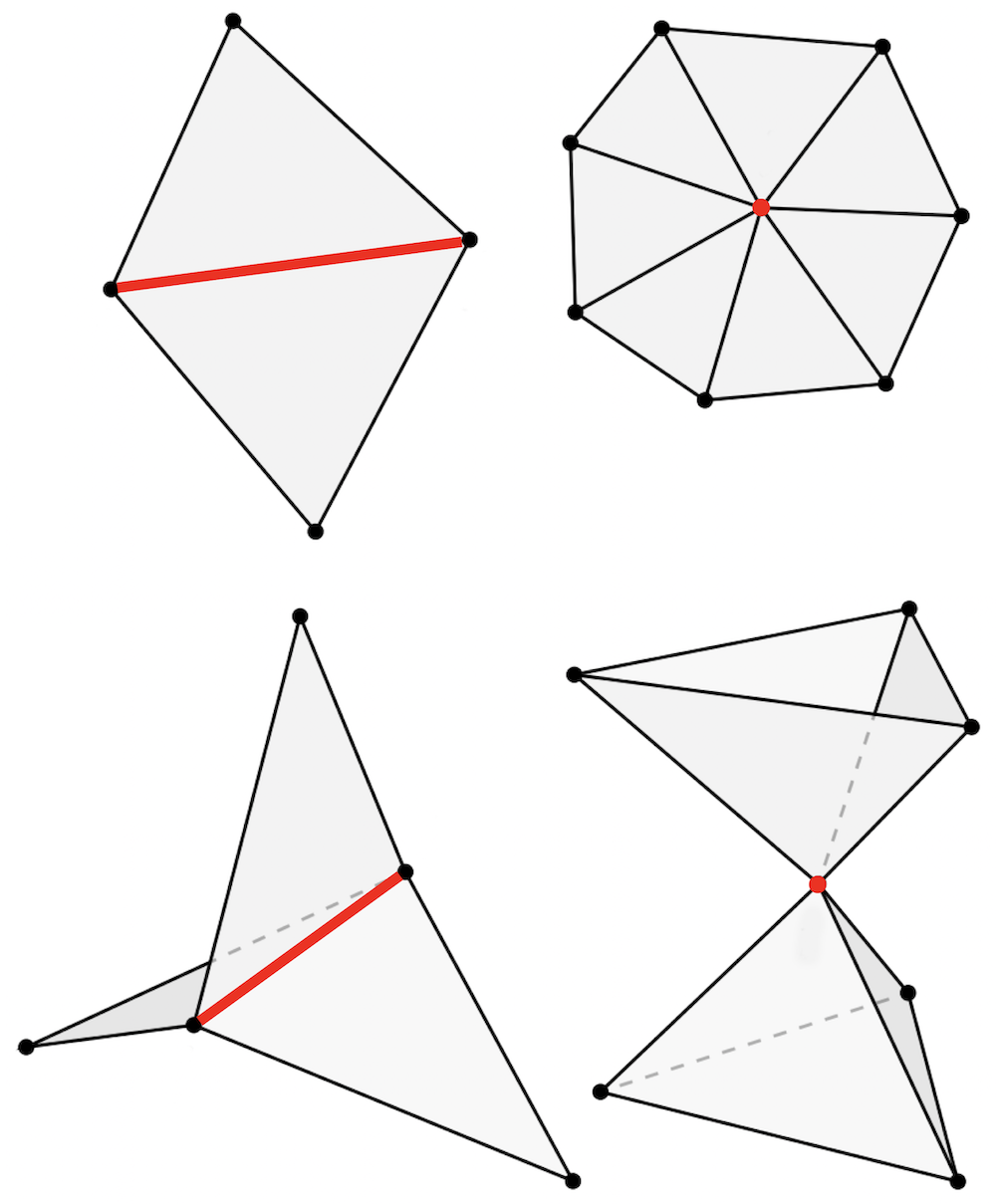}\\
Examples of manifold (top) and non-manifold (bottom) edges and nodes. 
For manifolds with boundary, one further defines {\em boundary edges} that belong to exactly one triangle. }
%

It is further assumed that that each edge is shared by exactly two triangles, and the boundary of all triangles incident on each node forms a single loop of edges. This condition guarantees that 1-hop neighbourhoods around each node are disk-like and the mesh thus constitutes a {\em discrete manifold} -- such meshes are referred to as {\em manifold meshes}. 
%
%
%
Similarly to Riemannian manifolds, we can define a {\em metric} on the mesh. In the simplest instance, it can be  induced from the embedding of the mesh nodes $\mathbf{x}_1, \hdots, \mathbf{x}_n$ and expressed through the Euclidean length of the edges, $\ell_{uv} = \| \mathbf{x}_u - \mathbf{x}_v \|$. 
A metric defined in this way automatically satisfies properties such as the {\em triangle inequality}, i.e., expressions of the form 
$
\ell_{uv} \leq \ell_{uq} + \ell_{vq}
$
for any $(u,v,q) \in \mathcal{F}$ and any combination of edges. %
Any property that can be expressed solely in terms of $\ell$ is {\em intrinsic}, and any deformation of the mesh preserving $\ell$ is an {\em isometry}  -- these notions are already familiar to the reader from our discussion in Section~\ref{sec:manifolds}.   

\paragraph{Laplacian matrices}
By analogy to our treatment of graphs, let us assume a (manifold) mesh with $n$ nodes, each associated with a $d$-dimensional feature vector, which we can arrange (assuming some arbitrary ordering) into an $n\times d$ matrix $\mathbf{X}$. 
The features can represent the geometric coordinates of the nodes as well as additional properties such as colors, normals, etc, or in specific applications such as chemistry where geometric graphs model molecules, properties such as the atomic number.

Let us first look at the 
spectral convolution~(\ref{eqn:conv_spectral}) on meshes, which we remind the readers, arises from the Laplacian operator. 
Considering the mesh as a discretisation of an underlying continuous surface, we can discretise the Laplacian  as 
\begin{equation}
(\boldsymbol{\Delta}\mathbf{X})_u = \sum_{v \in \mathcal{N}_u} w_{uv} (\mathbf{x}_u - \mathbf{x}_v),  
\label{eq:mesh_lap}
\end{equation}
or in matrix-vector notation, as an $n\times n$ symmetric matrix $\boldsymbol{\Delta} = \mathbf{D} - \mathbf{W}$, where $\mathbf{D} = \mathrm{diag}(d_1, \hdots, d_n)$ is called the {\em degree matrix} and $d_u = \sum_{v}w_{uv}$ the {\em degree} of node $u$.  
It is easy to see that equation~(\ref{eq:mesh_lap}) performs  local permutation-invariant aggregation of neighbour features $\phi(\mathbf{x}_u,\mathbf{X}_{\mathcal{N}_u}) =  d_u \mathbf{x}_u - \sum_{v \in \mathcal{N}_u} w_{uv} \mathbf{x}_v$, and $\mathbf{F}(\mathbf{X}) = \boldsymbol{\Delta}\mathbf{X}$ is in fact an instance of our general blueprint~(\ref{eq:graph_equivariant}) for constructing permutation-equivariant functions on graphs. 

Note that insofar there is nothing {\em specific to meshes} in our definition of Laplacian in~(\ref{eq:mesh_lap}); in fact, this construction is valid for arbitrary graphs as well, with edge weights identified with the adjacency matrix, $\mathbf{W} = \mathbf{A}$, i.e., $w_{uv} = 1$\marginnote{The degree in this case equals the number of neighbours. } if $(u,v) \in \mathcal{E}$ and zero otherwise.  
Laplacians constructed in this way are often called {\em combinatorial}, to reflect the fact that they merely capture the connectivity structure of the graph. \marginnote{If the graph is directed, the corresponding Laplacian is non-symmetric. }
For geometric graphs (which do not necessarily have the additional structure of meshes, but whose nodes do have spatial coordinates that induces a metric in the form of edge lengths), it is common to use weights inversely related to the metric, e.g. $w_{uv} \propto e^{-\ell_{uv}}$.

On meshes, we can exploit the additional structure afforded by the faces, and define the edge weights in equation~(\ref{eq:mesh_lap}) using the {\em cotangent formula} \citep{pinkall1993computing,meyer2003discrete}\marginnote{
\includegraphics[width=\linewidth]{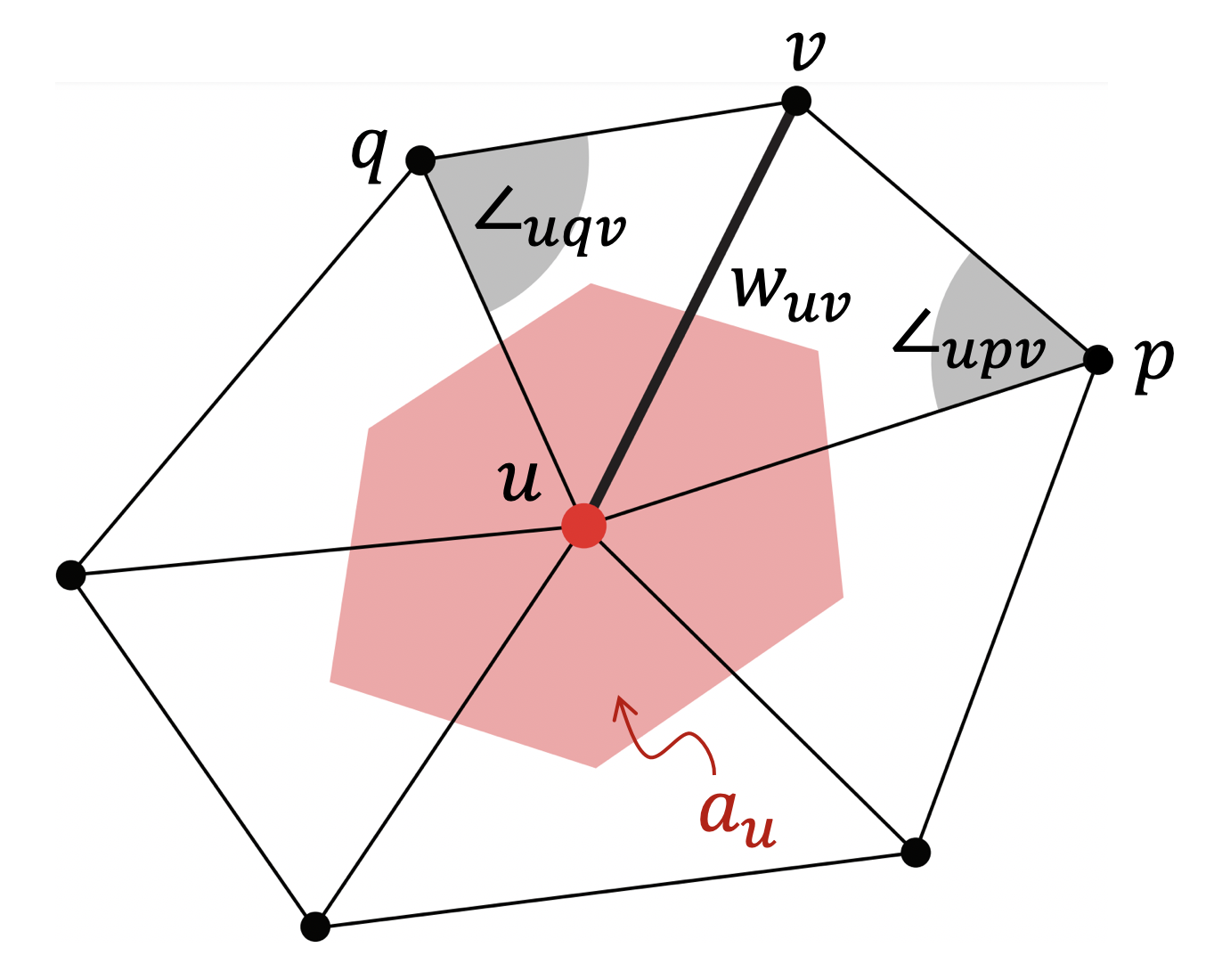}
The earliest use of this formula dates back to the PhD thesis of \cite{macneal1949solution}, who developed it to solve PDEs on the Caltech Electric Analog Computer.}
\begin{equation}
w_{uv} = \frac{\cot \angle_{uqv} + \cot \angle_{upv}}{2 a_u}
\label{eq:cotan}
\end{equation}
where $\angle_{uqv}$ and $\angle_{upv}$ are the two angles in the triangles $(u,q,v)$ and $(u,p,v)$ opposite the shared edge $(u,v)$, and $a_u$ is the local area element, typically computed as the area of the polygon constructed upon the barycenters of the triangles $(u,p,q)$ sharing the node $u$ and given by
$a_u = \frac{1}{3}\sum_{v,q : (u,v,q) \in \mathcal{F}} a_{uvq}$.

The cotangent Laplacian can be shown to have multiple convenient properties (see e.g. \cite{wardetzky2007discrete}): it is a {\em positive-semidefinite} matrix, $\boldsymbol{\Delta} \succcurlyeq 0$ and thus has non-negative eigenvalues $\lambda_1 \leq \hdots \leq \lambda_n$ that can be regarded as an analogy of frequency, it is symmetric and thus has orthogonal eigenvectors, and it is {\em local} (i.e., the value of $(\boldsymbol{\Delta}\mathbf{X})_u$ depends only on 1-hop neighbours, $\mathcal{N}_u$).
Perhaps the most important property is the convergence of the cotangent mesh Laplacian matrix $\boldsymbol{\Delta}$ to the continuous operator $\Delta$ when the mesh is infinitely refined \citep{wardetzky2008convergence}.  Equation~(\ref{eq:cotan}) constitutes thus an appropriate {\em discretisation}\marginnote{Some technical conditions must be imposed on the refinement, to avoid e.g. triangles becoming pathological. One such example is a bizarre triangulation of the cylinder known in German as the {\em Schwarzscher Stiefel} (Schwarz's boot) or in English literature as the `Schwarz lantern', proposed in 1880 by Hermann Schwarz, a German mathematician known from the Cauchy-Schwarz inequality fame. } of the Laplacian operator defined on Riemannian manifolds in Section~\ref{sec:manifolds}.

While one expects the Laplacian to be intrinsic, this is not very obvious from equation~(\ref{eq:cotan}), and it takes some effort to 
express the cotangent weights entirely in terms of the discrete metric $\ell$ as 
$$
w_{uv} = \frac{-\ell^2_{uv} + \ell^2_{vq} + \ell^2_{uq} }{8 a_{uvq}} + 
\frac{-\ell^2_{uv} + \ell^2_{vp} + \ell^2_{up} }{8 a_{uvp}}
$$
where the area of the triangles $a_{ijk}$ is given as  
$$
a_{uvq} = \sqrt{s_{uvq} (s_{uvq} - \ell_{uv}) (s_{uvq} - \ell_{vq}) (s_{uvq} - \ell_{uq}) }
$$
using {\em Heron's semiperimeter formula} with 
$s_{uvq} = \frac{1}{2}(\ell_{uv} + \ell_{uq} + \ell_{vq})$. 
This endows the Laplacian (and any quantities associated with it, such as its eigenvectors and eigenvalues) with {\em isometry invariance}, a property for which it is so loved in geometry processing and computer graphics (see an excellent review by \cite{wang2019intrinsic}): any deformation of the mesh that does not affect the metric $\ell$ (does not `stretch' or `squeeze' the edges of the mesh) does not change the Laplacian.

Finally, as we already noticed,\marginnote{
\includegraphics[width=0.9\linewidth]{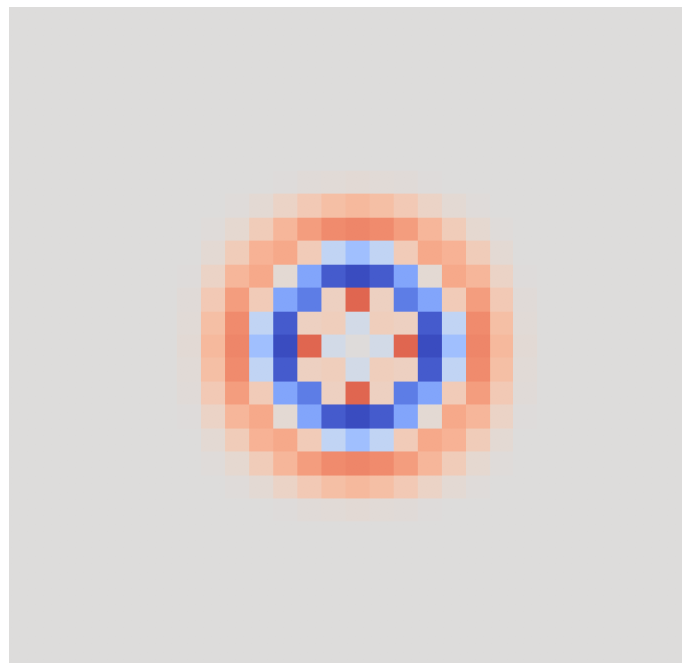}
Laplacian-based filters are isotropic. In the plane, such filters have radial symmetry. 
} the definition of the Laplacian~(\ref{eq:cotan}) is invariant to the permutation of nodes in $\mathcal{N}_u$, as it involves aggregation in the form of summation. 
While on general graphs this is a necessary evil due to the lack of canonical ordering of neighbours, on meshes we can order the 1-hop neighbours according to some orientation (e.g., clock-wise), and the only ambiguity is the selection of the first node. Thus, instead of any possible permutation we need to account for {\em cyclic shifts} (rotations), which intuitively corresponds to the ambiguity arising from $\mathrm{SO}(2)$ gauge transformations discussed in Section~\ref{sec:gauges}. 
For a fixed gauge, it is possible to define an {\em anisotropic Laplacian} that is sensitive to local directions and amounts to changing the metric or the weights $w_{uv}$. 
Constructions of this kind were used to design shape descriptors by \cite{andreux2014anisotropic,boscaini2016anisotropic} and in early Geometric Deep Learning architectures on meshes by \cite{boscaini2016learning}.

\paragraph{Spectral analysis on meshes}
The orthogonal eigenvectors $\boldsymbol{\Phi} = (\boldsymbol{\varphi}_1, \hdots, \boldsymbol{\varphi}_n)$ diagonalising the  Laplacian matrix ($\boldsymbol{\Delta} = \boldsymbol{\Phi} \boldsymbol{\Lambda} \boldsymbol{\Phi}^\top$, where $\boldsymbol{\Lambda} = \mathrm{diag}(\lambda_1, \hdots, \lambda_n)$ is the diagonal matrix of Laplacian eigenvalues), are used as the non-Euclidean analogy of the Fourier basis, allowing to perform spectral convolution on the mesh as the product of the respective Fourier transforms, 
$$
\mathbf{X} \star \boldsymbol{\theta} = 
\boldsymbol{\Phi} \, \mathrm{diag}(\boldsymbol{\Phi}^\top \boldsymbol{\theta}) (\boldsymbol{\Phi}^\top \mathbf{X}) 
=
\boldsymbol{\Phi}\, \mathrm{diag}(\hat{\boldsymbol{\theta}}) \hat{\mathbf{X}},
$$
where the filter $\hat{\boldsymbol{\theta}}$ is designed directly in the Fourier domain.  
Again, nothing in this formula is specific to meshes, and one can use the Laplacian matrix of a generic (undirected) graph.\marginnote{The fact that the graph is assumed to be undirected is important: in this case the Laplacian is symmetric and has orthogonal eigenvectors.}
It is tempting to exploit this spectral definition of convolution to generalise CNNs to graphs, which in fact was done by one of the authors of this text, \cite{bruna2013spectral}. 
However, it appears that the non-Euclidean Fourier transform 
is extremely sensitive to even minor perturbations of the underlying mesh or graph (see Figure~\ref{fig:mesh_horses} in Section~\ref{sec:manifolds}) and thus can only be used when one has to deal with different signals on a {\em fixed} domain, but not when one wishes to generalise across {\em different domains}. 
Unluckily, many computer graphics and vision problems fall into the latter category, where one trains a neural network on one set of 3D shapes (meshes) and test on a different set, making the Fourier transform-based approach inappropriate.  


As noted in Section~\ref{sec:manifolds}, it is preferable to use spectral filters of the form~(\ref{eqn:conv_spec}) applying some transfer function $\hat{p}(\lambda)$ to the Laplacian matrix, 
$$
\hat{p}(\boldsymbol{\Delta})\mathbf{X} = \boldsymbol{\Phi} \hat{p}(\boldsymbol{\Lambda})\boldsymbol{\Phi}^\top\mathbf{X} 
= \boldsymbol{\Phi}\, \mathrm{diag}(\hat{p}(\lambda_1), \hdots, \hat{p}(\lambda_n)) \hat{\mathbf{X}}. 
$$
%
%
When $\hat{p}$ can be expressed in terms of matrix-vector products, the eigendecomposition of the $n\times n$ matrix $\boldsymbol{\Delta}$ \marginnote{In the general case, the complexity of eigendecomposition is $\mathcal{O}(n^3)$.} can be avoided altogether.  
For example, \cite{defferrard2016convolutional} used  {\em polynomials} of degree $r$ as filter functions, 
$$
\hat{p}(\boldsymbol{\Delta})\mathbf{X} = \sum_{k=0}^r \alpha_k \boldsymbol{\Delta}^k \mathbf{X}  = \alpha_0 \mathbf{X} + \alpha_1 \boldsymbol{\Delta} \mathbf{X}  + \hdots + \alpha_r \boldsymbol{\Delta}^r \mathbf{X}, 
$$
amounting to the multiplication of the $n\times d$ feature matrix $\mathbf{X}$ by the $n\times n$ Laplacian matrix $r$ times. Since the Laplacian is typically sparse (with $\mathcal{O}(|\mathcal{E}|)$ non-zero elements) \marginnote{Meshes are nearly-regular graphs, with each node having $\mathcal{O}(1)$ neighbours, resulting in $\mathcal{O}(n)$ non-zeros in $\boldsymbol{\Delta}$.  }
this operation has low complexity of $\mathcal{O}(|\mathcal{E}|dr)\sim \mathcal{O}(|\mathcal{E}|)$. 
Furthermore, since the Laplacian is local,  
a polynomial filter of degree $r$ is localised in $r$-hop neighbourhood.

However, this exact property comes at a disadvantage when dealing with meshes, since the actual support of the filter (i.e., the radius it covers) depends on the {\em resolution} of the mesh. 
One has to bear in mind that meshes arise from the discretisation of some underlying continuous surface, and one may have two different meshes $\mathcal{T}$ and $\mathcal{T}'$ representing {\em the same object}. \marginnote{    \includegraphics[width=\linewidth]{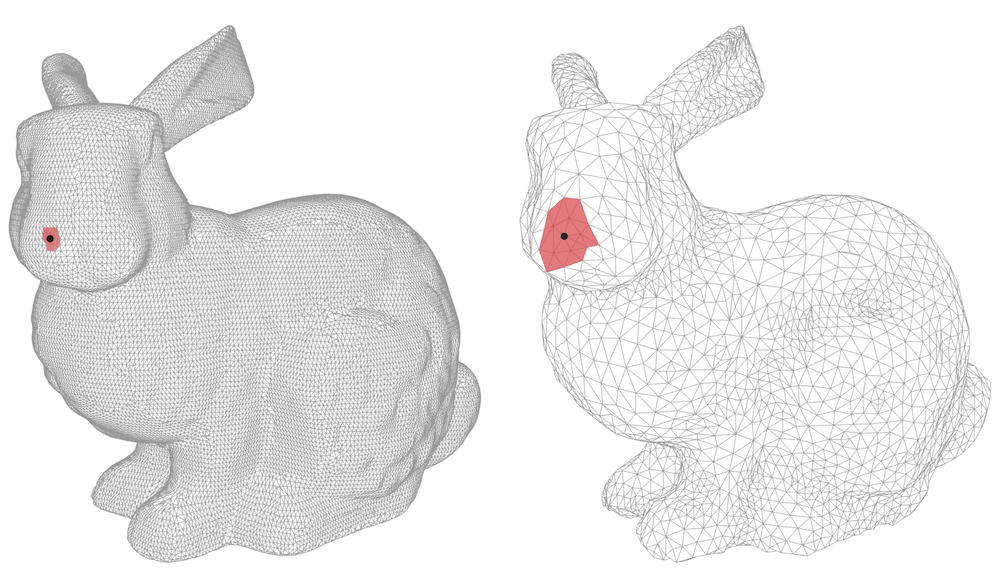}\\
Two-hop neighbourhoods on meshes of different resolution. } 
In a finer mesh, one might have to use larger neighbourhoods (thus, larger degree $r$ of the filter) than in a coarser one. 

For this reason, in computer graphics applications it is more common to use {\em rational filters}, since they are  resolution-independent. There are many ways to define such filters (see, e.g. \cite{patane2020fourier}), the most common being as a polynomial of some  rational function, e.g., $\frac{\lambda-1}{\lambda+1}$. 
More generally, one can use a complex function, such as the {\em Cayley transform} $\frac{\lambda - \mi}{ \lambda + \mi}$ that maps the real line into the unit circle in the complex plane. \marginnote{Cayley transform is a particular case of a {\em M{\"o}bius transformation}. When applied to the Laplacian (a positive-semindefinite matrix), it maps its non-negative eigenvalues to the complex  half-circle. }
\cite{levie2018cayleynets} used spectral filters expressed as {\em Cayley polynomials}, real rational functions with complex coefficients $\alpha_l \in \mathbb{C}$,   
$$
\hat{p}(\lambda) = \mathrm{Re} \left(  \sum_{l=0}^r 
\alpha_l \left( \frac{\lambda - \mi}{ \lambda + \mi}
\right)^{l}
\right).
$$
When applied to matrices, the computation of the Cayley polynomial requires matrix inversion, 
$$
\hat{p}(\boldsymbol{\Delta}) = \mathrm{Re} \left(  \sum_{l=0}^r
\alpha_l (\boldsymbol{\Delta} - \mi\mathbf{I})^{l} (\boldsymbol{\Delta} + \mi\mathbf{I})^{-l}
\right),\marginnote{In signal processing, polynomial filters are termed {\em finite impulse response} (FIR), whereas rational filters are {\em infinite impulse response} (IIR).}
$$
which can be carried out approximately with linear complexity. 
Unlike polynomial filters, rational filters do not have a local support, but have exponential decay 
\citep{levie2018cayleynets}.  
A crucial difference compared to the direct computation of the Fourier transform is that polynomial and rational filters are stable under approximate isometric deformations of the underlying graph or mesh -- various results of this kind were shown e.g. by \cite{levie2018cayleynets,levie2019transferability,gama2020stability,kenlay2021interpretable}.


\paragraph{Meshes as operators and Functional maps}

The paradigm of functional maps suggests thinking of meshes as {\em operators}. As we will show, this allows obtaining more interesting types of invariance exploiting the additional structure of meshes.  
For the purpose of our discussion, assume the mesh $\mathcal{T}$ is constructed upon embedded nodes with coordinates $\mathbf{X}$. 
If we construct an intrinsic operator like the Laplacian, it can be shown that it encodes completely the structure of the mesh, and one can recover the mesh (up to its isometric embedding, as shown by \cite{zeng2012discrete}). 
This is also true for some other operators (see e.g. \cite{boscaini2015shape,corman2017functional,chern2018shape}), so we will assume a general operator, or $n\times n$ matrix  $\mathbf{Q}(\mathcal{T}, \mathbf{X})$, as a representation of our mesh.

In this view, the discussion of Section~\ref{sec:proto-graphs} of learning functions of the form   $f(\mathbf{X},\mathcal{T})$ 
can be rephrased 
as learning functions of the form $f(\mathbf{Q})$. 
Similar to graphs and sets, the nodes of meshes also have no canonical ordering, i.e., functions on meshes must satisfy the permutation invariance or equivariance conditions, 
\begin{eqnarray*}
f(\mathbf{Q}) &=& f(\mathbf{P}\mathbf{Q}\mathbf{P}^\top) \\
\mathbf{P}\mathbf{F}(\mathbf{Q}) &=& \mathbf{F}(\mathbf{P}\mathbf{Q}\mathbf{P}^\top)
\end{eqnarray*}
for any permutation matrix $\mathbf{P}$. 
%
%
However, compared to general graphs we now have more structure: we can assume that our mesh arises from the discretisation of some underlying continuous surface $\Omega$. It is thus possible to have a different mesh  $\mathcal{T}'=(\mathcal{V}',\mathcal{E}',\mathcal{F}')$ with $n'$ nodes and coordinates $\mathbf{X}'$ representing the same object $\Omega$ as $\mathcal{T}$. 
%
%
Importantly, the meshes $\mathcal{T}$ and $\mathcal{T}'$ can have a different connectivity structure and even different number of nodes ($n'\neq n$). Therefore, we cannot think of these meshes as isomorphic graphs with mere reordering of nodes and consider the permutation matrix $\mathbf{P}$ as correspondence between them.

\begin{figure}[h!]
    \centering
    \includegraphics[width=0.45\linewidth]{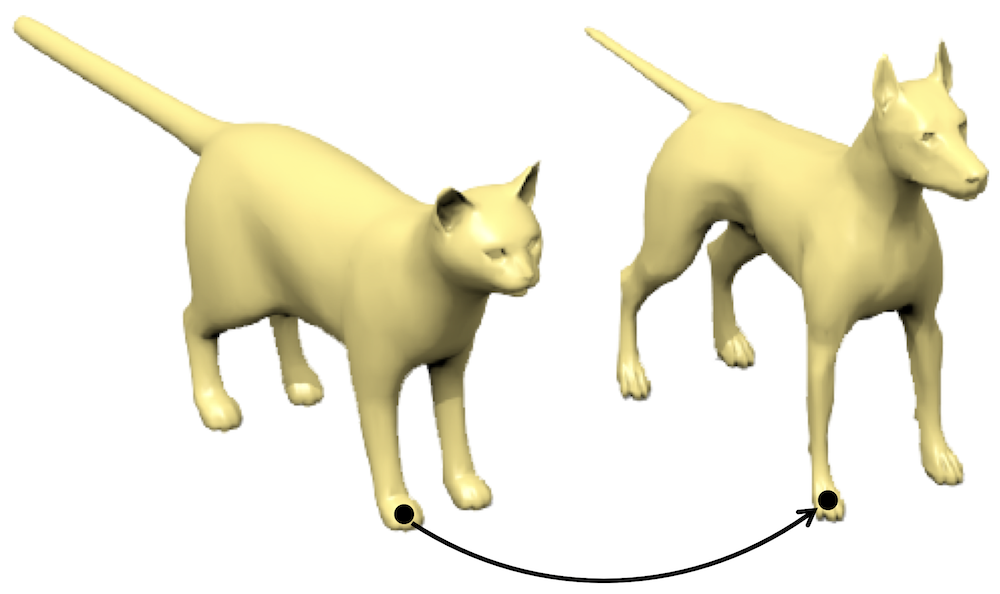}\hspace{5mm}    \includegraphics[width=0.45\linewidth]{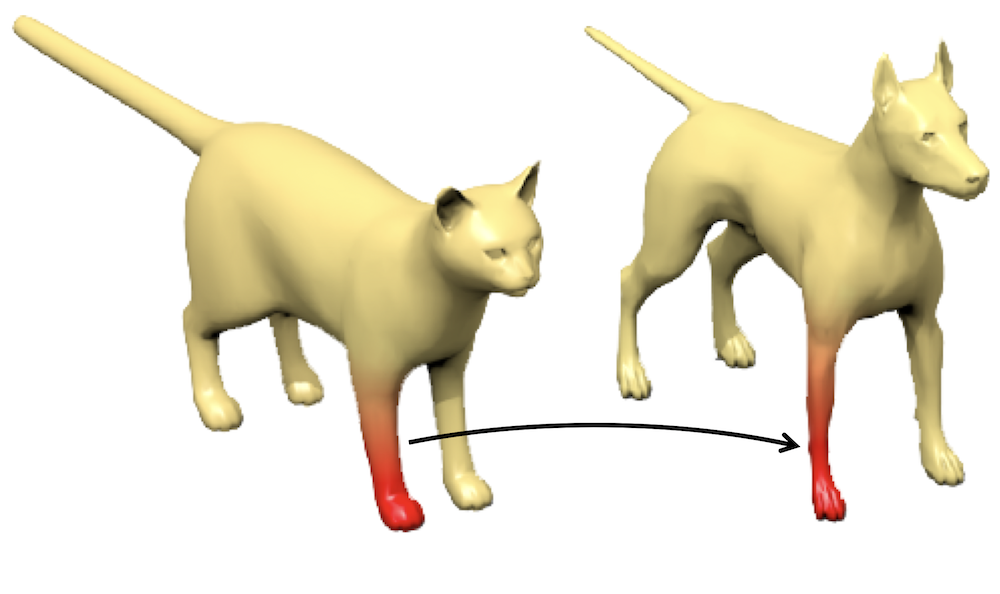}
    \caption{Pointwise map (left) vs functional map (right).   
    }
    \label{fig:func_maps}
\end{figure}

Functional maps were introduced by \cite{ovsjanikov2012functional} as a generalisation of the notion of correspondence to such settings, replacing the correspondence between {\em points} on two domains (a map $\eta : \Omega \rightarrow \Omega'$) with correspondence between {\em functions} (a map $\mathbf{C}:\mathcal{X}(\Omega) \rightarrow \mathcal{X}(\Omega')$, see  Figure~\ref{fig:func_maps}). 
A {\em functional map} is a linear operator 
$\mathbf{C}$, represented as a matrix $n'\times n$,  establishing correspondence between signals $\mathbf{x}'$
and $\mathbf{x}$ 
on the respective domains as 
$$
\mathbf{x}' = \mathbf{C}\mathbf{x}.\marginnote{
In most cases the functional map is implemented in the spectral domain, as a $k\times k$ map $\hat{\mathbf{C}}$ between the Fourier coefficients,
$
\mathbf{x}' = \boldsymbol{\Phi}'\hat{\mathbf{C}}\boldsymbol{\Phi}^\top\mathbf{x},
$
where $\boldsymbol{\Phi}$ and $\boldsymbol{\Phi}'$ are the respective $n\times k$ and $n'\times k$ matrices of the (truncated) Laplacian eigenbases, with $k\ll n,n'$. 
}
$$
%
%
\cite{rustamov2013map} showed that in order to guarantee {\em area-preserving} mapping, the functional map must be orthogonal, $\mathbf{C}^\top \mathbf{C} = \mathbf{I}$, i.e., be an element of the orthogonal group $\mathbf{C} \in \mathrm{O}(n)$. In this case, we can invert the map using $\mathbf{C}^{-1} = \mathbf{C}^\top$.

The functional map also establishes a relation between the operator representation of meshes, 
$$
\mathbf{Q}' = \mathbf{C} \mathbf{Q} \mathbf{C}^\top, \quad \quad \mathbf{Q} = \mathbf{C}^\top \mathbf{Q}' \mathbf{C}, 
$$
which we can interpret as follows: given an operator representation $\mathbf{Q}$ of $\mathcal{T}$ and a functional map $\mathbf{C}$, we can construct its representation $\mathbf{Q}'$ of $\mathcal{T}'$ by first mapping the signal from $\mathcal{T}'$ to $\mathcal{T}$ (using $\mathbf{C}^\top$), applying the operator $\mathbf{Q}$, and then mapping back to $\mathcal{T}'$ (using $\mathbf{C}$)\marginnote{Note that we read these operations \emph{right-to-left}. }
%
This leads us to a more general class of {\em remeshing  invariant} (or equivariant) functions on meshes, satisfying 
\begin{eqnarray*}
f(\mathbf{Q}) &=& f(\mathbf{C}\mathbf{Q}\mathbf{C}^\top) = f(\mathbf{Q}')\\
\mathbf{C}\mathbf{F}(\mathbf{Q}) &=& \mathbf{F}(\mathbf{C}\mathbf{Q}\mathbf{C}^\top) = \mathbf{F}(\mathbf{Q}')
\end{eqnarray*}
for any $\mathbf{C} \in \mathrm{O}(n)$. 
It is easy to see that the previous setting of permutation invariance and equivariance is a particular case,  \marginnote{This follows from the orthogonality of permutation matrices, $\mathbf{P}^\top \mathbf{P} = \mathbf{I}$.}
which can be thought of as a trivial remeshing in which only the order of nodes is changed.

\cite{wang2019learning} showed that given an eigendecomposition of the operator $\mathbf{Q} = \mathbf{V}\boldsymbol{\Lambda}\mathbf{V}^\top$, any remeshing invariant (or equivariant) function can be expressed as 
$f(\mathbf{Q}) = f(\boldsymbol{\Lambda})$ and 
$\mathbf{F}(\mathbf{Q}) = \mathbf{V} \mathbf{F}(\boldsymbol{\Lambda})$, 
or in other words, remeshing-invariant functions {\em involve only the spectrum of} $\mathbf{Q}$. 
Indeed, functions of Laplacian eigenvalues have been proven in practice to be robust to surface discretisation and perturbation, explaining the popularity of spectral constructions based on Laplacians in computer graphics, as well as in deep learning on graph \citep{defferrard2016convolutional,levie2018cayleynets}. 
%
%
%
Since this result refers to a generic operator $\mathbf{Q}$, multiple choices are available besides the ubiquitous Laplacian -- notable examples include the Dirac \citep{liu2017dirac,kostrikov2018surface} or Steklov \citep{wang2018steklov} operators, as well as learnable parametric operators \citep{wang2019learning}.



\section{Geometric Deep Learning Models}

Having thoroughly studied various instantiations of our Geometric Deep Learning blueprint (for different choices of domain, symmetry group, and notions of locality), we are ready to discuss how enforcing these prescriptions can yield some of the most popular deep learning architectures.

Our exposition, once again, will not be in strict order of generality. We initially cover three architectures for which the implementation follows nearly-directly from our preceding discussion: convolutional neural networks (CNNs), group-equivariant CNNs, and graph neural networks (GNNs).

We will then take a closer look into variants of GNNs for cases where a graph structure is not known upfront (i.e. unordered sets), and through our discussion we will describe the popular Deep Sets and Transformer architectures as instances of GNNs.

Following our discussion on geometric graphs and meshes, we first describe equivariant message passing networks, which introduce explicit geometric symmetries into GNN computations. Then, we show ways in which our theory of geodesics and gauge symmetries can be materialised within deep learning, recovering a family of intrinsic mesh CNNs (including Geodesic CNNs, MoNet and gauge-equivariant mesh CNNs).

Lastly, we look back on the grid domain from a \emph{temporal} angle. This discussion will lead us to recurrent neural networks (RNNs). We will demonstrate a manner in which RNNs are translation equivariant over temporal grids, but also study their stability to time warping transformations. This property is highly desirable for properly handling long-range dependencies, and enforcing class invariance to such transformations yields exactly the class of gated RNNs (including popular RNN models such as the LSTM or GRU). 

While we hope the above canvasses most of the key deep learning architectures in use at the time of writing, we are well aware that novel neural network instances are proposed daily. Accordingly, rather than aiming to cover every possible architecture, we hope that the following sections are illustrative enough, to the point that the reader is able to easily categorise any future Geometric Deep Learning developments using the lens of invariances and symmetries.

\subsection{Convolutional Neural Networks}
\label{sec:cnnsec}

Convolutional Neural Networks are perhaps the earliest and most well known  example of deep learning architectures following the blueprint of Geometric Deep Learning outlined in Section~\ref{sec:gdl_blueprint}. 
%
%
%
In Section \ref{sec:grids_euclidean} we have fully characterised the class of linear and local translation equivariant operators, given by 
convolutions ${\bf C(\thetab)}{\bf x} = {\bf x} \star \thetab$ with a localised filter $\thetab$\marginnote{Recall, ${\bf C}(\boldsymbol{\theta})$ is a \emph{circulant} matrix with parameters $\boldsymbol{\theta}$.}. 
Let us first focus on scalar-valued (`single-channel' or `grayscale') discretised images,  
where 
the domain is the grid $\Omega = [H] \times [W]$ with $\mathbf{u} = (u_1, u_2)$ and ${\bf x} \in \gX(\Omega, \R)$.

Any convolution with a compactly supported filter of size $H^f \times W^f$
can be written as a linear combination of 
generators $\thetab_{1,1}, \dots, \thetab_{{H^f,W^f}}$, given for example by the unit peaks $\boldsymbol{\theta}_{vw}(u_1, u_2) = \delta(u_1 - v, u_2 - w)$. Any local linear equivariant map is thus expressible as\marginnote{Note that we usually imagine $\vec{x}$ and $\boldsymbol{\theta}_{vw}$ as 2D matrices, but in this equation, both $\vec{x}$ and $\boldsymbol{\theta}_{vw}$ have their two coordinate dimensions \emph{flattened} into one---making $\vec{x}$ a vector, and $\vec{C}(\boldsymbol{\theta}_{vw})$ a matrix.} 
\begin{equation}
    \mathbf{F}(\mathbf{x}) = \sum_{v=1}^{H^f}\sum_{w=1}^{W^f} \alpha_{vw} \mathbf{C}(\thetab_{vw})\mathbf{x}~, 
\end{equation}
which, in coordinates, corresponds to the familiar 2D convolution (see Figure \ref{fig:2dconv} for an overview):
\begin{figure}
     \centering
     \includegraphics[width=0.7\linewidth]{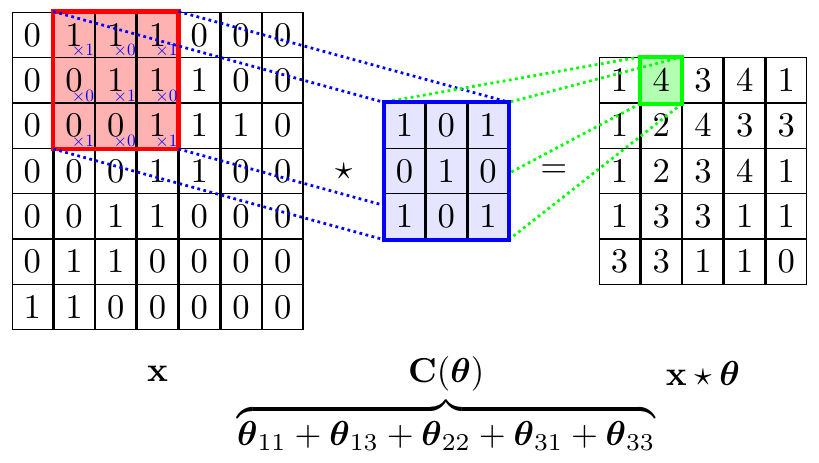}
     \caption{The process of convolving an image $\vec{x}$ with a filter $\vec{C}(\boldsymbol{\theta})$. The filter parameters $\boldsymbol{\theta}$ can be expressed as a linear combination of generators $\boldsymbol{\theta}_{vw}$.}
     \label{fig:2dconv}
\end{figure}
\begin{equation}\mathbf{F}(\mathbf{x})_{uv}= \sum_{a=1}^{H^f}\sum_{b=1}^{W^f} {\alpha}_{ab} x_{u+a, v+b}~.
\end{equation}
Other choices of the basis $\boldsymbol{\theta}_{vw}$ are also possible and will yield equivalent operations (for potentially different choices of $\alpha_{vw}$). A popular example are \emph{directional derivatives}: $\thetab_{vw}(u_1,u_2) = \delta(u_1, u_2) - \delta(u_1-v,u_2-w), (v,w) \neq (0,0) $ taken together with the local average $\thetab_{0}(u_1,u_2) = \frac{1}{H_fW_f}$. In fact, directional derivatives can be considered a grid-specific analogue of diffusion processes on graphs, which we recover if we assume each pixel to be a node connected to its immediate neighbouring pixels in the grid.

When the scalar input channel is replaced by multiple channels (e.g., RGB colours, or more generally an arbitrary number of \emph{feature maps}), the convolutional filter becomes a {\em convolutional tensor} expressing arbitrary linear combinations of input features into output feature maps. In coordinates, this can be expressed as 
\begin{equation}
\label{eq:basiccnnlayer}
\mathbf{F}(\mathbf{x})_{uvj}= \sum_{a=1}^{H^f}\sum_{b=1}^{W^f}\sum_{c=1}^M {\alpha}_{jabc} x_{u+a, v+b, c}~,~ j\in[N]~,
\end{equation}
where $M$ and $N$ are respectively the number of input and output channels. 
This basic operation encompasses a broad class of neural network architectures, which, as we will show in the next section, have had a profound impact across many areas of computer vision, signal processing, and beyond. Here, rather than dissecting the myriad of possible architectural variants of CNNs, we prefer to focus on some of the essential innovations that enabled their widespread use.

\paragraph{Efficient multiscale computation} 
As discussed in the GDL template for general symmetries, extracting translation invariant features out of the convolutional operator $\mathbf{F}$ requires a non-linear step.\marginnote{\includegraphics[width=\linewidth]{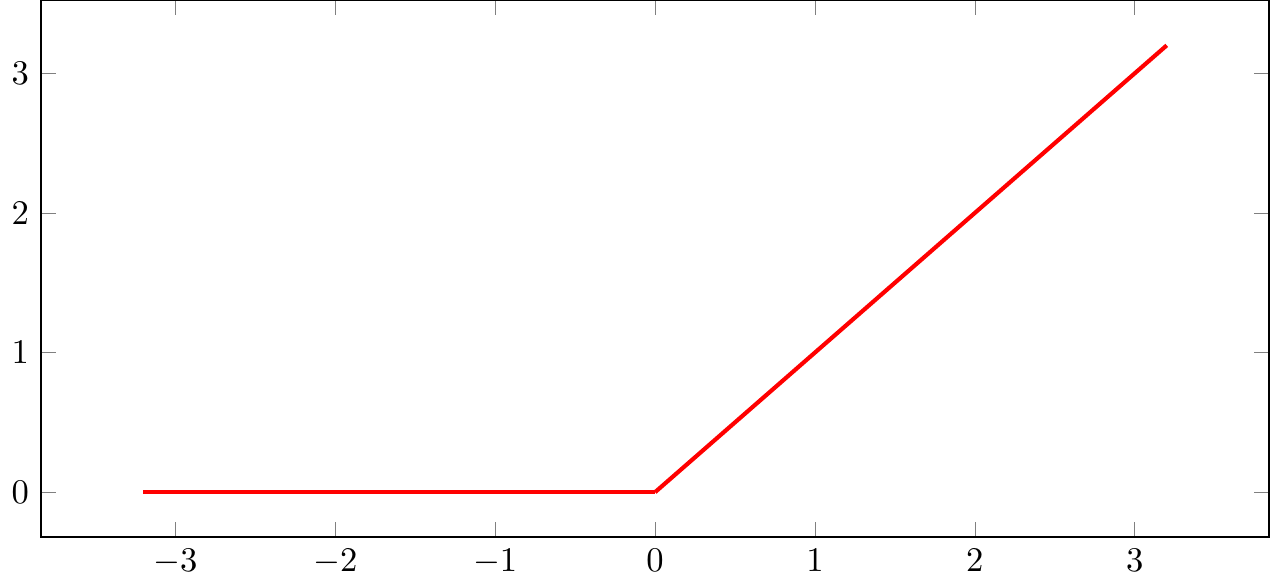}\\ ReLU, often considered a `modern' architectural choice, was already used in the Neocognitron \citep{fukushima1982neocognitron}. Rectification is equivalent to the principle of demodulation, which is fundamental in electrical engineering as the basis for many transmission protocols, such as FM radio; and also has a prominent role in models for neuronal activity.} 
Convolutional features are processed through a non-linear 
\emph{activation function} $\sigma$, acting element-wise on the input---i.e., $\sigma: \gX(\Omega) \to \gX(\Omega)$, as $\sigma(\mathbf{x})(u) = \sigma(\mathbf{x}(u))$. Perhaps the most popular example at the time of writing is the Rectified Linear Unit (ReLU): $\sigma(x) = \max(x, 0)$. 
This non-linearity effectively \emph{rectifies} the signals, pushing their energy towards lower frequencies, and enabling the computation of high-order interactions across scales by iterating the construction.

Already in the early works of \cite{fukushima1982neocognitron} and \cite{lecun1998gradient}, CNNs and similar architectures 
had a multiscale structure, where after each convolutional layer (\ref{eq:basiccnnlayer}) one performs a grid coarsening 
$\mathbf{P} : \gX(\Omega) \to \gX(\Omega')$, where the grid $\Omega'$ has coarser resolution than $\Omega$. 
This enables 
multiscale filters with effectively increasing receptive field, yet retaining a constant number of parameters per scale. 
Several signal coarsening 
strategies $\mathbf{P}$ (referred to as \emph{pooling}) may be used, the most common are 
applying a low-pass anti-aliasing filter (e.g. local average) followed by grid downsampling, or non-linear max-pooling. 
%
%

In summary, a `vanilla' CNN layer can be expressed as the composition of the basic objects already introduced in our Geometric Deep Learning blueprint:
\begin{equation}
    \mathbf{h} = \mathbf{P}(\sigma(\mathbf{F}( \mathbf{x})))~,
\end{equation}
i.e. an equivariant linear layer $\mathbf{F}$, a coarsening operation $\mathbf{P}$, and a non-linearity $\sigma$. It is also possible to perform translation invariant \emph{global} pooling operations within CNNs. Intuitively, this involves each pixel---which, after several convolutions, summarises a \emph{patch} centered around it---\emph{proposing} the final representation of the image\marginnote{CNNs which only consist of the operations mentioned in this paragraph are often dubbed ``all-convolutional''. In contrast, many CNNs \emph{flatten} the image across the spatial axes and pass them to an MLP classifier, once sufficient equivariant and coarsening layers have been applied. This loses translation invariance.}, with the ultimate choice being guided by a form of aggregation of these proposals. A popular choice here is the average function, as its outputs will retain similar magnitudes irrespective of the image size \citep{springenberg2014striving}.

\begin{figure}
    \centering
    \includegraphics[width=0.85\textwidth]{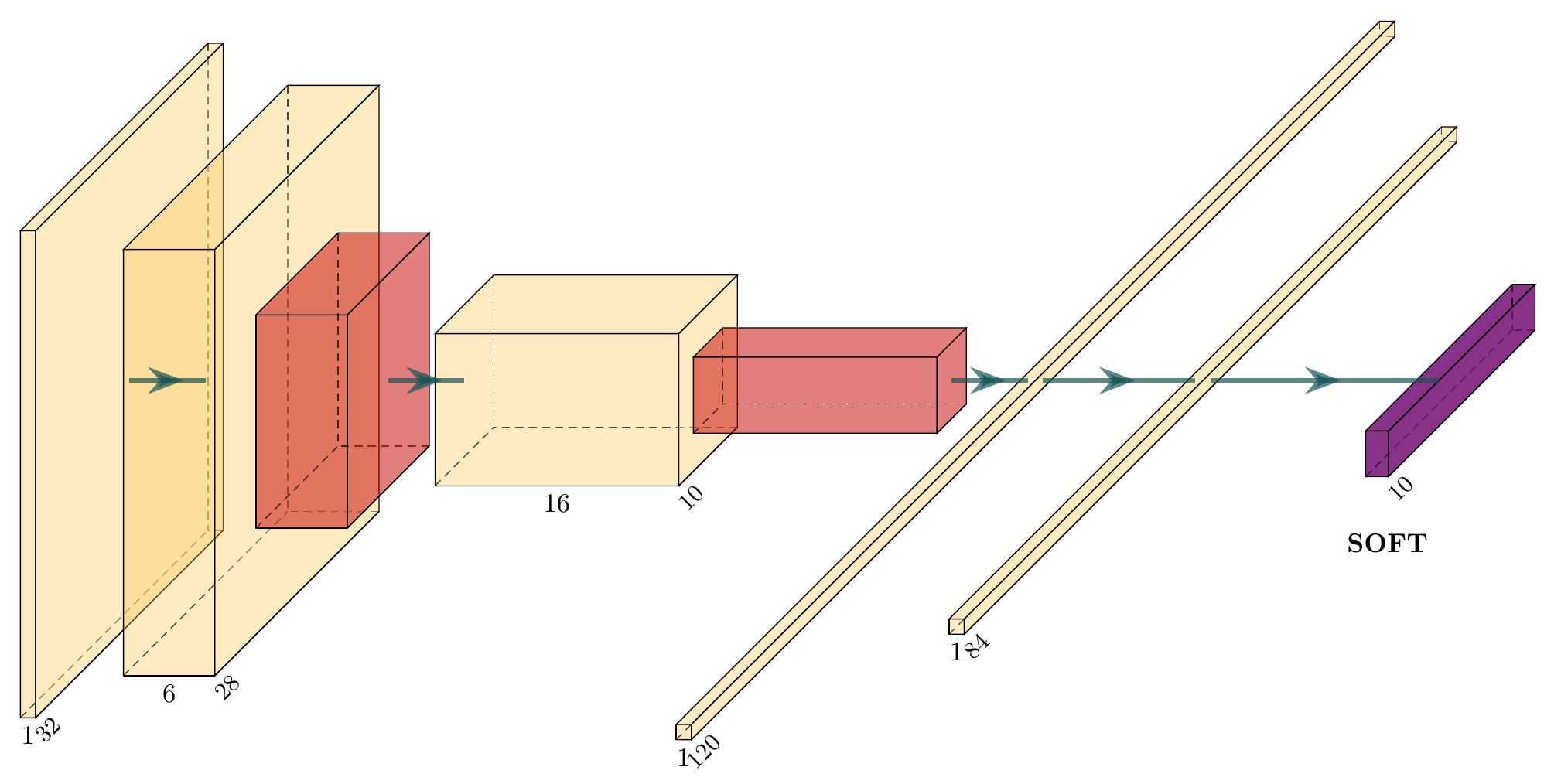}
    \includegraphics[width=0.85\textwidth]{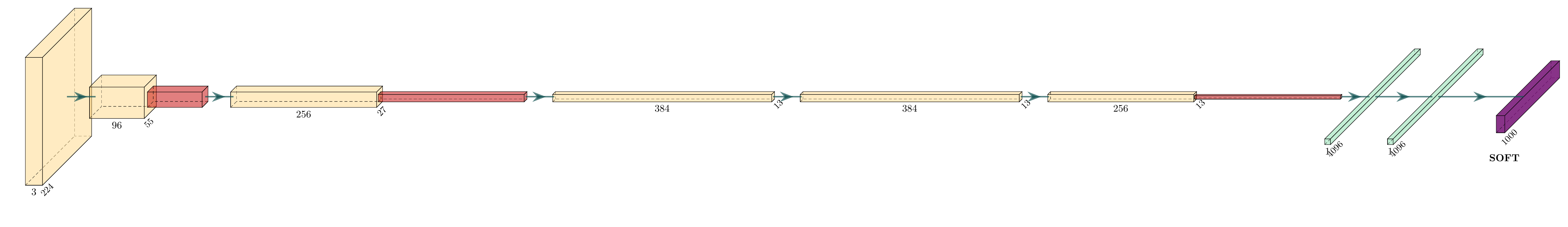}
    \includegraphics[width=0.85\textwidth]{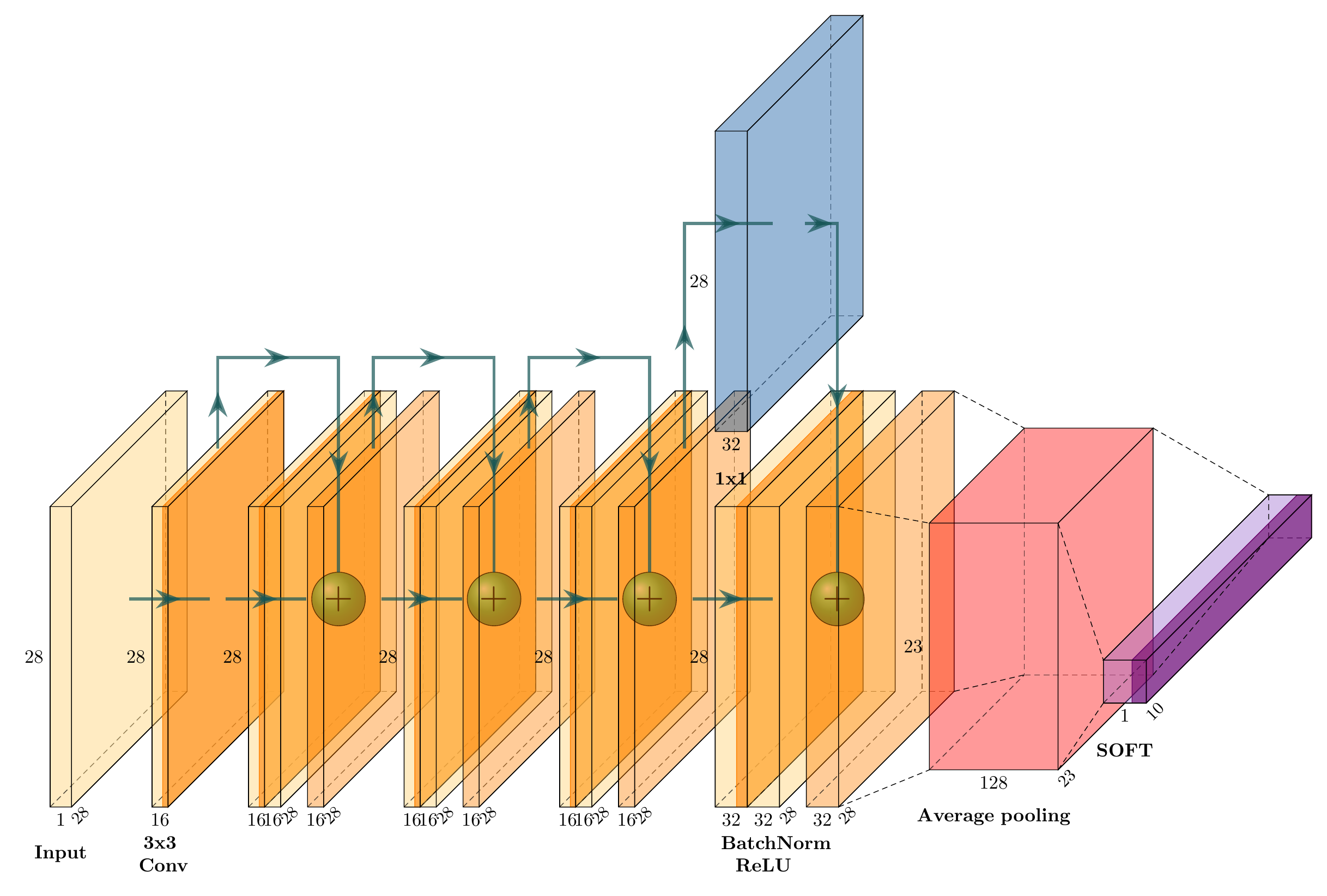}
    \includegraphics[width=0.85\textwidth]{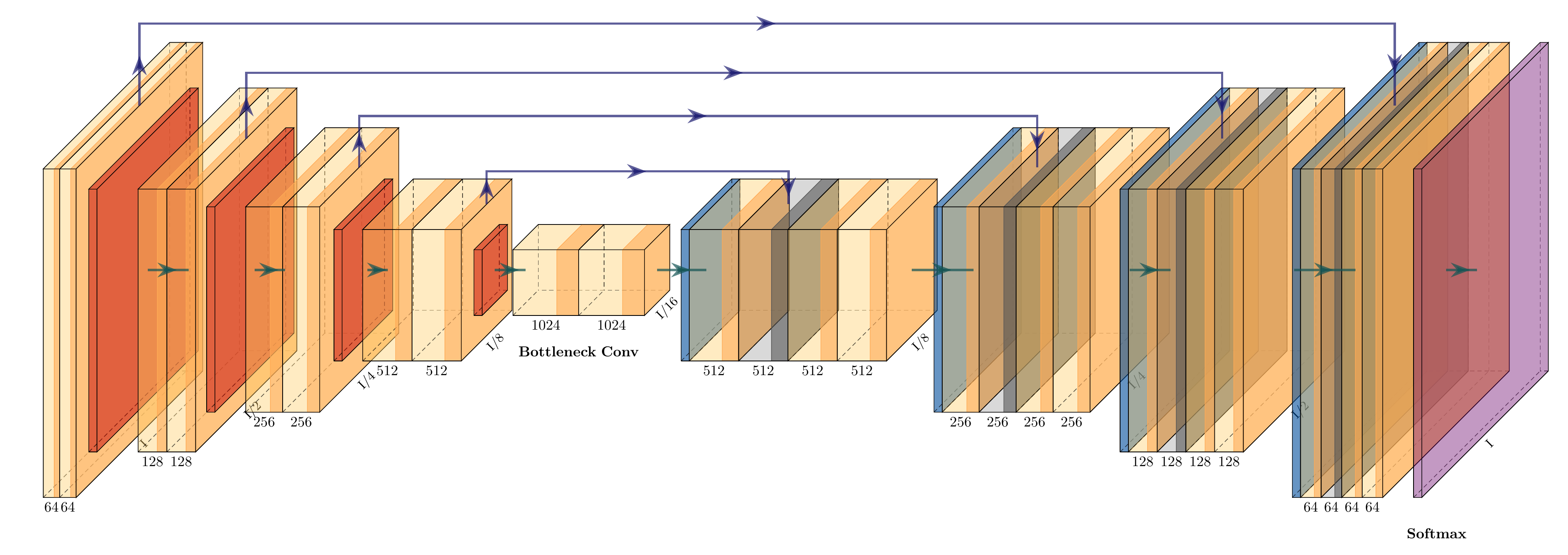}
    \caption{Prominent examples of CNN architectures. \textbf{Top-to-bottom}: LeNet \citep{lecun1998gradient}, AlexNet \citep{krizhevsky2012imagenet}, ResNet \citep{he2016deep} and U-Net \citep{ronneberger2015u}. Drawn using the PlotNeuralNet package \citep{haris2018}.}
    \label{fig:cnn_drawn_plot}
\end{figure}

Prominent examples following this CNN blueprint (some of which we will discuss next) are displayed in Figure \ref{fig:cnn_drawn_plot}.

\paragraph{Deep and Residual Networks}
A CNN architecture, in its simplest form, is therefore specified by hyperparameters $(H^f_k, W^f_k, N_k, p_k )_{k \leq K}$, with $M_{k+1} = N_k$ and $p_k=0,1$ indicating whether grid coarsening is performed or not. While all these hyperparameters are important in practice, a particularly important  question is to understand the role of depth $K$ in CNN architectures, and what are the fundamental tradeoffs involved in choosing such a key hyperparameter, especially in relation to the filter sizes ($H_k^f, W_k^f$). 

While a rigorous answer to this question is still elusive, mounting empirical evidence collected throughout the recent years suggests a favourable tradeoff towards deeper (large $K$) yet thinner (small $(H^f_k, W^f_k)$) models \marginnote{
Historically, ResNet models are predated by \emph{highway networks} \citep{srivastava2015highway}, which allow for more general \emph{gating} mechanisms to control the residual information flow.}. In this context, a crucial insight by \cite{he2016deep} was to reparametrise each convolutional layer to model a \emph{perturbation} of the previous features, rather than a generic non-linear transformation:
\begin{equation}
    \mathbf{h} = \mathbf{P}\left(\mathbf{x} + \sigma(\mathbf{F}(\mathbf{x})) \right)~.
\end{equation}
The resulting \emph{residual} networks provide several key advantages over the previous formulation. In essence, the residual parametrisation is consistent with the view that the deep network is a discretisation of an underlying continuous dynamical system, modelled as an ordinary differential equation (ODE)\marginnote{In this case, the ResNet is performing a Forward Euler discretisation of an ODE: $\dot{\mathbf{x}} = \sigma( \mathbf{F}(\mathbf{x}))$}. Crucially, learning a dynamical system by modeling its velocity turns out to be much easier than learning its position directly. In our learning setup, this translates into an optimisation landscape with more favorable geometry, leading to the ability to train much deeper architectures than was possible before. As will be discussed in future work, learning using deep neural networks defines a non-convex optimisation problem, which can be efficiently solved using gradient-descent methods under certain simplifying regimes. The key advantage of the ResNet parametrisation has been rigorously analysed in simple scenarios \citep{hardt2016identity}, and remains an active area of theoretical investigation. 
Finally, Neural ODEs \citep{chen2018neural} are a recent popular architecture that pushes the analogy with ODEs even further, by learning the parameters of the ODE $\dot{\mathbf{x}} = \sigma( \mathbf{F}(\mathbf{x}))$ directly and relying on standard numerical integration. 


\paragraph{Normalisation} 
Another important algorithmic innovation that boosted the empirical performance of CNNs significantly is the notion of \emph{normalisation}. In early models of neural activity, it was hypothesised that neurons perform some form of local `gain control', where the layer coefficients $\mathbf{x}_{k}$ are replaced by $\mathbf{\tilde{x}}_{k} = \sigma_k^{-1} \odot (\mathbf{x}_{k} - \mu_k)$. Here, $\mu_k$ and $\sigma_k$ encode the first and second-order moment information of $\mathbf{x}_k$, respectively. Further, they can be either computed globally or locally.

In the context of Deep Learning, this principle was widely adopted through the \emph{batch normalisation} layer \citep{ioffe2015batch}\marginnote{We note that normalising activations of neural networks has seen attention even before the advent of batch normalisation. See, e.g., \cite{lyu2008nonlinear}.}, followed by several variants \citep{ba2016layer,salimans2016weight,ulyanov2016instance,cooijmans2016recurrent,wu2018group}. 
Despite some attempts to rigorously explain the benefits of normalisation in terms of better conditioned optimisation landscapes \citep{santurkar2018does}, a general theory that can provide guiding principles is still missing at the time of writing.  

\paragraph{Data augmentation}
While CNNs encode the geometric priors associated with translation invariance and scale separation, they do not explicitly account for other known transformations that preserve semantic information, e.g lightning or color changes, or small rotations and dilations. A pragmatic approach to incorporate these priors with minimal architectural changes is to perform \emph{data augmentation}, where one manually performs said transformations to the input images and adds them into the training set.

Data augmentation has been empirically successful and is widely used---not only to train state-of-the-art vision architectures, but also to prop up several developments in self-supervised and causal representation learning \citep{chen2020simple,grill2020bootstrap,mitrovic2020representation}. However, it is provably sub-optimal in terms of sample complexity \citep{mei2021learning}; a more efficient strategy considers instead architectures with richer invariance groups---as we discuss next.

\subsection{Group-equivariant CNNs}

As discussed in Section \ref{sec:groups}, we can generalise the convolution operation from signals on a Euclidean space to signals on any \emph{homogeneous space} $\Omega$ acted upon by a group $\fG$\marginnote{Recall that a homogeneous space is a set $\Omega$ equipped with a transitive group action, meaning that for any $u,v \in \Omega$ there exists $\fg \in \fG$ such that $\fg. u = v$.}.
By analogy to the Euclidean convolution where a {\em translated} filter is matched with the signal, the idea of group convolution is to move the filter around the domain using the group action, e.g. by rotating and translating. 
By virtue of the \emph{transitivity} of the group action, we can move the filter to any position on $\Omega$.
In this section, we will discuss several concrete examples of the general idea of group convolution, including implementation aspects and architectural choices. 


\paragraph{Discrete group convolution}

\begin{figure}[h!]
    \centering
    \includegraphics[width=0.6\textwidth]{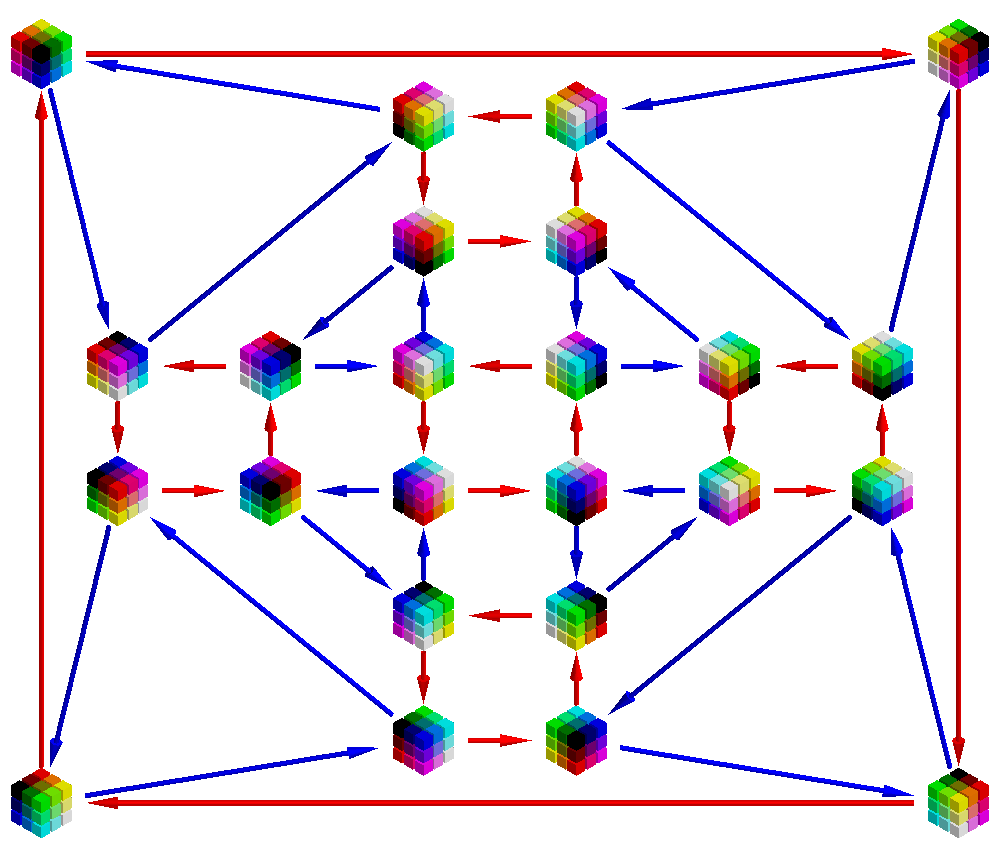}
    \caption{A $3 \times 3$ filter, rotated by all $24$ elements of the discrete rotation group $O_h$, generated by $90$-degree rotations about the vertical axis (red arrows), and $120$-degree rotations about a diagonal axis (blue arrows). 
    }
    \label{fig:cayley-diagram-Oh}
\end{figure}

We begin by considering the case where the domain $\Omega$ as well as the group $\fG$ are discrete.
As our first example, we consider medical volumetric images represented as signals of on 3D grids 
with discrete translation and rotation symmetries. 
%
The domain is the 3D cubical grid $\Omega = \Z^3$ and the images (e.g. 
MRI or CT 3D scans) are modelled as functions $x : \Z^3 \rightarrow \R$, i.e. $x \in \mathcal{X}(\Omega)$.
Although in practice such images have support on a finite cuboid $[W] \times [H] \times [D] \subset \Z^3$, we instead prefer to view them as functions on $\Z^3$ with appropriate zero padding. 
%
As our symmetry, we consider the group $\fG = \Z^3 \rtimes O_h$ of distance- and orientation-preserving transformations on $\Z^3$.
This group consists of translations ($\Z^3$) and the discrete rotations $O_h$ generated by $90$ degree rotations about the three axes (see Figure \ref{fig:cayley-diagram-Oh}).

As our second example, we consider DNA\marginnote{DNA is a biopolymer molecule 
made of four repeating units called {\em nucleotides} (Cytosine, Guanine, Adenine, and Thymine), arranged into two strands coiled around each other in a double helix, where each nucleotide occurs opposite of the complementary one ({\em base pairs} A/T and C/G). } sequences made up of four letters: C, G, A, and T. 
The sequences can be represented on the 1D grid $\Omega = \Z$ as signals $x : \Z \rightarrow \R^4$, where each letter is one-hot coded in $\R^4$. 
Naturally, we have a discrete 1D translation symmetry on the grid, but DNA sequences have an additional interesting symmetry.
This symmetry arises from the way DNA is physically embodied as a double helix, and the way it is read by the molecular machinery of the cell. 
%
Each strand of the double helix begins with what is called the $5'$-end and ends with a $3'$-end, with the $5'$ on one strand complemented by a $3'$ on the other strand.
In other words, the two strands have an opposite orientation.\marginnote{\includegraphics[width=\linewidth]{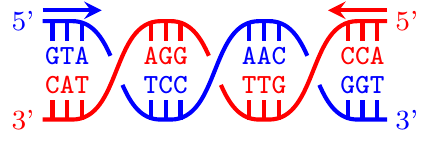}\\ A schematic of the DNA's double helix structure, with the two strands coloured in blue and red. Note how the sequences in the helices are complementary and read in reverse (from 5' to 3').}
Since the DNA molecule is always read off starting at the $5'$-end, but we do not know which one, a sequence such as ACCCTGG is equivalent to the reversed sequence with each letter replaced by its complement, CCAGGGT.
This is called {\em reverse-complement symmetry} of the letter sequence.
We thus have the two-element group $\Z_2 = \{0, 1\}$ corresponding to the identity $0$ and reverse-complement transformation $1$ (and composition $1 + 1 = 0 \mod{2}$).
The full group combines translations and reverse-complement transformations.

In our case, 
the group convolution~(\ref{eq:group-conv}) we defined in Section \ref{sec:groups} is given as 
\begin{equation}
    (x \star \theta)(\fg) = \sum_{u \in \Omega} x_u \rho(\fg) \theta_u,
\end{equation}
the inner product between the (single-channel) input signal $x$ and a filter $\theta$ transformed by $\fg \in \fG$ via $\rho(\fg) \theta_u = \theta_{\fg^{-1} u}$, and the output $x\star \theta$ is a function on $\fG$. 
Note that since $\Omega$ is discrete, we have replaced the integral from equation~(\ref{eq:group-conv}) by a sum. 

\paragraph{Transform+Convolve approach}
We will show that the group convolution can be implemented in two steps: a filter transformation step, and a translational convolution step.
The filter transformation step consists of creating rotated (or reverse-complement transformed) copies of a basic filter, while the translational convolution is the same as in standard CNNs and thus efficiently computable on hardware such as GPUs.  
To see this, note that in both of our examples we can write a general transformation $\fg \in \fG$ as a transformation $\fh \in \fH$ (e.g. a rotation or reverse-complement transformation) followed by a translation $\fk \in \Z^d$, i.e. $\fg = \fk \fh$ (with juxtaposition denoting the composition of the group elements $\fk$ and $\fh$).
By properties of the group representation, 
we have $\rho(\fg) = \rho(\fk \fh) = \rho(\fk) \rho(\fh)$.
Thus, 
\begin{equation}
    \begin{aligned}
        (x\star \theta )(\fk \fh) 
        &=
        \sum_{u \in \Omega} x_u \rho(\fk) \rho(\fh) \theta_{u} \\
        &=
        \sum_{u \in \Omega} x_u (\rho(\fh) \theta)_{u - \fk}
    \end{aligned}
\end{equation}
We recognise the last equation as the standard (planar Euclidean) convolution of the signal $x$ and the transformed filter $\rho(\fh) \theta$.
Thus, to implement group convolution for these groups, we take the canonical filter $\theta$, create transformed copies $\theta_\fh = \rho(\fh) \theta$ for each $\fh \in \fH$ (e.g. each rotation $\fh \in O_h$ or reverse-complement DNA symmetry $\fh \in \Z_2$), and then convolve $x$ with each of these filters: $(x\star \theta )(\fk \fh)  = (x\star \theta_{\fh} )(\fk)$.
For both of our examples, the symmetries act on filters by simply permuting the filter coefficients, as shown in Figure \ref{fig:cayley-diagram-Oh} for discrete rotations.
Hence, these operations can be implemented efficiently using an indexing operation with pre-computed indices.

While we defined the feature maps output by the group convolution $x \star \theta$ as functions on $\fG$, the fact that we can split $\fg$ into $\fh$ and $\fk$ means that we can also think of them as a stack of Euclidean feature maps (sometimes called \emph{orientation channels}), with one feature map per filter transformation / orientation $\fk$.
For instance, in our first example we would associate to each filter rotation (each node in Figure \ref{fig:cayley-diagram-Oh}) a feature map, which is obtained by 
convolving (in the traditional translational sense) the rotated filter. 
These feature maps can thus still be stored as a $W \times H \times C$ array, where the number of channels $C$ equals the number of independent filters times the number of transformations $\fh \in \fH$ (e.g. rotations).

As shown in Section \ref{sec:groups}, the group convolution is equivariant: $(\rho(\fg) x) \star \theta = \rho(\fg) (x \star \theta)$.
What this means in terms of orientation channels is that under the action of $\fh$, each orientation channel is transformed, and the orientation channels themselves are permuted.
For instance, if we associate one orientation channel per transformation in Figure  \ref{fig:cayley-diagram-Oh} and apply a rotation by $90$ degrees about the z-axis (corresponding to the red arrows), the feature maps will be permuted as shown by the red arrows.
This description makes it clear that a group convolutional neural network bears much similarity to a traditional CNN. 
Hence, many of the network design patterns discussed in the Section~\ref{sec:cnnsec}, such as residual networks, can be used with group convolutions as well.

\paragraph{Spherical CNNs in the Fourier domain}
For the continuous symmetry group of the sphere that we saw in Section~\ref{sec:groups}, it is possible to implement the convolution in the spectral domain, using the appropriate  Fourier transform (we remind the reader that the convolution on $\mathbb{S}^2$ is a function on $\mathrm{SO}(3)$, hence we need to define the Fourier transform on both these domains in order to implement multi-layer spherical CNNs). 
%
{\em Spherical harmonics} are an orthogonal basis on the 2D sphere,  analogous to the classical Fourier basis of complex exponential. 
On the special orthogonal group, the Fourier basis is known as the {\em Wigner D-functions}. %
In both cases, the Fourier transforms (coefficients) are computed as the inner product with the basis functions, and an analogy of the Convolution Theorem holds: one can compute the convolution in the Fourier domain as the element-wise product of the Fourier transforms. 
Furthermore, FFT-like algorithms exist for the efficient computation of Fourier transform on $\mathbb{S}^2$ and $\mathrm{SO}(3)$. We refer for further details to \cite{cohen2018spherical}.

%

\subsection{Graph Neural Networks}\label{sec:gnn-intro}



%


Graph Neural Networks (GNNs) are the realisation of our Geometric Deep Learning blueprint on graphs leveraging the properties of the permutation group. GNNs are among the most general class of deep learning architectures currently in existence, and as we will see in this text, most other deep learning architectures can be understood as a special case of the GNN with additional geometric structure.

As per our discussion in Section \ref{sec:proto-graphs}, we consider a graph to be specified with an adjacency matrix $\mathbf{A}$ and node features $\mathbf{X}$. 
We will study GNN architectures that are {\em permutation equivariant} functions $\mathbf{F}(\mathbf{X},\mathbf{A})$ constructed by applying shared {\em permutation invariant} functions $\phi(\mathbf{x}_u, \mathbf{X}_{\mathcal{N}_u})$ over local neighbourhoods. Under various guises, this local function $\phi$ can be referred to as ``diffusion'', ``propagation'', or ``message passing'', and the overall computation of such $\mathbf{F}$ as a ``GNN layer''.

\begin{figure}
    \centering
    \includegraphics[width=0.33\linewidth]{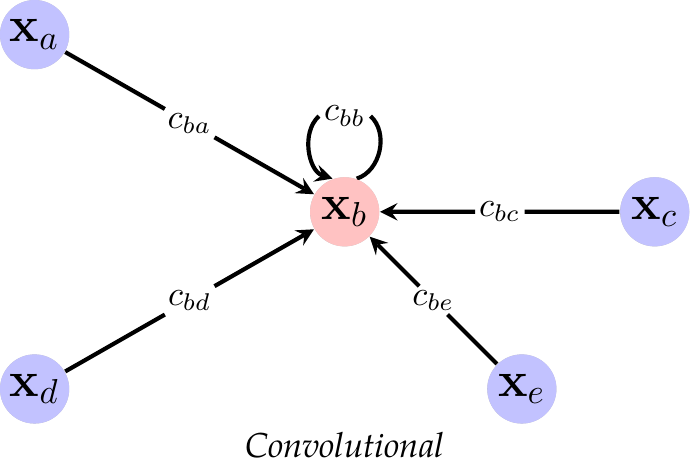}
    \hspace{-0.5em}
    \includegraphics[width=0.33\linewidth]{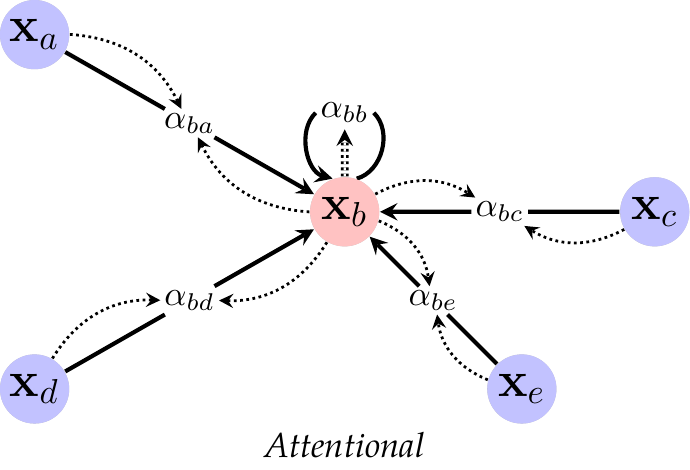}
    \hspace{-0.5em}
    \includegraphics[width=0.33\linewidth]{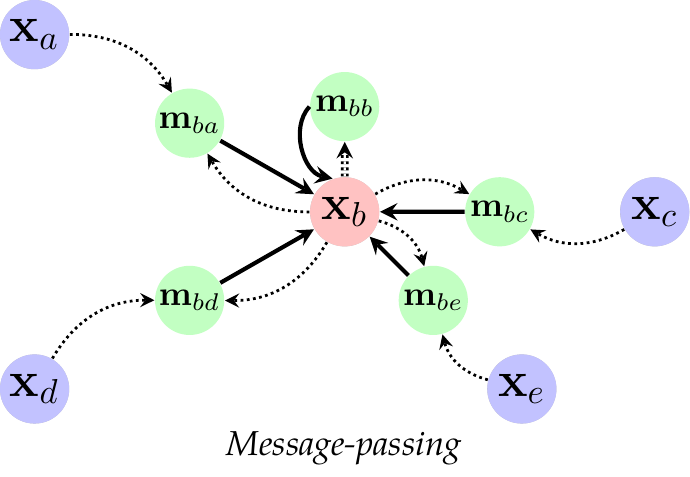}
    \caption{A visualisation of the dataflow for the three flavours of GNN layers, $g$. We use the neighbourhood of node $b$ from Figure \ref{fig:gc_gdl} to illustrate this. Left-to-right: \textbf{convolutional}, where sender node features are multiplied with a constant, $c_{uv}$; \textbf{attentional}, where this multiplier is \emph{implicitly} computed via an attention mechanism of the receiver over the sender: $\alpha_{uv}=a(\vec{x}_u, \vec{x}_v)$; and \textbf{message-passing}, where vector-based messages are computed based on both the sender and receiver: $\vec{m}_{uv}=\psi(\vec{x}_u, \vec{x}_v)$.}
    \label{fig:gc_flavours}
\end{figure}

The design and study of GNN layers is one of the most active areas of deep learning at the time of writing, making it a landscape that is challenging to navigate. Fortunately, we find that the vast majority of the literature may be derived from only three ``flavours'' of GNN layers (Figure \ref{fig:gc_flavours}), which we will present here. These flavours govern the extent to which $\phi$ transforms the neighbourhood features, allowing for varying degrees of complexity when modelling interactions across the graph.

In all three flavours, permutation invariance is ensured by \emph{aggregating} features from $\vec{X}_{\mathcal{N}_u}$ (potentially transformed, by means of some function $\psi$) with some permutation-invariant function $\bigoplus$, and then {\em updating} the features of node $u$, by means of some function $\phi$. Typically,\marginnote{Most commonly, $\psi$ and $\phi$ are learnable affine transformations with activation functions; e.g. $\psi(\vec{x}) = {\bf W}\vec{x} + \vec{b}$; $\phi(\vec{x}, \vec{z}) = \sigma\left({\bf W}\vec{x} + {\bf U}\vec{z} + \vec{b}\right)$, where ${\bf W}, {\bf U}, \vec{b}$ are learnable parameters and $\sigma$ is an activation function such as the rectified linear unit. The additional input of $\vec{x}_u$ to $\phi$ represents an optional {\em skip-connection}, which is often very useful.} $\psi$ and $\phi$ are learnable,
%
whereas $\bigoplus$ is realised as a nonparametric operation such as sum, mean, or maximum, though it can also be constructed e.g. using recurrent neural networks \citep{murphy2018janossy}.

In the \textbf{convolutional} flavour \citep{kipf2016semi,defferrard2016convolutional,wu2019simplifying}, the features of the neighbourhood nodes are  directly aggregated with fixed weights,
\begin{equation}
    \vec{h}_u = \phi\left(\vec{x}_u, \bigoplus\limits_{v\in\mathcal{N}_u}c_{uv}\psi(\vec{x}_v)\right).
\end{equation}
Here, $c_{uv}$ specifies the \emph{importance} of node $v$ to node $u$'s representation. It is a constant that often directly depends on the entries in ${\bf A}$ representing the structure of the graph. 
Note that when the aggregation operator $\bigoplus$ is chosen to be the summation, it can be considered as a linear diffusion or position-dependent linear filtering, a generalisation of convolution.\marginnote{It is worthy to note that this flavour does not express \emph{every} GNN layer that is convolutional (in the sense of commuting with the graph structure), but covers most such approaches proposed in practice. We will provide detailed discussion and extensions in future work.} 
In particular, the spectral filters we have seen in Sections~\ref{sec:manifolds} and~\ref{sec:meshes} fall under this category, as they amount to applying fixed local operators (e.g. the Laplacian matrix) to node-wise signals.


In the \textbf{attentional} flavour \citep{velickovic2018graph,monti2017geometric,zhang2018gaan}, the interactions are implicit 
\begin{equation}
    \vec{h}_u = \phi\left(\vec{x}_u, \bigoplus\limits_{v\in\mathcal{N}_u}a(\vec{x}_u, \vec{x}_v)\psi(\vec{x}_v)\right).
\end{equation}
Here, $a$ is a learnable \emph{self-attention mechanism}  that computes the importance coefficients $\alpha_{uv} = a(\vec{x}_u, \vec{x}_v)$ implicitly. They are often softmax-normalised across all neighbours. 
When $\bigoplus$ is the summation, the aggregation is still a linear combination of the neighbourhood node features, but now the weights are feature-dependent.

Finally, the \textbf{message-passing} flavour \citep{gilmer2017neural,battaglia2018relational} amounts to computing arbitrary vectors (``messages'') across edges, 
\begin{equation}
    \vec{h}_u = \phi\left(\vec{x}_u, \bigoplus\limits_{v\in\mathcal{N}_u}\psi(\vec{x}_u, \vec{x}_v)\right).
\end{equation}
Here, $\psi$ is a learnable \emph{message function}, computing $v$'s vector sent to $u$, and the aggregation can be considered as a form of message passing on the graph.

One important thing to note is a representational containment between these approaches: \emph{convolution} $\subseteq$ \emph{attention} $\subseteq$ \emph{message-passing}. Indeed, attentional GNNs can represent convolutional GNNs by an attention mechanism implemented as a look-up table $a(\vec{x}_u, \vec{x}_v) = c_{uv}$, and both convolutional and attentional GNNs are special cases of message-passing where the messages are only the sender nodes' features: $\psi(\vec{x}_u, \vec{x}_v) = c_{uv}\psi(\vec{x}_v)$ for convolutional GNNs and $\psi(\vec{x}_u, \vec{x}_v) = a(\vec{x}_u, \vec{x}_v)\psi(\vec{x}_v)$ for attentional GNNs.

This does not imply that message passing GNNs are always the most useful variant; as they have to compute vector-valued messages across edges, they are typically harder to train and require unwieldy amounts of memory. Further, on a wide range of naturally-occurring graphs, the graph's edges encode for downstream class similarity (i.e. an edge $(u, v)$ implies that $u$ and $v$ are likely to have the same output). For such graphs (often called \emph{homophilous}), convolutional aggregation across neighbourhoods is often a far better choice, both in terms of regularisation and scalability. Attentional GNNs offer a ``middle-ground'': they allow for modelling complex interactions within neighbourhoods while computing only scalar-valued quantities across the edges, making them more scalable than message-passing.

The ``three flavour'' categorisation presented here is provided with brevity in mind and inevitably neglects a wealth of nuances, insights, generalisations, and historical contexts to GNN models. Importantly, it excludes higher-dimensional GNN based on the Weisfeiler-Lehman hierarchy and spectral 
GNNs relying on the explicit computation of the graph Fourier transform. 

\subsection{Deep Sets, Transformers, and Latent Graph Inference}
\label{sec:deepset}

We close the discussion on GNNs by remarking on permutation-equivariant neural network architectures for learning representations of \emph{unordered sets}. While sets have the least structure among the domains we have discussed in this text, their importance has been recently highlighted by highly-popular architectures such as Transformers \citep{vaswani2017attention} and Deep Sets \citep{zaheer2017deep}. 
%
%
In the language of Section \ref{sec:proto-graphs}, we assume that we are given a matrix of node features, ${\bf X}$, but without any specified adjacency or ordering information between the nodes. The specific architectures will arise by deciding to what extent to model \emph{interactions} between the nodes.
 
\paragraph*{Empty edge set}

Unordered sets are provided without any additional structure or geometry whatsoever---hence, it could be argued that the most natural way to process them is to treat each set element entirely \emph{independently}. This translates to a permutation equivariant function over such inputs, which was already introduced in Section \ref{sec:proto-graphs}: a shared transformation applied to every node in isolation. Assuming the same notation as when describing GNNs (Section \ref{sec:gnn-intro}), such models can be represented as
\begin{equation*}
    \vec{h}_u = \psi(\vec{x}_u),
\end{equation*}
where $\psi$ is a learnable transformation. It may be observed that this is a special case of a convolutional GNN with $\mathcal{N}_u = \{u\}$---or, equivalently, ${\bf A} = {\bf I}$. Such an architecture is commonly referred to as Deep Sets, in recognition of the work of \cite{zaheer2017deep} that have theoretically proved several universal-approximation properties of such architectures. It should be noted that the need to process unordered sets commonly arises in computer vision and graphics when dealing with \emph{point clouds}; therein, such models are known as PointNets \citep{qi2017pointnet}.

\paragraph*{Complete edge set}

While assuming an empty edge set is a very efficient construct for building functions over unordered sets, often we would expect that elements of the set exhibit some form of relational structure---i.e., that there exists a \emph{latent graph} between the nodes. Setting ${\bf A} = {\bf I}$ discards any such structure, and may yield suboptimal performance.
Conversely, we could assume that, in absence of any other prior knowledge, we cannot upfront exclude \emph{any} possible links between nodes. In this approach we assume the \emph{complete} graph, ${\bf A} = {\bf 1}{\bf 1}^\top$; equivalently, $\mathcal{N}_u = \mathcal{V}$.
As we do not assume access to any coefficients of interaction, running {\em convolutional}-type GNNs over such a graph would amount to:
\begin{equation*}
\vec{h}_u = \phi\left(\vec{x}_u, \bigoplus_{v\in \mathcal{V}} \psi(\vec{x}_v)\right),
\end{equation*}
where the second input, $\bigoplus_{v\in \mathcal{V}} \psi(\vec{x}_v)$ is \emph{identical} for all nodes $u$\marginnote{This is a direct consequence of the permutation invariance of $\bigoplus$.}, and as such makes the model equivalently expressive to ignoring that input altogether; i.e. the ${\bf A} = {\bf I}$ case mentioned above. 

This motivates the use of a more expressive GNN flavour, the {\em attentional}, 
\begin{equation}
    \vec{h}_u = \phi\left(\vec{x}_u, \bigoplus_{v\in \mathcal{V}} a(\vec{x}_u, \vec{x}_v)\psi(\vec{x}_v)\right)
\end{equation}
which yields the \emph{self-attention} operator, the core of the Transformer architecture \citep{vaswani2017attention}. Assuming some kind of normalisation over the attentional coefficients (e.g. softmax), we can constrain all the scalars $a(\vec{x}_u, \vec{x}_v)$ to be in the range $[0, 1]$; as such, we can think of self-attention as inferring a \emph{soft adjacency matrix}, $a_{uv} = a(\vec{x}_u, \vec{x}_v)$, as a byproduct of gradient-based optimisation for some downstream task.

The above perspective means that we can pose Transformers exactly as attentional GNNs over a complete graph \citep{joshi2020transformers}.\marginnote{It is also appropriate to apply the message-passing flavour. 
While popular for physics simulations and relational reasoning (e.g. \cite{battaglia2016interaction,santoro2017simple}), they have not been as widely used as Transformers. This is likely due to the memory issues associated with computing vector messages over a complete graph, or the fact that vector-based messages are less interpretable than the ``soft adjacency'' provided by self-attention.} However, this is in apparent conflict with Transformers being initially proposed for modelling \emph{sequences}---the representations of $\vec{h}_u$ should be mindful of node $u$'s \emph{position} in the sequence, which complete-graph aggregation would ignore. Transformers address this issue by introducing \emph{positional encodings}: the node features $\vec{x}_u$ are augmented to encode node $u$'s position in the sequence, typically as samples from a sine wave whose frequency is dependent on $u$. 

On graphs, where no natural ordering of nodes exists, multiple alternatives were proposed to such positional encodings. While we defer discussing these alternatives for later, we note that one promising direction involves a realisation that the positional encodings used in Transformers can be directly related to the discrete Fourier transform (DFT), and hence to the eigenvectors of the graph Laplacian of a ``circular grid''. Hence, Transformers' positional encodings are implicitly representing our assumption that input nodes are connected in a grid. For more general graph structures, one may simply use the Laplacian eigenvectors of the (assumed) graph---an observation exploited by \citet{dwivedi2020generalization} within their empirically powerful Graph Transformer model.

\paragraph*{Inferred edge set}

Finally, one can try to learn the latent relational structure, leading to some general ${\bf A}$ that is neither ${\bf I}$ nor ${\bf 1}{\bf 1}^\top$. The problem of inferring a latent adjacency matrix ${\bf A}$ for a GNN to use (often called \emph{latent graph inference}) is of high interest for graph representation learning. This is due to the fact that assuming ${\bf A} = {\bf I}$ may be representationally inferior, and ${\bf A} = {\bf 1}{\bf 1}^\top$ may be challenging to implement due to memory requirements and large neighbourhoods to aggregate over. Additionally, it is closest to the ``true'' problem: inferring an adjacency matrix ${\bf A}$ implies detecting useful structure between the rows of ${\bf X}$, which may then help formulate hypotheses such as causal relations between variables.

Unfortunately, such a framing necessarily induces a step-up in modelling complexity. Specifically, it requires properly balancing a structure learning objective (which is \emph{discrete}, and hence challenging for gradient-based optimisation) with any downstream task the graph is used for. This makes latent graph inference a highly challenging and intricate problem. 

\subsection{Equivariant Message Passing Networks}

In many applications of Graph Neural Networks, node features (or parts thereof) are not just arbitrary vectors but \emph{coordinates} of geometric entities. This is the case, for example, when dealing with molecular graphs: the nodes representing atoms may contain information about the atom type as well as its 3D spatial coordinates. It is desirable to process the latter part of the features in a manner that would transform in the same way as the molecule is transformed in space, in other words, be equivariant to the Euclidean group $\mathrm{E}(3)$ of rigid motions (rotations, translations, and reflections) in addition to the standard permutation equivariance discussed before. 

To set the stage for our (slightly simplified) analysis, we will make a distinction between node {\em features} $\mathbf{f}_u \in \mathbb{R}^d$ and node {\em spatial coordinates} $\mathbf{x}_u\in \mathbb{R}^3$; the latter are endowed with Euclidean symmetry structure. In this setting, an equivariant layer explicitly transforms these two inputs separately, yielding modified node features $\vec{f}'_u$ and coordinates $\vec{x}'_u$.

We can now state our desirable equivariance property, following the Geometric Deep Learning blueprint. If the spatial component of the input is transformed by $\fg\in \mathrm{E}(3)$ (represented as $\rho(\fg)\mathbf{x} = \mathbf{R}\mathbf{x} + \mathbf{b}$, where $\mathbf{R}$ is an orthogonal matrix modeling rotations and reflections, and $\mathbf{b}$ is a translation vector), the spatial component of the output transforms in the same way (as $\mathbf{x}'_u \mapsto \mathbf{R}\mathbf{x}'_u + \mathbf{b}$), whereas $\mathbf{f}'_u$ remains invariant.

Much like the space of permutation equivariant functions we discussed before in the context of general graphs, there exists a vast amount of $\mathrm{E}(3)$-equivariant layers that would satisfy the constraints above---but not all of these layers would be geometrically stable, or easy to implement. In fact, the space of practically useful equivariant layers may well be easily described by a simple categorisation, not unlike our ``three flavours'' of spatial GNN layers.
One elegant solution was suggested by \cite{satorras2021n} in the form of {\em equivariant message passing}. Their model operates as follows:  
%
\begin{eqnarray*}
    \vec{f}'_u &=& \phi\left(\vec{f}_u, \bigoplus\limits_{v\in\mathcal{N}_u}\psi_\mathrm{f}(\vec{f}_u, \vec{f}_v, \| \vec{x}_u - \vec{x}_v \|^2)\right), \\
    \vec{x}'_u &=& \mathbf{x}_u + \sum_{v\neq u} (\vec{x}_u - \vec{x}_v) \psi_\mathrm{c}(\vec{f}_u, \vec{f}_v, \| \vec{x}_u - \vec{x}_v \|^2)
\end{eqnarray*}
where $\psi_\mathrm{f}$ and $\psi_\mathrm{c}$ are two distinct (learnable) functions. It can be shown that such an aggregation is equivariant under Euclidean transformations of the spatial coordinates. This is due to the fact that the only dependence of $\mathbf{f}'_u$ on $\mathbf{x}_u$ is through the distances $\| \vec{x}_u - \vec{x}_v \|^2$, and the action of $\mathrm{E}(3)$ necessarily leaves distances between nodes unchanged. Further, the computations of such a layer can be seen as a particular instance of the ``message-passing'' GNN flavour, hence they are efficient to implement.

To summarise, in contrast to ordinary GNNs, \cite{satorras2021n} enable the correct treatment of `coordinates' for each point in the graph. They are now treated as a member of the $\mathrm{E}(3)$ group, which means the network outputs behave correctly under rotations, reflections and translations of the input.\marginnote{\includegraphics[width=\linewidth]{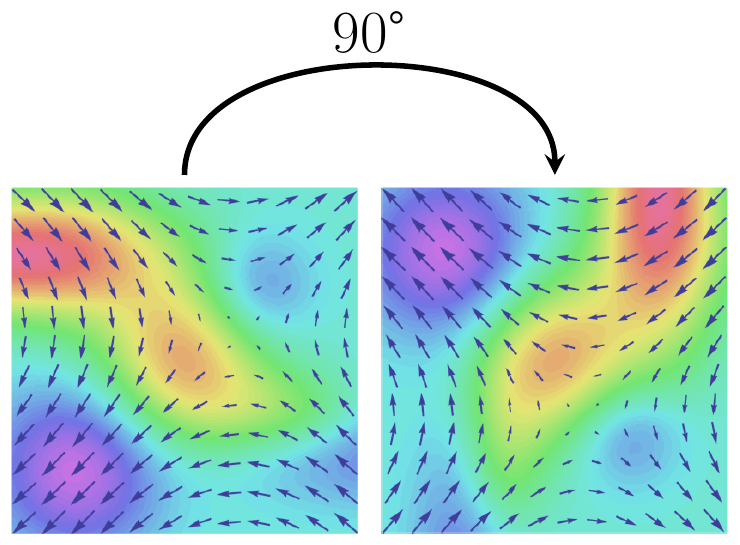}\\ While scalar features (heatmap) do not change under rotations, vector features (arrows) may change direction. The simple $\mathrm{E}(3)$ equivariant GNN given before does not take this into account.} The features, $\vec{f}_u$, however, are treated in a channel-wise manner and still assumed to be \emph{scalars} that do not change under these transformations. 
This limits the type of spatial information that can be captured within such a framework. For example, it may be desirable for some features to be encoded as \emph{vectors}---e.g. point velocities---which \emph{should} change direction under such transformations. 
\cite{satorras2021n} partially alleviate this issue by introducing the concept of velocities in one variant of their architecture. Velocities are a 3D vector property of each point which rotates appropriately. However, this is only a small subspace of the general representations that could be learned with an $\mathrm{E}(3)$ equivariant network. In general, node features may encode \emph{tensors} of arbitrary dimensionality that would still transform according to $\mathrm{E}(3)$ in a well-defined manner.

Hence, while the architecture discussed above already presents an elegant equivariant solution for many practical input representations, in some cases it may be desirable to explore a broader collection of functions that satisfy the equivariance property. Existing methods dealing with such settings can be categorised into two classes: \emph{irreducible representations} (of which the previously mentioned layer is a simplified instance) and \emph{regular representations}. We briefly survey them here, leaving detailed discussion to future work.

\paragraph{Irreducible representations}
Irreducible representations build on the finding that all elements of the roto-translation group can be brought into an irreducible form: a vector that is rotated by a block diagonal matrix. Crucially, each of those blocks is a \emph{Wigner D-matrix} (the aforementioned Fourier basis for Spherical CNNs). Approaches under this umbrella map from one set of irreducible representations to another using equivariant kernels. To find the full set of equivariant mappings, one can then directly solve the equivariance constraint over these kernels. The solutions form a linear combination of equivariant basis matrices derived by {\em Clebsch-Gordan matrices} and the spherical harmonics.

Early examples of the irreducible representations approach include Tensor Field Networks \citep{thomas2018tensor} and 3D Steerable CNNs \citep{weiler20183d}, both convolutional models operating on point clouds. The $\mathrm{SE}(3)$-Transformer of \citet{fuchs2020se} extends this framework to the graph domain, using an attentional layer rather than convolutional. Further, while our discussion focused on the special case solution of \citet{satorras2021n}, we note that the motivation for rotation or translation equivariant predictions over graphs had historically been explored in other fields, including architectures such as Dynamic Graph CNN \citep{wang2019dynamic} for point clouds and efficient message passing models for quantum chemistry, such as SchNet \citep{schutt2018schnet} and DimeNet \citep{klicpera2020directional}.

\paragraph{Regular representations}
While the approach of irreducible representations is attractive, it requires directly reasoning about the underlying group representations, which may be tedious, and only applicable to groups that are compact. Regular representation approaches are more general, but come with an additional computational burden -- for exact equivariance they require storing copies of latent feature embeddings for \emph{all} group elements\marginnote{This approach was, in fact, pioneered by the group convolutional neural networks we presented in previous sections.}. 

One promising approach in this space aims to observe equivariance to \emph{Lie groups}---through definitions of exponential and logarithmic maps---with the promise of rapid prototyping across various symmetry groups. While Lie groups are out of scope for this section, we refer the reader to two recent successful instances of this direction: the LieConv of \citet{finzi2020generalizing}, and the LieTransformer of \citet{hutchinson2020lietransformer}.

The approaches covered in this section represent popular ways of processing data on geometric graphs in an way that is explicitly equivariant to the underlying geometry. As discussed in Section~\ref{sec:meshes}, \emph{meshes} are a special instance of geometric graphs 
that can be understood as discretisations of continuous surfaces. 
We will study mesh-specific equivariant neural networks next.

\subsection{Intrinsic Mesh CNNs}
Meshes, in particular, triangular ones, are the `bread and butter' of computer graphics and perhaps the most common way of modeling 3D objects. The remarkable success of deep learning in general and CNNs in computer vision in particular has lead to a keen interest in the graphics and geometry processing  community around the mid-2010s\marginnote{
\includegraphics[width=0.9\linewidth]{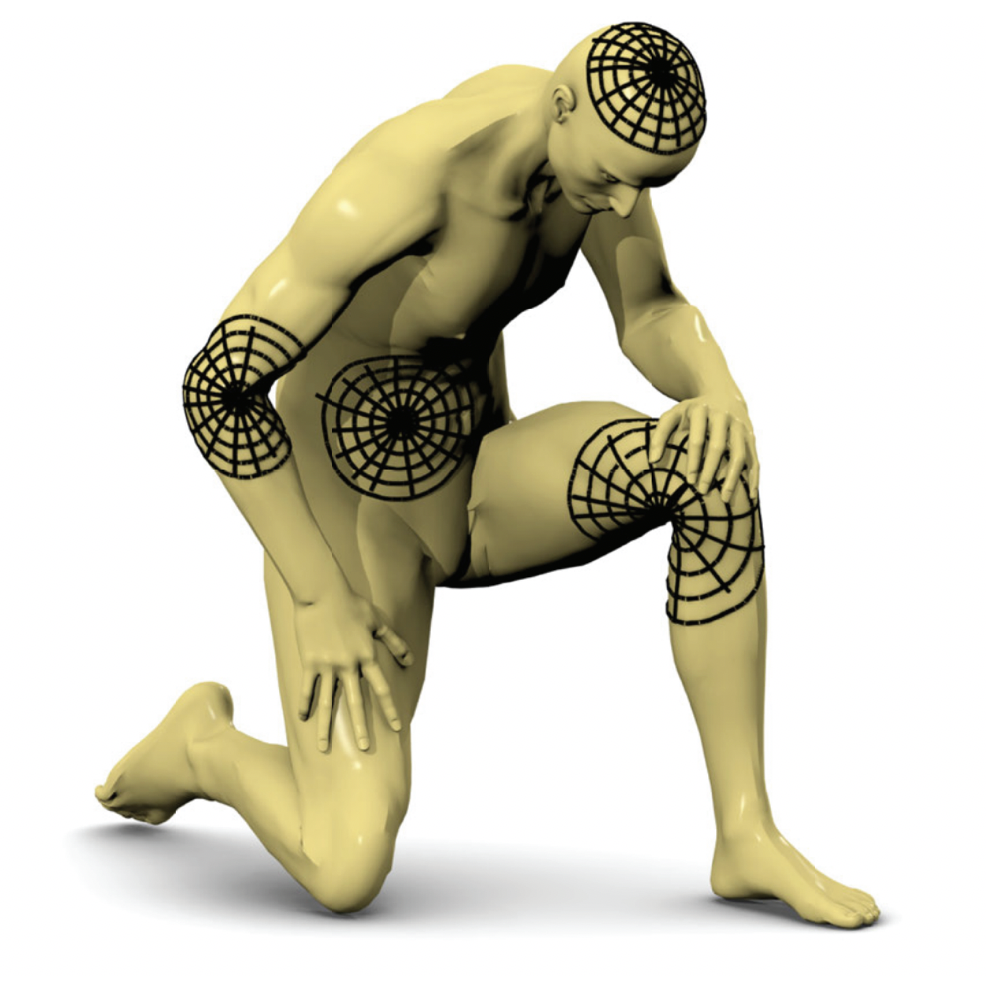}
Examples of geodesic patches. In order for the resulting patch to be a topological disk, its radius $R$ must be smaller than the injectivity radius. } to construct similar architectures for mesh data.

\paragraph{Geodesic patches}
Most of the architectures for deep learning on meshes  implement convolutional filters of the form~(\ref{eqn:conv_exp}) by 
discretising or approximating the exponential map and expressing the filter in a coordinate system of the tangent plane. 
Shooting a geodesic $\gamma:[0,T]\rightarrow \Omega$ from a point $u=\gamma(0)$ to nearby point $v=\gamma(T)$ defines a local system of {\em geodesic polar coordinates} $(r(u,v),\vartheta(u,v))$ where $r$ is the geodesic distance between $u$ and $v$ (length of the geodesic $\gamma$) and $\vartheta$ is the angle between $\gamma'(0)$ and some local reference direction. 
%
This allows to define a {\em geodesic patch} $x(u,r,\vartheta) = x(\exp_u \tilde{\omega}(r,\vartheta))$, where $\tilde{\omega}_u: [0,R]\times [0,2\pi) \rightarrow T_u\Omega$ is the local polar frame.

On a surface\marginnote{
\includegraphics[width=0.9\linewidth]{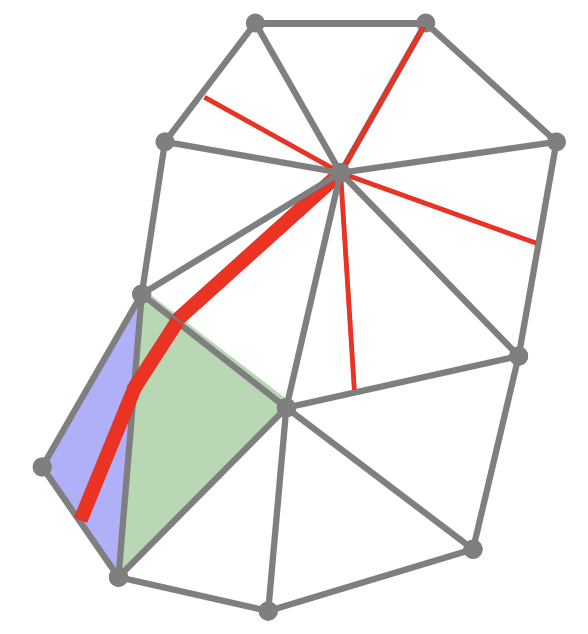}
Construction of discrete geodesics on a mesh. } discretised as a mesh, a geodesic is a poly-line that traverses the triangular faces. Traditionally, geodesics have been computed using the Fast Marching algorithm \cite{kimmel1998computing}, an efficient numerical approximation of a nonlinear PDE called the {\em eikonal equation} encountered in physical models of wave propagation in a medium. This scheme was adapted by \cite{kokkinos2012intrinsic} for the computation of local geodesic patches and later reused by \cite{masci2015geodesic} for the  construction of {\em Geodesic CNNs}, the first intrinsic CNN-like architectures on meshes.

\paragraph{Isotropic filters}
Importantly, in the definition of the geodesic patch we have ambiguity in the choice of the reference direction and the patch orientation. This is exactly the ambiguity of the choice of the gauge, and our local system of coordinates is defined up to arbitrary rotation (or a shift in the angular coordinate, $x(u,r,\vartheta+\vartheta_0)$), which can be different at every node. 
Perhaps the most straightforward solution is to use isotropic filters of the form $\theta(r)$ that perform a direction-independent aggregation of the neighbour features,  
$$
(x\star \theta)(u) = \int_0^R \int_0^{2\pi} x(u,r,\vartheta) \theta(r) \mathrm{d}r \mathrm{d}\vartheta.
$$
Spectral filters discussed in Sections~\ref{sec:manifolds}--\ref{sec:meshes} fall under this category: they are based on the Laplacian operator, which is isotropic. 
Such an approach, however, discards important directional information, and might fail to extract edge-like features.

\paragraph{Fixed gauge}
An alternative, to which we have already alluded in Section~\ref{sec:manifolds}, is to {\em fix some gauge}.  \cite{monti2017geometric} used the principal curvature directions: while this choice is not intrinsic and may ambiguous at flat points (where curvature vanishes) or uniform curvature (such as on a perfect sphere), the authors 
showed that it is reasonable for dealing with deformable human body shapes, which are approximately piecewise-rigid. 
Later works, e.g. \cite{melzi2019gframes}, showed reliable intrinsic construction of gauges on meshes, computed as  (intrinsic) gradients of intrinsic functions. While such tangent fields might have singularities (i.e., vanish at some points), the overall procedure is very robust to noise and remeshing.

\paragraph{Angular pooling}
Another approach, referred to as {\em angular max pooling}, was used by \cite{masci2015geodesic}. In this case, the filter $\theta(r,\vartheta)$ is anisotropic, but its matching with the function is performed over {\em all the possible rotations}, which are then aggregated:
$$
(x\star \theta)(u) = \max_{\vartheta_0 \in [0,2\pi) } \,\, \int_0^R \int_0^{2\pi} x(u,r,\vartheta) \theta(r,\vartheta+\vartheta_0) \mathrm{d}r \mathrm{d}\vartheta.
$$
Conceptually, this can be visualised as correlating geodesic patches with a rotating filter and collecting the strongest responses.

\begin{figure}[h!]
    \centering
    \includegraphics[width=\linewidth]{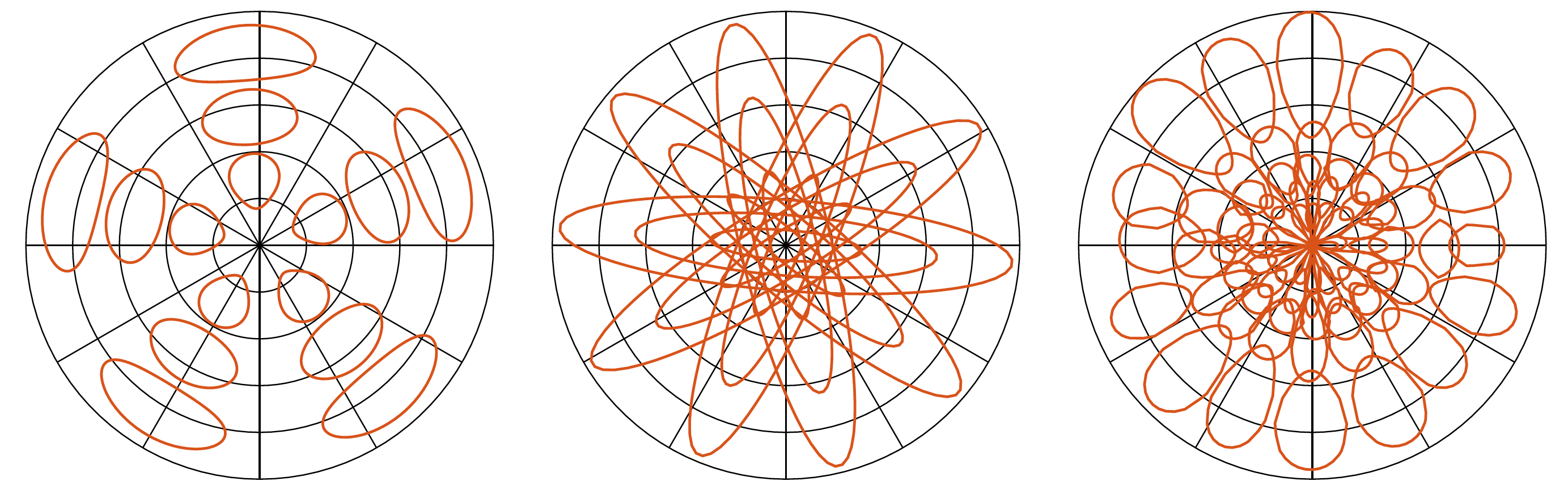}
    \caption{Left-to-right: examples of patch operators used in Geodesic CNN  \citep{masci2015geodesic}, Anisotropic CNN \citep{boscaini2016anisotropic} and MoNet \citep{monti2017geometric}, with the level sets of the weighting functions $w_k(r,\vartheta)$ shown in red. 
    }
    \label{fig:patches}
\end{figure}

On meshes, the continuous integrals can be discretised using a 
construction referred to as {\em patch operators} \citep{masci2015geodesic}. In a geodesic patch around node $u$, the neighbour nodes $\mathcal{N}_u$,\marginnote{Typically multi-hop neighbours are used. } represented in the local polar coordinates as $(r_{uv}, \vartheta_{uv})$, are weighted by a set of weighting functions $w_1(r,\vartheta), \hdots, w_K(r,\vartheta)$ (shown in Figure~\ref{fig:patches} and acting as `soft pixels') and aggregated,  
$$
(x\star \theta)_u = \frac{
\sum_{k=1}^K w_{k} \sum_{v\in \mathcal{N}_u} (r_{uv},\vartheta_{uv}) x_v \, \theta_k
}
{
\sum_{k=1}^K w_{k} \sum_{v\in \mathcal{N}_u} (r_{uv},\vartheta_{uv}) \theta_k
}
$$
  (here 
  $\theta_1,\hdots, \theta_K$ are the learnable coefficients of the filter). Multi-channel features are treated channel-wise, with a family of appropriate filters. 
  \cite{masci2015geodesic,boscaini2016learning} used pre-defined weighting functions $w$, while \cite{monti2017geometric} further allowed them to be learnable. 

\paragraph{Gauge-equivariant filters}
Both isotropic filters and angular max pooling lead to features that are {\em invariant} to gauge transformations; they transform according to the trivial representation $\rho(\fg) = 1$ (where $\fg \in \mathrm{SO}(2)$ is a rotation of the local coordinate frame).
This point of view suggests another approach, proposed by  \cite{cohen2019gauge, deHaan2020gauge} and discussed in Section~\ref{sec:gauges}, where the features computed by the network are associated with an arbitrary representation $\rho$ of the structure group $\fG$ (e.g. $\mathrm{SO(2)}$ or $\mathrm{O(2)}$ of rotations or rotations+reflections of the coordinate frame, respectively). 
%
Tangent vectors transform according to the standard representation $\rho(\fg) = \fg$.  
As another example, the feature vector obtained by matching $n$ rotated copies of the same filter transforms by cyclic shifts under rotations of the gauge; this is known as the regular representation of the cyclic group $C_n$.

As discussed in Section \ref{sec:gauges}, when dealing with such geometric features (associated to a non-trivial representation), we must first parallel transport them to the same vector space before applying the filter. 
On a mesh, this can be implemented via the following message passing mechanism described by  \cite{deHaan2020gauge}.
Let 
$\mathbf{x}_u \in \R^d$ be a $d$-dimensional input feature at mesh node $u$.
This feature is expressed relative to an (arbitrary) choice of gauge at $u$, and is assumed to transform according to a representation $\rho_{\textup{in}}$ of $\fG = \SO{2}$ under rotations of the gauge.
Similarly, the output features $\mathbf{h}_u$ 
of the mesh convolution are $d'$ dimensional and should transform according to $\rho_{\textup{out}}$ (which can be chosen at will by the network designer).


By analogy to Graph Neural Networks, we can 
implement the gauge-equivariant convolution~(\ref{eqn:gauge_eq_conv}) on meshes by sending messages from the neighbours $\mathcal{N}_u$ of $u$ (and from $u$ itself) to $u$:
\begin{equation}
    \mathbf{h}_u = \boldsymbol{\Theta}_{\textup{self}} \; \mathbf{x}_u + \sum_{v \in \mathcal{N}_u} \boldsymbol{\Theta}_{\textup{neigh}}(\vartheta_{uv}) \rho(\fg_{v \rightarrow u}) \mathbf{x}_v,
\end{equation}
where $\boldsymbol{\Theta}_{\textup{self}}, \boldsymbol{\Theta}_{\textup{neigh}}(\vartheta_{uv}) \in \R^{d' \times d}$ are learned filter matrices. 
The structure group element $\fg_{v \rightarrow u} \in \SO{2}$ denotes the effect of parallel transport from $v$ to $u$, expressed relative to the gauges at $u$ and $v$, and can be precomputed for each mesh. Its action is encoded by a \emph{transporter matrix} $\rho(\fg_{v\rightarrow u})\in\mathbb{R}^{d\times d}$.\marginnote{Note that $d$ is the feature dimension and is not necessarily equal to 2, the dimension of the mesh. }
The matrix $\boldsymbol{\Theta}_{\textup{neigh}}(\vartheta_{uv})$ depends on the angle $\vartheta_{uv}$ of the neighbour $v$ to the reference direction (e.g. first axis of the frame) at $u$, so this kernel is anisotropic: different neighbours are treated differently.

As explained in Section \ref{sec:gauges}, 
for $\mathbf{h}(u)$ to be a well-defined geometric quantity, it should transform as $\mathbf{h}(u) \mapsto \rho_{\textup{out}}(\fg^{-1}(u)) \mathbf{h}(u)$ under gauge transformations.
This will be the case when $\boldsymbol{\Theta}_{\textup{self}} \rho_{\textup{in}}(\vartheta) = \rho_{\textup{out}}(\vartheta) \boldsymbol{\Theta}_{\textup{self}}$ for all $\vartheta \in \SO{2}$,\marginnote{Here we abuse the notation, identifying 2D rotations with angles $\vartheta$. } and $\boldsymbol{\Theta}_{\textup{neigh}}(\vartheta_{uv} - \vartheta) \rho_{\textup{in}}(\vartheta) = \rho_{\textup{out}}(\vartheta) \boldsymbol{\Theta}_{\textup{neigh}}(\vartheta_{uv})$.
Since these constraints are linear, the space of matrices $\boldsymbol{\Theta}_{\textup{self}}$ and matrix-valued functions $\boldsymbol{\Theta}_{\textup{neigh}}$ satisfying these constraints is a linear subspace, and so we can parameterise them as a linear combination of basis kernels with learnable coefficients: $\boldsymbol{\Theta}_{\textup{self}} = \sum_i \alpha_i \boldsymbol{\Theta}_{\textup{self}}^i$ and 
$\boldsymbol{\Theta}_{\textup{neigh}} = \sum_i \beta_i \boldsymbol{\Theta}_{\textup{neigh}}^i$.

\subsection{Recurrent Neural Networks}

Our discussion has thus far always assumed the inputs to be solely \emph{spatial} across a given domain. However, in many common use cases, the inputs can also be considered \emph{sequential} (e.g. video, text or speech). In this case, we assume that the input consists of arbitrarily many \emph{steps}, wherein at each step $t$ we are provided with an input signal, which we represent as $\vec{X}^{(t)} \in \mathcal{X}(\Omega^{(t)})$.\marginnote{Whether the domain is considered static or dynamic concerns \emph{time scales}: e.g., a road network \emph{does} change over time (as new roads are built and old ones are demolished), but significantly slower compared to the flow of traffic. Similarly, in social networks, changes in engagement (e.g. Twitter users re-tweeting a tweet) happen at a much higher frequency than changes in the follow graph.}

While in general the domain can evolve in time together with the signals on it, it is typically assumed that the domain is kept fixed across all the $t$, i.e. $\Omega^{(t)} = \Omega$. Here, we will exclusively focus on this case, but note that exceptions are common. Social networks are an example where one often has to account for the domain changing through time, as new links are regularly created as well as erased. The domain in this  setting is often referred to as a \emph{dynamic graph} \citep{xu2020inductive,rossi2020temporal}.

Often, the individual $\vec{X}^{(t)}$ inputs will exhibit useful symmetries and hence may be nontrivially treated by any of our previously discussed architectures. Some common examples include: \emph{videos} ($\Omega$ is a fixed grid, and signals are a sequence of \emph{frames}); \emph{fMRI scans} ($\Omega$ is a fixed \emph{mesh} representing the geometry of the brain cortex, where different regions are activated at different times as a response to presented stimuli); and \emph{traffic flow networks} ($\Omega$ is a fixed \emph{graph} representing the road network, on which e.g. the average traffic speed is recorded at various nodes).

Let us assume an \emph{encoder} function $f(\vec{X}^{(t)})$ providing latent representations at the level of granularity appropriate for the problem and respectful of the symmetries of the input domain. 
As an example\marginnote{We do not lose generality in our example; equivalent analysis can be done e.g. for node-level outputs on a spatiotemporal graph; the only difference is in the choice of encoder $f$ (which will then be a permutation equivariant GNN).}, consider  processing video frames: that is, at each timestep, we are given a \emph{grid-structured input} represented as an $n\times d$ matrix $\mathbf{X}^{(t)}$, where $n$ is the number of pixels (fixed in time) and $d$ is the number of input channels (e.g. $d=3$ for RGB frames). 
Further, we are interested in analysis at the level of entire frames, in which case it is appropriate to implement $f$ as a translation invariant CNN, outputting a $k$-dimensional representation $\vec{z}^{(t)} = f(\vec{X}^{(t)})$ of the frame at time-step $t$. 

We are now left with the task of appropriately \emph{summarising} a sequence of vectors $\vec{z}^{(t)}$ {\em across all the steps.} A canonical way to \emph{dynamically} aggregate this information in a way that respects the temporal progression of inputs and also easily allows for \emph{online} arrival of novel data-points, is using a \emph{Recurrent Neural Network} (RNN). \marginnote{Note that the $\vec{z}^{(t)}$ vectors can be seen as points on a {\em temporal grid}: hence, processing them with a CNN is also viable in some cases. Transformers are also increasingly popular models for processing generic sequential inputs.}
What we will show here is that RNNs are 
an interesting geometric architecture to study in their own right, since they implement a rather unusual type of symmetry  over the inputs $\vec{z}^{(t)}$. 

\paragraph{SimpleRNNs}
\begin{figure}
    \centering
    \includegraphics[width=\linewidth]{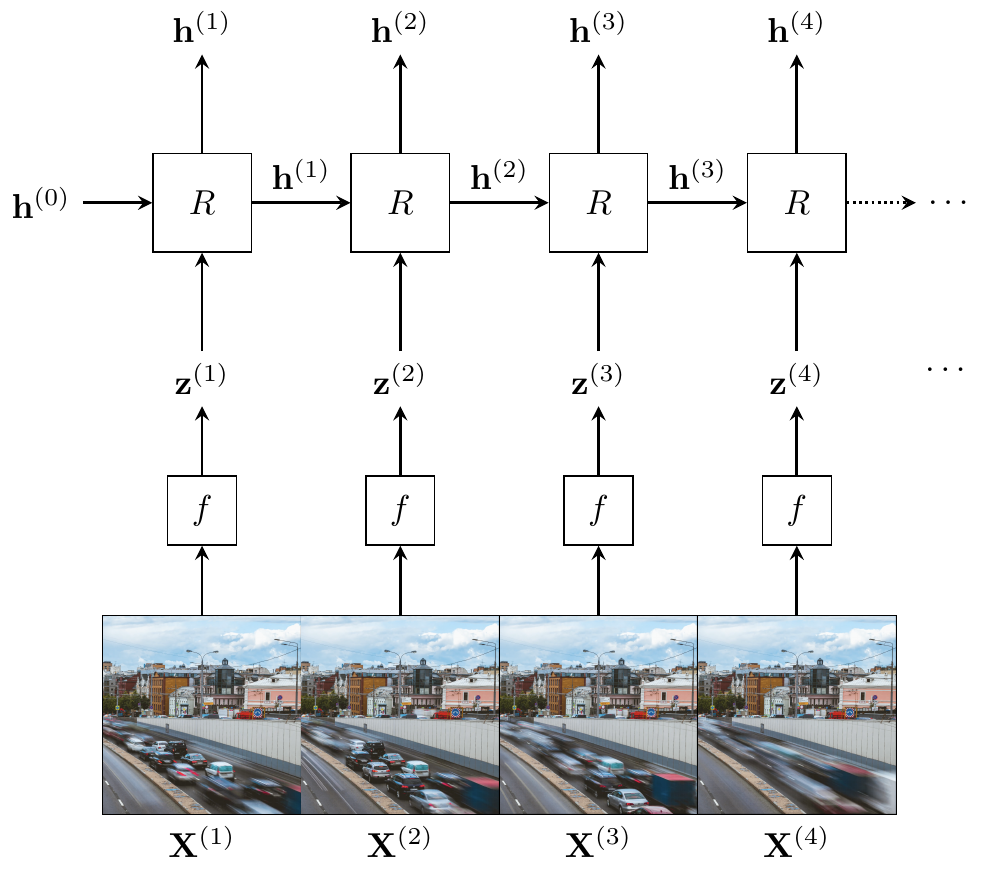}
    \caption{Illustration of processing video input with RNNs. Each input video frame $\vec{X}^{(t)}$ is processed using a shared function $f$---e.g. a translation invariant CNN---into a flat representation $\vec{z}^{(t)}$. Then the RNN update function $R$ is iterated across these vectors, iteratively updating a summary vector $\vec{h}^{(t)}$ which summarises all the inputs up to and including $\vec{z}^{(t)}$. The computation is seeded with an initial summary vector $\vec{h}^{(0)}$, which may be either pre-determined or learnable.}
    \label{fig:normal_rnn}
\end{figure}
At each step, the recurrent neural network computes an $m$-dimensional \emph{summary} vector $\vec{h}^{(t)}$ of all the input steps up to and
including $t$. This (partial) summary is computed conditional on the current step's
features and the previous step's summary, through a shared \emph{update} function, $R : \mathbb{R}^k \times \mathbb{R}^m \rightarrow \mathbb{R}^m$, as follows (see Figure \ref{fig:normal_rnn} for a summary):
\begin{equation}\label{eqn:rnn_upd}
    \vec{h}^{(t)} = R(\vec{z}^{(t)}, \vec{h}^{(t-1)})
\end{equation}
and, as both $\vec{z}^{(t)}$ and $\vec{h}^{(t-1)}$ are \emph{flat} vector representations, $R$ may be most easily expressed as a single fully-connected neural network layer (often known as \emph{SimpleRNN}\marginnote{In spite of their name, SimpleRNNs are remarkably expressive. For example, it was shown by \citet{siegelmann1995computational} that such models are \emph{Turing-complete}, meaning that they can likely represent \emph{any} computation we may ever be able to execute on computers.}; see \citet{elman1990finding,jordan1997serial}):
\begin{equation}\label{eqn:SimpleRNN}
     \vec{h}^{(t)} = \sigma(\vec{W}\vec{z}^{(t)} + \vec{U}\vec{h}^{(t-1)} + \vec{b})
\end{equation}
where $\vec{W}\in\mathbb{R}^{k\times m}$, $\vec{U}\in\mathbb{R}^{m\times m}$ and $\vec{b}\in\mathbb{R}^m$ are learnable parameters, and $\sigma$ is an activation function. While this introduces \emph{loops} in the network's computational graph, in practice
the network is unrolled for an appropriate number of steps, allowing for \emph{backpropagation through time} \citep{robinson1987utility,werbos1988generalization,mozer1989focused} to be applied.

The summary vectors may then be appropriately leveraged for the downstream task---if
a prediction is required at every step of the sequence, then a shared predictor may be applied
to each $\vec{h}^{(t)}$ individually. For classifying entire sequences, typically the final summary, $\vec{h}^{(T)}$, is passed to a classifier. Here, $T$ is the length of the sequence.

Specially, the initial summary vector is usually either set to the zero-vector, i.e. $\vec{h}^{(0)}=\vec{0}$, or it is made learnable. Analysing the manner in which the initial summary vector is set also allows us to deduce an interesting form of \emph{translation equivariance} exhibited by RNNs.

\paragraph{Translation equivariance in RNNs} Since we interpret the individual steps $t$ as \emph{discrete time-steps}, the input vectors $\vec{z}^{(t)}$ can be seen as living on a one-dimensional\marginnote{Note that this construction is extendable to grids in higher dimensions, allowing us to, e.g., process signals living on images in a \emph{scanline} fashion. Such a construction powered a popular series of models, such as the PixelRNN from \citet{van2016pixel}.} \emph{grid} of time-steps. While it might be attractive to attempt extending our translation equivariance analysis from CNNs here, it cannot be done in a trivial manner. 

To see why, let us assume that we have produced a new sequence $\vec{z}'^{(t)} = \vec{z}^{(t+1)}$ by performing a left-shift of our sequence by one step. 
It might be tempting to attempt showing $\vec{h}'^{(t)} = \vec{h}^{(t+1)}$, as one expects with translation equivariance; however, this will not generally hold. Consider $t = 1$; directly applying and expanding the update function, we recover the following:
\begin{align}\label{eqn:shiftrnn}
    \vec{h}'^{(1)} &= R(\vec{z}'^{(1)}, \vec{h}^{(0)}) = R(\vec{z}^{(2)}, \vec{h}^{(0)})\\ 
    \vec{h}^{(2)} &= R(\vec{z}^{(2)}, \vec{h}^{(1)}) = R(\vec{z}^{(2)}, R(\vec{z}^{(1)}, \vec{h}^{(0)}))\label{eqn:shiftrnn2} 
\end{align}
Hence, unless we can guarantee that $\vec{h}^{(0)} = R(\vec{z}^{(1)}, \vec{h}^{(0)})$, we will not recover translation equivariance. Similar analysis can then be done for steps $t > 1$. 

Fortunately, with a slight refactoring of how we represent $\vec{z}$, and for a suitable choice of $R$, it is possible to satisfy the equality above, and hence demonstrate a setting in which RNNs are equivariant to shifts.
Our problem was largely one of \emph{boundary conditions}: the equality above includes $\vec{z}^{(1)}$, which our left-shift operation destroyed. To abstract this problem away, we will observe how an RNN processes an appropriately \emph{left-padded} sequence, $\bar{\vec{z}}^{(t)}$, defined as follows: 
$$
    \bar{\vec{z}}^{(t)} = \begin{cases}
        \vec{0} & t \leq t'\\
        \vec{z}^{(t - t')} & t > t' 
    \end{cases}
$$
Such a sequence now allows for left-shifting\marginnote{Note that equivalent analyses will arise if we use a different padding vector than $\vec{0}$.} by up to $t'$ steps without destroying any of the original input elements. Further, note we do not need to handle right-shifting separately; indeed, equivariance to right shifts naturally follows from the RNN equations.

We can now again analyse the operation of the RNN over a left-shifted verson of $\bar{\vec{z}}^{(t)}$, which we denote by $\bar{\vec{z}}'^{(t)} = \bar{\vec{z}}^{(t+1)}$, as we did in Equations \ref{eqn:shiftrnn}--\ref{eqn:shiftrnn2}:
\begin{align*}
    \vec{h}'^{(1)} &= R(\bar{\vec{z}}'^{(1)}, \vec{h}^{(0)}) = R(\bar{\vec{z}}^{(2)}, \vec{h}^{(0)})\\ \vec{h}^{(2)} &= R(\bar{\vec{z}}^{(2)}, \vec{h}^{(1)}) = R(\bar{\vec{z}}^{(2)}, R(\bar{\vec{z}}^{(1)}, \vec{h}^{(0)})) = R(\bar{\vec{z}}^{(2)}, R(\vec{0}, \vec{h}^{(0)}))
\end{align*}
where the substitution $\bar{\vec{z}}^{(1)} = \vec{0}$ holds as long as $t' \geq 1$, i.e. as long as any padding is applied\marginnote{In a very similar vein, we can derive equivariance to left-shifting by $s$ steps as long as $t' \geq s$.}. Now, we can guarantee equivariance to left-shifting by one step ($\vec{h}'^{(t)} = \vec{h}^{(t+1)}$) as long as $\vec{h}^{(0)} = R(\vec{0}, \vec{h}^{(0)})$.

Said differently, $\vec{h}^{(0)}$ must be chosen to be a \emph{fixed point} of a function $\gamma(\vec{h}) = R(\vec{0}, \vec{h})$. If the update function $R$ is conveniently chosen, then not only can we guarantee existence of such fixed points, but we can even directly obtain them by iterating the application of $R$ until convergence; e.g., as follows:
\begin{equation}\label{eqn:iterated}
    \vec{h}_0 = \vec{0} \qquad \vec{h}_{k+1} = \gamma(\vec{h}_{k}),
\end{equation}
where the index $k$ refers to the iteration of $R$ in our computation, as opposed to the the index $(t)$ denoting the time step of the RNN.  
If we choose $R$ such that $\gamma$ is a \emph{contraction mapping}\marginnote{Contractions are functions $\gamma: \mathcal{X}\rightarrow \mathcal{X}$ such that, under some norm $\|\cdot\|$ on $\mathcal{X}$, applying $\gamma$ \emph{contracts} the distances between points: for all $\vec{x},\vec{y}\in\mathcal{X}$, and some $q\in [0,1)$, it holds that $\|\gamma(\vec{x}) - \gamma(\vec{y})\| \leq q\|\vec{x} - \vec{y}\|$. Iterating such a function then necessarily converges to a unique fixed point, as a direct consequence of \emph{Banach's Fixed Point Theorem} \citep{banach1922operations}.}, such an iteration will indeed converge to a \emph{unique} fixed point. Accordingly, we can then iterate Equation~(\ref{eqn:iterated}) until $\vec{h}_{k+1} = \vec{h}_k$, and we can set $\vec{h}^{(0)} = \vec{h}_k$. Note that this computation is equivalent to \emph{left-padding} the sequence with ``sufficiently many'' zero-vectors.

\paragraph{Depth in RNNs} 
It is also easy to stack multiple RNNs---simply use the $\vec{h}^{(t)}$ vectors as an input sequence for a second RNN. This kind of construction is occasionally called a ``deep RNN'', which is potentially misleading. Effectively, due to the repeated application of the recurrent operation, even a single RNN ``layer'' has depth \emph{equal to the number of input steps}.

This often introduces uniquely challenging learning dynamics when optimising RNNs, as each training example induces many gradient updates to the \emph{shared} parameters of the update network. Here we will focus on perhaps the most prominent such issue---that of \emph{vanishing} and \emph{exploding} gradients \citep{bengio1994learning}---which is especially problematic in RNNs, given their depth and parameter sharing. Further, it has single-handedly spurred some of the most influential research on RNNs. For a more detailed overview, we refer the reader to \citet{pascanu2013difficulty}, who have studied the training dynamics of RNNs in great detail, and exposed these challenges from a variety of perspectives: analytical, geometrical, and the lens of dynamical systems.

To illustrate vanishing gradients, consider a SimpleRNN with a sigmoidal activation function  $\sigma$\marginnote{\includegraphics[width=\linewidth]{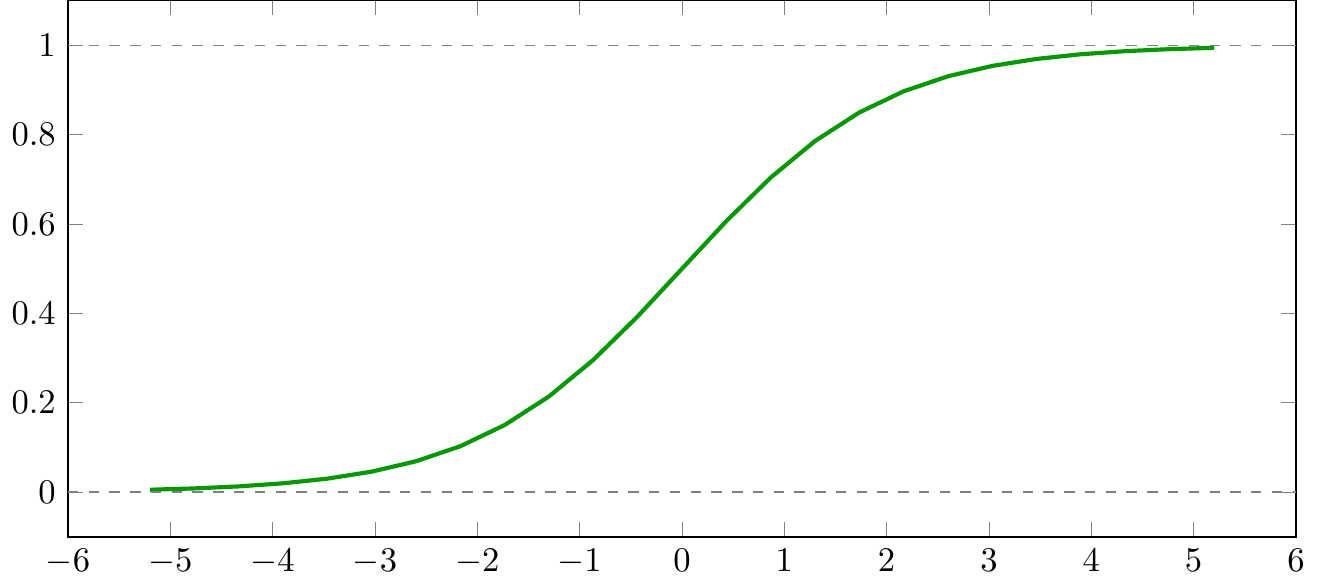}\\ Examples of such an activation include the \emph{logistic function}, $\sigma(x) = \frac{1}{1+\exp(-x)}$, and the \emph{hyperbolic tangent}, $\sigma(x)=\tanh x$. They are called sigmoidal due to the distinct S-shape of their plots.}, whose derivative magnitude $|\sigma'|$ is always between $0$ and $1$. Multiplying many such values results in gradients that quickly tend to zero, implying that early steps in the input sequence may not be able to have influence in updating the network parameters at all.

For example, consider the next‐word prediction task (common in e.g. predictive keyboards), and the input text \emph{``Petar is Serbian. He was born on \dots [long paragraph] \dots Petar currently lives in \rule{1.3cm}{0.15mm}''}. Here, predicting the next word as ``Serbia'' may only be reasonably concluded by considering the very start of the paragraph—but gradients have likely vanished by the time they reach this input step, making learning from such examples very challenging.

Deep feedforward neural networks have also suffered from the vanishing gradient problem, until the invention of the ReLU activation (which has gradients equal to \emph{exactly} zero or one---thus fixing the vanishing gradient problem). However, in RNNs, using ReLUs may easily lead to \emph{exploding} gradients, as the output space of the update function is now \emph{unbounded}, and gradient descent will update the cell once for every input step, quickly building up the scale of the updates.
Historically, the vanishing gradient phenomenon was recognised early on as a significant obstacle in the use of recurrent networks. Coping with this problem motivated the development of more sophisticated RNN layers, which we describe next.

\subsection{Long Short-Term Memory networks}
\label{sec:lstm}

\begin{figure}
    \centering
    \includegraphics[width=\linewidth]{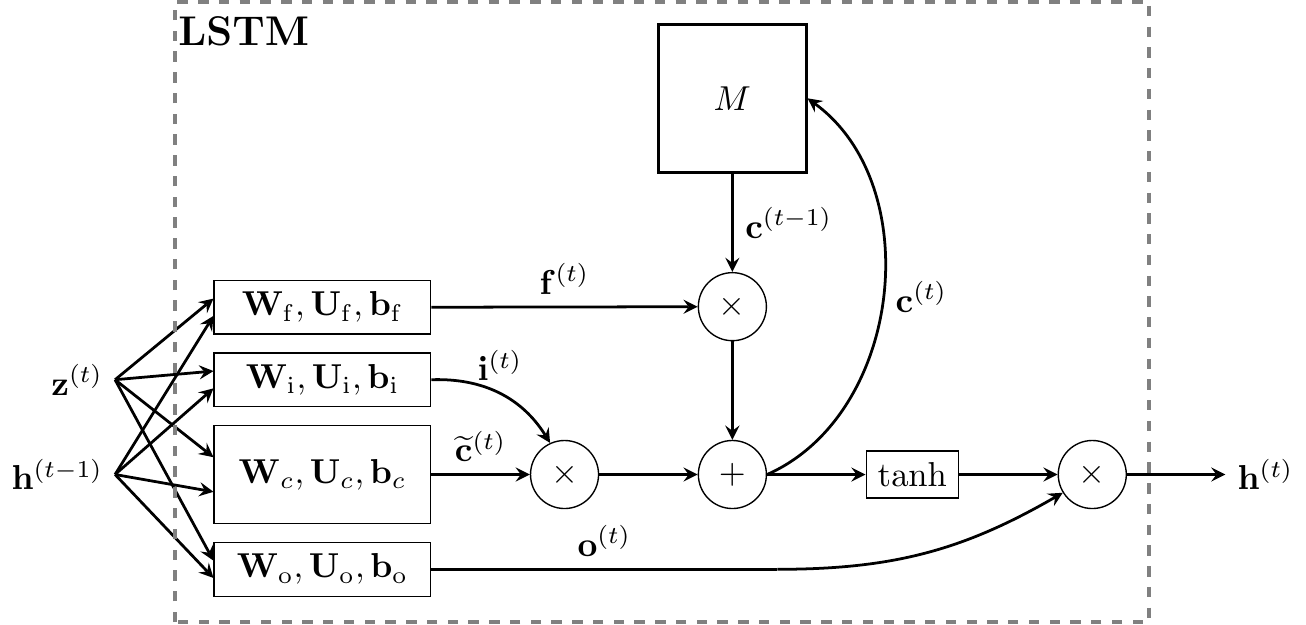}
    \caption{The dataflow of the long short-term memory (LSTM), with its components and memory cell ($M$) clearly highlighted. Based on the current input $\vec{z}^{(t)}$, previous summary $\vec{h}^{(t-1)}$ and previous cell state $\vec{c}^{(t-1)}$, the LSTM predicts the updated cell state $\vec{c}^{(t)}$ and summary $\vec{h}^{(t)}$.}
    \label{fig:lstm}
\end{figure}
A key invention that significantly reduced the effects of vanishing gradients in RNNs is that of \emph{gating mechanisms}, which allow the network to selectively \emph{overwrite} information in a data-driven way. Prominent examples of these \emph{gated RNNs} include the {\em Long Short-Term Memory} (LSTM; \citet{hochreiter1997long}) and the {\em Gated Recurrent Unit} (GRU; \citet{cho2014learning}). Here we will primarily discuss the LSTM---specifically, the variant presented by \citet{graves2013generating}---in order to illustrate the operations of such models. Concepts from LSTMs easily carry over to other gated RNNs.

Throughout this section, it will likely be useful to refer to Figure \ref{fig:lstm}, which illustrates all of the LSTM operations that we will discuss in text.

The LSTM augments the recurrent computation by introducing a \emph{memory cell}, which stores \emph{cell state} vectors, $\vec{c}^{(t)}\in\mathbb{R}^m$, that are \emph{preserved} between computational steps. The LSTM computes summary vectors, $\vec{h}^{(t)}$, directly based on $\vec{c}^{(t)}$, and $\vec{c}^{(t)}$ is, in turn, computed using $\vec{z}^{(t)}$, $\vec{h}^{(t-1)}$ and $\vec{c}^{(t-1)}$.
Critically, the cell is \textbf{not} completely overwritten based on $\vec{z}^{(t)}$ and $\vec{h}^{(t-1)}$, which would expose the network to the same issues as the SimpleRNN. Instead, a certain quantity of the previous cell state may be \emph{retained}---and the proportion by which this occurs is explicitly \emph{learned} from data.

Just like in SimpleRNN, we compute features by using a single fully-connected neural network layer over the current input step and previous summary:\marginnote{Note that we have set the activation function to $\tanh$ here; as LSTMs are designed to ameliorate the vanishing gradient problem, it is now appropriate to use a sigmoidal activation.}
\begin{equation}\label{eqn:lstmstart}
     \widetilde{\vec{c}}^{(t)} = \tanh(\vec{W}_c\vec{z}^{(t)} + \vec{U}_c\vec{h}^{(t-1)} + \vec{b}_c)
\end{equation}
But, as mentioned, we do not allow \emph{all} of this vector to enter the cell---hence why we call it the vector of \emph{candidate} features, and denote it as $\widetilde{\vec{c}}^{(t)}$. Instead, the LSTM directly learns \emph{gating vectors}, which are real-valued vectors in the range $[0, 1]$, and decide how much of the signal should be allowed to enter, exit, and overwrite the memory cell. 

Three such gates are computed, all based on $\vec{z}^{(t)}$ and $\vec{h}^{(t-1)}$: the \emph{input gate} $\vec{i}^{(t)}$, which computes the proportion of the candidate vector allowed to enter the cell; the \emph{forget gate} $\vec{f}^{(t)}$, which computes the proportion of the previous cell state to be retained, and the \emph{output gate} $\vec{o}^{(t)}$, which computes the proportion of the new cell state to be used for the final summary vector. Typically all of these gates are also derived using a single fully connected layer, albeit with the \emph{logistic sigmoid} activation $\mathrm{logistic}(x) = \frac{1}{1+\exp(-x)}$, in order to guarantee that the outputs are in the $[0, 1]$ range\marginnote{Note that the three gates are themselves \emph{vectors}, i.e. $\vec{i}^{(t)}, \vec{f}^{(t)}, \vec{o}^{(t)}\in [0,1]^m$. This allows them to control how much \emph{each} of the $m$ dimensions is allowed through the gate.}:
\begin{align}\label{eqn:lstm_gate}
     \vec{i}^{(t)} &= \mathrm{logistic}(\vec{W}_\mathrm{i}\vec{z}^{(t)} + \vec{U}_\mathrm{i}\vec{h}^{(t-1)} + \vec{b}_\mathrm{i})\\
     \vec{f}^{(t)} &= \mathrm{logistic}(\vec{W}_\mathrm{f}\vec{z}^{(t)} + \vec{U}_\mathrm{f}\vec{h}^{(t-1)} + \vec{b}_\mathrm{f})\\
     \vec{o}^{(t)} &= \mathrm{logistic}(\vec{W}_\mathrm{o}\vec{z}^{(t)} + \vec{U}_\mathrm{o}\vec{h}^{(t-1)} + \vec{b}_\mathrm{o})
\end{align}

Finally, these gates are appropriately applied to decode the \emph{new} cell state, $\vec{c}^{(t)}$, which is then modulated by the output gate to produce the summary vector $\vec{h}^{(t)}$, as follows:
\begin{align}
    \vec{c}^{(t)} &= \vec{i}^{(t)}\odot\widetilde{\vec{c}}^{(t)} + \vec{f}^{(t)}\odot\vec{c}^{(t-1)}\\
    \vec{h}^{(t)} &= \vec{o}^{(t)}\odot\tanh(\vec{c}^{(t)})\label{eqn:lstm_end}
\end{align}
where $\odot$ is element-wise vector multiplication. Applied together, Equations (\ref{eqn:lstmstart})--(\ref{eqn:lstm_end}) completely specify the \emph{update rule} for the LSTM, which now takes into account the cell vector $\vec{c}^{(t)}$ as well\marginnote{This is still compatible with the RNN update blueprint from Equation (\ref{eqn:rnn_upd}); simply consider the summary vector to be the \emph{concatenation} of $\vec{h}^{(t)}$ and $\vec{c}^{(t)}$; sometimes denoted by $\vec{h}^{(t)}\|\vec{c}^{(t)}$.}:
$$
    (\vec{h}^{(t)}, \vec{c}^{(t)}) = R(\vec{z}^{(t)}, (\vec{h}^{(t-1)}, \vec{c}^{(t-1)}))
$$
Note that, as the values of $\vec{f}^{(t)}$ are derived from $\vec{z}^{(t)}$ and $\vec{h}^{(t-1)}$---and therefore directly \emph{learnable} from data---the LSTM effectively learns how to appropriately forget past experiences. Indeed, the values of $\vec{f}^{(t)}$ directly appear in the backpropagation update for all the LSTM parameters ($\vec{W}_*, \vec{U}_*, \vec{b}_*$), allowing the network to explicitly \emph{control}, in a data-driven way, the degree of vanishing for the gradients across the time steps.

Besides tackling the vanishing gradient issue head-on, it turns out that gated RNNs also unlock a very useful form of invariance to \emph{time-warping} transformations, which remains out of reach of SimpleRNNs.

\paragraph{Time warping invariance of gated RNNs} We will start by illustrating, in a \emph{continuous-time} setting\marginnote{We focus on the continuous setting as it will be easier to reason about manipulations of time there.}, what does it mean to \emph{warp time}, and what is required of a recurrent model in order to achieve invariance to such transformations. Our exposition will largely follow the work of \citet{tallec2018can}, that initially described this phenomenon---and indeed, they were among the first to actually study RNNs from the lens of invariances.

Let us assume a continuous time-domain signal $z(t)$, on which we would like to apply an RNN. To align the RNN's discrete-time computation of summary vectors $\vec{h}^{(t)}$\marginnote{We will use $h(t)$ to denote a continuous signal at time $t$, and $\vec{h}^{(t)}$ to denote a discrete signal at time-step $t$.} with an analogue in the continuous domain, $h(t)$, we will observe its linear Taylor expansion:
\begin{equation}
    h(t + \delta) \approx h(t) + \delta\frac{\mathrm{d}h(t)}{\mathrm{d}t}
\end{equation}
and, setting $\delta = 1$, we recover a relationship between $h(t)$ and $h(t+1)$, which is exactly what the RNN update function $R$ (Equation \ref{eqn:rnn_upd}) computes. Namely, the RNN update function satisfies the following differential equation:
\begin{equation}\label{eqn:ode_pre_warp}
    \frac{\mathrm{d}h(t)}{\mathrm{d}t} = h(t+1) - h(t) = R(z(t+1), h(t)) - h(t)
\end{equation}

We would like the RNN to be resilient to the way in which the signal is sampled (e.g. by changing the time unit of measurement), in order to account for any imperfections or irregularities therein. Formally, we denote a \emph{time warping}\marginnote{\includegraphics[width=\linewidth]{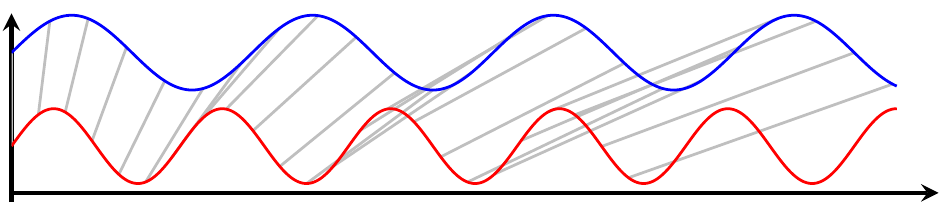}\\ Such warping operations can be \emph{simple}, such as time rescaling; e.g. $\tau(t) = 0.7t$ (displayed above), which, in a discrete setting, would amount to new inputs being received every $\sim 1.43$ steps. However, it also admits a wide spectrum of \emph{variably-changing} sampling rates, e.g. sampling may freely accelerate or decelerate throughout the time domain.} operation $\tau : \mathbb{R}^+\rightarrow\mathbb{R}^+$, as any monotonically increasing differentiable mapping between times. The notation $\tau$ is chosen because time warping represents an \emph{automorphism} of time.

Further, we state that a class of models is \emph{invariant} to time warping if, for any model of the class and any such $\tau$, there exists another (possibly the same) model from the class that processes the warped data in the same way as the original model did in the non-warped case.

This is a potentially very useful property. If we have an RNN class capable of modelling short-term dependencies well, and we can also show that this class is invariant to time warping, then we know it is possible to train such a model in a way that will usefully capture long-term dependencies as well (as they would correspond to a time dilation warping of a signal with short-term dependencies). 
As we will shortly see, it is no coincidence that \emph{gated} RNN models such as the LSTM were proposed to model long-range dependencies. Achieving time warping invariance is tightly coupled with presence of gating mechanisms, such as the input/forget/output gates of LSTMs.

When time gets warped by $\tau$, the signal observed by the RNN at time $t$ is $z(\tau(t))$ and, to remain invariant to such warpings, it should predict an equivalently-warped summary function $h(\tau(t))$. Using Taylor expansion arguments once more, we derive a form of Equation \ref{eqn:ode_pre_warp} for the warped time, that the RNN update $R$ should satisfy:
\begin{equation}\label{eqn:ode_post_warp}
    \frac{\mathrm{d}h(\tau(t))}{\mathrm{d}\tau(t)} = R(z(\tau(t+1)), h(\tau(t))) - h(\tau(t))
\end{equation}
However, the above derivative is computed with respect to the warped time $\tau(t)$, and hence does not take into account the original signal. To make our model take into account the warping transformation explicitly, we need to differentiate the warped summary function with respect to $t$. Applying the chain rule, this yields the following differential equation:
\begin{equation}\label{eqn:full_ode_warp}
    \frac{\mathrm{d}h(\tau(t))}{\mathrm{d}t} = \frac{\mathrm{d}h(\tau(t))}{\mathrm{d}\tau(t)}\frac{\mathrm{d}\tau(t)}{\mathrm{d}t} = \frac{\mathrm{d}\tau(t)}{\mathrm{d}t}R(z(\tau(t+1)), h(\tau(t))) - \frac{\mathrm{d}\tau(t)}{\mathrm{d}t}h(\tau(t))
\end{equation}
and, for our (continuous-time) RNN to remain invariant to \emph{any} time warping $\tau(t)$, it needs to be able to explicitly represent the derivative $\frac{\mathrm{d}\tau(t)}{\mathrm{d}t}$, which is not assumed known upfront! We need to introduce a \emph{learnable} function $\Gamma$ which approximates this derivative. For example, $\Gamma$ could be a neural network taking into account $z(t+1)$ and $h(t)$ and predicting scalar outputs.

Now, remark that, from the point of view of a \emph{discrete} RNN model under time warping, its input $\vec{z}^{(t)}$ will correspond to $z(\tau(t))$, and its summary $\vec{h}^{(t)}$ will correspond to $h(\tau(t))$. To obtain the required relationship of $\vec{h}^{(t)}$ to $\vec{h}^{(t+1)}$ in order to remain invariant to time warping, we will use a one-step Taylor expansion of $h(\tau(t))$:
\begin{equation*}
    h(\tau(t + \delta)) \approx h(\tau(t)) + \delta\frac{\mathrm{d}h(\tau(t))}{\mathrm{d}t}
\end{equation*}
and, once again, setting $\delta=1$ and substituting Equation \ref{eqn:full_ode_warp}, then discretising:
\begin{align*}
    \vec{h}^{(t+1)} &= \vec{h}^{(t)} + \frac{\mathrm{d}\tau(t)}{\mathrm{d}t}R(\vec{z}^{(t+1)}, \vec{h}^{(t)}) - \frac{\mathrm{d}\tau(t)}{\mathrm{d}t}\vec{h}^{(t)}\\
    &= \frac{\mathrm{d}\tau(t)}{\mathrm{d}t}R(\vec{z}^{(t+1)}, \vec{h}^{(t)}) + \left(1 - \frac{\mathrm{d}\tau(t)}{\mathrm{d}t}\right)\vec{h}^{(t)}
\end{align*}
Finally, we swap $\frac{\mathrm{d}\tau(t)}{\mathrm{d}t}$ with the aforementioned learnable function, $\Gamma$. This gives us the required form for our time warping-invariant RNN:
\begin{equation}\label{eqn:GaRNN}
    \vec{h}^{(t+1)} = \Gamma(\vec{z}^{(t+1)}, \vec{h}^{(t)})R(\vec{z}^{(t+1)}, \vec{h}^{(t)}) + (1 - \Gamma(\vec{z}^{(t+1)}, \vec{h}^{(t)}))\vec{h}^{(t)}
\end{equation}
We may quickly deduce that SimpleRNNs (Equation \ref{eqn:SimpleRNN}) are \emph{not} time warping invariant, given that they do not feature the second term in Equation \ref{eqn:GaRNN}. Instead, they fully overwrite $\vec{h}^{(t)}$ with $R(\vec{z}^{(t+1)}, \vec{h}^{(t)})$, which corresponds to assuming no time warping at all; $\frac{\mathrm{d}\tau(t)}{\mathrm{d}t} = 1$, i.e. $\tau(t) = t$.

Further, our link between continuous-time RNNs and the discrete RNN based on $R$ rested on the accuracy of the Taylor approximation, which holds only if the time-warping derivative is not too large, i.e., $\frac{\mathrm{d}\tau(t)}{\mathrm{d}t}\lesssim 1$. The intuitive explanation of this is: if our time warping operation ever \emph{contracts time} in a way that makes time increments ($t \rightarrow t + 1$) large enough that intermediate data changes are not sampled, the model can never hope to process time-warped inputs in the same way as original ones---it simply would not have access to the same information. Conversely, time \emph{dilations} of any form (which, in discrete terms, correspond to interspersing the input time-series with zeroes) are perfectly allowed within our framework.

Combined with our requirement of monotonically increasing $\tau$ ($\frac{\mathrm{d}\tau(t)}{\mathrm{d}t} > 0$), we can bound the output space of $\Gamma$ as $0 < \Gamma(\vec{z}^{(t+1)}, \vec{h}^{(t)}) < 1$, which motivates the use of the logistic sigmoid activation for $\Gamma$, e.g.:
$$
    \Gamma(\vec{z}^{(t+1)}, \vec{h}^{(t)}) = \mathrm{logistic}({\bf W}_\Gamma\vec{z}^{(t+1)} + {\bf U}_\Gamma\vec{h}^{(t)} + \vec{b}_\Gamma)
$$
\emph{exactly} matching the LSTM gating equations (e.g. Equation \ref{eqn:lstm_gate}). The main difference is that LSTMs compute gating \emph{vectors}, whereas Equation \ref{eqn:GaRNN} implies $\Gamma$ should output a scalar. Vectorised gates \citep{hochreiter1991untersuchungen} allow to fit a \emph{different} warping derivative in every dimension of $\vec{h}^{(t)}$, allowing for reasoning over \emph{multiple} time horizons simultaneously.

It is worth taking a pause here to summarise what we have done. By requiring that our RNN class is invariant to (non-destructive) time warping, we have derived the necessary form that it must have (Equation \ref{eqn:GaRNN}), and showed that it exactly corresponds to the class of \emph{gated} RNNs. The gates' primary role under this perspective is to accurately fit the \emph{derivative} $\frac{\mathrm{d}\tau(t)}{\mathrm{d}t}$ of the warping transformation.

The notion of \emph{class invariance} is somewhat distinct from the invariances we studied previously. Namely, once we train a gated RNN on a time-warped input with $\tau_1(t)$, we typically cannot zero-shot transfer\marginnote{One case where zero-shot transfer is possible is when the second time warping is assumed to be a \emph{time rescaling} of the first one ($\tau_2(t) = \alpha\tau_1(t)$). Transferring a gated RNN pre-trained on $\tau_1$ to a signal warped by $\tau_2$ merely requires \emph{rescaling the gates}: $\Gamma_2(\vec{z}^{(t+1)}, \vec{h}^{(t)}) = \alpha \Gamma_1(\vec{z}^{(t+1)}, \vec{h}^{(t)})$. $R$ can retain its parameters ($R_1 = R_2$).} it to a  signal warped by a different $\tau_2(t)$. Rather, class invariance only guarantees that gated RNNs are powerful enough to fit both of these signals in the same manner, but potentially with vastly different model parameters. That being said, the realisation that effective gating mechanisms are tightly related to fitting the warping derivative can yield useful prescriptions for gated RNN optimisation, as we now briefly demonstrate.

For example, we can often assume that the range of the dependencies we are interested in tracking within our signal will be in the range $[T_l, T_h]$ time-steps.

By analysing the analytic solutions to Equation \ref{eqn:full_ode_warp}, it can be shown that the characteristic \emph{forgetting time} of $\vec{h}^{(t)}$ by our gated RNN is proportional to $\frac{1}{\Gamma(\vec{z}^{(t+1)}, \vec{h}^{(t)})}$. Hence, we would like our gating values to lie between $\left[\frac{1}{T_h}, \frac{1}{T_m}\right]$ in order to effectively remember information within the assumed range.

Further, if we assume that $\vec{z}^{(t)}$ and $\vec{h}^{(t)}$ are roughly \emph{zero-centered}---which is a common by-product of applying transformations such as layer normalisation \citep{ba2016layer}---we can assume that $\mathbb{E}[\Gamma(\vec{z}^{(t+1)}, \vec{h}^{(t)})] \approx \mathrm{logistic}(\vec{b}_\Gamma)$. Controlling the \emph{bias} vector of the gating mechanism is hence a very powerful way of controlling the effective gate value\marginnote{This insight was already spotted by \citet{gers2000recurrent,jozefowicz2015empirical}, who empirically recommended initialising the forget-gate bias of LSTMs to a constant positive vector, such as $\vec{1}$.}.

Combining the two observations, we conclude that an appropriate range of gating values can be obtained by initialising $\vec{b}_\Gamma\sim-\log(\mathcal{U}(T_l, T_h) - 1)$, where $\mathcal{U}$ is the uniform real distribution. Such a recommendation was dubbed \emph{chrono initialisation} by \citet{tallec2018can}, and has been empirically shown to improve the long-range dependency modelling of gated RNNs.

\paragraph{Sequence-to-sequence learning with RNNs}
One prominent historical example of using RNN-backed computation are \emph{sequence-to-sequence} translation tasks, such as {\em machine translation} of natural languages. The pioneering \emph{seq2seq} work by \citet{sutskever2014sequence} achieved this by passing the summary vector, $\vec{h}^{(T)}$ as an initial input for a \emph{decoder} RNN, with outputs of RNN blocks being given as inputs for the next step.

\begin{figure}
    \centering
    \includegraphics[width=\linewidth]{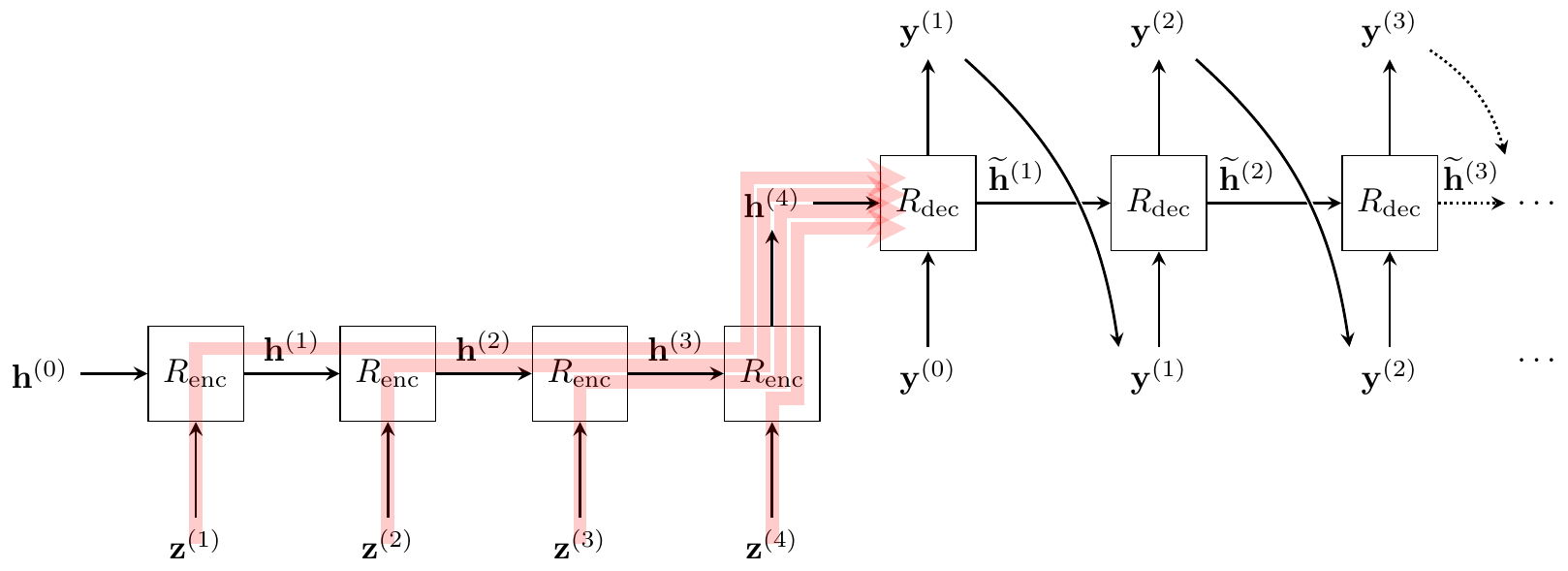}
    \caption{One typical example of a seq2seq architecture with an RNN encoder $R_\mathrm{enc}$ and RNN decoder $R_\mathrm{dec}$. The decoder is seeded with the final summary vector $\vec{h}^{(T)}$ coming out of the encoder, and then proceeds in an \emph{autoregressive} fashion: at each step, the predicted output from the previous step is fed back as input to $R_\mathrm{dec}$. The bottleneck problem is also illustrated with the red lines: the summary vector $\vec{h}^{(T)}$ is pressured to store \emph{all} relevant information for translating the input sequence, which becomes increasingly challenging as the input length grows.}
    \label{fig:rnn_bottleneck}
\end{figure}

This placed substantial representational pressure on the summary vector, $\vec{h}^{(T)}$. Within the context of deep learning, $\vec{h}^{(T)}$ is sometimes referred to as a \emph{bottleneck}\marginnote{\includegraphics[width=\linewidth]{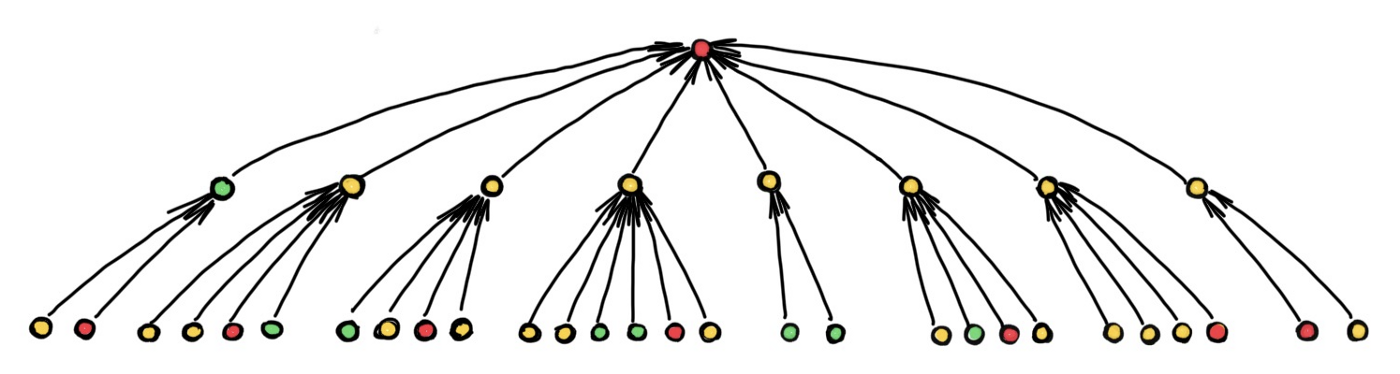}\\
The bottleneck effect has recently received substantial attention in the graph representation learning community \citep{alon2020bottleneck}, as well as neural algorithmic reasoning \citep{cappart2021combinatorial}.}. Its fixed capacity must be sufficient for representing the content of the entire input sequence, in a manner that is conducive to generating a corresponding sequence, while also supporting input sequences of substantially different lengths (Figure \ref{fig:rnn_bottleneck}).

In reality, different steps of the output may wish to focus (\emph{attend}) on different parts of the input, and all such choices are difficult to represent via a bottleneck vector. Following from this observation, the popular \emph{recurrent attention} model was proposed by \citet{bahdanau2014neural}. At every step of processing, a \emph{query vector} is generated by an RNN; this query vector then interacts with the representation of \emph{every} time-step $\vec{h}^{(t)}$, primarily by computing a weighted sum over them. This model pioneered neural content-based attention and predates the success of the Transformer model.

Lastly, while attending offers a \emph{soft} way to dynamically focus on parts of the input content, substantial work also learnt more \emph{explicit} ways to direct attention to the input. A powerful algorithmically grounded way of doing so is the \emph{pointer network} of \citet{vinyals2015pointer}, which proposes a simple modification of recurrent attention to allow for pointing over elements of \emph{variable-sized} inputs. These findings have then been generalised to the \emph{set2set} architecture \citep{vinyals2016order}, which generalises seq2seq models to unordered sets, supported by pointer network-backed LSTMs.

\section{Problems and Applications}

Invariances and symmetries arise all too commonly across data originating in the real world. Hence, it should come as no surprise that some of the most popular applications of machine learning in the 21st century have come about as a direct byproduct of Geometric Deep Learning, perhaps sometimes without fully realising this fact. 
We would like to provide readers with an overview---by no means comprehensive---of influential works in Geometric Deep Learning and exciting and promising new applications. Our motivation is twofold: to demonstrate specific instances of scientific and industrial problems where the five geometric domains commonly arise, and to serve additional motivation for further study of Geometric Deep Learning principles and architectures.

\paragraph{Chemistry and Drug Design} One of the most promising applications of representation learning on graphs is in computational chemistry and \emph{drug development}.\marginnote{Many drugs are not designed but discovered, often serendipitously. The historic source of a number of drugs from the plant kingdom is reflected in their names: e.g., 
the acetylsalicylic acid, commonly known as {\em aspirin}, is contained in the bark of the willow tree ({\em Salix alba}), whose medicinal properties are known since antiquity.
} 
Traditional drugs are small molecules that are designed
to chemically attach (`bind') to some target molecule, typically a protein, in order to activate or disrupt some disease-related chemical process. 
Unfortunately, drug development is an extremely long and expensive process: 
at the time of writing, 
bringing a new drug to the market typically takes more than a decade and costs more than a billion dollars. 
One of the reasons is the cost of testing 
where many drugs fail at different stages -- less than 5\% of candidates make it to the last stage (see e.g. \cite{gaudelet2020utilising}).

Since the space of chemically synthesisable molecules is very large (estimated around $10^{60}$), the search for candidate molecules with the right combination of properties such as target binding affinity, low toxicity, solubility, etc. cannot be done experimentally, and {\em virtual} or {\em in silico screening} (i.e., the use of computational techniques to identify promising molecules), is employed. Machine learning techniques play an increasingly more prominent role in this task. 
A prominent example of the use of Geometric Deep Learning for virtual drug screening was recently shown by \citet{stokes2020deep} 
using a graph neural network trained to predict whether or not candidate molecules inhibit growth\marginnote{ \includegraphics[width=\linewidth]{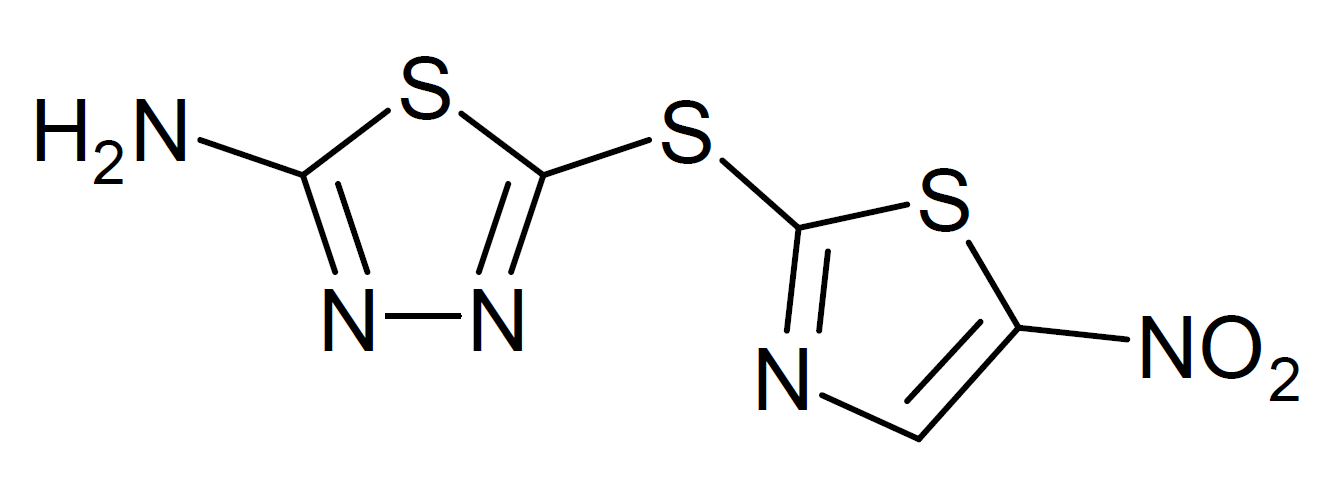}\\ Molecular graph of Halicin.} in the model bacterium \emph{Escherichia coli}, they were able to effectively discover that \emph{Halicin}, a molecule originally indicated for treating diabetes, is a highly potent antibiotic, even against bacteria strains with known antibiotic resistance.  This discovery was widely covered in both scientific and popular press.

Speaking more broadly, the application of graph neural networks to molecules modeled as graphs has been a very active field, with multiple specialised architectures proposed recently that are inspired by physics and e.g. incorporate equivariance to rotations and translations (see e.g. \cite{thomas2018tensor,anderson2019cormorant,fuchs2020se,satorras2021n}). Further, \citet{bapst2020unveiling} have successfully demonstrated the utility of GNNs for predictively modelling the dynamics of glass, in a manner that outperformed the previously available physics-based models.
Historically, many works in computational chemistry were precursors of modern graph neural network architectures sharing many common traits with them.

\paragraph{Drug Repositioning} 
%
While generating entirely novel drug candidates is a potentially viable approach, a faster and cheaper avenue for developing new therapies is \emph{drug repositioning}, which seeks to evaluate already-approved drugs (either alone or in combinations) for a novel purpose. 
This often significantly decreases the amount of clinical evaluation that is necessary to release the drug to the market. %
At some level of abstraction, the action of drugs on the body biochemistry and their interactions between each other and other biomolecules can be modeled as a graph, giving rise to the concept of `network medicine' coined by the prominent network scientist Albert-L{\'a}szl{\'o} Barab{\'a}si and advocating the use of biological networks (such as protein-protein interactions and metabolic pathways) to develop new therapies \citep{barabasi2011network}.

Geometric Deep Learning offers a modern take on this class of  approaches. A prominent early example is the work of \cite{zitnik2018modeling}, who used graph neural networks to predict side effects in a form of drug repositioning known as {\em combinatorial therapy} or {\em polypharmacy}, formulated as edge prediction in a drug-drug interaction graph. 
The novel coronavirus pandemic, which is largely ongoing at the time of writing this text, has sparked a particular interest in 
attempting to apply such approaches against COVID-19 \citep{gysi2020network}.
Finally, we should note that drug repositioning is not necessarily limited to synthetic molecules: \cite{veselkov2019hyperfoods} applied similar approaches to drug-like molecules contained in food (since, as we mentioned, many plant-based foods contain biological analogues of compounds used in oncological therapy). 
One of the authors of this text is involved in a collaboration adding a creative twist to this research, by partnering with a molecular chef that designs exciting recipes based on the  `hyperfood' ingredients rich in such drug-like molecules.

\paragraph{Protein biology} 
Since we have already mentioned proteins as drug targets, lets us spend a few more moments on this topic. 
Proteins are arguably among the most important biomolecules 
that have myriads of functions in our body, including protection against pathogens (antibodies), giving structure to our skin (collagen), transporting oxygen to cells (haemoglobin), catalysing chemical reactions (enzymes), and signaling (many hormones are proteins). 
Chemically speaking, a protein is a biopolymer, or a chain of small building blocks called {\em aminoacids} that under the influence of electrostatic forces fold into a complex 3D structure. It is this structure that endows the protein with its functions,\marginnote{A common metaphor, dating back to the chemistry Nobel laureate Emil Fischer is the {\em Schl{\"u}ssel-Schloss-Prinzip} (`key-lock principle', 1894): two proteins often only interact if they have geometrically and chemically complementary structures. } and hence it is crucial to the understanding of how proteins work and what they do. Since proteins are common targets for drug therapies, 
the pharmaceutical industry has a keen interest in this field.


A typical hierarchy of problems in protein bioinformatics is going from protein {\em sequence} (a 1D string over an alphabet of of 20 different amino acids) to 3D {\em structure} (a problem known as `protein folding') to {\em function} (`protein function prediction'). 
Recent approaches such as DeepMind's AlphaFold by  \cite{senior2020improved} used {\em contact graphs} to represent the protein structure. \cite{gligorijevic2020structure} showed that applying graph neural networks on such graphs allows to achieve better function prediction than using purely sequence-based methods.

\cite{gainza2020deciphering} developed\marginnote{ \includegraphics[width=\linewidth]{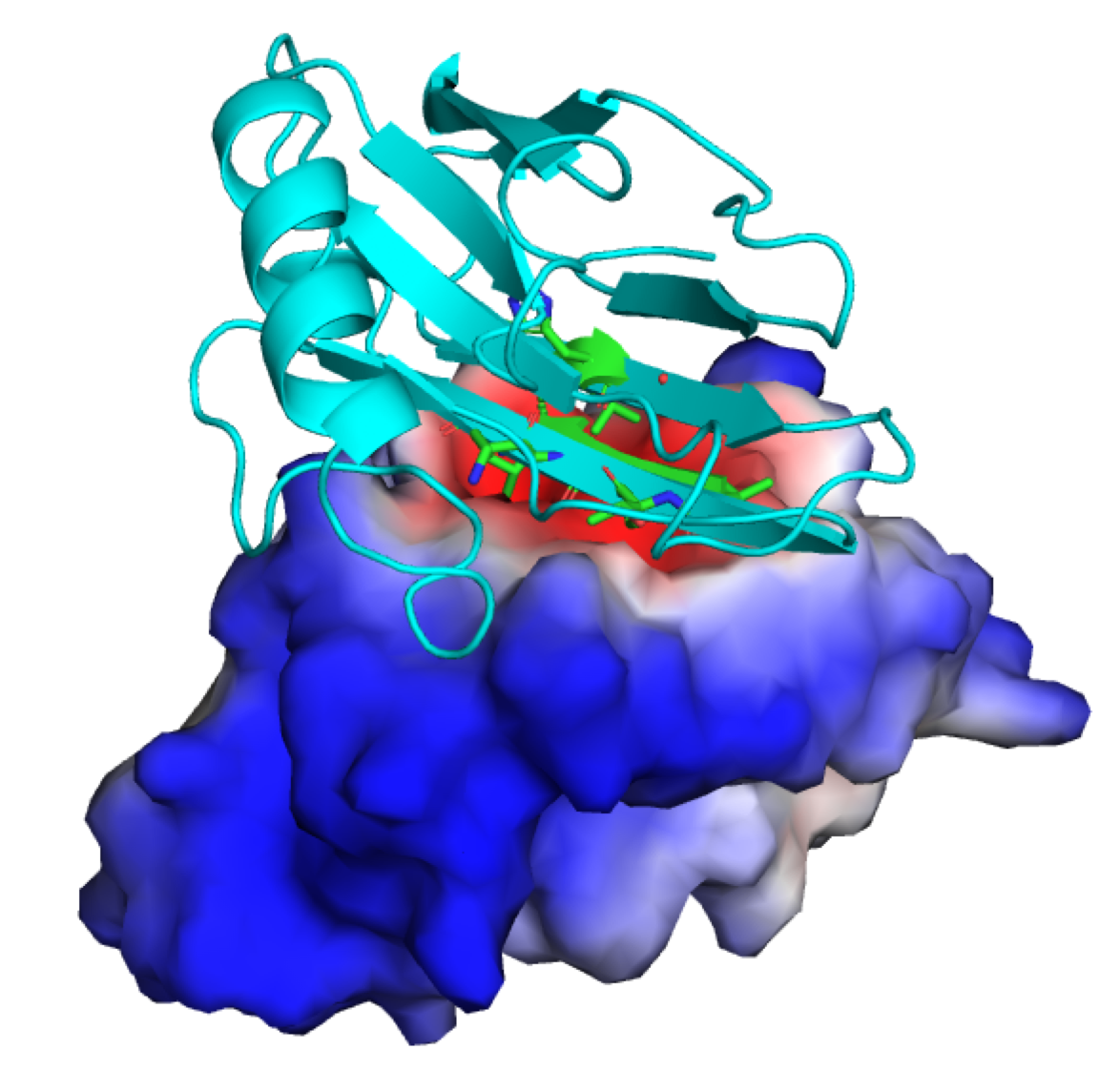}\\ Oncologial target PD-L1 protein surface (heat map indicated the predicted binding site) and the designed binder (shown as ribbon diagram).} a Geometric Deep Learning pipeline called MaSIF predicting interactions between proteins from their 3D structure. MaSIF models the protein as a molecular surface discretised as a mesh, arguing that this representation is advantageous when dealing with interactions as it allows to abstract the internal fold structure. The architecture was based on mesh convolutional neural network operating on pre-computed chemical and geometric features in small local geodesic patches. %
The network was trained using a few thousand co-crystal protein 3D structures from the Protein Data Bank to address multiple tasks, including interface prediction, ligand classification, and docking, and allowed to do {\em de novo} (`from scratch') design of proteins that could in principle act as biological immunotherapy drug against cancer -- such proteins are designed to inhibit protein-protein interactions (PPI) between parts of the programmed cell death protein complex (PD-1/PD-L1) and give the immune system the ability to attack the tumor cells.


\paragraph{Recommender Systems and Social Networks} The first popularised large-scale applications of graph representation learning have occurred within \emph{social networks}\marginnote{\includegraphics[width=\linewidth]{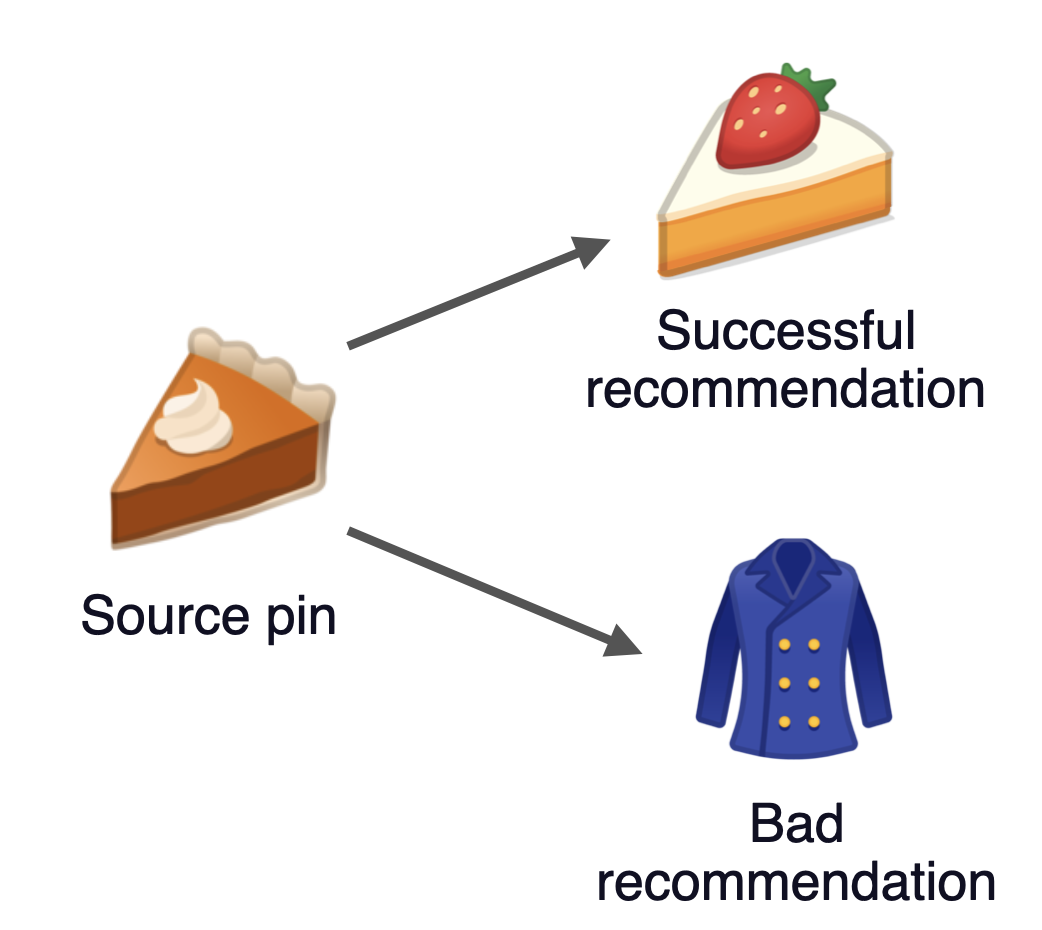}}, primarily in the context of \emph{recommender systems}. Recommenders are tasked with deciding which content to serve to users, potentially depending on their previous history of interactions on the service. This is typically realised through a \emph{link prediction} objective: supervise the embeddings of various nodes (pieces of content) such that they are kept close together if they are deemed \emph{related} (e.g. commonly viewed together). Then the \emph{proximity} of two embeddings (e.g. their inner product) can be interpreted as the probability that they are linked by an edge in the content graph, and hence for any content queried by users, one approach could serve its $k$ nearest neighbours in the embedding space.

Among the pioneers of this methodology is the American image sharing and social media company Pinterest: besides presenting one of the first successful deployments of GNNs in production, their method, PinSage\marginnote{Pinterest had also presented follow-up work, PinnerSage \citep{pal2020pinnersage}, which effectively integrates user-specific contextual information into the recommender.}, successfully made graph representation learning \emph{scalable} to graphs of millions of nodes and billions of edges \citep{ying2018graph}. Related applications, particularly in the space of product recommendations, soon followed. Popular GNN-backed recommenders that are currently deployed in production include Alibaba's Aligraph \citep{zhu2019aligraph} and Amazon's P-Companion \citep{hao2020p}. In this way, graph deep  learning is influencing millions of people on a daily level.

Within the context of content analysis on social networks, another noteworthy effort is Fabula AI, which is among the first GNN-based startups to be acquired (in 2019, by Twitter). The startup, founded by one of the authors of the text and his team, developed novel technology for detecting misinformation on social networks \citep{monti2019fake}. Fabula's solution consists of modelling the spread of a particular news item by the network of users who shared it. The users are connected if one of them re-shared the information from the other, but also if they follow each other on the social network. This graph is then fed into a graph neural network, which classifies the entire graph as either `true' or `fake' content -- with labels based on agreement between fact-checking bodies. Besides demonstrating strong predictive power which stabilises quickly (often within a few hours of the news spreading), analysing the embeddings of individual user nodes revealed clear clustering of users who tend to share incorrect information, exemplifying the well-known \emph{`echo chamber'} effect.

\paragraph{Traffic forecasting} Transportation networks are another area\marginnote{\includegraphics[width=\linewidth]{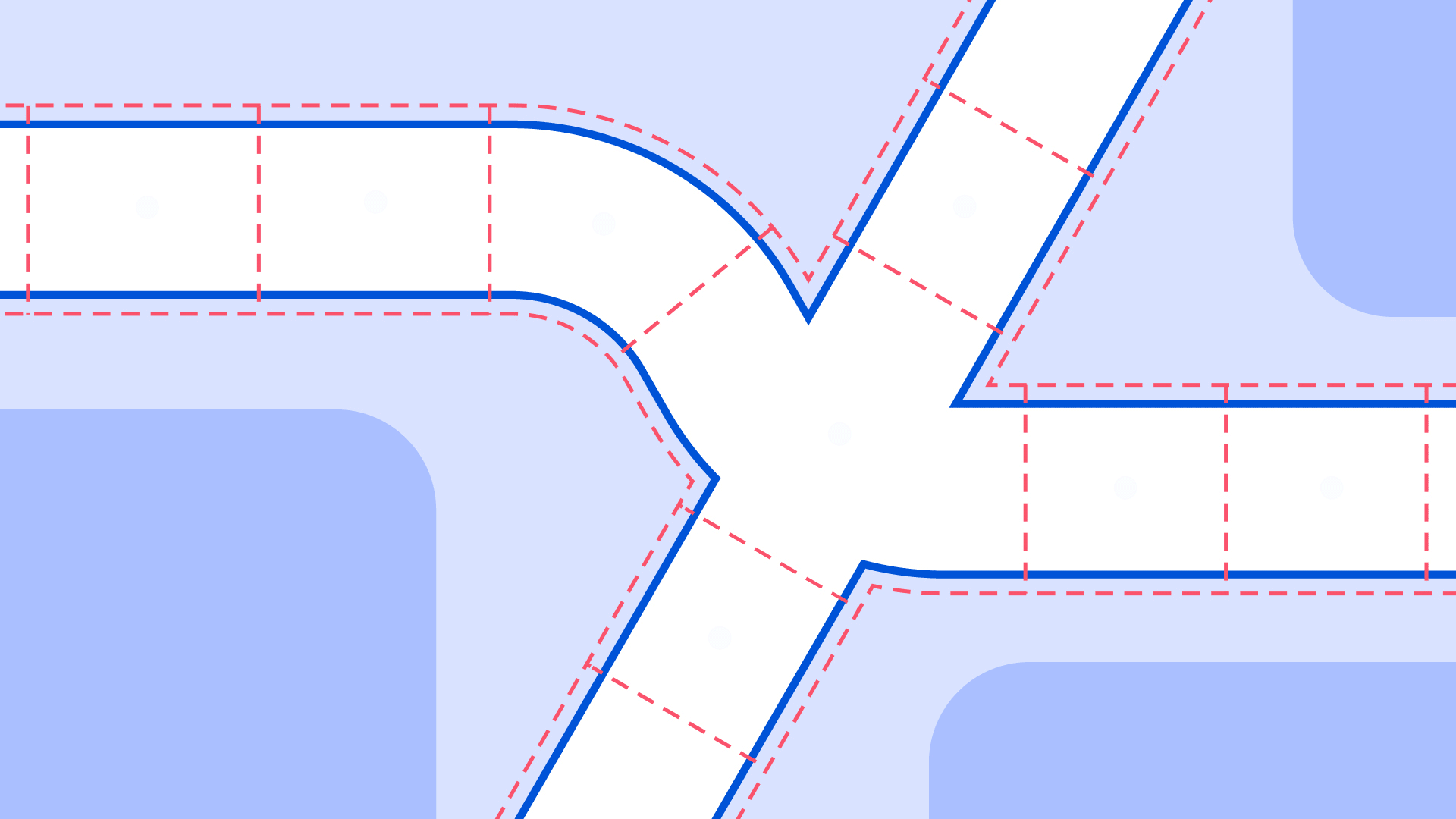}\\ \includegraphics[width=\linewidth]{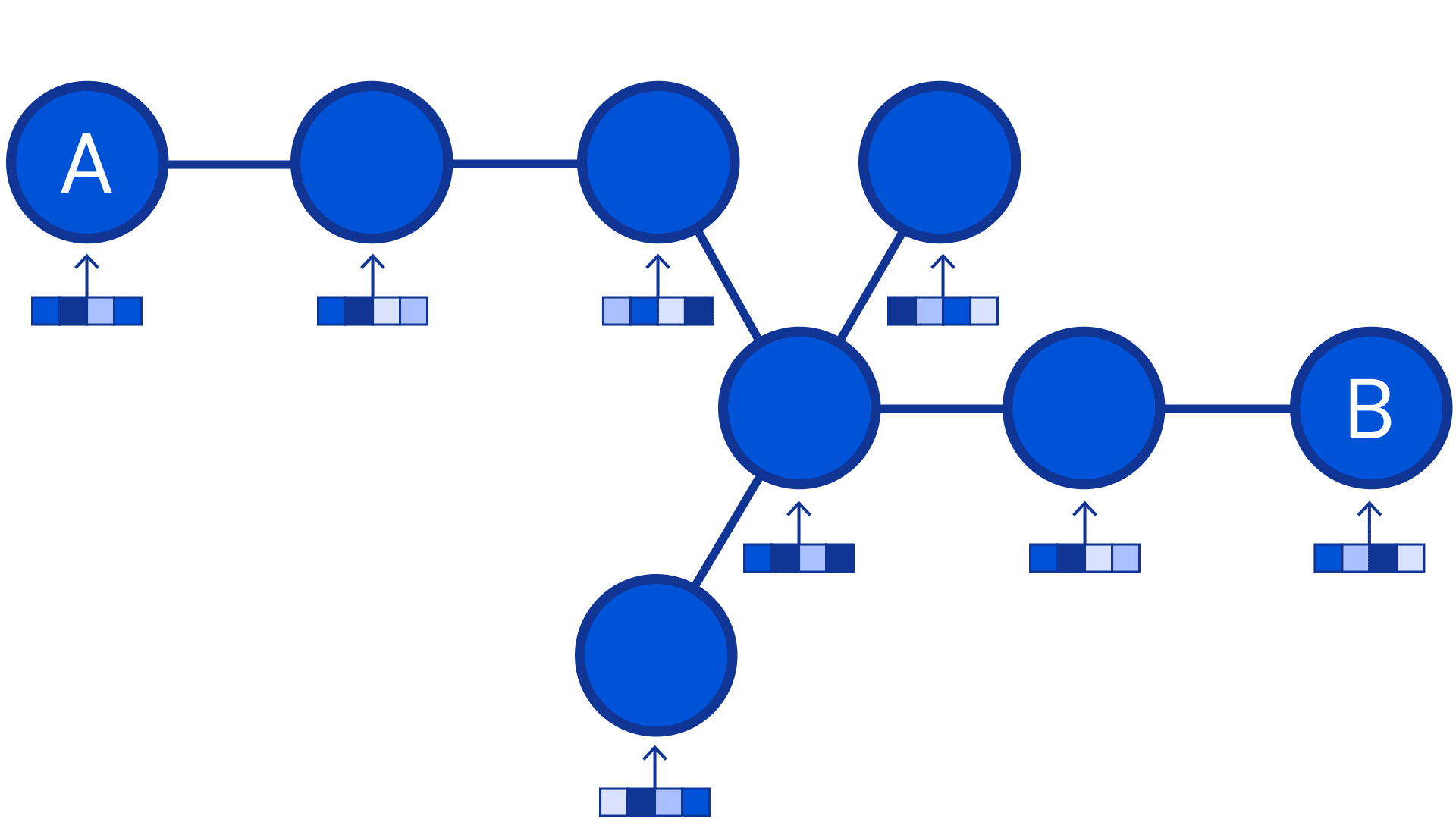}\\ A road network (top) with its corresponding graph representation (bottom).} where Geometric Deep Learning techniques are already making an actionable impact over billions of users worldwide. For example, on road networks, we can observe intersections as nodes, and road segments as edges connecting them---these edges can then be featurised by the road length, current or historical speeds along their segment, and the like. 

One standard prediction problem in this space is predicting the \emph{estimated time of arrival} (ETA): for a given candidate route, providing the expected travel time necessary to traverse it. Such a problem is essential in this space, not only for user-facing traffic recommendation apps, but also for enterprises (such as food delivery or ride-sharing services) that leverage these predictions within their own operations.

Graph neural networks\marginnote{\includegraphics[width=\linewidth]{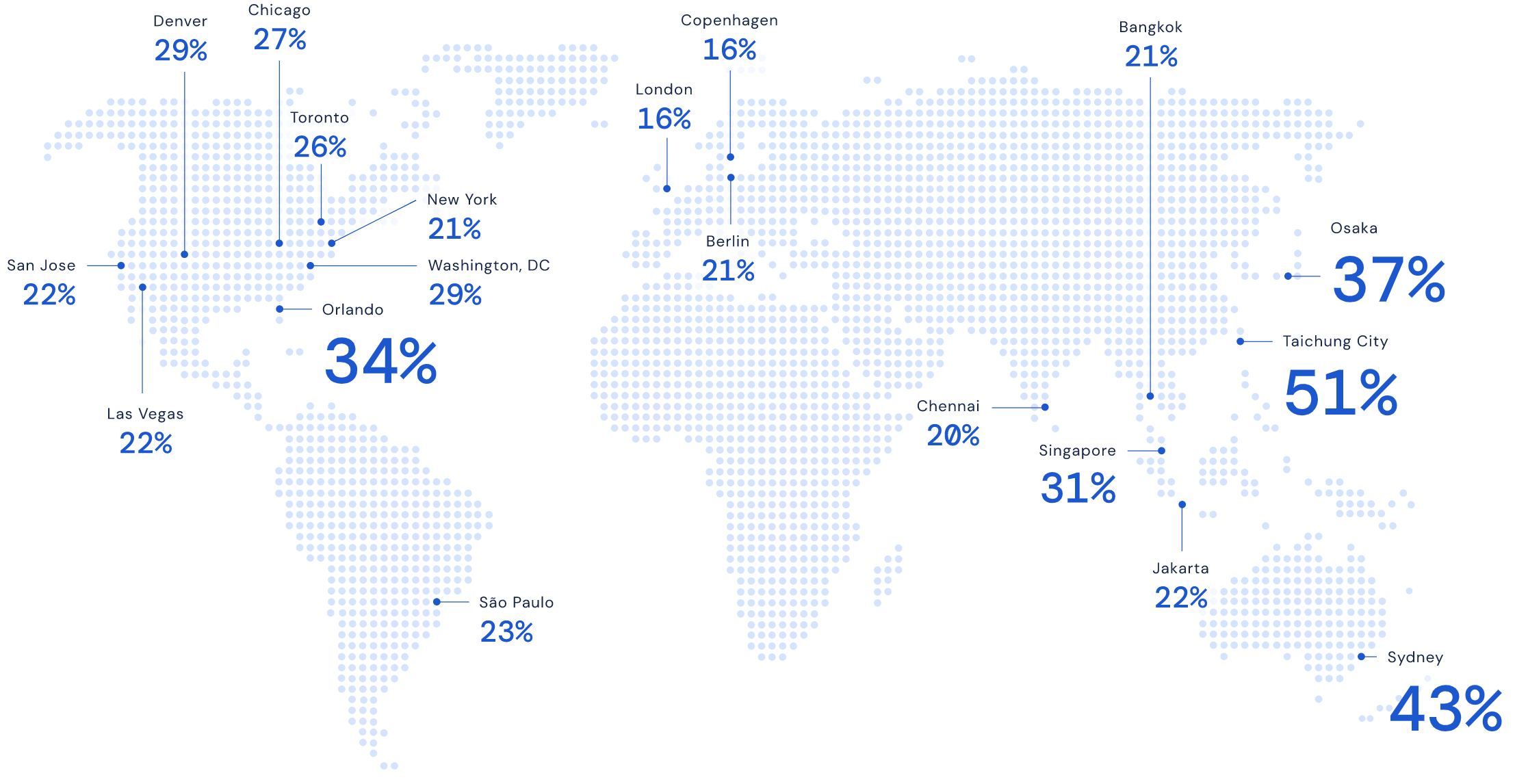}\\ Several of the metropolitan areas where GNNs are serving queries within Google Maps, with indicated relative improvements in prediction quality (40+\% in cities like Sydney).} have shown immense promise in this space as well: they can, for example, be used to directly predict the ETA for a relevant subgraph of the road network (effectively, a \emph{graph regression} task). Such an approach was successfully leveraged by DeepMind, yielding a GNN-based ETA predictor which is now deployed in production at Google Maps \citep{derrowpinion2021traffic}, serving ETA queries in several major metropolitan areas worldwide. Similar returns have been observed by the Baidu Maps team, where travel time predictions are currently served by the ConSTGAT model, which is itself based on a spatio-temporal variant of the graph attention network model \citep{fang2020constgat}.

\paragraph{Object recognition} A principal benchmark for machine learning\marginnote{\includegraphics[width=\linewidth]{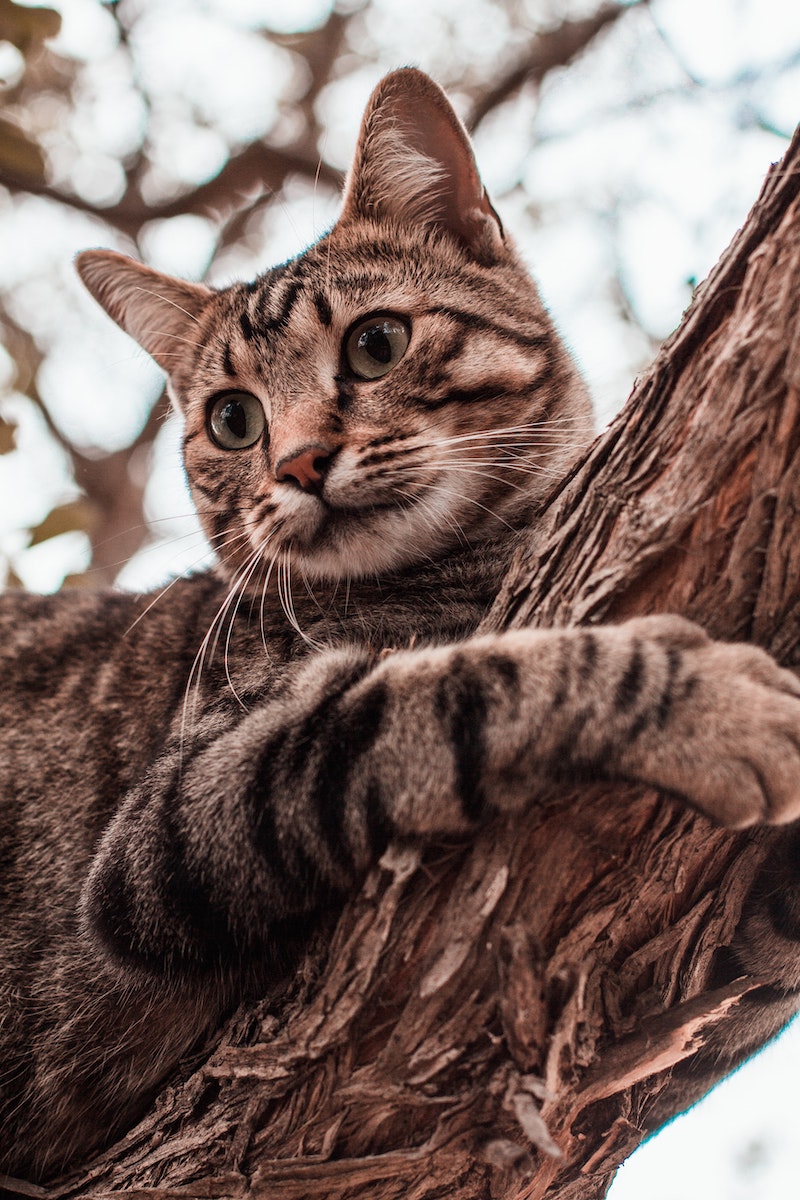}\\ One example input image, the likes of which can be found in ImageNet, representing the \emph{``tabby cat''} class.} techniques in computer vision is the ability to \emph{classify} a central object within a provided image. The ImageNet large scale visual recognition challenge \citep[ILSVRC]{russakovsky2015imagenet} was an annual object classification challenge that propelled much of the early development in Geometric Deep Learning. ImageNet requires models to classify realistic images scraped from the Web into one of 1000 categories: such categories are at the same time diverse (covering both animate and inanimate objects), and specific (with many classes focused on distinguishing various cat and dog breeds). Hence, good performance on ImageNet often implies a solid level of feature extraction from general photographs, which formed a foundation for various \emph{transfer learning} setups from pre-trained ImageNet models.

The success of convolutional neural networks on ImageNet---particularly the AlexNet model of \citet{krizhevsky2012imagenet}, which swept ILSVRC 2012 by a large margin---has in a large way spearheaded the adoption of deep learning as a whole, both in academia and in industry. Since then, CNNs have consistently ranked on top of the ILSVRC, spawning many popular architectures such as VGG-16 \citep{simonyan2014very}\marginnote{Interestingly, the VGG-16 architecture has sixteen convolutional layers and is denoted as ``very deep'' by the authors. Subsequent developments quickly scaled up such models to hundreds or even thousands of layers.}, Inception \citep{szegedy2015going} and ResNets \citep{he2016deep}, which have successfully surpassed human-level performance on this task. The design decisions and regularisation techniques employed by these architectures (such as rectified linear activations \citep{nair2010rectified}, dropout \citep{srivastava2014dropout}, skip connections \citep{he2016deep} and batch normalisation \citep{ioffe2015batch}) form the backbone of many of the effective CNN models in use today.

Concurrently with object classification, significant progress had been made on object \emph{detection}; that is, isolating all objects of interest within an image, and tagging them with certain classes. Such a task is relevant in a variety of downstream problems, from image captioning all the way to autonomous vehicles. It necessitates a more fine-grained approach, as the predictions need to be \emph{localised}; as such, often, translation equivariant models have proven their worth in this domain. One impactful example in this space includes the R-CNN family of models \citep{girshick2014rich,girshick2015fast,ren2015faster,he2017mask} whereas, in the related field of \emph{semantic segmentation}, the SegNet model of \citet{badrinarayanan2017segnet} proved influential, with its encoder-decoder architecture relying on the VGG-16 backbone.

\paragraph{Game playing} Convolutional neural networks also play a prominent role as translation-invariant feature extractors in \emph{reinforcement learning} (RL) environments, whenever the observed state can be represented in a grid domain; e.g. this is the case when learning to play video games from pixels. 
In this case, the CNN is responsible for reducing the input to a flat vector representation, which is then used for deriving {\em policy} or {\em value functions} that drive the RL agent's behaviour. While the specifics of reinforcement learning are not the focus of this section, we do note that some of the most impactful results of deep learning in the past decade have come about through CNN-backed reinforcement learning.

One particular example that is certainly worth mentioning here is DeepMind's \emph{AlphaGo} \citep{silver2016mastering}. It encodes the current state within a game of Go by applying a CNN to the $19\times 19$ grid representing the current positions of the placed stones. Then, through a combination of learning from previous expert moves, Monte Carlo tree search, and self-play, it had successfully reached a level of Go mastery that was sufficient to outperform Lee Sedol, one of the strongest Go players of all time, in a five-round challenge match that was widely publicised worldwide. 

While this already represented a significant milestone for broader artificial intelligence---with Go having a substantially more complex state-space than, say, chess\marginnote{\includegraphics[width=\linewidth]{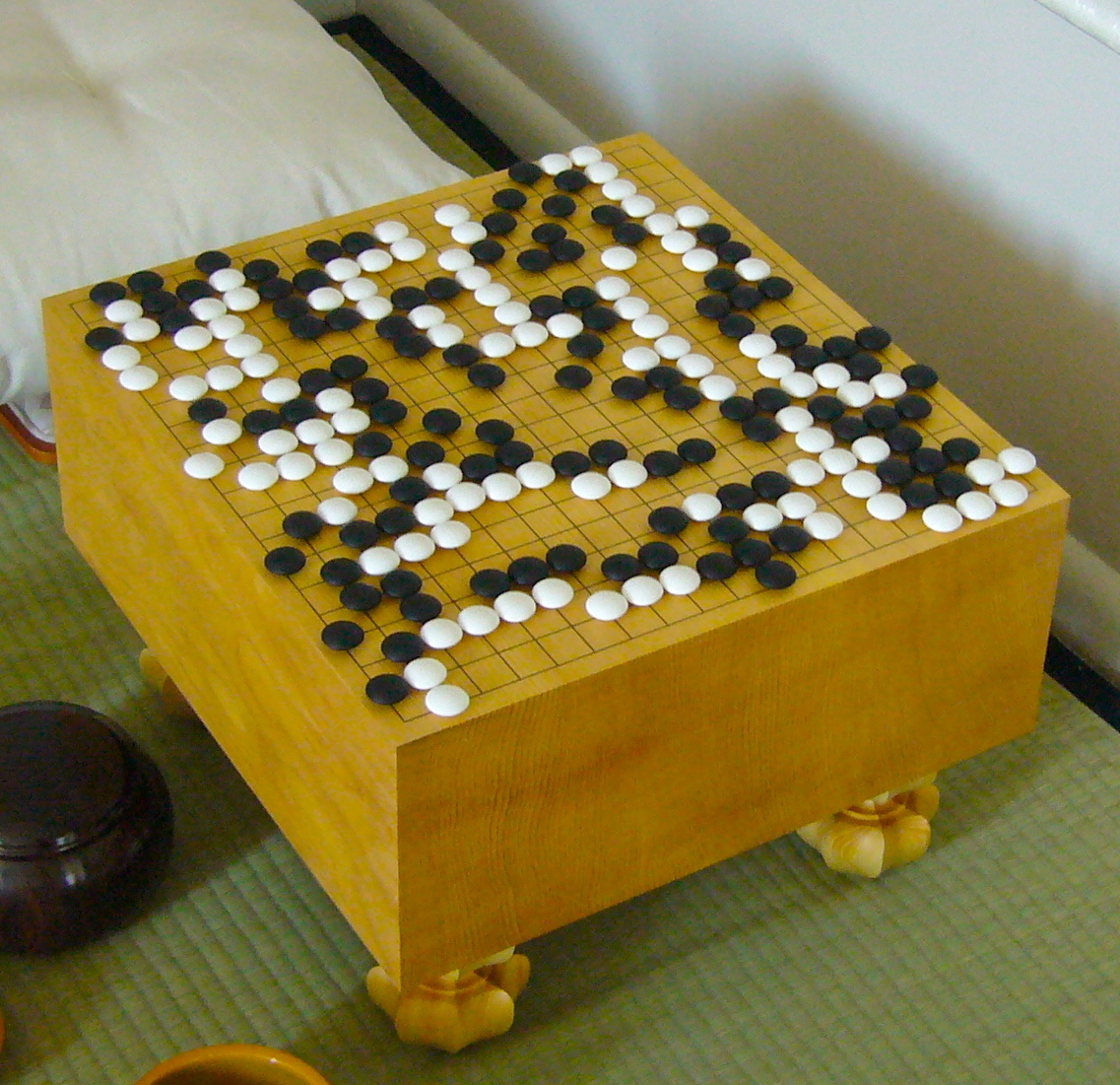} The game of Go is played on a $19\times 19$ board, with two players placing white and black \emph{stones} on empty fields. The number of legal states has been estimated at $\approx2\times 10^{170}$ \citep{tromp2006combinatorics}, vastly outnumbering the number of atoms in the universe.}---the development of AlphaGo did not stop there. The authors gradually removed more and more Go-specific biases from the architecture, with \emph{AlphaGo Zero} removing human biases, optimising purely through self-play \citep{silver2017mastering}, \emph{AlphaZero} expands this algorithm to related two-player games, such as Chess and Shogi; lastly, \emph{MuZero} \citep{schrittwieser2020mastering} incorporates a model that enables \emph{learning} the rules of the game on-the-fly, which allows reaching strong performance in the Atari 2600 console, as well as Go, Chess and Shogi, without any upfront knowledge of the rules. Throughout all of these developments, CNNs remained the backbone behind these models' representation of the input.

While several high-performing RL agents were proposed for the Atari 2600 platform over the years \citep{mnih2015human,mnih2016asynchronous,schulman2017proximal}, for a long time they were unable to reach human-level performance on \emph{all} of the 57 games provided therein. This barrier was finally broken with Agent57 \citep{badia2020agent57}, which used a parametric family of policies, ranging from strongly exploratory to purely exploitative, and prioritising them in different ways during different stages of training. It, too, powers most of its computations by a CNN applied to the video game's framebuffer.

\paragraph{Text and speech synthesis} Besides images (which naturally map to a \emph{two}-dimensional grid), several of (geometric) deep learning's strongest successes have happened on one-dimensional grids. Natural examples of this are \emph{text} and \emph{speech}, folding the Geometric Deep Learning blueprint within diverse areas such as natural language processing and digital signal processing.

Some of the most widely applied and publicised works in this space focus on \emph{synthesis}: being able to generate speech or text, either unconditionally or conditioned on a particular \emph{prompt}. Such a setup can support a plethora of useful tasks, such as  \emph{text-to-speech} (TTS), predictive text completion, and machine translation. Various neural architectures for text and speech generation have been proposed over the past decade, initially mostly based on \emph{recurrent} neural networks (e.g. the aforementioned seq2seq model \citep{sutskever2014sequence} or recurrent attention \citep{bahdanau2014neural}). However, in recent times, they have been gradually replaced by convolutional neural networks and Transformer-based architectures.

One particular limitation of simple 1D convolutions in this setting is their linearly growing \emph{receptive field}, requiring many layers in order to cover the sequence generated so far. \emph{Dilated}\marginnote{Dilated convolution is also referred to as \emph{\`{a} trous} convolution (literally \emph{``holed''} in French).} convolutions, instead, offer an \emph{exponentially} growing receptive field with an equivalent number of parameters. Owing to this, they proved a very strong alternative, eventually becoming competitive with RNNs on machine translation \citep{kalchbrenner2016neural}, while drastically reducing the computational complexity, owing to their parallelisability across all input positions.\marginnote{Such techniques have also outperformed RNNs on problems as diverse as protein-protein interaction \citep{deac2019attentive}.} The most well-known application of dilated convolutions is the \emph{WaveNet} model from \cite{oord2016wavenet}. WaveNets demonstrated that, using dilations, it is possible to synthesise speech at the level of \emph{raw waveform} (typically 16,000 samples per second or more), producing speech samples that were significantly more ``human-like'' than the best previous text-to-speech (TTS) systems\marginnote{Besides this, the WaveNet model proved capable of generating piano pieces.}. Subsequently, it was further demonstrated that the computations of WaveNets can be distilled in a much simpler model, the \emph{WaveRNN} \citep{kalchbrenner2018efficient}---and this model enabled effectively deploying this technology at an industrial scale. This allowed not only its deployment for large-scale speech generation for services such as the Google Assistant, but also allowing for efficient on-device computations; e.g. for Google Duo, which uses end-to-end encryption.

Transformers \citep{vaswani2017attention} have managed to surpass the limitations of both recurrent and convolutional architectures, showing that \emph{self-attention} is sufficient for achieving state-of-the-art performance in machine translation. Subsequently, they have revolutionised natural language processing. Through the pre-trained embeddings provided by models such as BERT \citep{devlin2018bert}, Transformer computations have become enabled for a large amount of downstream applications of natural language processing---for example, Google uses BERT embeddings to power its search engine. 

Arguably the most widely publicised application of Transformers in the past years is text generation, spurred primarily by the \emph{Generative Pre-trained Transformer} (GPT, \cite{radford2018improving,radford2019language,brown2020language}) family of models from OpenAI. In particular, GPT-3 \citep{brown2020language} successfully scaled language model learning to 175 billion learnable parameters, trained on next-word prediction on web-scale amounts of scraped textual corpora. This allowed it not only to become a highly-potent few-shot learner on a variety of language-based tasks, but also a text generator with capability to produce coherent and human-sounding pieces of text. This capability not only implied a large amount of downstream applications, but also induced a vast media coverage.

\paragraph{Healthcare} 
Applications in the medical domain are another promising area for Geometric Deep Learning. There are multiple ways in which these methods are being used. 
First, more traditional architectures such as CNNs have been applied to grid-structured data, for example, for the 
prediction of length of stay in Intensive Care Units
\citep{rocheteau2020temporal}, 
or diagnosis of sight-threatening diseases from retinal scans \citep{de2018clinically}. 
\cite{winkels2019pulmonary} showed that using 3D roto-translation group convolutional networks improves 
the accuracy of pulmonary nodule detection compared to conventional CNNs.

Second, modelling organs as geometric surfaces, mesh convolutional neural networks were shown to be able to address a diverse range of tasks, from reconstructing facial structure from genetics-related information \citep{mahdi20203d} to 
brain cortex parcellation \citep{cucurull2018convolutional} to 
regressing demographic properties from cortical surface structures \citep{besson2020geometric}. 
%
%
The latter examples represent an increasing trend in  neuroscience to consider the brain as a surface with complex folds\marginnote{Such structure of the brain cortex are called {\em sulci} and {\em gyri} in anatomical literature.} giving rise to highly non-Euclidean structures. 

At the same time, 
neuroscientists often try construct and analyse {\em functional networks} of the brain representing the various regions of the brain that are activated together when performing some cognitive function; these networks are often constructed using functional magnetic resonance imaging (fMRI) that shows in real time which areas of the brain consume more blood.\marginnote{Typically, Blood Oxygen-Level Dependent (BOLD) contrast imaging is used. } 
These functional networks can reveal patient demographics (e.g., telling apart males from females, \cite{arslan2018graph}), as well as used for neuropathology diagnosis, which is the third area of application of Geometric Deep Learning in medicine we would like to highlight here. In this context, \cite{ktena2017distance} pioneered the use of graph neural networks for the prediction of neurological conditions such as Autism Spectrum Disorder. 
The geometric and functional structure of the brain appears to be intimately related, and recently \cite{itani2021combining} pointed to the benefits of exploiting them jointly in neurological disease analysis.  

Fourth, {\em patient networks} are becoming more prominent in ML-based medical diagnosis. The rationale behind these methods is that the information of patient demographic, genotypic, and phenotypic similarity could improve predicting their disease.  
 \cite{parisot2018disease} 
 applied graph neural networks on networks of patients created from demographic features for neurological disease diagnosis, showing that the use of the graph improves prediction results.
\cite{cosmo2020latent} showed the benefits of latent graph learning (by which the network {\em learns} an unknown patient graph) in this setting.  The latter work used data from the UK Biobank, a large-scale collection of medical data including brain imaging \citep{miller2016multimodal}.

A wealth of data about hospital patients may be found in \emph{electronic health records} (EHRs)\marginnote{Publicly available anonymised critical-care EHR datasets include MIMIC-III \citep{johnson2016mimic} and eICU \citep{pollard2018eicu}.}. Besides giving a comprehensive view of the patient's progression, EHR analysis allows for \emph{relating} similar patients together. This aligns with the \emph{pattern recognition method}, which is commonly used in diagnostics. Therein, the clinician uses \emph{experience} to recognise a pattern of clinical characteristics, and it may be the primary method used when the clinician's experience may enable them to diagnose the condition quickly. Along these lines, several works attempt to construct a patient graph based on EHR data, either by analysing the embeddings of their doctor's notes \citep{malone2018learning}, diagnosis similarity on admission \citep{rocheteau2021predicting}, or even assuming a fully-connected graph \citep{zhu2019graph}. In all cases, promising results have been shown in favour of using graph representation learning for processing EHRs.

\paragraph{Particle physics and astrophysics} 
High energy physicists were perhaps among the first domain experts in the field of natural sciences to embrace the new shiny tool, graph neural networks. 
In a recent review paper, \cite{shlomi2020graph}\marginnote{\includegraphics[width=\linewidth]{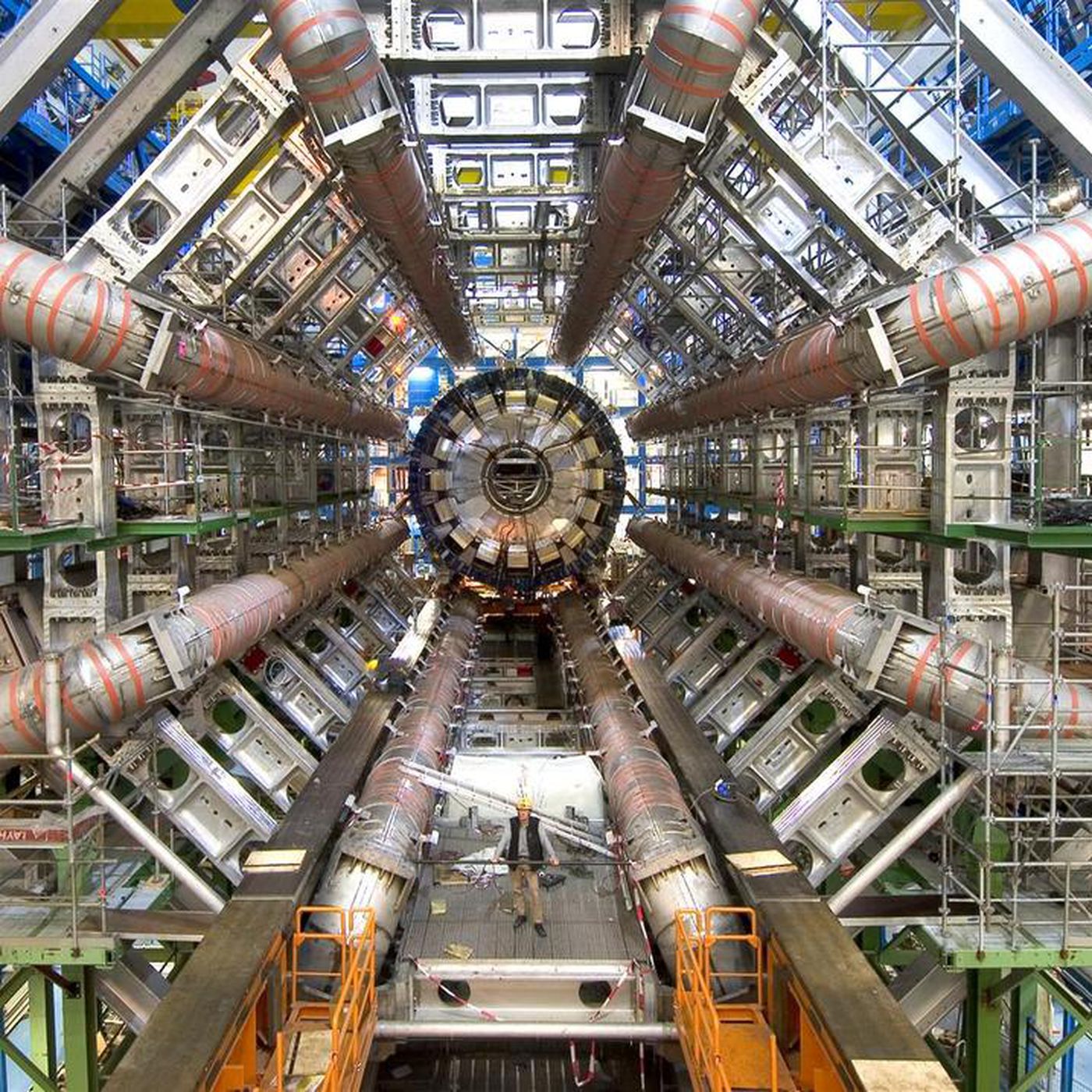} 
Part of the Large Hadron Collider detectors. 
} note that machine learning has historically been heavily used in particle physics experiments, either to learn complicated inverse functions allowing to infer the underlying physics process from the information measured in the detector, or to perform classification and regression tasks. For the latter, it was often necessary to force the data into an unnatural representation such as grid, in order to be able to used standard deep learning architectures such as CNN. Yet, many problems in physics involve data in the form of unordered sets with rich relations and interactions, which can be naturally represented as graphs.

One important application in high-energy physics is the reconstruction and classification of {\em particle jets} -- sprays of stable particles 
arising from multiple successive interaction and decays of particles originating from a single
initial event. 
In the Large Hardon Collider, the largest and best-known particle accelerator built at CERN, such jet are the result of collisions of protons at nearly the speed of light. 
These collisions produce massive particles, such as the long though-for Higgs boson or the top quark.  
%
%
The identification and classification of collision events is of crucial importance, as it might provide experimental evidence to the existence of new particles.

Multiple Geometric Deep Learning approaches\marginnote{\includegraphics[width=\linewidth]{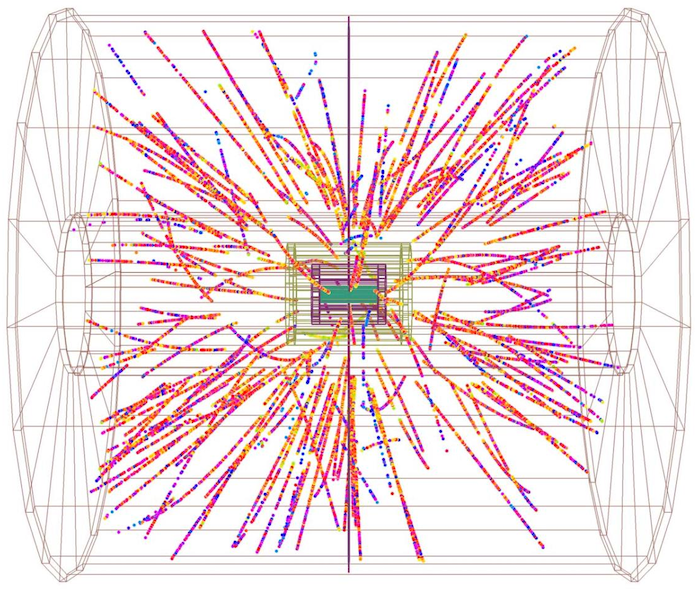} 
Example of a particle jet. 
} have recently been proposed for particle jet classification task, e.g. by \cite{komiske2019energy} and \cite{qu1902particlenet},  based on DeepSet and Dynamic Graph CNN architectures, respectively. 
%
More recently, there has also been interest in developing specialsed architectures derived from physics consideration and incorporating inductive biases consistent with Hamiltonian or Lagrangian mechanics (see e.g. \cite{sanchez2019hamiltonian,cranmer2020lagrangian}), equivariant to the Lorentz group (a fundamental symmetry of space and time in physics) \citep{bogatskiy2020lorentz}, or even incorporating symbolic reasoning \citep{cranmer2019learning} and capable of learning physical laws from data. Such approaches are more interpretable (and thus considered more `trustworthy' by domain experts) and also offer better generalisation.

Besides particle accelerators,
 particle detectors are now being used by astrophysicist for {\em multi-messenger astronomy} --  a new way of coordinated observation of disparate signals, such as electromagnetic radiation, gravitational waves, and neutrinos, coming from the same source. 
Neutrino astronomy is of particular interest, since neutrinos interact only very rarely with matter, and thus travel enormous distances practically unaffected.\marginnote{\includegraphics[width=\linewidth]{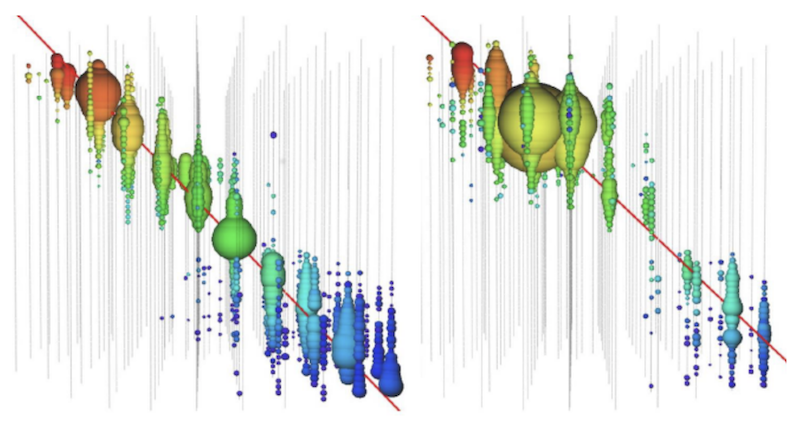} The characteristic pattern of light deposition in IceCube detector from background events (muon bundles, left)
and astrophysical neutrinos (high-energy single muon, right). \cite{choma2018graph}
} Detecting neutrinos allows to observe objects inaccessible to optical telescopes, but requires enormously-sized detectors -- the IceCube neutrino observatory uses a cubic kilometer of Antarctic ice shelf on the South Pole as its detector. 
Detecting high-energy neutrinos can possibly shed lights on some of the most mysterious objects in the Universe, such as blazars and black holes. 
\cite{choma2018graph} used a Geometric neural network to model the irregular geometry of the IceCube neutrino detector, showing significantly better performance in detecting neutrinos coming from astrophysical sources and separating them from background events.

While neutrino astronomy offers a big promise in the study of the Cosmos, traditional optical and radio telescopes are still the `battle horses' of astronomers. 
With these traditional instruments, Geometric Deep Learning can still offer new methodologies for data analysis. For example, \cite{scaife2021fanaroff} used rotationally-equivariant CNNs for the classification of radio galaxies, and \cite{mcewen2021scattering} used 
spherical CNNs for the analysis of cosmic microwave background radiation, a relic from the Big Bang that might shed light on the formation of the primordial Universe. As we already mentioned, such signals are naturally represented on the sphere and equivariant neural networks are an appropriate tool to study them.




\paragraph{Virtual and Augmented Reality} 
Another field of applications which served as the motivation for the development of a large class of Geometric Deep Learning methods is computer vision and graphics, in particular, dealing with 3D body models for virtual and augmented reality. 
Motion capture technology used to produce special effects in movies like Avatar often operates in two stages: first, the input from a 3D scanner capturing the motions of the body or the face of the actor is put into correspondence with some canonical shape, typically modelled as a discrete manifold or a mesh (this problem is often called `analysis'). Second, a new shape is generated to repeat the motion of the input (`synthesis').  
Initial works on Geometric Deep Learning in computer graphics and vision \citep{masci2015geodesic,boscaini2016learning,monti2017geometric} developed mesh convolutional neural networks to address the analysis problem, or more specifically, deformable shape correspondence.

First geometric autoencoder architectures for 3D shape synthesis were proposed independently by \cite{litany2018deformable} and  \cite{ranjan2018generating}. In these architectures, a canonical mesh (of the body, face, or hand) was assumed to be known and the synthesis task consisted of regressing the 3D coordinates of the nodes (the embedding of the surface, using the jargon of differential geometry). 
\cite{kulon2020weakly} showed \marginnote{ \includegraphics[width=\linewidth]{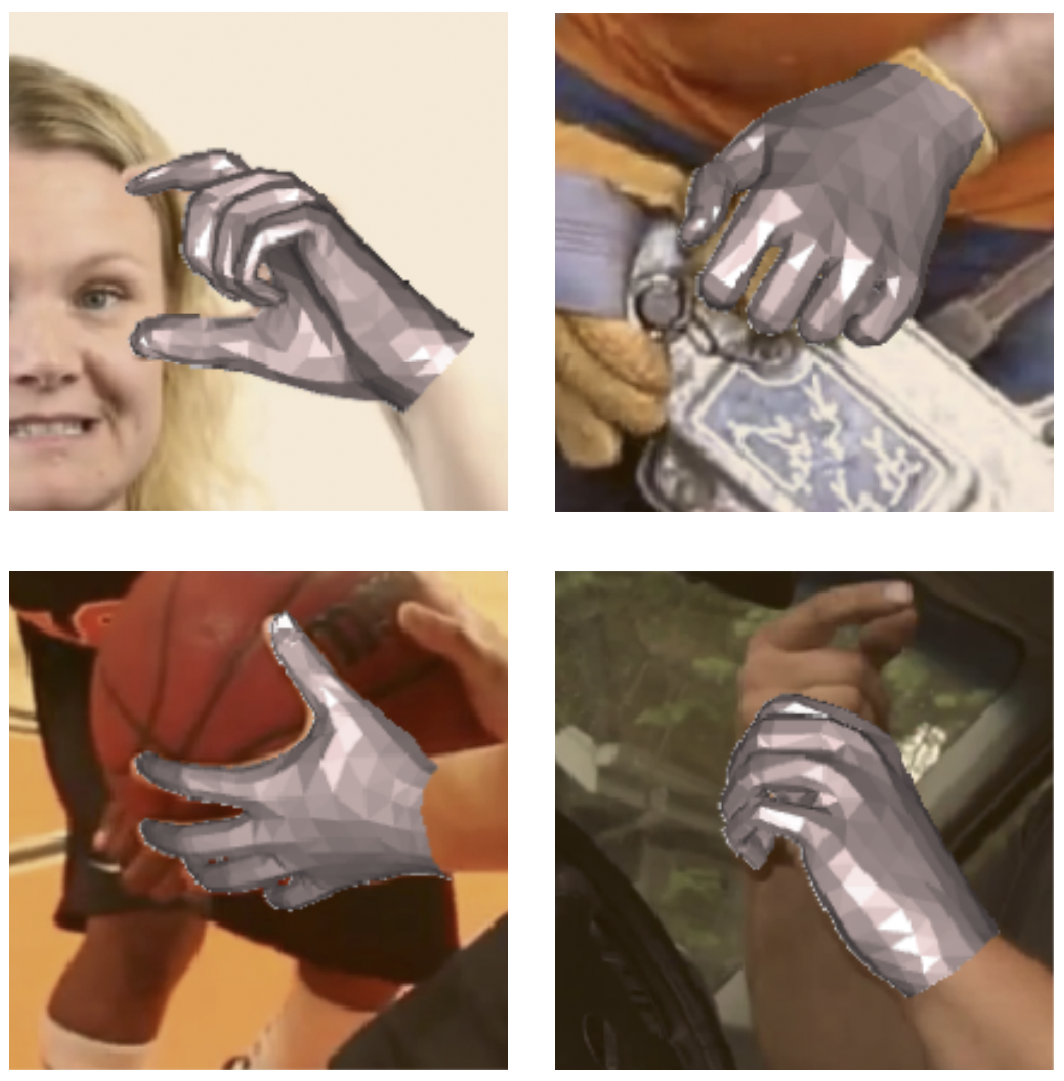}\\ Examples of complex 3D hand poses reconstructed from 2D images in the wild \citep{kulon2020weakly}.}a hybrid pipeline for 3D hand pose estimation with an image CNN-based encoder and a geometric decoder. A demo of this system, developed in collaboration with a British startup company Ariel AI and presented at CVPR 2020, allowed to create realistic body avatars with fully articulated hands from video input on a mobile phone faster than real-time.  %
Ariel AI was acquired by Snap in 2020, and at the time of writing its technology is used in Snap's augmented reality products. 

\section{Historic Perspective} 

``Symmetry, as wide or as narrow as you may define its meaning, is one idea by which man through the ages has tried to comprehend and create order, beauty, and perfection.'' \marginnote{
\includegraphics[width=0.9\linewidth]{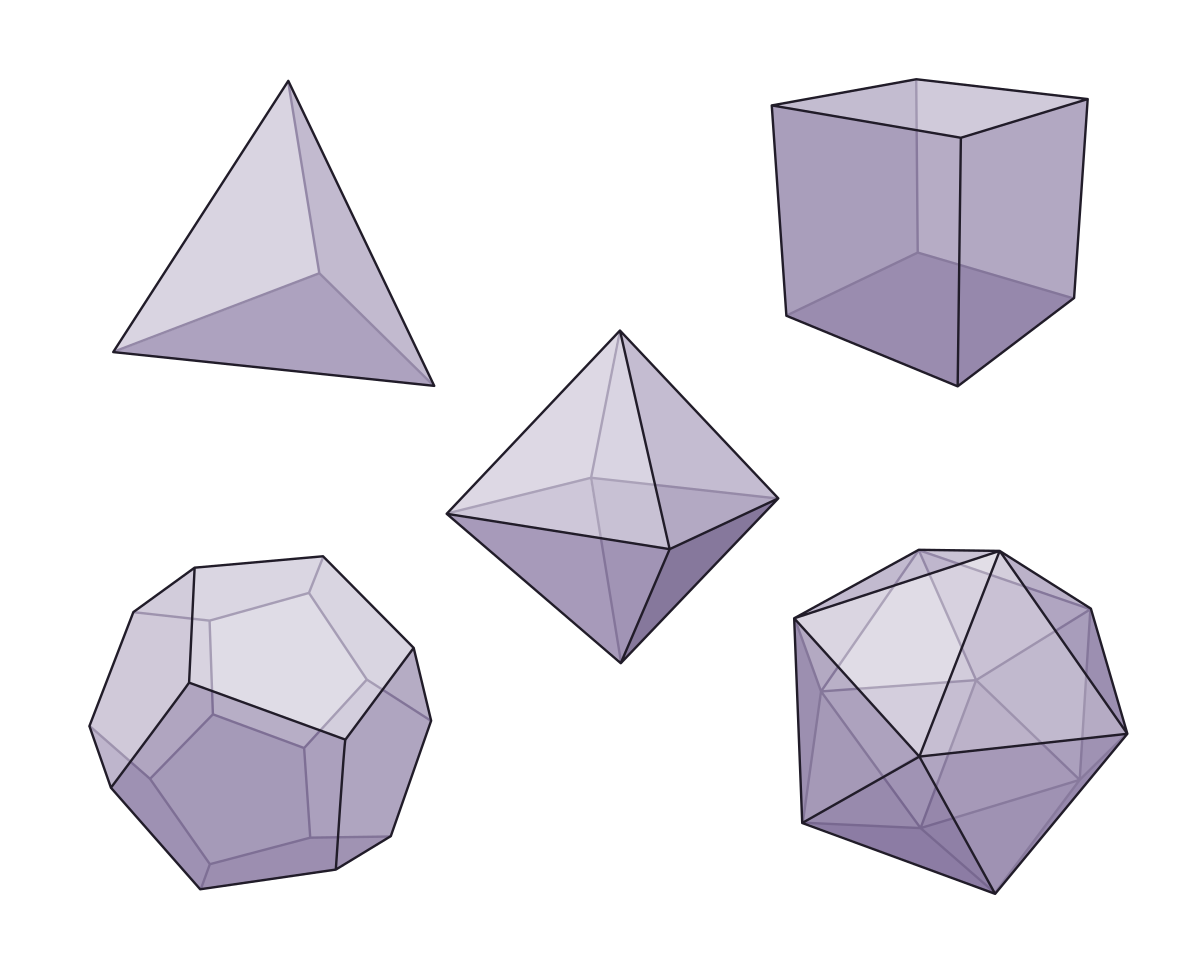}
The tetrahedron, cube, octahedron, dodecahedron, and icosahedron are called {\em Platonic solids}.}
This somewhat poetic definition of symmetry is given in the eponymous book of the great mathematician Hermann \citet{weyl2015symmetry}, his {\em Schwanengesang} on the eve of retirement from the Institute for Advanced Study in Princeton. 
Weyl traces the special place symmetry has occupied in science and art to the ancient times, from Sumerian symmetric designs to the Pythagoreans who believed the circle to be perfect due to its rotational symmetry. 
Plato considered the five regular polyhedra bearing his name today  so fundamental that they must be the basic building blocks shaping the material world. 
Yet, though Plato is credited with coining the term \textgreek{summetr'ia}, which literally translates as `same measure', he used it only vaguely to convey the beauty of proportion in art and harmony in music. 
It was the astronomer and mathematician Johannes Kepler to attempt the first rigorous analysis of the symmetric shape of water crystals. In his treatise (`On the Six-Cornered Snowflake'),\marginnote{
Fully titled 
{\em Strena, Seu De Nive Sexangula} ('New Year's gift, or on the Six-Cornered Snowflake') was, as suggested by the title, a small booklet sent by Kepler in 1611 as a Christmas gift to his patron and friend Johannes Matthäus Wackher von Wackenfels. } he attributed the six-fold dihedral structure of snowflakes to hexagonal packing of particles -- an idea that though preceded the clear understanding of how matter is formed, still holds today as the basis of crystallography \citep{ball2011retrospect}.


\paragraph{Symmetry in Mathematics and Physics}
In modern mathematics, symmetry is almost univocally expressed in the language of group theory. %
The origins of this theory are usually attributed to {\'E}variste Galois, who coined the term and used it to study solvability of polynomial equations in the 1830s. 
Two other names associated with group theory are those of Sophus Lie and Felix Klein, who met and worked fruitfully together for a period of time \citep{tobies2019felix}. The former would develop the theory of continuous symmetries that today bears his name; the latter proclaimed group theory to be the organising principle of geometry in his Erlangen Program, which we mentioned in the beginning of this text. Riemannian geometry was explicitly excluded from Klein's unified geometric picture, and it took another fifty years before it was integrated, largely thanks to the work of {\'E}lie Cartan in the 1920s.   



Emmy Noether, Klein's colleague in G\"ottingen, proved that every differentiable symmetry of the action of a physical system has a corresponding conservation law \citep{variationsprobleme1918nachr}. In physics, it was a stunning result: beforehand, meticulous experimental observation was required to discover fundamental laws such as the conservation of energy, and even then, it was an empirical result not coming from anywhere. Noether’s Theorem --- ``a guiding star to 20th and 21st century physics'', in the words of the Nobel laureate Frank Wilczek --- showed that the conservation of energy emerges from the translational symmetry of time, a rather intuitive idea that the results of an experiment should not depend on whether it is conducted today or tomorrow. 


The symmetry\marginnote{Weyl first conjectured (incorrectly) in 1919 that invariance under the change of scale or ``gauge'' was a local symmetry of electromagnetism. The term {\em gauge}, or {\em Eich} in German, was chosen by analogy to the various track gauges of railroads. 
After the development of quantum mechanics, \cite{weyl1929elektron} modified the gauge choice by replacing the scale factor with a change of wave phase. See \cite{straumann1996early}.
}  associated with charge conservation is the global {\em gauge invariance} of the electromagnetic field, first appearing in Maxwell's formulation of electrodynamics  \citep{maxwell1865viii};  
however, its importance initially remained unnoticed. 
%
The same Hermann Weyl 
who wrote so dithyrambically about symmetry 
is the one who first introduced the concept of gauge invariance in physics in the early 20th century, emphasizing its role as a 
principle from which electromagnetism can be {\em derived}. 
It took several decades until this fundamental principle — in its generalised form 
developed by \cite{yang1954conservation} — 
proved successful in providing a unified framework to describe the quantum-mechanical behavior of electromagnetism and the weak and strong forces, finally culminating in the Standard Model that captures all the fundamental forces of nature but gravity. 
We can thus join another Nobel-winning physicist, Philip \cite{anderson1972more}, in concluding that ``it is only slightly overstating the case to say that physics is the study of symmetry.'' 



\paragraph{Early Use of Symmetry in Machine Learning}
In machine learning and its applications to pattern recognition and computer vision, the importance of symmetry has long been recognised. 
Early work on designing equivariant feature detectors for pattern recognition was done by \citet{amari1978feature},\marginnote{Shun'ichi Amari is credited as the creator of the field of {\em information geometry} that applies Riemannian geometry models to probability. The main object studied by information geometry is a {\em statistical manifold}, where each point corresponds to a probability distribution. } \citet{kanatani2012group}, and \citet{lenz1990group}. 
In the 
neural networks literature, the famous Group Invariance Theorem for Perceptrons
by \citet{minsky2017perceptrons} puts fundamental limitations on the capabilities
of (single-layer) perceptrons to learn invariants. This was one of the primary motivations for studying multi-layer architectures \citep{sejnowski1986learning,shawe1989building,shawe1993symmetries}, which ultimately led to deep learning. 

In the neural network community, {\em Neocognitron} \citep{fukushima1982neocognitron} is credited as the first  implementation of shift invariance in a neural network for ``pattern recognition unaffected by shift in position''. 
%
His solution came in the form of hierarchical neural network with local connectivity, drawing inspiration from the receptive fields discovered in the visual cortex by the neuroscientists David Hubel and Torsten Wiesel two decades earlier \citep{hubel1959receptive}. \marginnote{This classical work was recognised by the Nobel Prize in Medicine in 1981, which Hubel and Wiesel shared with Roger Sperry. }
These ideas culminated in Convolutional Neural Networks in the seminal work of Yann LeCun and co-authors \citep{lecun1998gradient}. 
The first work to take a representation-theoretical view on invariant and equivariant neural networks was performed by \citet{wood1996representation}, unfortunately rarely cited. 
More recent incarnations of these ideas include the works of \citet{makadia2007correspondence,esteves2020spin}
and one of the authors of this text \citep{cohen2016group}.


\paragraph{Graph Neural Networks}
It is difficult to pinpoint exactly when the concept of Graph Neural Networks began to emerge---partly due to the fact that most of the early work  did not place graphs as a first-class citizen, partly since GNNs became practical only in the late 2010s, and partly because this field emerged from the confluence of several research areas. That being said, early forms of graph neural networks can be traced back at least to the 1990s, with examples including Alessandro Sperduti's Labeling RAAM \citep{sperduti1994encoding}, the ``backpropagation through structure'' of 
\cite{goller1996learning}, and adaptive processing of data structures 
\citep{sperduti1997supervised,frasconi1998general}. While these works were primarily concerned with operating over ``structures'' (often trees or directed acyclic graphs), many of the invariances preserved in their architectures are reminiscent of the GNNs more commonly in use today.

The first proper treatment of the processing of generic graph structures (and the coining of the term \emph{``graph neural network''}) happened after the turn of the 21st century.\marginnote{Concurrently, Alessio Micheli had proposed the \emph{neural network for graphs} (NN4G) model, which focused on a \emph{feedforward} rather than recurrent paradigm \citep{micheli2009neural}.} Within the Artificial Intelligence lab at the Universit\`{a} degli Studi di Siena (Italy), papers led by Marco Gori and Franco Scarselli have proposed the first ``GNN'' \citep{gori2005new,scarselli2008graph}. They relied on recurrent mechanisms, required the neural network parameters to specify \emph{contraction mappings}, and thus computing node representations by searching for a fixed point---this in itself necessitated a special form of backpropagation \citep{almeida1990learning,pineda1988generalization} and did not depend on node features at all. All of the above issues were rectified by the Gated GNN (GGNN) model of \cite{li2015gated}. GGNNs brought many benefits of modern RNNs, such as gating mechanisms \citep{cho2014learning} and backpropagation through time, to the GNN model, and remain popular today.

\paragraph{Computational chemistry}
It is also very important to note an independent and concurrent line of development for GNNs: one that was entirely driven by the needs of computational chemistry, where 
 molecules are most naturally expressed as graphs of atoms (nodes) connected by chemical bonds (edges). 
 This invited computational techniques for molecular property prediction that operate directly over such a graph structure, which had become present in machine learning in the 1990s: this includes the ChemNet model of 
 \cite{kireev1995chemnet} and the work of 
 \cite{baskin1997neural}. Strikingly, the ``molecular graph networks'' of \cite{merkwirth2005automatic} explicitly proposed many of the elements commonly found in contemporary GNNs---such as edge type-conditioned weights or global pooling---as early as 2005. The chemical motivation continued to drive GNN development into the 2010s, with two significant GNN advancements centered around improving molecular fingerprinting \citep{duvenaud2015convolutional} and predicting quantum-chemical properties \citep{gilmer2017neural} from small molecules. 
 At the time of writing this text, molecular property prediction is one of the most successful applications of GNNs, with impactful results in virtual screening of new antibiotic drugs \citep{stokes2020deep}.

\paragraph{Node embeddings} Some of the earliest success stories of deep learning on graphs involve learning representations of nodes in an unsupervised fashion, based on the graph structure. Given their structural inspiration, this direction also provides one of the most direct links between graph representation learning and network science communities. The key early approaches in this space relied on \emph{random walk}-based embeddings: learning node representations in a way that brings them closer together if the nodes co-occur in a short random walk. Representative methods in this space include DeepWalk \citep{perozzi2014deepwalk}, node2vec \citep{grover2016node2vec} and LINE \citep{tang2015line}, which are all purely self-supervised. Planetoid \citep{yang2016revisiting} was the first in this space to incorporate supervision label information, when it is available.

Unifying random walk objectives with GNN encoders\marginnote{Recently, a theoretical framework was developed by \citet{srinivasan2019equivalence} in which the equivalence of structural and positional representations was demonstrated. Additionally, \citet{qiu2018network} have demonstrated that all random-walk based embedding techniques are equivalent to an appropriately-posed matrix factorisation task.} was attempted on several occasions, with representative approaches including Variational Graph Autoencoder (VGAE,  \cite{kipf2016variational}), embedding propagation \citep{garcia2017learning}, and unsupervised variants of GraphSAGE \citep{hamilton2017inductive}. However, this was met with mixed results, and it was shortly discovered that pushing neighbouring node representations together is already a key part of GNNs' inductive bias. Indeed, it was shown that an \emph{untrained} GNN was already showing performance that is competitive with DeepWalk, in settings where node features are available \citep{velickovic2019deep,wu2019simplifying}. This launched a direction that moves away from combining random walk objectives with GNNs and shifting towards \emph{contrastive} approaches inspired by mutual information maximisation and aligning to successful methods in the image domain. Prominent examples of this direction include Deep Graph Informax (DGI,  \cite{velickovic2019deep}), GRACE \citep{zhu2020deep}, BERT-like objectives \citep{hu2020strategies} and BGRL \citep{thakoor2021bootstrapped}.

\paragraph{Probabilistic graphical models} 
Graph neural networks have also, concurrently, resurged through embedding the computations of \emph{probabilistic graphical models} (PGMs, \cite{wainwright2008graphical}).
PGMs are a powerful tool for processing graphical data, and their utility arises from their probabilistic perspective on the graph's edges: namely, the nodes are treated as \emph{random variables}, while the graph structure encodes \emph{conditional independence} assumptions, allowing for significantly simplifying the calculation and sampling from the joint distribution. Indeed, many algorithms for (exactly or approximately) supporting learning and inference on PGMs rely on forms of passing messages over their edges \citep{pearl2014probabilistic}, with examples including variational mean-field inference and loopy belief propagation \citep{yedidia2001bethe,murphy2013loopy}.

This connection between PGMs and message passing was subsequently developed into GNN architectures, with early theoretical links established by the authors of structure2vec \citep{dai2016discriminative}. Namely, by posing a graph representation learning setting as a Markov random field (of nodes corresponding to input features and latent representations), the authors directly align the computation of both mean-field inference and loopy belief propagation to a model not unlike the GNNs commonly in use today. 

The key ``trick'' which allowed for relating the latent representations of a GNN to probability distributions maintained by a PGM was the usage of \emph{Hilbert-space embeddings} of distributions \citep{smola2007hilbert}. Given $\phi$, an appropriately chosen embedding function for features $\vec{x}$, it is possible to embed their probability distribution $p(\vec{x})$ as the \emph{expected} embedding $\mathbb{E}_{\vec{x}\sim p(\vec{x})}\phi(\vec{x})$. Such a correspondence allows us to perform GNN-like computations, knowing that the representations computed by the GNN will always correspond to an embedding of \emph{some} probability distribution over the node features.

The structure2vec model itself is, ultimately, a GNN architecture which easily sits within our framework, but its setup has inspired a series of GNN architectures which more directly incorporate computations found in PGMs. Emerging examples have successfully combined GNNs with conditional random fields \citep{gao2019conditional,spalevic2020hierachial}, relational Markov networks \citep{qu2019gmnn} and Markov logic networks \citep{zhang2020efficient}.

\paragraph{The Weisfeiler-Lehman formalism} The resurgence of graph neural networks was followed closely by a drive to understand their fundamental limitations, especially in terms of expressive power. While it was becoming evident that GNNs are a strong modelling tool of graph-structured data, it was also clear that they wouldn't be able to solve \emph{any} task specified on a graph perfectly.\marginnote{Due to their permutation invariance, GNNs will attach identical representations to two isomorphic graphs, so this case is trivially solved.} A canonical illustrative example of this is deciding \emph{graph isomorphism}: is our GNN able to attach different representations to two given non-isomorphic graphs? This is a useful framework for two reasons. If the GNN is unable to do this, then it will be hopeless on any task requiring the discrimination of these two graphs. Further, it is currently not known if deciding graph isomorphism is in \textsc{P}\marginnote{The best currently known algorithm for deciding graph isomorphism is due to \citet{babai1983canonical}, though a recent (not fully reviewed) proposal by \citet{babai2016graph} implies a quasi-polynomial time solution.}, the complexity class in which all GNN computations typically reside.

The key framework which binds GNNs to graph isomorphism is the \emph{Weisfeiler-Lehman} (WL) graph isomorphism test \citep{weisfeiler1968reduction}. This test generates a graph representation by iteratively passing node features along the edges of the graph, then \emph{randomly hashing} their sums across neighbourhoods. Connections to \emph{randomly-initialised} convolutional GNNs are apparent, and have been observed early on: for example, within the GCN model of \citet{kipf2016semi}. Aside from this connection, the WL iteration was previously introduced in the domain of \emph{graph kernels} by \citet{shervashidze2011weisfeiler}, and it still presents a strong baseline for unsupervised learning of whole-graph representations.

While\marginnote{\includegraphics[width=\linewidth]{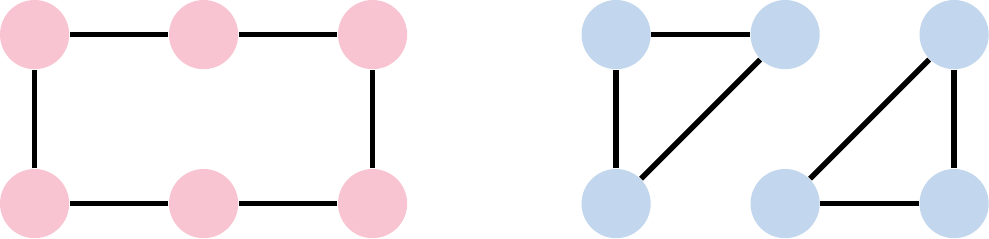} One simple example: the WL test cannot distinguish a \emph{6-cycle} from \emph{two triangles}.} the WL test is conceptually simple, and there are many simple examples of non-isomorphic graphs it cannot distinguish, its expressive power is ultimately strongly tied to GNNs. Analyses by \citet{morris2019weisfeiler} and \citet{xu2018powerful} have both reached a striking conclusion: \emph{any} GNN conforming to one of the three flavours we outlined in Section \ref{sec:gnn-intro} cannot be more powerful than the WL test!

In order to exactly reach this level of representational power, certain constraints must exist on the GNN update rule. \citet{xu2018powerful} have shown that, in the discrete-feature domain, the aggregation function the GNN uses must be \emph{injective}, with \emph{summation} being a key representative\marginnote{Popular aggregators such as maximisation and averaging fall short in this regard, because they would not be able to distinguish e.g. the neighbour multisets $\ldblbrace \vec{a}, \vec{b}\rdblbrace$ and $\ldblbrace \vec{a}, \vec{a}, \vec{b}, \vec{b}\rdblbrace$.}. Based on the outcome of their analysis, \citet{xu2018powerful} propose the Graph Isomorphism Network (GIN), which is a simple but powerful example of a maximally-expressive GNN under this framework. It is also expressible under the convolutional GNN flavour we propose.

Lastly, it is worth noting that these findings do not generalise to \emph{continuous} node feature spaces. In fact, using the Borsuk-Ulam theorem \citep{borsuk1933drei}, \citet{corso2020principal} have demonstrated that, assuming real-valued node features, obtaining injective aggregation functions requires \emph{multiple} aggregators (specifically, equal to the \emph{degree} of the receiver node)\marginnote{One example of such aggregators are the \emph{moments} of the multiset of neighbours.}. Their findings have driven the Principal Neighbourhood Aggregation (PNA) architecture, which proposes a multiple-aggregator GNN that is empirically powerful and stable.

\paragraph{Higher-order methods} 

The findings of the previous paragraphs do not contradict the practical utility of GNNs. Indeed, in many real-world applications the input features are sufficiently \emph{rich} to support useful discriminative computations over the graph structure, despite of the above limitations\marginnote{Which, in contrast, almost always consider \emph{featureless} or \emph{categorically-featured} graphs.}. 

However, one key corollary is that GNNs are relatively quite weak at detecting some rudimentary \emph{structures} within a graph. Guided by the specific limitations or failure cases of the WL test, several works have provided \emph{stronger} variants of GNNs that are \emph{provably} more powerful than the WL test, and hence likely to be useful on tasks that require such structural detection\marginnote{One prominent example is computational chemistry, wherein a molecule's chemical function can be strongly influenced by the presence of aromatic \emph{rings} in its molecular graph.}.

Perhaps the most direct place to hunt for more expressive GNNs is the WL test itself. Indeed, the strength of the original WL test can be enhanced by considering a \emph{hierarchy} of WL tests, such that $k$-WL tests attach representations to $k$\emph{-tuples} of nodes \citep{morris2017glocalized}. The $k$-WL test has been directly translated into a \emph{higher-order} $k$-GNN architecture by \citet{morris2019weisfeiler},\marginnote{There have been efforts, such as the $\delta$-$k$-LGNN \citep{morris2020weisfeiler}, to sparsify the computation of the $k$-GNN.} which is provably more powerful than the GNN flavours we considered before. However, its requirement to maintain tuple representations implies that, in practice, it is hard to scale beyond $k=3$.

Concurrently, \citet{maron2018invariant,maron2019provably} have studied the characterisation of invariant and equivariant graph networks over $k$-tuples of nodes. Besides demonstrating the surprising result of \emph{any} invariant or equivariant graph network being expressible as a linear combination of a finite number of generators---the amount of which only depends on $k$---the authors showed that the expressive power of such layers is equivalent to the $k$-WL test, and proposed an empirically scalable variant which is provably 3-WL powerful.

Besides generalising the domain over which representations are computed, significant effort had also went into analysing specific failure cases of 1-WL and augmenting GNN \emph{inputs} to help them distinguish such cases. One common example is attaching \emph{identifying features} to the nodes, which can help detecting structure\marginnote{For example, if a node sees its own identifier $k$ hops away, it is a direct  indicator that it is within a $k$-cycle.}. Proposals to do this include \emph{one-hot} representations \citep{murphy2019relational}, as well as purely \emph{random} features \citep{sato2020random}.

More broadly, there have been many efforts to incorporate \emph{structural} information within the message passing process, either by modulating the message function or the graph that the computations are carried over\marginnote{In the computational chemistry domain, it is often assumed that molecular function is driven by substructures (the \emph{functional groups}), which have directly inspired the modelling of molecules at a \emph{motif} level. For references, consider \cite{jin2018junction,jin2020hierarchical,fey2020hierarchical}.}. Several interesting lines of work here involve sampling \emph{anchor node sets} \citep{you2019position}, aggregating based on \emph{Laplacian eigenvectors} \citep{stachenfeld2020graph,beaini2020directional,dwivedi2020generalization}, or performing \emph{topological data analysis}, either for positional embeddings \citep{bouritsas2020improving} or driving message passing \citep{bodnar2021weisfeiler}.


\paragraph{Signal processing and Harmonic analysis}
Since the early successes of Convolutional Neural Networks, researchers have resorted to tools from harmonic analysis, image processing, and computational neuroscience trying to provide a theoretical framework that explains their efficiency. 
$M$-theory is a framework inspired by the visual cortex, pioneered by Tomaso Poggio and collaborators \citep{riesenhuber1999hierarchical, serre2007feedforward}, based on the notion of templates that can be manipulated under certain symmetry groups. Another notable model arising from computational neuroscience were {\em steerable pyramids}, a form of multiscale wavelet decompositions  with favorable properties against certain input transformations, developed by \cite{simoncelli1995steerable}. They were a central element in early generative models for textures \citep{portilla2000parametric}, which were subsequently improved by replacing steerable wavelet features with deep CNN features \cite{gatys2015texture}. 
Finally, Scattering transforms, introduced by St{\'e}phane \cite{mallat2012group} and developed by \cite{bruna2013invariant}, provided a framework to understand CNNs by replacing trainable filters with multiscale wavelet decompositions, also showcasing the deformation stability and the role of depth in the architecture.






\paragraph{Signal Processing on Graph and Meshes}
Another important class of graph neural networks, often referred to as {\em spectral}, has emerged from the work of one of the authors of this text \citep{bruna2013spectral}, using the notion of the \emph{Graph Fourier transform}.  
The roots of this construction are in the signal processing and computational harmonic analysis communities, where dealing with non-Euclidean signals has become prominent in the late 2000s and early 2010s. Influential papers from the groups of Pierre Vandergheynst \citep{shuman2013emerging} and Jos{\'e} Moura \citep{sandryhaila2013discrete} popularised the notion of ``Graph Signal Processing'' (GSP) and the generalisation of Fourier transforms based on the eigenvectors of graph adjacency and Laplacian matrices. 
The graph convolutional neural networks relying on spectral filters by  \citet{defferrard2016convolutional} and \cite{kipf2016semi} are among the most cited in the field and can likely be credited) as ones  reigniting the interest in machine learning on graphs in recent years.

It is worth noting that, in the field of computer graphics and geometry processing, non-Euclidean harmonic analysis predates Graph Signal Processing by at least a decade. We can trace spectral filters on manifolds and meshes to the works of \cite{taubin1996optimal}. These methods became mainstream in the 2000s following the influential papers of 
\cite{karni2000spectral} on spectral geometry compression and of \cite{levy2006laplace} on using the Laplacian eigenvectors as a non-Euclidean Fourier basis. 
Spectral methods have been used for a range of applications,\marginnote{Learnable shape descriptors similar to spectral graph CNNs were proposed by Roee Litman and Alex Bronstein (\citeyear{litman2013learning}), the latter being a twin brother of the author of this text. } most prominent of which is the construction of shape descriptors \citep{sun2009concise} and functional maps \citep{ovsjanikov2012functional}; these methods are still broadly used in computer graphics at the time of writing.


\paragraph{Computer Graphics and Geometry Processing}
Models for shape analysis based on intrinsic metric invariants were introduced by various authors in the field of computer graphics and geometry processing \citep{elad2003bending,memoli2005theoretical,bronstein2006generalized}, and are discussed in depth by one of the authors in his earlier book \citep{bronstein2008numerical}. The notions of intrinsic symmetries were also explored in the same field  \cite{raviv2007symmetries,ovsjanikov2008global}. 
The first architecture for deep learning on meshes, Geodesic CNNs, was developed in the team of one of the authors of the text \citep{masci2015geodesic}. This model used local filters with shared weights, applied to geodesic radial patches. It was a particular setting of gauge-equivariant CNNs developed later by another author of the text \citep{cohen2019gauge}. 
A generalisation of Geodesic CNNs with learnable aggregation operations, MoNet, proposed by Federico  \cite{monti2017geometric} from the same team, used an attention-like mechanism over the local structural features of the mesh, that was demonstrated to work on general graphs as well. 
The graph attention network (GAT), which technically speaking can be considered a particular instance of MoNet, was introduced by another author of this text \citep{velickovic2018graph}. GATs generalise MoNet's attention mechanism to also incorporate node feature information, breaking away from the purely structure-derived relevance of prior work. It is one of the most popular GNN architectures currently in use.

In the context of computer graphics, it is also worthwhile to mention that the idea of learning on sets \citep{zaheer2017deep} was concurrently developed in the group of Leo Guibas at Stanford under the name PointNet \citep{qi2017pointnet} for the analysis of 3D point clouds. This architecture has lead to multiple follow-up works, including one by an author of this text called Dynamic Graph CNN (DGCNN, \cite{wang2019dynamic}). DGCNN used a nearest-neighbour graph to capture the local structure of the point cloud to allow exchange of information across the nodes; the key characteristic of this architecture was that the graph was constructed on-the-fly and updated between the layers of the neural network in relation to the downstream task. 
This latter property made DGCNN one of the first incarnations of `latent graph learning', which in its turn has had significant follow up. Extensions to DGCNN's $k$-nearest neighbour graph proposal include more explicit control over these graphs' edges, either through bilevel optimisation \citep{franceschi2019learning}, reinforcement learning \citep{kazi2020differentiable} or direct supervision \citep{velivckovic2020pointer}. Independently, a variational direction (which probabilistically samples edges from a computed \emph{posterior} distribution) has emerged through the NRI model \citep{kipf2018neural}. While it still relies on quadratic computation in the number of nodes, it allows for explicitly encoding uncertainty about the chosen edges.

Another very popular direction in learning on graphs without a provided graph relies on performing GNN-style computations over a \emph{complete} graph, letting the network infer its own way to exploit the connectivity. The need for this arisen particularly in natural language processing, where various words in a sentence interact in highly nontrivial and non-sequential ways. Operating over a complete graph of words brought about the first incarnation of the Transformer model \citep{vaswani2017attention}, which de-throned both recurrent and convolutional models as state-of-the-art in neural machine translation, and kicked off an avalanche of related work, transcending the boundaries between NLP and other fields. Fully-connected GNN computation has also concurrently emerged on simulation \citep{battaglia2016interaction}, reasoning \citep{santoro2017simple}, and multi-agent \citep{hoshen2017vain} applications, and still represents a popular choice when the number of nodes is reasonably small.
%



\paragraph{Algorithmic reasoning} 
For most of the discussion we posed in this section, we have given examples of \emph{spatially} induced geometries, which in turn shape the underlying domain, and its invariances and symmetries. However, plentiful examples of invariances and symmetries also arise in a \emph{computational} setting. One critical difference to many common settings of Geometric Deep Learning is that links no longer need to encode for any kind of similarity, proximity, or types of relations---they merely specify the ``recipe'' for the dataflow between data points they connect.

Instead, the computations of the neural network mimic the reasoning process of an \emph{algorithm} \citep{cormen2009introduction}, with additional invariances induced by the algorithm's control flow and intermediate results\marginnote{For example, one invariant of the Bellman-Ford pathfinding algorithm \citep{bellman1958routing} is that, after $k$ steps, it will always compute the shortest paths to the source node that use no more than $k$ edges.}. In the space of algorithms the assumed input invariants are often referred to as \emph{preconditions}, while the invariants preserved by the algorithm are known as \emph{postconditions}.

Eponymously, the research direction of \emph{algorithmic reasoning} \citep[Section 3.3.]{cappart2021combinatorial} seeks to produce neural network architectures that appropriately preserve algorithmic invariants. The area has investigated the construction of general-purpose neural computers, e.g., the \emph{neural Turing machine} \citep{graves2014neural} and the \emph{differentiable neural computer} \citep{graves2016hybrid}. While such architectures have all the hallmarks of general computation, they introduced several components at once, making them often challenging to optimise, and in practice, they are almost always outperformed by simple relational reasoners, such as the ones proposed by \citet{santoro2017simple,santoro2018relational}. 

As modelling complex postconditions is challenging, plentiful work on inductive biases for learning to execute \citep{zaremba2014learning} has focused on primitive algorithms (e.g. simple arithmetic). Prominent examples in this space include the \emph{neural GPU} \citep{kaiser2015neural}, \emph{neural RAM} \citep{kurach2015neural}, \emph{neural programmer-interpreters} \citep{reed2015neural}, \emph{neural arithmetic-logic units} \citep{trask2018neural,madsen2020neural} and \emph{neural execution engines} \citep{yan2020neural}.

Emulating combinatorial algorithms of \emph{superlinear} complexity was made possible with the rapid development of GNN architectures. The \emph{algorithmic alignment} framework pioneered by \citet{xu2019can} demonstrated, theoretically,  that GNNs \emph{align} with dynamic programming \citep{bellman1966dynamic}, which is a language in which most algorithms can be expressed. It was concurrently empirically shown, by one of the authors of this text, that it is possible to design and train GNNs that align with algorithmic invariants in practice \citep{velivckovic2019neural}. Onwards, alignment was achieved with \emph{iterative algorithms} \citep{tang2020towards}, \emph{linearithmic algorithms} \citep{freivalds2019neural}, \emph{data structures} \citep{velivckovic2020pointer} and \emph{persistent memory} \citep{strathmann2021persistent}. Such models have also seen practical use in \emph{implicit planners} \citep{deac2020xlvin}, breaking into the space of \emph{reinforcement learning} algorithms.

Concurrently, significant progress has been made on using GNNs for \emph{physics simulations} \citep{sanchez2020learning,pfaff2020learning}. This direction yielded much of the same recommendations for the design of generalising GNNs. Such a correspondence is to be expected: given that algorithms can be phrased as discrete-time simulations, and simulations are typically implemented as step-wise algorithms, both directions will need to preserve similar kinds of invariants.

Tightly bound with the study of algorithmic reasoning are measures of \emph{extrapolation}. This is a notorious pain-point for neural networks, given that most of their success stories are obtained when  generalising \emph{in-distribution}; i.e. when the patterns found in the training data properly anticipate the ones found in the test data. However, algorithmic invariants must be preserved irrespective of, e.g., the size or generative distribution of the input, meaning that the training set will likely not cover any possible scenario encountered in practice. \citet{xu2020neural} have proposed a geometric argument for what is required of an extrapolating GNN backed by rectifier activations: its components and featurisation would need to be designed so as to make its constituent modules (e.g. message function) learn only \emph{linear} target functions. \citet{bevilacqua2021size} propose observing extrapolation under the lens of \emph{causal reasoning}, yielding \emph{environment-invariant} representations of graphs.

\paragraph{Geometric Deep Learning}
Our final historical remarks regard the very name of this text.
The term `Geometric Deep Learning' was first introduced by one of the authors of this text in his ERC grant in 2015 and popularised in the eponymous IEEE Signal Processing Magazine paper \citep{bronstein2017geometric}. This paper proclaimed, albeit ``with some caution'', the signs of ``a new field being born.'' Given the recent popularity of graph neural networks, the increasing use of ideas of invariance and equivariance in a broad range of machine learning applications, and the very fact of us writing this text, it is probably right to consider this prophecy at least partially fulfilled. 
The name ``4G: Grids, Graphs, Groups, and Gauges'' was coined by Max Welling for the ELLIS Program on Geometric Deep Learning, co-directed by two authors of the text. Admittedly, the last `G' is somewhat of a stretch, since the underlying structures are manifolds and bundles rather than gauges. 
For this text, we added another `G', Geodesics, in reference to metric invariants and intrinsic symmetries of manifolds.


\section*{Acknowledgements}

This text
represents a humble attempt to summarise and synthesise decades of existing knowledge in 
deep learning architectures, through the geometric lens of invariance and symmetry. 
We hope that our perspective 
will make it easier both for newcomers and practitioners to navigate the field, and for researchers to synthesise novel architectures, as instances of our blueprint. In a way, we hope to have presented \emph{``all you need to build the architectures that are all you need''}---a play on words inspired by \citet{vaswani2017attention}.

The bulk of the text was written during late 2020 and early 2021. As it often happens, we had thousands of doubts whether  the whole picture makes sense, and used opportunities provided by our colleagues to  
%
help us break our ``stage fright'' and present early versions of our work, which saw the  
light of day in Petar's talk at Cambridge (courtesy of Pietro Li{\`o}) and Michael's talks at Oxford (courtesy of Xiaowen Dong) and Imperial College (hosted by Michael Huth and Daniel Rueckert). Petar was also able to present our work at Friedrich-Alexander-Universit\"{a}t Erlangen-N\"{u}rnberg---the birthplace of the Erlangen Program!---owing to a kind invitation from Andreas Maier.
%
%
The feedback we received for these talks was enormously invaluable to keeping our spirits high, as well as polishing the work further. 
Last, but certainly not least, we thank the organising committee of ICLR 2021, where our work will be featured in a keynote talk, delivered by Michael.



We should note that 
reconciling such a vast quantity of research is seldom enabled by the expertise of only four people. Accordingly, we would like to give due credit to all of the researchers who have carefully studied aspects of our text as it evolved, and provided us with careful comments and references: Yoshua Bengio, Charles Blundell, Andreea Deac, Fabian Fuchs, Francesco di Giovanni, Marco Gori, Raia Hadsell, Will Hamilton, Maksym Korablyov, Christian Merkwirth, Razvan Pascanu, Bruno Ribeiro, Anna Scaife, J\"{u}rgen Schmidhuber, Marwin Segler, Corentin Tallec, Ng\^{a}n V\~{u}, Peter Wirnsberger and David Wong. Their expert feedback was invaluable to solidifying our unification efforts and making them more useful to various niches. Though, of course, any irregularities within this text are our responsibility alone. It is currently very much a work-in-progress, and we are very happy to receive comments at any stage. Please contact us if you spot any errors or omissions.

\bibliographystyle{plainnat}
\bibliography{references}

\begin{thebibliography}{340}
\providecommand{\natexlab}[1]{#1}
\providecommand{\url}[1]{\texttt{#1}}
\expandafter\ifx\csname urlstyle\endcsname\relax
  \providecommand{\doi}[1]{doi: #1}\else
  \providecommand{\doi}{doi: \begingroup \urlstyle{rm}\Url}\fi

\bibitem[Aflalo and Kimmel(2013)]{aflalo2013spectral}
Yonathan Aflalo and Ron Kimmel.
\newblock Spectral multidimensional scaling.
\newblock \emph{PNAS}, 110\penalty0 (45):\penalty0 18052--18057, 2013.

\bibitem[Aflalo et~al.(2015)Aflalo, Brezis, and Kimmel]{aflalo2015optimality}
Yonathan Aflalo, Haim Brezis, and Ron Kimmel.
\newblock On the optimality of shape and data representation in the spectral
  domain.
\newblock \emph{SIAM J. Imaging Sciences}, 8\penalty0 (2):\penalty0 1141--1160,
  2015.

\bibitem[Almeida(1990)]{almeida1990learning}
Luis~B Almeida.
\newblock A learning rule for asynchronous perceptrons with feedback in a
  combinatorial environment.
\newblock In \emph{Artificial neural networks: concept learning}, pages
  102--111. 1990.

\bibitem[Alon and Yahav(2020)]{alon2020bottleneck}
Uri Alon and Eran Yahav.
\newblock On the bottleneck of graph neural networks and its practical
  implications.
\newblock \emph{arXiv:2006.05205}, 2020.

\bibitem[Amari(1978)]{amari1978feature}
Sl~Amari.
\newblock Feature spaces which admit and detect invariant signal
  transformations.
\newblock In \emph{Joint Conference on Pattern Recognition}, 1978.

\bibitem[Anderson et~al.(2019)Anderson, Hy, and Kondor]{anderson2019cormorant}
Brandon Anderson, Truong-Son Hy, and Risi Kondor.
\newblock Cormorant: Covariant molecular neural networks.
\newblock \emph{arXiv:1906.04015}, 2019.

\bibitem[Anderson(1972)]{anderson1972more}
Philip~W Anderson.
\newblock More is different.
\newblock \emph{Science}, 177\penalty0 (4047):\penalty0 393--396, 1972.

\bibitem[Andreux et~al.(2014)Andreux, Rodola, Aubry, and
  Cremers]{andreux2014anisotropic}
Mathieu Andreux, Emanuele Rodola, Mathieu Aubry, and Daniel Cremers.
\newblock Anisotropic {Laplace-Beltrami} operators for shape analysis.
\newblock In \emph{ECCV}, 2014.

\bibitem[Arslan et~al.(2018)Arslan, Ktena, Glocker, and
  Rueckert]{arslan2018graph}
Salim Arslan, Sofia~Ira Ktena, Ben Glocker, and Daniel Rueckert.
\newblock Graph saliency maps through spectral convolutional networks:
  Application to sex classification with brain connectivity.
\newblock In \emph{Graphs in Biomedical Image Analysis and Integrating Medical
  Imaging and Non-Imaging Modalities}, pages 3--13. 2018.

\bibitem[Ba et~al.(2016)Ba, Kiros, and Hinton]{ba2016layer}
Jimmy~Lei Ba, Jamie~Ryan Kiros, and Geoffrey~E Hinton.
\newblock Layer normalization.
\newblock \emph{arXiv:1607.06450}, 2016.

\bibitem[Babai(2016)]{babai2016graph}
L{\'a}szl{\'o} Babai.
\newblock Graph isomorphism in quasipolynomial time.
\newblock In \emph{ACM Symposium on Theory of Computing}, 2016.

\bibitem[Babai and Luks(1983)]{babai1983canonical}
L{\'a}szl{\'o} Babai and Eugene~M Luks.
\newblock Canonical labeling of graphs.
\newblock In \emph{ACM Symposium on Theory of computing}, 1983.

\bibitem[Bach(2017)]{bach2017breaking}
Francis Bach.
\newblock Breaking the curse of dimensionality with convex neural networks.
\newblock \emph{JMLR}, 18\penalty0 (1):\penalty0 629--681, 2017.

\bibitem[Badia et~al.(2020)Badia, Piot, Kapturowski, Sprechmann, Vitvitskyi,
  Guo, and Blundell]{badia2020agent57}
Adri{\`a}~Puigdom{\`e}nech Badia, Bilal Piot, Steven Kapturowski, Pablo
  Sprechmann, Alex Vitvitskyi, Zhaohan~Daniel Guo, and Charles Blundell.
\newblock Agent57: Outperforming the atari human benchmark.
\newblock In \emph{ICML}, 2020.

\bibitem[Badrinarayanan et~al.(2017)Badrinarayanan, Kendall, and
  Cipolla]{badrinarayanan2017segnet}
Vijay Badrinarayanan, Alex Kendall, and Roberto Cipolla.
\newblock Segnet: A deep convolutional encoder-decoder architecture for image
  segmentation.
\newblock \emph{Trans. PAMI}, 39\penalty0 (12):\penalty0 2481--2495, 2017.

\bibitem[Bahdanau et~al.(2014)Bahdanau, Cho, and Bengio]{bahdanau2014neural}
Dzmitry Bahdanau, Kyunghyun Cho, and Yoshua Bengio.
\newblock Neural machine translation by jointly learning to align and
  translate.
\newblock \emph{arXiv:1409.0473}, 2014.

\bibitem[Ball(2011)]{ball2011retrospect}
Philip Ball.
\newblock In retrospect: On the six-cornered snowflake.
\newblock \emph{Nature}, 480\penalty0 (7378):\penalty0 455--455, 2011.

\bibitem[Bamieh(2018)]{bamieh2018discovering}
Bassam Bamieh.
\newblock Discovering transforms: A tutorial on circulant matrices, circular
  convolution, and the discrete fourier transform.
\newblock \emph{arXiv:1805.05533}, 2018.

\bibitem[Banach(1922)]{banach1922operations}
Stefan Banach.
\newblock Sur les op{\'e}rations dans les ensembles abstraits et leur
  application aux {\'e}quations int{\'e}grales.
\newblock \emph{Fundamenta Mathematicae}, 3\penalty0 (1):\penalty0 133--181,
  1922.

\bibitem[Bapst et~al.(2020)Bapst, Keck, Grabska-Barwi{\'n}ska, Donner, Cubuk,
  Schoenholz, Obika, Nelson, Back, Hassabis, et~al.]{bapst2020unveiling}
Victor Bapst, Thomas Keck, A~Grabska-Barwi{\'n}ska, Craig Donner, Ekin~Dogus
  Cubuk, Samuel~S Schoenholz, Annette Obika, Alexander~WR Nelson, Trevor Back,
  Demis Hassabis, et~al.
\newblock Unveiling the predictive power of static structure in glassy systems.
\newblock \emph{Nature Physics}, 16\penalty0 (4):\penalty0 448--454, 2020.

\bibitem[Barab{\'a}si et~al.(2011)Barab{\'a}si, Gulbahce, and
  Loscalzo]{barabasi2011network}
Albert-L{\'a}szl{\'o} Barab{\'a}si, Natali Gulbahce, and Joseph Loscalzo.
\newblock Network medicine: a network-based approach to human disease.
\newblock \emph{Nature Reviews Genetics}, 12\penalty0 (1):\penalty0 56--68,
  2011.

\bibitem[Barron(1993)]{barron1993universal}
Andrew~R Barron.
\newblock Universal approximation bounds for superpositions of a sigmoidal
  function.
\newblock \emph{IEEE Trans. Information Theory}, 39\penalty0 (3):\penalty0
  930--945, 1993.

\bibitem[Baskin et~al.(1997)Baskin, Palyulin, and Zefirov]{baskin1997neural}
Igor~I Baskin, Vladimir~A Palyulin, and Nikolai~S Zefirov.
\newblock A neural device for searching direct correlations between structures
  and properties of chemical compounds.
\newblock \emph{J. Chemical Information and Computer Sciences}, 37\penalty0
  (4):\penalty0 715--721, 1997.

\bibitem[Battaglia et~al.(2016)Battaglia, Pascanu, Lai, Rezende, and
  Kavukcuoglu]{battaglia2016interaction}
Peter~W Battaglia, Razvan Pascanu, Matthew Lai, Danilo Rezende, and Koray
  Kavukcuoglu.
\newblock Interaction networks for learning about objects, relations and
  physics.
\newblock \emph{arXiv:1612.00222}, 2016.

\bibitem[Battaglia et~al.(2018)Battaglia, Hamrick, Bapst, Sanchez-Gonzalez,
  Zambaldi, Malinowski, Tacchetti, Raposo, Santoro, Faulkner,
  et~al.]{battaglia2018relational}
Peter~W Battaglia, Jessica~B Hamrick, Victor Bapst, Alvaro Sanchez-Gonzalez,
  Vinicius Zambaldi, Mateusz Malinowski, Andrea Tacchetti, David Raposo, Adam
  Santoro, Ryan Faulkner, et~al.
\newblock Relational inductive biases, deep learning, and graph networks.
\newblock \emph{arXiv:1806.01261}, 2018.

\bibitem[Beaini et~al.(2020)Beaini, Passaro, L{\'e}tourneau, Hamilton, Corso,
  and Li{\`o}]{beaini2020directional}
Dominique Beaini, Saro Passaro, Vincent L{\'e}tourneau, William~L Hamilton,
  Gabriele Corso, and Pietro Li{\`o}.
\newblock Directional graph networks.
\newblock \emph{arXiv:2010.02863}, 2020.

\bibitem[Bellman(1958)]{bellman1958routing}
Richard Bellman.
\newblock On a routing problem.
\newblock \emph{Quarterly of Applied Mathematics}, 16\penalty0 (1):\penalty0
  87--90, 1958.

\bibitem[Bellman(1966)]{bellman1966dynamic}
Richard Bellman.
\newblock Dynamic programming.
\newblock \emph{Science}, 153\penalty0 (3731):\penalty0 34--37, 1966.

\bibitem[Bengio et~al.(1994)Bengio, Simard, and Frasconi]{bengio1994learning}
Yoshua Bengio, Patrice Simard, and Paolo Frasconi.
\newblock Learning long-term dependencies with gradient descent is difficult.
\newblock \emph{IEEE Trans. Neural Networks}, 5\penalty0 (2):\penalty0
  157--166, 1994.

\bibitem[Berger(2012)]{berger2012panoramic}
Marcel Berger.
\newblock \emph{A panoramic view of {R}iemannian geometry}.
\newblock Springer, 2012.

\bibitem[Besson et~al.(2020)Besson, Parrish, Katsaggelos, and
  Bandt]{besson2020geometric}
Pierre Besson, Todd Parrish, Aggelos~K Katsaggelos, and S~Kathleen Bandt.
\newblock Geometric deep learning on brain shape predicts sex and age.
\newblock \emph{BioRxiv:177543}, 2020.

\bibitem[Bevilacqua et~al.(2021)Bevilacqua, Zhou, and
  Ribeiro]{bevilacqua2021size}
Beatrice Bevilacqua, Yangze Zhou, and Bruno Ribeiro.
\newblock Size-invariant graph representations for graph classification
  extrapolations.
\newblock \emph{arXiv:2103.05045}, 2021.

\bibitem[Blanc et~al.(2020)Blanc, Gupta, Valiant, and
  Valiant]{blanc2020implicit}
Guy Blanc, Neha Gupta, Gregory Valiant, and Paul Valiant.
\newblock Implicit regularization for deep neural networks driven by an
  ornstein-uhlenbeck like process.
\newblock In \emph{COLT}, 2020.

\bibitem[Bodnar et~al.(2021)Bodnar, Frasca, Wang, Otter, Mont{\'u}far, Li{\`o},
  and Bronstein]{bodnar2021weisfeiler}
Cristian Bodnar, Fabrizio Frasca, Yu~Guang Wang, Nina Otter, Guido
  Mont{\'u}far, Pietro Li{\`o}, and Michael Bronstein.
\newblock Weisfeiler and lehman go topological: Message passing simplicial
  networks.
\newblock \emph{arXiv:2103.03212}, 2021.

\bibitem[Bogatskiy et~al.(2020)Bogatskiy, Anderson, Offermann, Roussi, Miller,
  and Kondor]{bogatskiy2020lorentz}
Alexander Bogatskiy, Brandon Anderson, Jan Offermann, Marwah Roussi, David
  Miller, and Risi Kondor.
\newblock Lorentz group equivariant neural network for particle physics.
\newblock In \emph{ICML}, 2020.

\bibitem[Borsuk(1933)]{borsuk1933drei}
Karol Borsuk.
\newblock Drei s{\"a}tze {\"u}ber die n-dimensionale euklidische sph{\"a}re.
\newblock \emph{Fundamenta Mathematicae}, 20\penalty0 (1):\penalty0 177--190,
  1933.

\bibitem[Boscaini et~al.(2015)Boscaini, Eynard, Kourounis, and
  Bronstein]{boscaini2015shape}
Davide Boscaini, Davide Eynard, Drosos Kourounis, and Michael~M Bronstein.
\newblock Shape-from-operator: Recovering shapes from intrinsic operators.
\newblock \emph{Computer Graphics Forum}, 34\penalty0 (2):\penalty0 265--274,
  2015.

\bibitem[Boscaini et~al.(2016{\natexlab{a}})Boscaini, Masci, Rodoi{\`a}, and
  Bronstein]{boscaini2016learning}
Davide Boscaini, Jonathan Masci, Emanuele Rodoi{\`a}, and Michael Bronstein.
\newblock Learning shape correspondence with anisotropic convolutional neural
  networks.
\newblock In \emph{NIPS}, 2016{\natexlab{a}}.

\bibitem[Boscaini et~al.(2016{\natexlab{b}})Boscaini, Masci, Rodol{\`a},
  Bronstein, and Cremers]{boscaini2016anisotropic}
Davide Boscaini, Jonathan Masci, Emanuele Rodol{\`a}, Michael~M Bronstein, and
  Daniel Cremers.
\newblock Anisotropic diffusion descriptors.
\newblock \emph{Computer Graphics Forum}, 35\penalty0 (2):\penalty0 431--441,
  2016{\natexlab{b}}.

\bibitem[Bougleux et~al.(2015)Bougleux, Brun, Carletti, Foggia, Ga{\"u}zere,
  and Vento]{bougleux2015quadratic}
S{\'e}bastien Bougleux, Luc Brun, Vincenzo Carletti, Pasquale Foggia, Benoit
  Ga{\"u}zere, and Mario Vento.
\newblock A quadratic assignment formulation of the graph edit distance.
\newblock \emph{arXiv:1512.07494}, 2015.

\bibitem[Bouritsas et~al.(2020)Bouritsas, Frasca, Zafeiriou, and
  Bronstein]{bouritsas2020improving}
Giorgos Bouritsas, Fabrizio Frasca, Stefanos Zafeiriou, and Michael~M
  Bronstein.
\newblock Improving graph neural network expressivity via subgraph isomorphism
  counting.
\newblock \emph{arXiv:2006.09252}, 2020.

\bibitem[Bronstein et~al.(2006)Bronstein, Bronstein, and
  Kimmel]{bronstein2006generalized}
Alexander~M Bronstein, Michael~M Bronstein, and Ron Kimmel.
\newblock Generalized multidimensional scaling: a framework for
  isometry-invariant partial surface matching.
\newblock \emph{PNAS}, 103\penalty0 (5):\penalty0 1168--1172, 2006.

\bibitem[Bronstein et~al.(2008)Bronstein, Bronstein, and
  Kimmel]{bronstein2008numerical}
Alexander~M Bronstein, Michael~M Bronstein, and Ron Kimmel.
\newblock \emph{Numerical geometry of non-rigid shapes}.
\newblock Springer, 2008.

\bibitem[Bronstein et~al.(2017)Bronstein, Bruna, LeCun, Szlam, and
  Vandergheynst]{bronstein2017geometric}
Michael~M Bronstein, Joan Bruna, Yann LeCun, Arthur Szlam, and Pierre
  Vandergheynst.
\newblock Geometric deep learning: going beyond {E}uclidean data.
\newblock \emph{IEEE Signal Processing Magazine}, 34\penalty0 (4):\penalty0
  18--42, 2017.

\bibitem[Brown et~al.(2020)Brown, Mann, Ryder, Subbiah, Kaplan, Dhariwal,
  Neelakantan, Shyam, Sastry, Askell, et~al.]{brown2020language}
Tom~B Brown, Benjamin Mann, Nick Ryder, Melanie Subbiah, Jared Kaplan, Prafulla
  Dhariwal, Arvind Neelakantan, Pranav Shyam, Girish Sastry, Amanda Askell,
  et~al.
\newblock Language models are few-shot learners.
\newblock \emph{arXiv:2005.14165}, 2020.

\bibitem[Bruna and Mallat(2013)]{bruna2013invariant}
Joan Bruna and St{\'e}phane Mallat.
\newblock Invariant scattering convolution networks.
\newblock \emph{IEEE transactions on pattern analysis and machine
  intelligence}, 35\penalty0 (8):\penalty0 1872--1886, 2013.

\bibitem[Bruna et~al.(2013)Bruna, Zaremba, Szlam, and LeCun]{bruna2013spectral}
Joan Bruna, Wojciech Zaremba, Arthur Szlam, and Yann LeCun.
\newblock Spectral networks and locally connected networks on graphs.
\newblock In \emph{ICLR}, 2013.

\bibitem[Cappart et~al.(2021)Cappart, Ch{\'e}telat, Khalil, Lodi, Morris, and
  Veli{\v{c}}kovi{\'c}]{cappart2021combinatorial}
Quentin Cappart, Didier Ch{\'e}telat, Elias Khalil, Andrea Lodi, Christopher
  Morris, and Petar Veli{\v{c}}kovi{\'c}.
\newblock Combinatorial optimization and reasoning with graph neural networks.
\newblock \emph{arXiv:2102.09544}, 2021.

\bibitem[Chen et~al.(2018)Chen, Rubanova, Bettencourt, and
  Duvenaud]{chen2018neural}
Ricky~TQ Chen, Yulia Rubanova, Jesse Bettencourt, and David Duvenaud.
\newblock Neural ordinary differential equations.
\newblock \emph{arXiv:1806.07366}, 2018.

\bibitem[Chen et~al.(2020)Chen, Kornblith, Norouzi, and Hinton]{chen2020simple}
Ting Chen, Simon Kornblith, Mohammad Norouzi, and Geoffrey Hinton.
\newblock A simple framework for contrastive learning of visual
  representations.
\newblock In \emph{ICML}, 2020.

\bibitem[Chern et~al.(2018)Chern, Kn{\"o}ppel, Pinkall, and
  Schr{\"o}der]{chern2018shape}
Albert Chern, Felix Kn{\"o}ppel, Ulrich Pinkall, and Peter Schr{\"o}der.
\newblock Shape from metric.
\newblock \emph{ACM Trans. Graphics}, 37\penalty0 (4):\penalty0 1--17, 2018.

\bibitem[Cho et~al.(2014)Cho, Van~Merri{\"e}nboer, Gulcehre, Bahdanau,
  Bougares, Schwenk, and Bengio]{cho2014learning}
Kyunghyun Cho, Bart Van~Merri{\"e}nboer, Caglar Gulcehre, Dzmitry Bahdanau,
  Fethi Bougares, Holger Schwenk, and Yoshua Bengio.
\newblock Learning phrase representations using rnn encoder-decoder for
  statistical machine translation.
\newblock \emph{arXiv:1406.1078}, 2014.

\bibitem[Choma et~al.(2018)Choma, Monti, Gerhardt, Palczewski, Ronaghi,
  Prabhat, Bhimji, Bronstein, Klein, and Bruna]{choma2018graph}
Nicholas Choma, Federico Monti, Lisa Gerhardt, Tomasz Palczewski, Zahra
  Ronaghi, Prabhat Prabhat, Wahid Bhimji, Michael~M Bronstein, Spencer~R Klein,
  and Joan Bruna.
\newblock Graph neural networks for icecube signal classification.
\newblock In \emph{ICMLA}, 2018.

\bibitem[Cohen and Welling(2016)]{cohen2016group}
Taco Cohen and Max Welling.
\newblock Group equivariant convolutional networks.
\newblock In \emph{ICML}, 2016.

\bibitem[Cohen et~al.(2019)Cohen, Weiler, Kicanaoglu, and
  Welling]{cohen2019gauge}
Taco Cohen, Maurice Weiler, Berkay Kicanaoglu, and Max Welling.
\newblock Gauge equivariant convolutional networks and the icosahedral {CNN}.
\newblock In \emph{ICML}, 2019.

\bibitem[Cohen et~al.(2018)Cohen, Geiger, K{\"o}hler, and
  Welling]{cohen2018spherical}
Taco~S Cohen, Mario Geiger, Jonas K{\"o}hler, and Max Welling.
\newblock Spherical cnns.
\newblock \emph{arXiv:1801.10130}, 2018.

\bibitem[Cooijmans et~al.(2016)Cooijmans, Ballas, Laurent, G{\"u}l{\c{c}}ehre,
  and Courville]{cooijmans2016recurrent}
Tim Cooijmans, Nicolas Ballas, C{\'e}sar Laurent, {\c{C}}a{\u{g}}lar
  G{\"u}l{\c{c}}ehre, and Aaron Courville.
\newblock Recurrent batch normalization.
\newblock \emph{arXiv:1603.09025}, 2016.

\bibitem[Corman et~al.(2017)Corman, Solomon, Ben-Chen, Guibas, and
  Ovsjanikov]{corman2017functional}
Etienne Corman, Justin Solomon, Mirela Ben-Chen, Leonidas Guibas, and Maks
  Ovsjanikov.
\newblock Functional characterization of intrinsic and extrinsic geometry.
\newblock \emph{ACM Trans. Graphics}, 36\penalty0 (2):\penalty0 1--17, 2017.

\bibitem[Cormen et~al.(2009)Cormen, Leiserson, Rivest, and
  Stein]{cormen2009introduction}
Thomas~H Cormen, Charles~E Leiserson, Ronald~L Rivest, and Clifford Stein.
\newblock \emph{Introduction to algorithms}.
\newblock MIT press, 2009.

\bibitem[Corso et~al.(2020)Corso, Cavalleri, Beaini, Li{\`o}, and
  Veli{\v{c}}kovi{\'c}]{corso2020principal}
Gabriele Corso, Luca Cavalleri, Dominique Beaini, Pietro Li{\`o}, and Petar
  Veli{\v{c}}kovi{\'c}.
\newblock Principal neighbourhood aggregation for graph nets.
\newblock \emph{arXiv:2004.05718}, 2020.

\bibitem[Cosmo et~al.(2020)Cosmo, Kazi, Ahmadi, Navab, and
  Bronstein]{cosmo2020latent}
Luca Cosmo, Anees Kazi, Seyed-Ahmad Ahmadi, Nassir Navab, and Michael
  Bronstein.
\newblock Latent-graph learning for disease prediction.
\newblock In \emph{MICCAI}, 2020.

\bibitem[Cranmer et~al.(2020)Cranmer, Greydanus, Hoyer, Battaglia, Spergel, and
  Ho]{cranmer2020lagrangian}
Miles Cranmer, Sam Greydanus, Stephan Hoyer, Peter Battaglia, David Spergel,
  and Shirley Ho.
\newblock Lagrangian neural networks.
\newblock \emph{arXiv:2003.04630}, 2020.

\bibitem[Cranmer et~al.(2019)Cranmer, Xu, Battaglia, and
  Ho]{cranmer2019learning}
Miles~D Cranmer, Rui Xu, Peter Battaglia, and Shirley Ho.
\newblock Learning symbolic physics with graph networks.
\newblock \emph{arXiv:1909.05862}, 2019.

\bibitem[Cucurull et~al.(2018)Cucurull, Wagstyl, Casanova,
  Veli{\v{c}}kovi{\'c}, Jakobsen, Drozdzal, Romero, Evans, and
  Bengio]{cucurull2018convolutional}
Guillem Cucurull, Konrad Wagstyl, Arantxa Casanova, Petar Veli{\v{c}}kovi{\'c},
  Estrid Jakobsen, Michal Drozdzal, Adriana Romero, Alan Evans, and Yoshua
  Bengio.
\newblock Convolutional neural networks for mesh-based parcellation of the
  cerebral cortex.
\newblock 2018.

\bibitem[Cybenko(1989)]{cybenko1989approximation}
George Cybenko.
\newblock Approximation by superpositions of a sigmoidal function.
\newblock \emph{Mathematics of Control, Signals and Systems}, 2\penalty0
  (4):\penalty0 303--314, 1989.

\bibitem[Dai et~al.(2016)Dai, Dai, and Song]{dai2016discriminative}
Hanjun Dai, Bo~Dai, and Le~Song.
\newblock Discriminative embeddings of latent variable models for structured
  data.
\newblock In \emph{ICML}, 2016.

\bibitem[De~Fauw et~al.(2018)De~Fauw, Ledsam, Romera-Paredes, Nikolov, Tomasev,
  Blackwell, Askham, Glorot, O’Donoghue, Visentin, et~al.]{de2018clinically}
Jeffrey De~Fauw, Joseph~R Ledsam, Bernardino Romera-Paredes, Stanislav Nikolov,
  Nenad Tomasev, Sam Blackwell, Harry Askham, Xavier Glorot, Brendan
  O’Donoghue, Daniel Visentin, et~al.
\newblock Clinically applicable deep learning for diagnosis and referral in
  retinal disease.
\newblock \emph{Nature Medicine}, 24\penalty0 (9):\penalty0 1342--1350, 2018.

\bibitem[de~Haan et~al.(2020)de~Haan, Weiler, Cohen, and
  Welling]{deHaan2020gauge}
Pim de~Haan, Maurice Weiler, Taco Cohen, and Max Welling.
\newblock Gauge equivariant mesh {CNNs}: Anisotropic convolutions on geometric
  graphs.
\newblock In \emph{NeurIPS}, 2020.

\bibitem[Deac et~al.(2019)Deac, Veli{\v{c}}kovi{\'c}, and
  Sormanni]{deac2019attentive}
Andreea Deac, Petar Veli{\v{c}}kovi{\'c}, and Pietro Sormanni.
\newblock Attentive cross-modal paratope prediction.
\newblock \emph{Journal of Computational Biology}, 26\penalty0 (6):\penalty0
  536--545, 2019.

\bibitem[Deac et~al.(2020)Deac, Veli{\v{c}}kovi{\'c}, Milinkovi{\'c}, Bacon,
  Tang, and Nikoli{\'c}]{deac2020xlvin}
Andreea Deac, Petar Veli{\v{c}}kovi{\'c}, Ognjen Milinkovi{\'c}, Pierre-Luc
  Bacon, Jian Tang, and Mladen Nikoli{\'c}.
\newblock Xlvin: executed latent value iteration nets.
\newblock \emph{arXiv:2010.13146}, 2020.

\bibitem[Defferrard et~al.(2016)Defferrard, Bresson, and
  Vandergheynst]{defferrard2016convolutional}
Micha{\"e}l Defferrard, Xavier Bresson, and Pierre Vandergheynst.
\newblock Convolutional neural networks on graphs with fast localized spectral
  filtering.
\newblock \emph{NIPS}, 2016.

\bibitem[Derrow-Pinion et~al.(2021)Derrow-Pinion, She, Wong, Lange, Hester,
  Perez, Nunkesser, Lee, Guo, Battaglia, Gupta, Li, Xu, Sanchez-Gonzalez, Li,
  and Veli\v{c}kovi\'{c}]{derrowpinion2021traffic}
Austin Derrow-Pinion, Jennifer She, David Wong, Oliver Lange, Todd Hester, Luis
  Perez, Marc Nunkesser, Seongjae Lee, Xueying Guo, Peter~W Battaglia, Vishal
  Gupta, Ang Li, Zhongwen Xu, Alvaro Sanchez-Gonzalez, Yujia Li, and Petar
  Veli\v{c}kovi\'{c}.
\newblock {Traffic Prediction with Graph Neural Networks in Google Maps}.
\newblock 2021.

\bibitem[Devlin et~al.(2018)Devlin, Chang, Lee, and Toutanova]{devlin2018bert}
Jacob Devlin, Ming-Wei Chang, Kenton Lee, and Kristina Toutanova.
\newblock Bert: Pre-training of deep bidirectional transformers for language
  understanding.
\newblock \emph{arXiv:1810.04805}, 2018.

\bibitem[Duvenaud et~al.(2015)Duvenaud, Maclaurin, Iparraguirre, Bombarell,
  Hirzel, Aspuru-Guzik, and Adams]{duvenaud2015convolutional}
David~K Duvenaud, Dougal Maclaurin, Jorge Iparraguirre, Rafael Bombarell,
  Timothy Hirzel, Al{\'a}n Aspuru-Guzik, and Ryan~P Adams.
\newblock Convolutional networks on graphs for learning molecular fingerprints.
\newblock \emph{NIPS}, 2015.

\bibitem[Dwivedi and Bresson(2020)]{dwivedi2020generalization}
Vijay~Prakash Dwivedi and Xavier Bresson.
\newblock A generalization of transformer networks to graphs.
\newblock \emph{arXiv:2012.09699}, 2020.

\bibitem[Elad and Kimmel(2003)]{elad2003bending}
Asi Elad and Ron Kimmel.
\newblock On bending invariant signatures for surfaces.
\newblock \emph{Trans. PAMI}, 25\penalty0 (10):\penalty0 1285--1295, 2003.

\bibitem[Elman(1990)]{elman1990finding}
Jeffrey~L Elman.
\newblock Finding structure in time.
\newblock \emph{Cognitive Science}, 14\penalty0 (2):\penalty0 179--211, 1990.

\bibitem[Esteves et~al.(2020)Esteves, Makadia, and Daniilidis]{esteves2020spin}
Carlos Esteves, Ameesh Makadia, and Kostas Daniilidis.
\newblock Spin-weighted spherical {CNN}s.
\newblock \emph{arXiv:2006.10731}, 2020.

\bibitem[Fang et~al.(2020)Fang, Huang, Wang, Zeng, Liang, and
  Wang]{fang2020constgat}
Xiaomin Fang, Jizhou Huang, Fan Wang, Lingke Zeng, Haijin Liang, and Haifeng
  Wang.
\newblock {ConSTGAT}: Contextual spatial-temporal graph attention network for
  travel time estimation at baidu maps.
\newblock In \emph{KDD}, 2020.

\bibitem[Fey et~al.(2020)Fey, Yuen, and Weichert]{fey2020hierarchical}
Matthias Fey, Jan-Gin Yuen, and Frank Weichert.
\newblock Hierarchical inter-message passing for learning on molecular graphs.
\newblock \emph{arXiv:2006.12179}, 2020.

\bibitem[Finzi et~al.(2020)Finzi, Stanton, Izmailov, and
  Wilson]{finzi2020generalizing}
Marc Finzi, Samuel Stanton, Pavel Izmailov, and Andrew~Gordon Wilson.
\newblock Generalizing convolutional neural networks for equivariance to lie
  groups on arbitrary continuous data.
\newblock In \emph{ICML}, 2020.

\bibitem[Folkman(1967)]{folkman1967regular}
Jon Folkman.
\newblock Regular line-symmetric graphs.
\newblock \emph{Journal of Combinatorial Theory}, 3\penalty0 (3):\penalty0
  215--232, 1967.

\bibitem[Franceschi et~al.(2019)Franceschi, Niepert, Pontil, and
  He]{franceschi2019learning}
Luca Franceschi, Mathias Niepert, Massimiliano Pontil, and Xiao He.
\newblock Learning discrete structures for graph neural networks.
\newblock In \emph{ICML}, 2019.

\bibitem[Frasconi et~al.(1998)Frasconi, Gori, and
  Sperduti]{frasconi1998general}
Paolo Frasconi, Marco Gori, and Alessandro Sperduti.
\newblock A general framework for adaptive processing of data structures.
\newblock \emph{IEEE Trans. Neural Networks}, 9\penalty0 (5):\penalty0
  768--786, 1998.

\bibitem[Freivalds et~al.(2019)Freivalds, Ozoli{\c{n}}{\v{s}}, and
  {\v{S}}ostaks]{freivalds2019neural}
K{\=a}rlis Freivalds, Em{\=\i}ls Ozoli{\c{n}}{\v{s}}, and Agris {\v{S}}ostaks.
\newblock Neural shuffle-exchange networks--sequence processing in o (n log n)
  time.
\newblock \emph{arXiv:1907.07897}, 2019.

\bibitem[Fuchs et~al.(2020)Fuchs, Worrall, Fischer, and Welling]{fuchs2020se}
Fabian~B Fuchs, Daniel~E Worrall, Volker Fischer, and Max Welling.
\newblock {SE}(3)-transformers: {3D} roto-translation equivariant attention
  networks.
\newblock \emph{arXiv:2006.10503}, 2020.

\bibitem[Fukushima and Miyake(1982)]{fukushima1982neocognitron}
Kunihiko Fukushima and Sei Miyake.
\newblock Neocognitron: A self-organizing neural network model for a mechanism
  of visual pattern recognition.
\newblock In \emph{Competition and Cooperation in Neural Nets}, pages 267--285.
  Springer, 1982.

\bibitem[Gainza et~al.(2020)Gainza, Sverrisson, Monti, Rodola, Boscaini,
  Bronstein, and Correia]{gainza2020deciphering}
Pablo Gainza, Freyr Sverrisson, Frederico Monti, Emanuele Rodola, D~Boscaini,
  MM~Bronstein, and BE~Correia.
\newblock Deciphering interaction fingerprints from protein molecular surfaces
  using geometric deep learning.
\newblock \emph{Nature Methods}, 17\penalty0 (2):\penalty0 184--192, 2020.

\bibitem[Gama et~al.(2019)Gama, Ribeiro, and Bruna]{gama2019diffusion}
Fernando Gama, Alejandro Ribeiro, and Joan Bruna.
\newblock Diffusion scattering transforms on graphs.
\newblock In \emph{ICLR}, 2019.

\bibitem[Gama et~al.(2020)Gama, Bruna, and Ribeiro]{gama2020stability}
Fernando Gama, Joan Bruna, and Alejandro Ribeiro.
\newblock Stability properties of graph neural networks.
\newblock \emph{IEEE Trans. Signal Processing}, 68:\penalty0 5680--5695, 2020.

\bibitem[Gao et~al.(2019)Gao, Pei, and Huang]{gao2019conditional}
Hongchang Gao, Jian Pei, and Heng Huang.
\newblock Conditional random field enhanced graph convolutional neural
  networks.
\newblock In \emph{KDD}, 2019.

\bibitem[Garc{\'\i}a-Dur{\'a}n and Niepert(2017)]{garcia2017learning}
Alberto Garc{\'\i}a-Dur{\'a}n and Mathias Niepert.
\newblock Learning graph representations with embedding propagation.
\newblock \emph{arXiv:1710.03059}, 2017.

\bibitem[Gatys et~al.(2015)Gatys, Ecker, and Bethge]{gatys2015texture}
Leon~A Gatys, Alexander~S Ecker, and Matthias Bethge.
\newblock Texture synthesis using convolutional neural networks.
\newblock \emph{arXiv preprint arXiv:1505.07376}, 2015.

\bibitem[Gaudelet et~al.(2020)Gaudelet, Day, Jamasb, Soman, Regep, Liu, Hayter,
  Vickers, Roberts, Tang, et~al.]{gaudelet2020utilising}
Thomas Gaudelet, Ben Day, Arian~R Jamasb, Jyothish Soman, Cristian Regep,
  Gertrude Liu, Jeremy~BR Hayter, Richard Vickers, Charles Roberts, Jian Tang,
  et~al.
\newblock Utilising graph machine learning within drug discovery and
  development.
\newblock \emph{arXiv:2012.05716}, 2020.

\bibitem[Gers and Schmidhuber(2000)]{gers2000recurrent}
Felix~A Gers and J{\"u}rgen Schmidhuber.
\newblock Recurrent nets that time and count.
\newblock In \emph{IJCNN}, 2000.

\bibitem[Gilmer et~al.(2017)Gilmer, Schoenholz, Riley, Vinyals, and
  Dahl]{gilmer2017neural}
Justin Gilmer, Samuel~S Schoenholz, Patrick~F Riley, Oriol Vinyals, and
  George~E Dahl.
\newblock Neural message passing for quantum chemistry.
\newblock \emph{arXiv:1704.01212}, 2017.

\bibitem[Girshick(2015)]{girshick2015fast}
Ross Girshick.
\newblock Fast {R-CNN}.
\newblock In \emph{CVPR}, 2015.

\bibitem[Girshick et~al.(2014)Girshick, Donahue, Darrell, and
  Malik]{girshick2014rich}
Ross Girshick, Jeff Donahue, Trevor Darrell, and Jitendra Malik.
\newblock Rich feature hierarchies for accurate object detection and semantic
  segmentation.
\newblock In \emph{CVPR}, 2014.

\bibitem[Gligorijevic et~al.(2020)Gligorijevic, Renfrew, Kosciolek, Leman,
  Berenberg, Vatanen, Chandler, Taylor, Fisk, Vlamakis,
  et~al.]{gligorijevic2020structure}
Vladimir Gligorijevic, P~Douglas Renfrew, Tomasz Kosciolek, Julia~Koehler
  Leman, Daniel Berenberg, Tommi Vatanen, Chris Chandler, Bryn~C Taylor, Ian~M
  Fisk, Hera Vlamakis, et~al.
\newblock Structure-based function prediction using graph convolutional
  networks.
\newblock \emph{bioRxiv:786236}, 2020.

\bibitem[Goller and Kuchler(1996)]{goller1996learning}
Christoph Goller and Andreas Kuchler.
\newblock Learning task-dependent distributed representations by
  backpropagation through structure.
\newblock In \emph{ICNN}, 1996.

\bibitem[Goodfellow et~al.(2014)Goodfellow, Pouget-Abadie, Mirza, Xu,
  Warde-Farley, Ozair, Courville, and Bengio]{goodfellow2014generative}
Ian~J Goodfellow, Jean Pouget-Abadie, Mehdi Mirza, Bing Xu, David Warde-Farley,
  Sherjil Ozair, Aaron Courville, and Yoshua Bengio.
\newblock Generative adversarial networks.
\newblock \emph{arXiv:1406.2661}, 2014.

\bibitem[Gori et~al.(2005)Gori, Monfardini, and Scarselli]{gori2005new}
Marco Gori, Gabriele Monfardini, and Franco Scarselli.
\newblock A new model for learning in graph domains.
\newblock In \emph{IJCNN}, 2005.

\bibitem[Graves(2013)]{graves2013generating}
Alex Graves.
\newblock Generating sequences with recurrent neural networks.
\newblock \emph{arXiv:1308.0850}, 2013.

\bibitem[Graves et~al.(2014)Graves, Wayne, and Danihelka]{graves2014neural}
Alex Graves, Greg Wayne, and Ivo Danihelka.
\newblock Neural turing machines.
\newblock \emph{arXiv:1410.5401}, 2014.

\bibitem[Graves et~al.(2016)Graves, Wayne, Reynolds, Harley, Danihelka,
  Grabska-Barwi{\'n}ska, Colmenarejo, Grefenstette, Ramalho, Agapiou,
  et~al.]{graves2016hybrid}
Alex Graves, Greg Wayne, Malcolm Reynolds, Tim Harley, Ivo Danihelka, Agnieszka
  Grabska-Barwi{\'n}ska, Sergio~G{\'o}mez Colmenarejo, Edward Grefenstette,
  Tiago Ramalho, John Agapiou, et~al.
\newblock Hybrid computing using a neural network with dynamic external memory.
\newblock \emph{Nature}, 538\penalty0 (7626):\penalty0 471--476, 2016.

\bibitem[Grill et~al.(2020)Grill, Strub, Altch{\'e}, Tallec, Richemond,
  Buchatskaya, Doersch, Pires, Guo, Azar, et~al.]{grill2020bootstrap}
Jean-Bastien Grill, Florian Strub, Florent Altch{\'e}, Corentin Tallec,
  Pierre~H Richemond, Elena Buchatskaya, Carl Doersch, Bernardo~Avila Pires,
  Zhaohan~Daniel Guo, Mohammad~Gheshlaghi Azar, et~al.
\newblock Bootstrap your own latent: A new approach to self-supervised
  learning.
\newblock \emph{arXiv:2006.07733}, 2020.

\bibitem[Gromov(1981)]{gromov1981structures}
Mikhael Gromov.
\newblock \emph{Structures m{\'e}triques pour les vari{\'e}t{\'e}s
  riemanniennes}.
\newblock Cedic, 1981.

\bibitem[Grover and Leskovec(2016)]{grover2016node2vec}
Aditya Grover and Jure Leskovec.
\newblock node2vec: Scalable feature learning for networks.
\newblock In \emph{KDD}, 2016.

\bibitem[Gunasekar et~al.(2017)Gunasekar, Woodworth, Bhojanapalli, Neyshabur,
  and Srebro]{gunasekar2017implicit}
Suriya Gunasekar, Blake~E Woodworth, Srinadh Bhojanapalli, Behnam Neyshabur,
  and Nati Srebro.
\newblock Implicit regularization in matrix factorization.
\newblock In \emph{NIPS}, 2017.

\bibitem[Gysi et~al.(2020)Gysi, Do~Valle, Zitnik, Ameli, Gan, Varol, Sanchez,
  Baron, Ghiassian, Loscalzo, et~al.]{gysi2020network}
Deisy~Morselli Gysi, {\'I}talo Do~Valle, Marinka Zitnik, Asher Ameli, Xiao Gan,
  Onur Varol, Helia Sanchez, Rebecca~Marlene Baron, Dina Ghiassian, Joseph
  Loscalzo, et~al.
\newblock Network medicine framework for identifying drug repurposing
  opportunities for {COVID}-19.
\newblock \emph{arXiv:2004.07229}, 2020.

\bibitem[Hamilton et~al.(2017)Hamilton, Ying, and
  Leskovec]{hamilton2017inductive}
Will Hamilton, Zhitao Ying, and Jure Leskovec.
\newblock Inductive representation learning on large graphs.
\newblock In \emph{NIPS}, 2017.

\bibitem[Hao et~al.(2020)Hao, Zhao, Li, Dong, Faloutsos, Sun, and
  Wang]{hao2020p}
Junheng Hao, Tong Zhao, Jin Li, Xin~Luna Dong, Christos Faloutsos, Yizhou Sun,
  and Wei Wang.
\newblock P-companion: A principled framework for diversified complementary
  product recommendation.
\newblock In \emph{Information \& Knowledge Management}, 2020.

\bibitem[Hardt and Ma(2016)]{hardt2016identity}
Moritz Hardt and Tengyu Ma.
\newblock Identity matters in deep learning.
\newblock \emph{arXiv:1611.04231}, 2016.

\bibitem[He et~al.(2016)He, Zhang, Ren, and Sun]{he2016deep}
Kaiming He, Xiangyu Zhang, Shaoqing Ren, and Jian Sun.
\newblock Deep residual learning for image recognition.
\newblock In \emph{CVPR}, 2016.

\bibitem[He et~al.(2017)He, Gkioxari, Doll{\'a}r, and Girshick]{he2017mask}
Kaiming He, Georgia Gkioxari, Piotr Doll{\'a}r, and Ross Girshick.
\newblock Mask r-cnn.
\newblock In \emph{CVPR}, 2017.

\bibitem[Helv{\'e}tius(1759)]{helvetius1759esprit}
Claude~Adrien Helv{\'e}tius.
\newblock \emph{De l'esprit}.
\newblock Durand, 1759.

\bibitem[Hjelm et~al.(2019)Hjelm, Fedorov, Lavoie-Marchildon, Grewal, Bachman,
  Trischler, and Bengio]{hjelm2018learning}
R~Devon Hjelm, Alex Fedorov, Samuel Lavoie-Marchildon, Karan Grewal, Phil
  Bachman, Adam Trischler, and Yoshua Bengio.
\newblock Learning deep representations by mutual information estimation and
  maximization.
\newblock In \emph{ICLR}, 2019.

\bibitem[Hochreiter(1991)]{hochreiter1991untersuchungen}
Sepp Hochreiter.
\newblock \emph{Untersuchungen zu dynamischen neuronalen {N}etzen}.
\newblock PhD thesis, Technische Universit{\"a}t M{\"u}nchen, 1991.

\bibitem[Hochreiter and Schmidhuber(1997)]{hochreiter1997long}
Sepp Hochreiter and J{\"u}rgen Schmidhuber.
\newblock Long short-term memory.
\newblock \emph{Neural Computation}, 9\penalty0 (8):\penalty0 1735--1780, 1997.

\bibitem[Hornik(1991)]{hornik1991approximation}
Kurt Hornik.
\newblock Approximation capabilities of multilayer feedforward networks.
\newblock \emph{Neural Networks}, 4\penalty0 (2):\penalty0 251--257, 1991.

\bibitem[Hoshen(2017)]{hoshen2017vain}
Yedid Hoshen.
\newblock Vain: Attentional multi-agent predictive modeling.
\newblock \emph{arXiv:1706.06122}, 2017.

\bibitem[Hu et~al.(2020)Hu, Liu, Gomes, Zitnik, Liang, Pande, and
  Leskovec]{hu2020strategies}
Weihua Hu, Bowen Liu, Joseph Gomes, Marinka Zitnik, Percy Liang, Vijay Pande,
  and Jure Leskovec.
\newblock Strategies for pre-training graph neural networks.
\newblock In \emph{ICLR}, 2020.

\bibitem[Hubel and Wiesel(1959)]{hubel1959receptive}
David~H Hubel and Torsten~N Wiesel.
\newblock Receptive fields of single neurones in the cat's striate cortex.
\newblock \emph{J. Physiology}, 148\penalty0 (3):\penalty0 574--591, 1959.

\bibitem[Hutchinson et~al.(2020)Hutchinson, Lan, Zaidi, Dupont, Teh, and
  Kim]{hutchinson2020lietransformer}
Michael Hutchinson, Charline~Le Lan, Sheheryar Zaidi, Emilien Dupont, Yee~Whye
  Teh, and Hyunjik Kim.
\newblock {LieTransformer}: Equivariant self-attention for {L}ie groups.
\newblock \emph{arXiv:2012.10885}, 2020.

\bibitem[Ioffe and Szegedy(2015)]{ioffe2015batch}
Sergey Ioffe and Christian Szegedy.
\newblock Batch normalization: Accelerating deep network training by reducing
  internal covariate shift.
\newblock In \emph{ICML}, 2015.

\bibitem[Iqbal(2018)]{haris2018}
Haris Iqbal.
\newblock Harisiqbal88/plotneuralnet v1.0.0, December 2018.
\newblock URL \url{https://doi.org/10.5281/zenodo.2526396}.

\bibitem[Itani and Thanou(2021)]{itani2021combining}
Sarah Itani and Dorina Thanou.
\newblock Combining anatomical and functional networks for neuropathology
  identification: A case study on autism spectrum disorder.
\newblock \emph{Medical Image Analysis}, 69:\penalty0 101986, 2021.

\bibitem[Jin et~al.(2018)Jin, Barzilay, and Jaakkola]{jin2018junction}
Wengong Jin, Regina Barzilay, and Tommi Jaakkola.
\newblock Junction tree variational autoencoder for molecular graph generation.
\newblock In \emph{ICML}, 2018.

\bibitem[Jin et~al.(2020)Jin, Barzilay, and Jaakkola]{jin2020hierarchical}
Wengong Jin, Regina Barzilay, and Tommi Jaakkola.
\newblock Hierarchical generation of molecular graphs using structural motifs.
\newblock In \emph{ICML}, 2020.

\bibitem[Johnson et~al.(2016)Johnson, Pollard, Shen, Li-Wei, Feng, Ghassemi,
  Moody, Szolovits, Celi, and Mark]{johnson2016mimic}
Alistair~EW Johnson, Tom~J Pollard, Lu~Shen, H~Lehman Li-Wei, Mengling Feng,
  Mohammad Ghassemi, Benjamin Moody, Peter Szolovits, Leo~Anthony Celi, and
  Roger~G Mark.
\newblock Mimic-iii, a freely accessible critical care database.
\newblock \emph{Scientific Data}, 3\penalty0 (1):\penalty0 1--9, 2016.

\bibitem[Jordan(1997)]{jordan1997serial}
Michael~I Jordan.
\newblock Serial order: A parallel distributed processing approach.
\newblock In \emph{Advances in Psychology}, volume 121, pages 471--495. 1997.

\bibitem[Joshi(2020)]{joshi2020transformers}
Chaitanya Joshi.
\newblock Transformers are graph neural networks.
\newblock \emph{The Gradient}, 2020.

\bibitem[Jozefowicz et~al.(2015)Jozefowicz, Zaremba, and
  Sutskever]{jozefowicz2015empirical}
Rafal Jozefowicz, Wojciech Zaremba, and Ilya Sutskever.
\newblock An empirical exploration of recurrent network architectures.
\newblock In \emph{ICML}, 2015.

\bibitem[Kaiser and Sutskever(2015)]{kaiser2015neural}
{\L}ukasz Kaiser and Ilya Sutskever.
\newblock Neural {GPU}s learn algorithms.
\newblock \emph{arXiv:1511.08228}, 2015.

\bibitem[Kalchbrenner et~al.(2016)Kalchbrenner, Espeholt, Simonyan, van~den
  Oord, Graves, and Kavukcuoglu]{kalchbrenner2016neural}
Nal Kalchbrenner, Lasse Espeholt, Karen Simonyan, Aaron van~den Oord, Alex
  Graves, and Koray Kavukcuoglu.
\newblock Neural machine translation in linear time.
\newblock \emph{arXiv:1610.10099}, 2016.

\bibitem[Kalchbrenner et~al.(2018)Kalchbrenner, Elsen, Simonyan, Noury,
  Casagrande, Lockhart, Stimberg, van~den Oord, Dieleman, and
  Kavukcuoglu]{kalchbrenner2018efficient}
Nal Kalchbrenner, Erich Elsen, Karen Simonyan, Seb Noury, Norman Casagrande,
  Edward Lockhart, Florian Stimberg, Aaron van~den Oord, Sander Dieleman, and
  Koray Kavukcuoglu.
\newblock Efficient neural audio synthesis.
\newblock In \emph{ICML}, 2018.

\bibitem[Kanatani(2012)]{kanatani2012group}
Ken-Ichi Kanatani.
\newblock \emph{Group-theoretical methods in image understanding}.
\newblock Springer, 2012.

\bibitem[Karni and Gotsman(2000)]{karni2000spectral}
Zachi Karni and Craig Gotsman.
\newblock Spectral compression of mesh geometry.
\newblock In \emph{Proc. Computer Graphics and Interactive Techniques}, 2000.

\bibitem[Kazi et~al.(2020)Kazi, Cosmo, Navab, and
  Bronstein]{kazi2020differentiable}
Anees Kazi, Luca Cosmo, Nassir Navab, and Michael Bronstein.
\newblock Differentiable graph module ({DGM}) graph convolutional networks.
\newblock \emph{arXiv:2002.04999}, 2020.

\bibitem[Kenlay et~al.(2021)Kenlay, Thanou, and Dong]{kenlay2021interpretable}
Henry Kenlay, Dorina Thanou, and Xiaowen Dong.
\newblock Interpretable stability bounds for spectral graph filters.
\newblock \emph{arXiv:2102.09587}, 2021.

\bibitem[Kimmel and Sethian(1998)]{kimmel1998computing}
Ron Kimmel and James~A Sethian.
\newblock Computing geodesic paths on manifolds.
\newblock \emph{PNAS}, 95\penalty0 (15):\penalty0 8431--8435, 1998.

\bibitem[Kingma and Ba(2014)]{kingma2014adam}
Diederik~P Kingma and Jimmy Ba.
\newblock Adam: A method for stochastic optimization.
\newblock \emph{arXiv:1412.6980}, 2014.

\bibitem[Kingma and Welling(2013)]{kingma2013auto}
Diederik~P Kingma and Max Welling.
\newblock Auto-encoding variational bayes.
\newblock \emph{arXiv:1312.6114}, 2013.

\bibitem[Kipf et~al.(2018)Kipf, Fetaya, Wang, Welling, and
  Zemel]{kipf2018neural}
Thomas Kipf, Ethan Fetaya, Kuan-Chieh Wang, Max Welling, and Richard Zemel.
\newblock Neural relational inference for interacting systems.
\newblock In \emph{ICML}, 2018.

\bibitem[Kipf and Welling(2016{\natexlab{a}})]{kipf2016semi}
Thomas~N Kipf and Max Welling.
\newblock Semi-supervised classification with graph convolutional networks.
\newblock \emph{arXiv:1609.02907}, 2016{\natexlab{a}}.

\bibitem[Kipf and Welling(2016{\natexlab{b}})]{kipf2016variational}
Thomas~N Kipf and Max Welling.
\newblock Variational graph auto-encoders.
\newblock \emph{arXiv:1611.07308}, 2016{\natexlab{b}}.

\bibitem[Kireev(1995)]{kireev1995chemnet}
Dmitry~B Kireev.
\newblock Chemnet: a novel neural network based method for graph/property
  mapping.
\newblock \emph{J. Chemical Information and Computer Sciences}, 35\penalty0
  (2):\penalty0 175--180, 1995.

\bibitem[Klicpera et~al.(2020)Klicpera, Gro{\ss}, and
  G{\"u}nnemann]{klicpera2020directional}
Johannes Klicpera, Janek Gro{\ss}, and Stephan G{\"u}nnemann.
\newblock Directional message passing for molecular graphs.
\newblock \emph{arXiv:2003.03123}, 2020.

\bibitem[Kokkinos et~al.(2012)Kokkinos, Bronstein, Litman, and
  Bronstein]{kokkinos2012intrinsic}
Iasonas Kokkinos, Michael~M Bronstein, Roee Litman, and Alex~M Bronstein.
\newblock Intrinsic shape context descriptors for deformable shapes.
\newblock In \emph{CVPR}, 2012.

\bibitem[Komiske et~al.(2019)Komiske, Metodiev, and Thaler]{komiske2019energy}
Patrick~T Komiske, Eric~M Metodiev, and Jesse Thaler.
\newblock Energy flow networks: deep sets for particle jets.
\newblock \emph{Journal of High Energy Physics}, 2019\penalty0 (1):\penalty0
  121, 2019.

\bibitem[Kostrikov et~al.(2018)Kostrikov, Jiang, Panozzo, Zorin, and
  Bruna]{kostrikov2018surface}
Ilya Kostrikov, Zhongshi Jiang, Daniele Panozzo, Denis Zorin, and Joan Bruna.
\newblock Surface networks.
\newblock In \emph{CVPR}, 2018.

\bibitem[Krizhevsky et~al.(2012)Krizhevsky, Sutskever, and
  Hinton]{krizhevsky2012imagenet}
Alex Krizhevsky, Ilya Sutskever, and Geoffrey~E Hinton.
\newblock Imagenet classification with deep convolutional neural networks.
\newblock In \emph{NIPS}, 2012.

\bibitem[Ktena et~al.(2017)Ktena, Parisot, Ferrante, Rajchl, Lee, Glocker, and
  Rueckert]{ktena2017distance}
Sofia~Ira Ktena, Sarah Parisot, Enzo Ferrante, Martin Rajchl, Matthew Lee, Ben
  Glocker, and Daniel Rueckert.
\newblock Distance metric learning using graph convolutional networks:
  Application to functional brain networks.
\newblock In \emph{MICCAI}, 2017.

\bibitem[Kulon et~al.(2020)Kulon, Guler, Kokkinos, Bronstein, and
  Zafeiriou]{kulon2020weakly}
Dominik Kulon, Riza~Alp Guler, Iasonas Kokkinos, Michael~M Bronstein, and
  Stefanos Zafeiriou.
\newblock Weakly-supervised mesh-convolutional hand reconstruction in the wild.
\newblock In \emph{CVPR}, 2020.

\bibitem[Kurach et~al.(2015)Kurach, Andrychowicz, and
  Sutskever]{kurach2015neural}
Karol Kurach, Marcin Andrychowicz, and Ilya Sutskever.
\newblock Neural random-access machines.
\newblock \emph{arXiv:1511.06392}, 2015.

\bibitem[LeCun et~al.(1998)LeCun, Bottou, Bengio, and
  Haffner]{lecun1998gradient}
Yann LeCun, L{\'e}on Bottou, Yoshua Bengio, and Patrick Haffner.
\newblock Gradient-based learning applied to document recognition.
\newblock \emph{Proc. IEEE}, 86\penalty0 (11):\penalty0 2278--2324, 1998.

\bibitem[Lenz(1990)]{lenz1990group}
Reiner Lenz.
\newblock \emph{Group theoretical methods in image processing}.
\newblock Springer, 1990.

\bibitem[Leshno et~al.(1993)Leshno, Lin, Pinkus, and
  Schocken]{leshno1993multilayer}
Moshe Leshno, Vladimir~Ya Lin, Allan Pinkus, and Shimon Schocken.
\newblock Multilayer feedforward networks with a nonpolynomial activation
  function can approximate any function.
\newblock \emph{Neural Networks}, 6\penalty0 (6):\penalty0 861--867, 1993.

\bibitem[Levie et~al.(2018)Levie, Monti, Bresson, and
  Bronstein]{levie2018cayleynets}
Ron Levie, Federico Monti, Xavier Bresson, and Michael~M Bronstein.
\newblock Cayleynets: Graph convolutional neural networks with complex rational
  spectral filters.
\newblock \emph{IEEE Trans. Signal Processing}, 67\penalty0 (1):\penalty0
  97--109, 2018.

\bibitem[Levie et~al.(2019)Levie, Isufi, and
  Kutyniok]{levie2019transferability}
Ron Levie, Elvin Isufi, and Gitta Kutyniok.
\newblock On the transferability of spectral graph filters.
\newblock In \emph{Sampling Theory and Applications}, 2019.

\bibitem[L{\'e}vy(2006)]{levy2006laplace}
Bruno L{\'e}vy.
\newblock {Laplace-Beltrami} eigenfunctions towards an algorithm that
  ``understands'' geometry.
\newblock In \emph{Proc. Shape Modeling and Applications}, 2006.

\bibitem[Li et~al.(2015)Li, Tarlow, Brockschmidt, and Zemel]{li2015gated}
Yujia Li, Daniel Tarlow, Marc Brockschmidt, and Richard Zemel.
\newblock Gated graph sequence neural networks.
\newblock \emph{arXiv:1511.05493}, 2015.

\bibitem[Litany et~al.(2018)Litany, Bronstein, Bronstein, and
  Makadia]{litany2018deformable}
Or~Litany, Alex Bronstein, Michael Bronstein, and Ameesh Makadia.
\newblock Deformable shape completion with graph convolutional autoencoders.
\newblock In \emph{CVPR}, 2018.

\bibitem[Litman and Bronstein(2013)]{litman2013learning}
Roee Litman and Alexander~M Bronstein.
\newblock Learning spectral descriptors for deformable shape correspondence.
\newblock \emph{Trans. PAMI}, 36\penalty0 (1):\penalty0 171--180, 2013.

\bibitem[Liu et~al.(2017)Liu, Jacobson, and Crane]{liu2017dirac}
Hsueh-Ti~Derek Liu, Alec Jacobson, and Keenan Crane.
\newblock A {D}irac operator for extrinsic shape analysis.
\newblock \emph{Computer Graphics Forum}, 36\penalty0 (5):\penalty0 139--149,
  2017.

\bibitem[Lyu and Simoncelli(2008)]{lyu2008nonlinear}
Siwei Lyu and Eero~P Simoncelli.
\newblock Nonlinear image representation using divisive normalization.
\newblock In \emph{CVPR}, 2008.

\bibitem[MacNeal(1949)]{macneal1949solution}
Richard~H MacNeal.
\newblock \emph{The solution of partial differential equations by means of
  electrical networks}.
\newblock PhD thesis, California Institute of Technology, 1949.

\bibitem[Madsen and Johansen(2020)]{madsen2020neural}
Andreas Madsen and Alexander~Rosenberg Johansen.
\newblock Neural arithmetic units.
\newblock \emph{arXiv:2001.05016}, 2020.

\bibitem[Mahdi et~al.(2020)Mahdi, Nauwelaers, Joris, Bouritsas, Gong, Bokhnyak,
  Walsh, Shriver, Bronstein, and Claes]{mahdi20203d}
Soha~Sadat Mahdi, Nele Nauwelaers, Philip Joris, Giorgos Bouritsas, Shunwang
  Gong, Sergiy Bokhnyak, Susan Walsh, Mark Shriver, Michael Bronstein, and
  Peter Claes.
\newblock 3d facial matching by spiral convolutional metric learning and a
  biometric fusion-net of demographic properties.
\newblock \emph{arXiv:2009.04746}, 2020.

\bibitem[Maiorov(1999)]{maiorov1999best}
VE~Maiorov.
\newblock On best approximation by ridge functions.
\newblock \emph{Journal of Approximation Theory}, 99\penalty0 (1):\penalty0
  68--94, 1999.

\bibitem[Makadia et~al.(2007)Makadia, Geyer, and
  Daniilidis]{makadia2007correspondence}
Ameesh Makadia, Christopher Geyer, and Kostas Daniilidis.
\newblock Correspondence-free structure from motion.
\newblock \emph{IJCV}, 75\penalty0 (3):\penalty0 311--327, 2007.

\bibitem[Mallat(1999)]{mallat1999wavelet}
St{\'e}phane Mallat.
\newblock \emph{A wavelet tour of signal processing}.
\newblock Elsevier, 1999.

\bibitem[Mallat(2012)]{mallat2012group}
St{\'e}phane Mallat.
\newblock Group invariant scattering.
\newblock \emph{Communications on Pure and Applied Mathematics}, 65\penalty0
  (10):\penalty0 1331--1398, 2012.

\bibitem[Malone et~al.(2018)Malone, Garcia-Duran, and
  Niepert]{malone2018learning}
Brandon Malone, Alberto Garcia-Duran, and Mathias Niepert.
\newblock Learning representations of missing data for predicting patient
  outcomes.
\newblock \emph{arXiv:1811.04752}, 2018.

\bibitem[Maron et~al.(2018)Maron, Ben-Hamu, Shamir, and
  Lipman]{maron2018invariant}
Haggai Maron, Heli Ben-Hamu, Nadav Shamir, and Yaron Lipman.
\newblock Invariant and equivariant graph networks.
\newblock \emph{arXiv:1812.09902}, 2018.

\bibitem[Maron et~al.(2019)Maron, Ben-Hamu, Serviansky, and
  Lipman]{maron2019provably}
Haggai Maron, Heli Ben-Hamu, Hadar Serviansky, and Yaron Lipman.
\newblock Provably powerful graph networks.
\newblock \emph{arXiv:1905.11136}, 2019.

\bibitem[Marquis(2009)]{marquis2009category}
Jean-Pierre Marquis.
\newblock Category theory and klein’s erlangen program.
\newblock In \emph{From a Geometrical Point of View}, pages 9--40. Springer,
  2009.

\bibitem[Masci et~al.(2015)Masci, Boscaini, Bronstein, and
  Vandergheynst]{masci2015geodesic}
Jonathan Masci, Davide Boscaini, Michael Bronstein, and Pierre Vandergheynst.
\newblock Geodesic convolutional neural networks on {R}iemannian manifolds.
\newblock In \emph{CVPR Workshops}, 2015.

\bibitem[Maxwell(1865)]{maxwell1865viii}
James~Clerk Maxwell.
\newblock A dynamical theory of the electromagnetic field.
\newblock \emph{Philosophical Transactions of the Royal Society of London},
  \penalty0 (155):\penalty0 459--512, 1865.

\bibitem[McEwen et~al.(2021)McEwen, Wallis, and
  Mavor-Parker]{mcewen2021scattering}
Jason~D McEwen, Christopher~GR Wallis, and Augustine~N Mavor-Parker.
\newblock Scattering networks on the sphere for scalable and rotationally
  equivariant spherical cnns.
\newblock \emph{arXiv:2102.02828}, 2021.

\bibitem[Mei et~al.(2021)Mei, Misiakiewicz, and Montanari]{mei2021learning}
Song Mei, Theodor Misiakiewicz, and Andrea Montanari.
\newblock Learning with invariances in random features and kernel models.
\newblock \emph{arXiv:2102.13219}, 2021.

\bibitem[Melzi et~al.(2019)Melzi, Spezialetti, Tombari, Bronstein, Stefano, and
  Rodol{\`a}]{melzi2019gframes}
Simone Melzi, Riccardo Spezialetti, Federico Tombari, Michael~M Bronstein,
  Luigi~Di Stefano, and Emanuele Rodol{\`a}.
\newblock Gframes: Gradient-based local reference frame for 3d shape matching.
\newblock In \emph{CVPR}, 2019.

\bibitem[M{\'e}moli and Sapiro(2005)]{memoli2005theoretical}
Facundo M{\'e}moli and Guillermo Sapiro.
\newblock A theoretical and computational framework for isometry invariant
  recognition of point cloud data.
\newblock \emph{Foundations of Computational Mathematics}, 5\penalty0
  (3):\penalty0 313--347, 2005.

\bibitem[Merkwirth and Lengauer(2005)]{merkwirth2005automatic}
Christian Merkwirth and Thomas Lengauer.
\newblock Automatic generation of complementary descriptors with molecular
  graph networks.
\newblock \emph{J. Chemical Information and Modeling}, 45\penalty0
  (5):\penalty0 1159--1168, 2005.

\bibitem[Meyer et~al.(2003)Meyer, Desbrun, Schr{\"o}der, and
  Barr]{meyer2003discrete}
Mark Meyer, Mathieu Desbrun, Peter Schr{\"o}der, and Alan~H Barr.
\newblock Discrete differential-geometry operators for triangulated
  2-manifolds.
\newblock In \emph{Visualization and Mathematics III}, pages 35--57. 2003.

\bibitem[Micheli(2009)]{micheli2009neural}
Alessio Micheli.
\newblock Neural network for graphs: A contextual constructive approach.
\newblock \emph{IEEE Trans. Neural Networks}, 20\penalty0 (3):\penalty0
  498--511, 2009.

\bibitem[Miller et~al.(2016)Miller, Alfaro-Almagro, Bangerter, Thomas, Yacoub,
  Xu, Bartsch, Jbabdi, Sotiropoulos, Andersson, et~al.]{miller2016multimodal}
Karla~L Miller, Fidel Alfaro-Almagro, Neal~K Bangerter, David~L Thomas, Essa
  Yacoub, Junqian Xu, Andreas~J Bartsch, Saad Jbabdi, Stamatios~N Sotiropoulos,
  Jesper~LR Andersson, et~al.
\newblock Multimodal population brain imaging in the uk biobank prospective
  epidemiological study.
\newblock \emph{Nature Neuroscience}, 19\penalty0 (11):\penalty0 1523--1536,
  2016.

\bibitem[Minsky and Papert(2017)]{minsky2017perceptrons}
Marvin Minsky and Seymour~A Papert.
\newblock \emph{Perceptrons: An introduction to computational geometry}.
\newblock MIT Press, 2017.

\bibitem[Mitrovic et~al.(2020)Mitrovic, McWilliams, Walker, Buesing, and
  Blundell]{mitrovic2020representation}
Jovana Mitrovic, Brian McWilliams, Jacob Walker, Lars Buesing, and Charles
  Blundell.
\newblock Representation learning via invariant causal mechanisms.
\newblock \emph{arXiv:2010.07922}, 2020.

\bibitem[Mnih et~al.(2015)Mnih, Kavukcuoglu, Silver, Rusu, Veness, Bellemare,
  Graves, Riedmiller, Fidjeland, Ostrovski, et~al.]{mnih2015human}
Volodymyr Mnih, Koray Kavukcuoglu, David Silver, Andrei~A Rusu, Joel Veness,
  Marc~G Bellemare, Alex Graves, Martin Riedmiller, Andreas~K Fidjeland, Georg
  Ostrovski, et~al.
\newblock Human-level control through deep reinforcement learning.
\newblock \emph{Nature}, 518\penalty0 (7540):\penalty0 529--533, 2015.

\bibitem[Mnih et~al.(2016)Mnih, Badia, Mirza, Graves, Lillicrap, Harley,
  Silver, and Kavukcuoglu]{mnih2016asynchronous}
Volodymyr Mnih, Adria~Puigdomenech Badia, Mehdi Mirza, Alex Graves, Timothy
  Lillicrap, Tim Harley, David Silver, and Koray Kavukcuoglu.
\newblock Asynchronous methods for deep reinforcement learning.
\newblock In \emph{ICML}, 2016.

\bibitem[Monti et~al.(2017)Monti, Boscaini, Masci, Rodola, Svoboda, and
  Bronstein]{monti2017geometric}
Federico Monti, Davide Boscaini, Jonathan Masci, Emanuele Rodola, Jan Svoboda,
  and Michael~M Bronstein.
\newblock Geometric deep learning on graphs and manifolds using mixture model
  cnns.
\newblock In \emph{CVPR}, 2017.

\bibitem[Monti et~al.(2019)Monti, Frasca, Eynard, Mannion, and
  Bronstein]{monti2019fake}
Federico Monti, Fabrizio Frasca, Davide Eynard, Damon Mannion, and Michael~M
  Bronstein.
\newblock Fake news detection on social media using geometric deep learning.
\newblock \emph{arXiv:1902.06673}, 2019.

\bibitem[Morris et~al.(2017)Morris, Kersting, and Mutzel]{morris2017glocalized}
Christopher Morris, Kristian Kersting, and Petra Mutzel.
\newblock Glocalized {Weisfeiler-Lehman} graph kernels: Global-local feature
  maps of graphs.
\newblock In \emph{ICDM}, 2017.

\bibitem[Morris et~al.(2019)Morris, Ritzert, Fey, Hamilton, Lenssen, Rattan,
  and Grohe]{morris2019weisfeiler}
Christopher Morris, Martin Ritzert, Matthias Fey, William~L Hamilton, Jan~Eric
  Lenssen, Gaurav Rattan, and Martin Grohe.
\newblock Weisfeiler and leman go neural: Higher-order graph neural networks.
\newblock In \emph{AAAI}, 2019.

\bibitem[Morris et~al.(2020)Morris, Rattan, and Mutzel]{morris2020weisfeiler}
Christopher Morris, Gaurav Rattan, and Petra Mutzel.
\newblock {W}eisfeiler and {L}eman go sparse: Towards scalable higher-order
  graph embeddings.
\newblock In \emph{NeurIPS}, 2020.

\bibitem[Mozer(1989)]{mozer1989focused}
Michael~C Mozer.
\newblock A focused back-propagation algorithm for temporal pattern
  recognition.
\newblock \emph{Complex Systems}, 3\penalty0 (4):\penalty0 349--381, 1989.

\bibitem[Murphy et~al.(2013)Murphy, Weiss, and Jordan]{murphy2013loopy}
Kevin Murphy, Yair Weiss, and Michael~I Jordan.
\newblock Loopy belief propagation for approximate inference: An empirical
  study.
\newblock \emph{arXiv:1301.6725}, 2013.

\bibitem[Murphy et~al.(2019)Murphy, Srinivasan, Rao, and
  Ribeiro]{murphy2019relational}
Ryan Murphy, Balasubramaniam Srinivasan, Vinayak Rao, and Bruno Ribeiro.
\newblock Relational pooling for graph representations.
\newblock In \emph{ICML}, 2019.

\bibitem[Murphy et~al.(2018)Murphy, Srinivasan, Rao, and
  Ribeiro]{murphy2018janossy}
Ryan~L Murphy, Balasubramaniam Srinivasan, Vinayak Rao, and Bruno Ribeiro.
\newblock Janossy pooling: Learning deep permutation-invariant functions for
  variable-size inputs.
\newblock \emph{arXiv:1811.01900}, 2018.

\bibitem[Nair and Hinton(2010)]{nair2010rectified}
Vinod Nair and Geoffrey~E Hinton.
\newblock Rectified linear units improve restricted boltzmann machines.
\newblock In \emph{ICML}, 2010.

\bibitem[Nash(1956)]{nash1971imbedding}
John Nash.
\newblock The imbedding problem for {R}iemannian manifolds.
\newblock \emph{Annals of Mathematics}, 63\penalty0 (1):\penalty0 20--–63,
  1956.

\bibitem[Neyshabur et~al.(2015)Neyshabur, Tomioka, and
  Srebro]{neyshabur2015norm}
Behnam Neyshabur, Ryota Tomioka, and Nathan Srebro.
\newblock Norm-based capacity control in neural networks.
\newblock In \emph{COLT}, 2015.

\bibitem[Noether(1918)]{variationsprobleme1918nachr}
Emmy Noether.
\newblock Invariante variationsprobleme.
\newblock In \emph{K{\"o}nig Gesellsch. d. Wiss. zu G{\"o}ttingen, Math-Phys.
  Klassc}, pages 235--257. 1918.

\bibitem[Ovsjanikov et~al.(2008)Ovsjanikov, Sun, and
  Guibas]{ovsjanikov2008global}
Maks Ovsjanikov, Jian Sun, and Leonidas Guibas.
\newblock Global intrinsic symmetries of shapes.
\newblock \emph{Computer Graphics Forum}, 27\penalty0 (5):\penalty0 1341--1348,
  2008.

\bibitem[Ovsjanikov et~al.(2012)Ovsjanikov, Ben-Chen, Solomon, Butscher, and
  Guibas]{ovsjanikov2012functional}
Maks Ovsjanikov, Mirela Ben-Chen, Justin Solomon, Adrian Butscher, and Leonidas
  Guibas.
\newblock Functional maps: a flexible representation of maps between shapes.
\newblock \emph{ACM Trans. Graphics}, 31\penalty0 (4):\penalty0 1--11, 2012.

\bibitem[Pal et~al.(2020)Pal, Eksombatchai, Zhou, Zhao, Rosenberg, and
  Leskovec]{pal2020pinnersage}
Aditya Pal, Chantat Eksombatchai, Yitong Zhou, Bo~Zhao, Charles Rosenberg, and
  Jure Leskovec.
\newblock Pinnersage: Multi-modal user embedding framework for recommendations
  at pinterest.
\newblock In \emph{KDD}, 2020.

\bibitem[Parisot et~al.(2018)Parisot, Ktena, Ferrante, Lee, Guerrero, Glocker,
  and Rueckert]{parisot2018disease}
Sarah Parisot, Sofia~Ira Ktena, Enzo Ferrante, Matthew Lee, Ricardo Guerrero,
  Ben Glocker, and Daniel Rueckert.
\newblock Disease prediction using graph convolutional networks: application to
  autism spectrum disorder and alzheimer’s disease.
\newblock \emph{Medical Image Analysis}, 48:\penalty0 117--130, 2018.

\bibitem[Pascanu et~al.(2013)Pascanu, Mikolov, and
  Bengio]{pascanu2013difficulty}
Razvan Pascanu, Tomas Mikolov, and Yoshua Bengio.
\newblock On the difficulty of training recurrent neural networks.
\newblock In \emph{ICML}, 2013.

\bibitem[Patan{\`e}(2020)]{patane2020fourier}
Giuseppe Patan{\`e}.
\newblock Fourier-based and rational graph filters for spectral processing.
\newblock \emph{arXiv:2011.04055}, 2020.

\bibitem[Pearl(2014)]{pearl2014probabilistic}
Judea Pearl.
\newblock \emph{Probabilistic reasoning in intelligent systems: networks of
  plausible inference}.
\newblock Elsevier, 2014.

\bibitem[Penrose(2005)]{penrose2005road}
Roger Penrose.
\newblock \emph{The road to reality: A complete guide to the laws of the
  universe}.
\newblock Random House, 2005.

\bibitem[Perozzi et~al.(2014)Perozzi, Al-Rfou, and Skiena]{perozzi2014deepwalk}
Bryan Perozzi, Rami Al-Rfou, and Steven Skiena.
\newblock Deepwalk: Online learning of social representations.
\newblock In \emph{KDD}, 2014.

\bibitem[Pfaff et~al.(2020)Pfaff, Fortunato, Sanchez-Gonzalez, and
  Battaglia]{pfaff2020learning}
Tobias Pfaff, Meire Fortunato, Alvaro Sanchez-Gonzalez, and Peter~W Battaglia.
\newblock Learning mesh-based simulation with graph networks.
\newblock \emph{arXiv:2010.03409}, 2020.

\bibitem[Pineda(1988)]{pineda1988generalization}
Fernando~J Pineda.
\newblock Generalization of back propagation to recurrent and higher order
  neural networks.
\newblock In \emph{NIPS}, 1988.

\bibitem[Pinkall and Polthier(1993)]{pinkall1993computing}
Ulrich Pinkall and Konrad Polthier.
\newblock Computing discrete minimal surfaces and their conjugates.
\newblock \emph{Experimental Mathematics}, 2\penalty0 (1):\penalty0 15--36,
  1993.

\bibitem[Pinkus(1999)]{pinkus1999approximation}
Allan Pinkus.
\newblock Approximation theory of the mlp model in neural networks.
\newblock \emph{Acta Numerica}, 8:\penalty0 143--195, 1999.

\bibitem[Pollard et~al.(2018)Pollard, Johnson, Raffa, Celi, Mark, and
  Badawi]{pollard2018eicu}
Tom~J Pollard, Alistair~EW Johnson, Jesse~D Raffa, Leo~A Celi, Roger~G Mark,
  and Omar Badawi.
\newblock The eicu collaborative research database, a freely available
  multi-center database for critical care research.
\newblock \emph{Scientific Data}, 5\penalty0 (1):\penalty0 1--13, 2018.

\bibitem[Portilla and Simoncelli(2000)]{portilla2000parametric}
Javier Portilla and Eero~P Simoncelli.
\newblock A parametric texture model based on joint statistics of complex
  wavelet coefficients.
\newblock \emph{International journal of computer vision}, 40\penalty0
  (1):\penalty0 49--70, 2000.

\bibitem[Qi et~al.(2017)Qi, Su, Mo, and Guibas]{qi2017pointnet}
Charles~R Qi, Hao Su, Kaichun Mo, and Leonidas~J Guibas.
\newblock Pointnet: Deep learning on point sets for 3d classification and
  segmentation.
\newblock In \emph{CVPR}, 2017.

\bibitem[Qiu et~al.(2018)Qiu, Dong, Ma, Li, Wang, and Tang]{qiu2018network}
Jiezhong Qiu, Yuxiao Dong, Hao Ma, Jian Li, Kuansan Wang, and Jie Tang.
\newblock Network embedding as matrix factorization: Unifying deepwalk, line,
  pte, and node2vec.
\newblock In \emph{WSDM}, 2018.

\bibitem[Qu and Gouskos(2019)]{qu1902particlenet}
H~Qu and L~Gouskos.
\newblock Particlenet: jet tagging via particle clouds.
\newblock \emph{arXiv:1902.08570}, 2019.

\bibitem[Qu et~al.(2019)Qu, Bengio, and Tang]{qu2019gmnn}
Meng Qu, Yoshua Bengio, and Jian Tang.
\newblock {GMNN}: Graph {M}arkov neural networks.
\newblock In \emph{ICML}, 2019.

\bibitem[Radford et~al.(2018)Radford, Narasimhan, Salimans, and
  Sutskever]{radford2018improving}
Alec Radford, Karthik Narasimhan, Tim Salimans, and Ilya Sutskever.
\newblock Improving language understanding by generative pre-training.
\newblock 2018.

\bibitem[Radford et~al.(2019)Radford, Wu, Child, Luan, Amodei, and
  Sutskever]{radford2019language}
Alec Radford, Jeffrey Wu, Rewon Child, David Luan, Dario Amodei, and Ilya
  Sutskever.
\newblock Language models are unsupervised multitask learners.
\newblock \emph{OpenAI blog}, 1\penalty0 (8):\penalty0 9, 2019.

\bibitem[Ranjan et~al.(2018)Ranjan, Bolkart, Sanyal, and
  Black]{ranjan2018generating}
Anurag Ranjan, Timo Bolkart, Soubhik Sanyal, and Michael~J Black.
\newblock Generating {3D} faces using convolutional mesh autoencoders.
\newblock In \emph{ECCV}, 2018.

\bibitem[Raviv et~al.(2007)Raviv, Bronstein, Bronstein, and
  Kimmel]{raviv2007symmetries}
Dan Raviv, Alexander~M Bronstein, Michael~M Bronstein, and Ron Kimmel.
\newblock Symmetries of non-rigid shapes.
\newblock In \emph{ICCV}, 2007.

\bibitem[Razin and Cohen(2020)]{razin2020implicit}
Noam Razin and Nadav Cohen.
\newblock Implicit regularization in deep learning may not be explainable by
  norms.
\newblock \emph{arXiv:2005.06398}, 2020.

\bibitem[Reed and De~Freitas(2015)]{reed2015neural}
Scott Reed and Nando De~Freitas.
\newblock Neural programmer-interpreters.
\newblock \emph{arXiv:1511.06279}, 2015.

\bibitem[Ren et~al.(2015)Ren, He, Girshick, and Sun]{ren2015faster}
Shaoqing Ren, Kaiming He, Ross Girshick, and Jian Sun.
\newblock Faster r-cnn: Towards real-time object detection with region proposal
  networks.
\newblock \emph{arXiv:1506.01497}, 2015.

\bibitem[Rezende and Mohamed(2015)]{rezende2015variational}
Danilo Rezende and Shakir Mohamed.
\newblock Variational inference with normalizing flows.
\newblock In \emph{ICML}, 2015.

\bibitem[Riesenhuber and Poggio(1999)]{riesenhuber1999hierarchical}
Maximilian Riesenhuber and Tomaso Poggio.
\newblock Hierarchical models of object recognition in cortex.
\newblock \emph{Nature neuroscience}, 2\penalty0 (11):\penalty0 1019--1025,
  1999.

\bibitem[Robinson and Fallside(1987)]{robinson1987utility}
AJ~Robinson and Frank Fallside.
\newblock \emph{The utility driven dynamic error propagation network}.
\newblock University of Cambridge, 1987.

\bibitem[Rocheteau et~al.(2020)Rocheteau, Li{\`o}, and
  Hyland]{rocheteau2020temporal}
Emma Rocheteau, Pietro Li{\`o}, and Stephanie Hyland.
\newblock Temporal pointwise convolutional networks for length of stay
  prediction in the intensive care unit.
\newblock \emph{arXiv:2007.09483}, 2020.

\bibitem[Rocheteau et~al.(2021)Rocheteau, Tong, Veli{\v{c}}kovi{\'c}, Lane, and
  Li{\`o}]{rocheteau2021predicting}
Emma Rocheteau, Catherine Tong, Petar Veli{\v{c}}kovi{\'c}, Nicholas Lane, and
  Pietro Li{\`o}.
\newblock Predicting patient outcomes with graph representation learning.
\newblock \emph{arXiv:2101.03940}, 2021.

\bibitem[Ronneberger et~al.(2015)Ronneberger, Fischer, and
  Brox]{ronneberger2015u}
Olaf Ronneberger, Philipp Fischer, and Thomas Brox.
\newblock U-net: Convolutional networks for biomedical image segmentation.
\newblock In \emph{MICCAI}, 2015.

\bibitem[Rosenblatt(1958)]{rosenblatt1958perceptron}
Frank Rosenblatt.
\newblock The perceptron: a probabilistic model for information storage and
  organization in the brain.
\newblock \emph{Psychological Review}, 65\penalty0 (6):\penalty0 386, 1958.

\bibitem[Rossi et~al.(2020)Rossi, Chamberlain, Frasca, Eynard, Monti, and
  Bronstein]{rossi2020temporal}
Emanuele Rossi, Ben Chamberlain, Fabrizio Frasca, Davide Eynard, Federico
  Monti, and Michael Bronstein.
\newblock Temporal graph networks for deep learning on dynamic graphs.
\newblock \emph{arXiv:2006.10637}, 2020.

\bibitem[Russakovsky et~al.(2015)Russakovsky, Deng, Su, Krause, Satheesh, Ma,
  Huang, Karpathy, Khosla, Bernstein, et~al.]{russakovsky2015imagenet}
Olga Russakovsky, Jia Deng, Hao Su, Jonathan Krause, Sanjeev Satheesh, Sean Ma,
  Zhiheng Huang, Andrej Karpathy, Aditya Khosla, Michael Bernstein, et~al.
\newblock Imagenet large scale visual recognition challenge.
\newblock \emph{IJCV}, 115\penalty0 (3):\penalty0 211--252, 2015.

\bibitem[Rustamov et~al.(2013)Rustamov, Ovsjanikov, Azencot, Ben-Chen, Chazal,
  and Guibas]{rustamov2013map}
Raif~M Rustamov, Maks Ovsjanikov, Omri Azencot, Mirela Ben-Chen,
  Fr{\'e}d{\'e}ric Chazal, and Leonidas Guibas.
\newblock Map-based exploration of intrinsic shape differences and variability.
\newblock \emph{ACM Trans. Graphics}, 32\penalty0 (4):\penalty0 1--12, 2013.

\bibitem[Salimans and Kingma(2016)]{salimans2016weight}
Tim Salimans and Diederik~P Kingma.
\newblock Weight normalization: A simple reparameterization to accelerate
  training of deep neural networks.
\newblock \emph{arXiv:1602.07868}, 2016.

\bibitem[Sanchez-Gonzalez et~al.(2019)Sanchez-Gonzalez, Bapst, Cranmer, and
  Battaglia]{sanchez2019hamiltonian}
Alvaro Sanchez-Gonzalez, Victor Bapst, Kyle Cranmer, and Peter Battaglia.
\newblock Hamiltonian graph networks with {ODE} integrators.
\newblock \emph{arXiv:1909.12790}, 2019.

\bibitem[Sanchez-Gonzalez et~al.(2020)Sanchez-Gonzalez, Godwin, Pfaff, Ying,
  Leskovec, and Battaglia]{sanchez2020learning}
Alvaro Sanchez-Gonzalez, Jonathan Godwin, Tobias Pfaff, Rex Ying, Jure
  Leskovec, and Peter Battaglia.
\newblock Learning to simulate complex physics with graph networks.
\newblock In \emph{ICML}, 2020.

\bibitem[Sandryhaila and Moura(2013)]{sandryhaila2013discrete}
Aliaksei Sandryhaila and Jos{\'e}~MF Moura.
\newblock Discrete signal processing on graphs.
\newblock \emph{IEEE Trans. Signal Processing}, 61\penalty0 (7):\penalty0
  1644--1656, 2013.

\bibitem[Santoro et~al.(2017)Santoro, Raposo, Barrett, Malinowski, Pascanu,
  Battaglia, and Lillicrap]{santoro2017simple}
Adam Santoro, David Raposo, David~G Barrett, Mateusz Malinowski, Razvan
  Pascanu, Peter Battaglia, and Timothy Lillicrap.
\newblock A simple neural network module for relational reasoning.
\newblock In \emph{NIPS}, 2017.

\bibitem[Santoro et~al.(2018)Santoro, Faulkner, Raposo, Rae, Chrzanowski,
  Weber, Wierstra, Vinyals, Pascanu, and Lillicrap]{santoro2018relational}
Adam Santoro, Ryan Faulkner, David Raposo, Jack Rae, Mike Chrzanowski,
  Theophane Weber, Daan Wierstra, Oriol Vinyals, Razvan Pascanu, and Timothy
  Lillicrap.
\newblock Relational recurrent neural networks.
\newblock \emph{arXiv:1806.01822}, 2018.

\bibitem[Santurkar et~al.(2018)Santurkar, Tsipras, Ilyas, and
  Madry]{santurkar2018does}
Shibani Santurkar, Dimitris Tsipras, Andrew Ilyas, and Aleksander Madry.
\newblock How does batch normalization help optimization?
\newblock \emph{arXiv:1805.11604}, 2018.

\bibitem[Sato et~al.(2020)Sato, Yamada, and Kashima]{sato2020random}
Ryoma Sato, Makoto Yamada, and Hisashi Kashima.
\newblock Random features strengthen graph neural networks.
\newblock \emph{arXiv:2002.03155}, 2020.

\bibitem[Satorras et~al.(2021)Satorras, Hoogeboom, and Welling]{satorras2021n}
Victor~Garcia Satorras, Emiel Hoogeboom, and Max Welling.
\newblock {E}(n) equivariant graph neural networks.
\newblock \emph{arXiv:2102.09844}, 2021.

\bibitem[Scaife and Porter(2021)]{scaife2021fanaroff}
Anna~MM Scaife and Fiona Porter.
\newblock {Fanaroff-Riley} classification of radio galaxies using
  group-equivariant convolutional neural networks.
\newblock \emph{Monthly Notices of the Royal Astronomical Society}, 2021.

\bibitem[Scarselli et~al.(2008)Scarselli, Gori, Tsoi, Hagenbuchner, and
  Monfardini]{scarselli2008graph}
Franco Scarselli, Marco Gori, Ah~Chung Tsoi, Markus Hagenbuchner, and Gabriele
  Monfardini.
\newblock The graph neural network model.
\newblock \emph{IEEE Trans. Neural Networks}, 20\penalty0 (1):\penalty0 61--80,
  2008.

\bibitem[Schrittwieser et~al.(2020)Schrittwieser, Antonoglou, Hubert, Simonyan,
  Sifre, Schmitt, Guez, Lockhart, Hassabis, Graepel,
  et~al.]{schrittwieser2020mastering}
Julian Schrittwieser, Ioannis Antonoglou, Thomas Hubert, Karen Simonyan,
  Laurent Sifre, Simon Schmitt, Arthur Guez, Edward Lockhart, Demis Hassabis,
  Thore Graepel, et~al.
\newblock Mastering atari, go, chess and shogi by planning with a learned
  model.
\newblock \emph{Nature}, 588\penalty0 (7839):\penalty0 604--609, 2020.

\bibitem[Schulman et~al.(2017)Schulman, Wolski, Dhariwal, Radford, and
  Klimov]{schulman2017proximal}
John Schulman, Filip Wolski, Prafulla Dhariwal, Alec Radford, and Oleg Klimov.
\newblock Proximal policy optimization algorithms.
\newblock \emph{arXiv:1707.06347}, 2017.

\bibitem[Sch{\"u}tt et~al.(2018)Sch{\"u}tt, Sauceda, Kindermans, Tkatchenko,
  and M{\"u}ller]{schutt2018schnet}
Kristof~T Sch{\"u}tt, Huziel~E Sauceda, P-J Kindermans, Alexandre Tkatchenko,
  and K-R M{\"u}ller.
\newblock Schnet--a deep learning architecture for molecules and materials.
\newblock \emph{The Journal of Chemical Physics}, 148\penalty0 (24):\penalty0
  241722, 2018.

\bibitem[Sejnowski et~al.(1986)Sejnowski, Kienker, and
  Hinton]{sejnowski1986learning}
Terrence~J Sejnowski, Paul~K Kienker, and Geoffrey~E Hinton.
\newblock Learning symmetry groups with hidden units: Beyond the perceptron.
\newblock \emph{Physica D: Nonlinear Phenomena}, 22\penalty0 (1-3):\penalty0
  260--275, 1986.

\bibitem[Senior et~al.(2020)Senior, Evans, Jumper, Kirkpatrick, Sifre, Green,
  Qin, {\v{Z}}{\'\i}dek, Nelson, Bridgland, et~al.]{senior2020improved}
Andrew~W Senior, Richard Evans, John Jumper, James Kirkpatrick, Laurent Sifre,
  Tim Green, Chongli Qin, Augustin {\v{Z}}{\'\i}dek, Alexander~WR Nelson, Alex
  Bridgland, et~al.
\newblock Improved protein structure prediction using potentials from deep
  learning.
\newblock \emph{Nature}, 577\penalty0 (7792):\penalty0 706--710, 2020.

\bibitem[Serre et~al.(2007)Serre, Oliva, and Poggio]{serre2007feedforward}
Thomas Serre, Aude Oliva, and Tomaso Poggio.
\newblock A feedforward architecture accounts for rapid categorization.
\newblock \emph{Proceedings of the national academy of sciences}, 104\penalty0
  (15):\penalty0 6424--6429, 2007.

\bibitem[Shamir and Vardi(2020)]{shamir2020implicit}
Ohad Shamir and Gal Vardi.
\newblock Implicit regularization in relu networks with the square loss.
\newblock \emph{arXiv:2012.05156}, 2020.

\bibitem[Shawe-Taylor(1989)]{shawe1989building}
John Shawe-Taylor.
\newblock Building symmetries into feedforward networks.
\newblock In \emph{ICANN}, 1989.

\bibitem[Shawe-Taylor(1993)]{shawe1993symmetries}
John Shawe-Taylor.
\newblock Symmetries and discriminability in feedforward network architectures.
\newblock \emph{IEEE Trans. Neural Networks}, 4\penalty0 (5):\penalty0
  816--826, 1993.

\bibitem[Shervashidze et~al.(2011)Shervashidze, Schweitzer, Van~Leeuwen,
  Mehlhorn, and Borgwardt]{shervashidze2011weisfeiler}
Nino Shervashidze, Pascal Schweitzer, Erik~Jan Van~Leeuwen, Kurt Mehlhorn, and
  Karsten~M Borgwardt.
\newblock Weisfeiler-lehman graph kernels.
\newblock \emph{JMLR}, 12\penalty0 (9), 2011.

\bibitem[Shlomi et~al.(2020)Shlomi, Battaglia, and Vlimant]{shlomi2020graph}
Jonathan Shlomi, Peter Battaglia, and Jean-Roch Vlimant.
\newblock Graph neural networks in particle physics.
\newblock \emph{Machine Learning: Science and Technology}, 2\penalty0
  (2):\penalty0 021001, 2020.

\bibitem[Shuman et~al.(2013)Shuman, Narang, Frossard, Ortega, and
  Vandergheynst]{shuman2013emerging}
David~I Shuman, Sunil~K Narang, Pascal Frossard, Antonio Ortega, and Pierre
  Vandergheynst.
\newblock The emerging field of signal processing on graphs: Extending
  high-dimensional data analysis to networks and other irregular domains.
\newblock \emph{IEEE Signal Processing Magazine}, 30\penalty0 (3):\penalty0
  83--98, 2013.

\bibitem[Siegelmann and Sontag(1995)]{siegelmann1995computational}
Hava~T Siegelmann and Eduardo~D Sontag.
\newblock On the computational power of neural nets.
\newblock \emph{Journal of Computer and System Sciences}, 50\penalty0
  (1):\penalty0 132--150, 1995.

\bibitem[Silver et~al.(2016)Silver, Huang, Maddison, Guez, Sifre, Van
  Den~Driessche, Schrittwieser, Antonoglou, Panneershelvam, Lanctot,
  et~al.]{silver2016mastering}
David Silver, Aja Huang, Chris~J Maddison, Arthur Guez, Laurent Sifre, George
  Van Den~Driessche, Julian Schrittwieser, Ioannis Antonoglou, Veda
  Panneershelvam, Marc Lanctot, et~al.
\newblock Mastering the game of go with deep neural networks and tree search.
\newblock \emph{Nature}, 529\penalty0 (7587):\penalty0 484--489, 2016.

\bibitem[Silver et~al.(2017)Silver, Schrittwieser, Simonyan, Antonoglou, Huang,
  Guez, Hubert, Baker, Lai, Bolton, et~al.]{silver2017mastering}
David Silver, Julian Schrittwieser, Karen Simonyan, Ioannis Antonoglou, Aja
  Huang, Arthur Guez, Thomas Hubert, Lucas Baker, Matthew Lai, Adrian Bolton,
  et~al.
\newblock Mastering the game of go without human knowledge.
\newblock \emph{Nature}, 550\penalty0 (7676):\penalty0 354--359, 2017.

\bibitem[Simoncelli and Freeman(1995)]{simoncelli1995steerable}
Eero~P Simoncelli and William~T Freeman.
\newblock The steerable pyramid: A flexible architecture for multi-scale
  derivative computation.
\newblock In \emph{Proceedings., International Conference on Image Processing},
  volume~3, pages 444--447. IEEE, 1995.

\bibitem[Simonyan and Zisserman(2014)]{simonyan2014very}
Karen Simonyan and Andrew Zisserman.
\newblock Very deep convolutional networks for large-scale image recognition.
\newblock \emph{arXiv:1409.1556}, 2014.

\bibitem[Smola et~al.(2007)Smola, Gretton, Song, and
  Sch{\"o}lkopf]{smola2007hilbert}
Alex Smola, Arthur Gretton, Le~Song, and Bernhard Sch{\"o}lkopf.
\newblock A {H}ilbert space embedding for distributions.
\newblock In \emph{ALT}, 2007.

\bibitem[Spalevi{\'c} et~al.(2020)Spalevi{\'c}, Veli{\v{c}}kovi{\'c},
  Kova{\v{c}}evi{\'c}, and Nikoli{\'c}]{spalevic2020hierachial}
Stefan Spalevi{\'c}, Petar Veli{\v{c}}kovi{\'c}, Jovana Kova{\v{c}}evi{\'c},
  and Mladen Nikoli{\'c}.
\newblock Hierachial protein function prediction with tail-{GNN}s.
\newblock \emph{arXiv:2007.12804}, 2020.

\bibitem[Sperduti(1994)]{sperduti1994encoding}
Alessandro Sperduti.
\newblock Encoding labeled graphs by labeling {RAAM}.
\newblock In \emph{NIPS}, 1994.

\bibitem[Sperduti and Starita(1997)]{sperduti1997supervised}
Alessandro Sperduti and Antonina Starita.
\newblock Supervised neural networks for the classification of structures.
\newblock \emph{IEEE Trans. Neural Networks}, 8\penalty0 (3):\penalty0
  714--735, 1997.

\bibitem[Springenberg et~al.(2014)Springenberg, Dosovitskiy, Brox, and
  Riedmiller]{springenberg2014striving}
Jost~Tobias Springenberg, Alexey Dosovitskiy, Thomas Brox, and Martin
  Riedmiller.
\newblock Striving for simplicity: The all convolutional net.
\newblock \emph{arXiv:1412.6806}, 2014.

\bibitem[Srinivasan and Ribeiro(2019)]{srinivasan2019equivalence}
Balasubramaniam Srinivasan and Bruno Ribeiro.
\newblock On the equivalence between positional node embeddings and structural
  graph representations.
\newblock \emph{arXiv:1910.00452}, 2019.

\bibitem[Srivastava et~al.(2014)Srivastava, Hinton, Krizhevsky, Sutskever, and
  Salakhutdinov]{srivastava2014dropout}
Nitish Srivastava, Geoffrey Hinton, Alex Krizhevsky, Ilya Sutskever, and Ruslan
  Salakhutdinov.
\newblock Dropout: a simple way to prevent neural networks from overfitting.
\newblock \emph{JMLR}, 15\penalty0 (1):\penalty0 1929--1958, 2014.

\bibitem[Srivastava et~al.(2015)Srivastava, Greff, and
  Schmidhuber]{srivastava2015highway}
Rupesh~Kumar Srivastava, Klaus Greff, and J{\"u}rgen Schmidhuber.
\newblock Highway networks.
\newblock \emph{arXiv:1505.00387}, 2015.

\bibitem[Stachenfeld et~al.(2020)Stachenfeld, Godwin, and
  Battaglia]{stachenfeld2020graph}
Kimberly Stachenfeld, Jonathan Godwin, and Peter Battaglia.
\newblock Graph networks with spectral message passing.
\newblock \emph{arXiv:2101.00079}, 2020.

\bibitem[Stokes et~al.(2020)Stokes, Yang, Swanson, Jin, Cubillos-Ruiz, Donghia,
  MacNair, French, Carfrae, Bloom-Ackerman, et~al.]{stokes2020deep}
Jonathan~M Stokes, Kevin Yang, Kyle Swanson, Wengong Jin, Andres Cubillos-Ruiz,
  Nina~M Donghia, Craig~R MacNair, Shawn French, Lindsey~A Carfrae, Zohar
  Bloom-Ackerman, et~al.
\newblock A deep learning approach to antibiotic discovery.
\newblock \emph{Cell}, 180\penalty0 (4):\penalty0 688--702, 2020.

\bibitem[Strathmann et~al.(2021)Strathmann, Barekatain, Blundell, and
  Veli{\v{c}}kovi{\'c}]{strathmann2021persistent}
Heiko Strathmann, Mohammadamin Barekatain, Charles Blundell, and Petar
  Veli{\v{c}}kovi{\'c}.
\newblock Persistent message passing.
\newblock \emph{arXiv:2103.01043}, 2021.

\bibitem[Straumann(1996)]{straumann1996early}
Norbert Straumann.
\newblock Early history of gauge theories and weak interactions.
\newblock \emph{hep-ph/9609230}, 1996.

\bibitem[Sun et~al.(2009)Sun, Ovsjanikov, and Guibas]{sun2009concise}
Jian Sun, Maks Ovsjanikov, and Leonidas Guibas.
\newblock A concise and provably informative multi-scale signature based on
  heat diffusion.
\newblock \emph{Computer Graphics Forum}, 28\penalty0 (5):\penalty0 1383--1392,
  2009.

\bibitem[Sutskever et~al.(2014)Sutskever, Vinyals, and
  Le]{sutskever2014sequence}
Ilya Sutskever, Oriol Vinyals, and Quoc~V Le.
\newblock Sequence to sequence learning with neural networks.
\newblock \emph{arXiv:1409.3215}, 2014.

\bibitem[Szegedy et~al.(2015)Szegedy, Liu, Jia, Sermanet, Reed, Anguelov,
  Erhan, Vanhoucke, and Rabinovich]{szegedy2015going}
Christian Szegedy, Wei Liu, Yangqing Jia, Pierre Sermanet, Scott Reed, Dragomir
  Anguelov, Dumitru Erhan, Vincent Vanhoucke, and Andrew Rabinovich.
\newblock Going deeper with convolutions.
\newblock In \emph{CVPR}, 2015.

\bibitem[Tallec and Ollivier(2018)]{tallec2018can}
Corentin Tallec and Yann Ollivier.
\newblock Can recurrent neural networks warp time?
\newblock \emph{arXiv:1804.11188}, 2018.

\bibitem[Tang et~al.(2020)Tang, Huang, Gu, Lu, and Su]{tang2020towards}
Hao Tang, Zhiao Huang, Jiayuan Gu, Bao-Liang Lu, and Hao Su.
\newblock Towards scale-invariant graph-related problem solving by iterative
  homogeneous gnns.
\newblock In \emph{NeurIPS}, 2020.

\bibitem[Tang et~al.(2015)Tang, Qu, Wang, Zhang, Yan, and Mei]{tang2015line}
Jian Tang, Meng Qu, Mingzhe Wang, Ming Zhang, Jun Yan, and Qiaozhu Mei.
\newblock Line: Large-scale information network embedding.
\newblock In \emph{WWW}, 2015.

\bibitem[Taubin et~al.(1996)Taubin, Zhang, and Golub]{taubin1996optimal}
Gabriel Taubin, Tong Zhang, and Gene Golub.
\newblock Optimal surface smoothing as filter design.
\newblock In \emph{ECCV}, 1996.

\bibitem[Thakoor et~al.(2021)Thakoor, Tallec, Azar, Munos,
  Veli{\v{c}}kovi{\'c}, and Valko]{thakoor2021bootstrapped}
Shantanu Thakoor, Corentin Tallec, Mohammad~Gheshlaghi Azar, R{\'e}mi Munos,
  Petar Veli{\v{c}}kovi{\'c}, and Michal Valko.
\newblock Bootstrapped representation learning on graphs.
\newblock \emph{arXiv:2102.06514}, 2021.

\bibitem[Thomas et~al.(2018)Thomas, Smidt, Kearnes, Yang, Li, Kohlhoff, and
  Riley]{thomas2018tensor}
Nathaniel Thomas, Tess Smidt, Steven Kearnes, Lusann Yang, Li~Li, Kai Kohlhoff,
  and Patrick Riley.
\newblock Tensor field networks: Rotation-and translation-equivariant neural
  networks for {3D} point clouds.
\newblock \emph{arXiv:1802.08219}, 2018.

\bibitem[Tobies(2019)]{tobies2019felix}
Renate Tobies.
\newblock Felix {K}lein—-mathematician, academic organizer, educational
  reformer.
\newblock In \emph{The Legacy of Felix Klein}, pages 5--21. Springer, 2019.

\bibitem[Trask et~al.(2018)Trask, Hill, Reed, Rae, Dyer, and
  Blunsom]{trask2018neural}
Andrew Trask, Felix Hill, Scott Reed, Jack Rae, Chris Dyer, and Phil Blunsom.
\newblock Neural arithmetic logic units.
\newblock \emph{arXiv:1808.00508}, 2018.

\bibitem[Tromp and Farneb{\"a}ck(2006)]{tromp2006combinatorics}
John Tromp and Gunnar Farneb{\"a}ck.
\newblock Combinatorics of go.
\newblock In \emph{International Conference on Computers and Games}, 2006.

\bibitem[Tsybakov(2008)]{tsybakov2008introduction}
Alexandre~B Tsybakov.
\newblock \emph{Introduction to nonparametric estimation}.
\newblock Springer, 2008.

\bibitem[Ulyanov et~al.(2016)Ulyanov, Vedaldi, and
  Lempitsky]{ulyanov2016instance}
Dmitry Ulyanov, Andrea Vedaldi, and Victor Lempitsky.
\newblock Instance normalization: The missing ingredient for fast stylization.
\newblock \emph{arXiv:1607.08022}, 2016.

\bibitem[van~den Oord et~al.(2016{\natexlab{a}})van~den Oord, Dieleman, Zen,
  Simonyan, Vinyals, Graves, Kalchbrenner, Senior, and
  Kavukcuoglu]{oord2016wavenet}
Aaron van~den Oord, Sander Dieleman, Heiga Zen, Karen Simonyan, Oriol Vinyals,
  Alex Graves, Nal Kalchbrenner, Andrew Senior, and Koray Kavukcuoglu.
\newblock Wavenet: A generative model for raw audio.
\newblock \emph{arXiv:1609.03499}, 2016{\natexlab{a}}.

\bibitem[van~den Oord et~al.(2016{\natexlab{b}})van~den Oord, Kalchbrenner, and
  Kavukcuoglu]{van2016pixel}
Aaron van~den Oord, Nal Kalchbrenner, and Koray Kavukcuoglu.
\newblock Pixel recurrent neural networks.
\newblock In \emph{ICML}, 2016{\natexlab{b}}.

\bibitem[Vaswani et~al.(2017)Vaswani, Shazeer, Parmar, Uszkoreit, Jones, Gomez,
  Kaiser, and Polosukhin]{vaswani2017attention}
Ashish Vaswani, Noam Shazeer, Niki Parmar, Jakob Uszkoreit, Llion Jones,
  Aidan~N Gomez, {\L}ukasz Kaiser, and Illia Polosukhin.
\newblock Attention is all you need.
\newblock In \emph{NIPS}, 2017.

\bibitem[Veli{\v{c}}kovi{\'{c}} et~al.(2018)Veli{\v{c}}kovi{\'{c}}, Cucurull,
  Casanova, Romero, Li{\`{o}}, and Bengio]{velickovic2018graph}
Petar Veli{\v{c}}kovi{\'{c}}, Guillem Cucurull, Arantxa Casanova, Adriana
  Romero, Pietro Li{\`{o}}, and Yoshua Bengio.
\newblock {Graph Attention Networks}.
\newblock \emph{ICLR}, 2018.

\bibitem[Veli{\v{c}}kovi{\'c} et~al.(2019)Veli{\v{c}}kovi{\'c}, Ying, Padovano,
  Hadsell, and Blundell]{velivckovic2019neural}
Petar Veli{\v{c}}kovi{\'c}, Rex Ying, Matilde Padovano, Raia Hadsell, and
  Charles Blundell.
\newblock Neural execution of graph algorithms.
\newblock \emph{arXiv:1910.10593}, 2019.

\bibitem[Veli{\v{c}}kovi{\'c} et~al.(2020)Veli{\v{c}}kovi{\'c}, Buesing,
  Overlan, Pascanu, Vinyals, and Blundell]{velivckovic2020pointer}
Petar Veli{\v{c}}kovi{\'c}, Lars Buesing, Matthew~C Overlan, Razvan Pascanu,
  Oriol Vinyals, and Charles Blundell.
\newblock Pointer graph networks.
\newblock \emph{arXiv:2006.06380}, 2020.

\bibitem[Veli\v{c}kovi\'{c} et~al.(2019)Veli\v{c}kovi\'{c}, Fedus, Hamilton,
  Li\`{o}, Bengio, and Hjelm]{velickovic2019deep}
Petar Veli\v{c}kovi\'{c}, Wiliam Fedus, William~L. Hamilton, Pietro Li\`{o},
  Yoshua Bengio, and R~Devon Hjelm.
\newblock {Deep Graph Infomax}.
\newblock In \emph{ICLR}, 2019.

\bibitem[Veselkov et~al.(2019)Veselkov, Gonzalez, Aljifri, Galea, Mirnezami,
  Youssef, Bronstein, and Laponogov]{veselkov2019hyperfoods}
Kirill Veselkov, Guadalupe Gonzalez, Shahad Aljifri, Dieter Galea, Reza
  Mirnezami, Jozef Youssef, Michael Bronstein, and Ivan Laponogov.
\newblock Hyperfoods: Machine intelligent mapping of cancer-beating molecules
  in foods.
\newblock \emph{Scientific Reports}, 9\penalty0 (1):\penalty0 1--12, 2019.

\bibitem[Vinyals et~al.(2015)Vinyals, Fortunato, and
  Jaitly]{vinyals2015pointer}
Oriol Vinyals, Meire Fortunato, and Navdeep Jaitly.
\newblock Pointer networks.
\newblock \emph{arXiv:1506.03134}, 2015.

\bibitem[Vinyals et~al.(2016)Vinyals, Bengio, and Kudlur]{vinyals2016order}
Oriol Vinyals, Samy Bengio, and Manjunath Kudlur.
\newblock {Order matters: Sequence to sequence for sets}.
\newblock In \emph{ICLR}, 2016.

\bibitem[von Luxburg and Bousquet(2004)]{von2004distance}
Ulrike von Luxburg and Olivier Bousquet.
\newblock Distance-based classification with lipschitz functions.
\newblock \emph{JMLR}, 5:\penalty0 669--695, 2004.

\bibitem[Wainwright and Jordan(2008)]{wainwright2008graphical}
Martin~J Wainwright and Michael~Irwin Jordan.
\newblock \emph{Graphical models, exponential families, and variational
  inference}.
\newblock Now Publishers Inc, 2008.

\bibitem[Wang and Solomon(2019)]{wang2019intrinsic}
Yu~Wang and Justin Solomon.
\newblock Intrinsic and extrinsic operators for shape analysis.
\newblock In \emph{Handbook of Numerical Analysis}, volume~20, pages 41--115.
  Elsevier, 2019.

\bibitem[Wang et~al.(2018)Wang, Ben-Chen, Polterovich, and
  Solomon]{wang2018steklov}
Yu~Wang, Mirela Ben-Chen, Iosif Polterovich, and Justin Solomon.
\newblock Steklov spectral geometry for extrinsic shape analysis.
\newblock \emph{ACM Trans. Graphics}, 38\penalty0 (1):\penalty0 1--21, 2018.

\bibitem[Wang et~al.(2019{\natexlab{a}})Wang, Kim, Bronstein, and
  Solomon]{wang2019learning}
Yu~Wang, Vladimir Kim, Michael Bronstein, and Justin Solomon.
\newblock Learning geometric operators on meshes.
\newblock In \emph{ICLR Workshops}, 2019{\natexlab{a}}.

\bibitem[Wang et~al.(2019{\natexlab{b}})Wang, Sun, Liu, Sarma, Bronstein, and
  Solomon]{wang2019dynamic}
Yue Wang, Yongbin Sun, Ziwei Liu, Sanjay~E Sarma, Michael~M Bronstein, and
  Justin~M Solomon.
\newblock Dynamic graph {CNN} for learning on point clouds.
\newblock \emph{ACM Trans. Graphics}, 38\penalty0 (5):\penalty0 1--12,
  2019{\natexlab{b}}.

\bibitem[Wardetzky(2008)]{wardetzky2008convergence}
Max Wardetzky.
\newblock Convergence of the cotangent formula: An overview.
\newblock \emph{Discrete Differential Geometry}, pages 275--286, 2008.

\bibitem[Wardetzky et~al.(2007)Wardetzky, Mathur, K{\"a}lberer, and
  Grinspun]{wardetzky2007discrete}
Max Wardetzky, Saurabh Mathur, Felix K{\"a}lberer, and Eitan Grinspun.
\newblock Discrete {L}aplace operators: no free lunch.
\newblock In \emph{Symposium on Geometry Processing}, 2007.

\bibitem[Weiler et~al.(2018)Weiler, Geiger, Welling, Boomsma, and
  Cohen]{weiler20183d}
Maurice Weiler, Mario Geiger, Max Welling, Wouter Boomsma, and Taco Cohen.
\newblock 3d steerable cnns: Learning rotationally equivariant features in
  volumetric data.
\newblock \emph{arXiv:1807.02547}, 2018.

\bibitem[Weisfeiler and Leman(1968)]{weisfeiler1968reduction}
Boris Weisfeiler and Andrei Leman.
\newblock The reduction of a graph to canonical form and the algebra which
  appears therein.
\newblock \emph{NTI Series}, 2\penalty0 (9):\penalty0 12--16, 1968.

\bibitem[Werbos(1988)]{werbos1988generalization}
Paul~J Werbos.
\newblock Generalization of backpropagation with application to a recurrent gas
  market model.
\newblock \emph{Neural Networks}, 1\penalty0 (4):\penalty0 339--356, 1988.

\bibitem[Weyl(1929)]{weyl1929elektron}
Hermann Weyl.
\newblock Elektron und gravitation. i.
\newblock \emph{Zeitschrift f{\"u}r Physik}, 56\penalty0 (5-6):\penalty0
  330--352, 1929.

\bibitem[Weyl(2015)]{weyl2015symmetry}
Hermann Weyl.
\newblock \emph{Symmetry}.
\newblock Princeton University Press, 2015.

\bibitem[Winkels and Cohen(2019)]{winkels2019pulmonary}
Marysia Winkels and Taco~S Cohen.
\newblock Pulmonary nodule detection in ct scans with equivariant cnns.
\newblock \emph{Medical Image Analysis}, 55:\penalty0 15--26, 2019.

\bibitem[Wood and Shawe-Taylor(1996)]{wood1996representation}
Jeffrey Wood and John Shawe-Taylor.
\newblock Representation theory and invariant neural networks.
\newblock \emph{Discrete Applied Mathematics}, 69\penalty0 (1-2):\penalty0
  33--60, 1996.

\bibitem[Wu et~al.(2019)Wu, Souza, Zhang, Fifty, Yu, and
  Weinberger]{wu2019simplifying}
Felix Wu, Amauri Souza, Tianyi Zhang, Christopher Fifty, Tao Yu, and Kilian
  Weinberger.
\newblock Simplifying graph convolutional networks.
\newblock In \emph{ICML}, 2019.

\bibitem[Wu and He(2018)]{wu2018group}
Yuxin Wu and Kaiming He.
\newblock Group normalization.
\newblock In \emph{ECCV}, 2018.

\bibitem[Xu et~al.(2020{\natexlab{a}})Xu, Ruan, Korpeoglu, Kumar, and
  Achan]{xu2020inductive}
Da~Xu, Chuanwei Ruan, Evren Korpeoglu, Sushant Kumar, and Kannan Achan.
\newblock Inductive representation learning on temporal graphs.
\newblock \emph{arXiv:2002.07962}, 2020{\natexlab{a}}.

\bibitem[Xu et~al.(2018)Xu, Hu, Leskovec, and Jegelka]{xu2018powerful}
Keyulu Xu, Weihua Hu, Jure Leskovec, and Stefanie Jegelka.
\newblock How powerful are graph neural networks?
\newblock \emph{arXiv:1810.00826}, 2018.

\bibitem[Xu et~al.(2019)Xu, Li, Zhang, Du, Kawarabayashi, and
  Jegelka]{xu2019can}
Keyulu Xu, Jingling Li, Mozhi Zhang, Simon~S Du, Ken-ichi Kawarabayashi, and
  Stefanie Jegelka.
\newblock What can neural networks reason about?
\newblock \emph{arXiv:1905.13211}, 2019.

\bibitem[Xu et~al.(2020{\natexlab{b}})Xu, Li, Zhang, Du, Kawarabayashi, and
  Jegelka]{xu2020neural}
Keyulu Xu, Jingling Li, Mozhi Zhang, Simon~S Du, Ken-ichi Kawarabayashi, and
  Stefanie Jegelka.
\newblock How neural networks extrapolate: From feedforward to graph neural
  networks.
\newblock \emph{arXiv:2009.11848}, 2020{\natexlab{b}}.

\bibitem[Yan et~al.(2020)Yan, Swersky, Koutra, Ranganathan, and
  Heshemi]{yan2020neural}
Yujun Yan, Kevin Swersky, Danai Koutra, Parthasarathy Ranganathan, and Milad
  Heshemi.
\newblock Neural execution engines: Learning to execute subroutines.
\newblock \emph{arXiv:2006.08084}, 2020.

\bibitem[Yang and Mills(1954)]{yang1954conservation}
Chen-Ning Yang and Robert~L Mills.
\newblock Conservation of isotopic spin and isotopic gauge invariance.
\newblock \emph{Physical Review}, 96\penalty0 (1):\penalty0 191, 1954.

\bibitem[Yang et~al.(2016)Yang, Cohen, and Salakhudinov]{yang2016revisiting}
Zhilin Yang, William Cohen, and Ruslan Salakhudinov.
\newblock Revisiting semi-supervised learning with graph embeddings.
\newblock In \emph{ICML}, 2016.

\bibitem[Yedidia et~al.(2001)Yedidia, Freeman, and Weiss]{yedidia2001bethe}
Jonathan~S Yedidia, William~T Freeman, and Yair Weiss.
\newblock Bethe free energy, kikuchi approximations, and belief propagation
  algorithms.
\newblock \emph{NIPS}, 2001.

\bibitem[Ying et~al.(2018)Ying, He, Chen, Eksombatchai, Hamilton, and
  Leskovec]{ying2018graph}
Rex Ying, Ruining He, Kaifeng Chen, Pong Eksombatchai, William~L Hamilton, and
  Jure Leskovec.
\newblock Graph convolutional neural networks for web-scale recommender
  systems.
\newblock In \emph{KDD}, 2018.

\bibitem[You et~al.(2019)You, Ying, and Leskovec]{you2019position}
Jiaxuan You, Rex Ying, and Jure Leskovec.
\newblock Position-aware graph neural networks.
\newblock In \emph{ICML}, 2019.

\bibitem[Zaheer et~al.(2017)Zaheer, Kottur, Ravanbakhsh, Poczos, Salakhutdinov,
  and Smola]{zaheer2017deep}
Manzil Zaheer, Satwik Kottur, Siamak Ravanbakhsh, Barnabas Poczos, Russ~R
  Salakhutdinov, and Alexander~J Smola.
\newblock Deep sets.
\newblock In \emph{NIPS}, 2017.

\bibitem[Zaremba and Sutskever(2014)]{zaremba2014learning}
Wojciech Zaremba and Ilya Sutskever.
\newblock Learning to execute.
\newblock \emph{arXiv:1410.4615}, 2014.

\bibitem[Zeng et~al.(2012)Zeng, Guo, Luo, and Gu]{zeng2012discrete}
Wei Zeng, Ren Guo, Feng Luo, and Xianfeng Gu.
\newblock Discrete heat kernel determines discrete riemannian metric.
\newblock \emph{Graphical Models}, 74\penalty0 (4):\penalty0 121--129, 2012.

\bibitem[Zhang et~al.(2018)Zhang, Shi, Xie, Ma, King, and Yeung]{zhang2018gaan}
Jiani Zhang, Xingjian Shi, Junyuan Xie, Hao Ma, Irwin King, and Dit-Yan Yeung.
\newblock Gaan: Gated attention networks for learning on large and
  spatiotemporal graphs.
\newblock \emph{arXiv:1803.07294}, 2018.

\bibitem[Zhang et~al.(2020)Zhang, Chen, Yang, Ramamurthy, Li, Qi, and
  Song]{zhang2020efficient}
Yuyu Zhang, Xinshi Chen, Yuan Yang, Arun Ramamurthy, Bo~Li, Yuan Qi, and
  Le~Song.
\newblock Efficient probabilistic logic reasoning with graph neural networks.
\newblock \emph{arXiv:2001.11850}, 2020.

\bibitem[Zhu et~al.(2019)Zhu, Zhao, Yang, Lin, Zhou, Ai, Li, and
  Zhou]{zhu2019aligraph}
Rong Zhu, Kun Zhao, Hongxia Yang, Wei Lin, Chang Zhou, Baole Ai, Yong Li, and
  Jingren Zhou.
\newblock Aligraph: A comprehensive graph neural network platform.
\newblock \emph{arXiv:1902.08730}, 2019.

\bibitem[Zhu and Razavian(2019)]{zhu2019graph}
Weicheng Zhu and Narges Razavian.
\newblock Variationally regularized graph-based representation learning for
  electronic health records.
\newblock \emph{arXiv:1912.03761}, 2019.

\bibitem[Zhu et~al.(2020)Zhu, Xu, Yu, Liu, Wu, and Wang]{zhu2020deep}
Yanqiao Zhu, Yichen Xu, Feng Yu, Qiang Liu, Shu Wu, and Liang Wang.
\newblock Deep graph contrastive representation learning.
\newblock \emph{arXiv:2006.04131}, 2020.

\bibitem[Zitnik et~al.(2018)Zitnik, Agrawal, and Leskovec]{zitnik2018modeling}
Marinka Zitnik, Monica Agrawal, and Jure Leskovec.
\newblock Modeling polypharmacy side effects with graph convolutional networks.
\newblock \emph{Bioinformatics}, 34\penalty0 (13):\penalty0 i457--i466, 2018.

\end{thebibliography}

\end{document}